\documentclass[a4,10pt,twoside,openright,italian,english]{book}

\usepackage[twoside=true]{geometry}
\usepackage[cam,center,a4,pdflatex,axes]{crop}

\usepackage{phdthesis}

\usepackage{color}

\usepackage{array}
\usepackage{mdwmath}
\usepackage{mdwtab}
\usepackage{amsmath,amssymb}
\usepackage[numbers]{natbib}

\usepackage{listings}
\usepackage{subcaption}
\usepackage{booktabs}
\usepackage{latexsym}
\usepackage{bnf}
\usepackage{hyperref}
\hypersetup{
    colorlinks=true,
    linkcolor=blue,
    filecolor=magenta,      
    urlcolor=cyan,
}
\usepackage{cleveref} 
\usepackage{rotating}
\usepackage{multirow}
\usepackage{phdtitle}
\usepackage{paralist}
\usepackage{bibentry}
\usepackage{lscape}
\usepackage{longtable}
\usepackage[T1]{fontenc}
\usepackage[utf8]{inputenc}
\usepackage{tabularx}
\usepackage[dvipsnames]{xcolor}
\usepackage{collcell,xfp}

\usepackage{graphbox}
\usepackage{caption}
\usepackage{tablefootnote}

\usepackage{eurosym}
\usepackage{siunitx}
\usepackage{xfrac}
\usepackage{dcolumn}

\usepackage{algorithm}
\usepackage[noend]{algpseudocode}

\algnewcommand\algorithmicforeach{\textbf{for each}}
\algdef{S}[FOR]{ForEach}[1]{\algorithmicforeach\ #1\ \algorithmicdo}
\DeclareMathOperator*{\argmax}{arg\,max}

\usepackage{amsfonts}
\usepackage{graphicx}
\usepackage{textcomp}
\usepackage{mathtools}
\DeclarePairedDelimiter\abs{\lvert}{\rvert}%
\DeclarePairedDelimiter\norm{\lVert}{\rVert}%
\makeatletter
\let\oldabs\abs
\def\abs{\@ifstar{\oldabs}{\oldabs*}}
\let\oldnorm\norm
\def\norm{\@ifstar{\oldnorm}{\oldnorm*}}
\makeatother

\usepackage[italian,main=english]{babel}
\usepackage{ulem}
\hypersetup{pdftitle={Deep Learning Techniques for Visual Counting}, pdfauthor={Luca Ciampi}}

\nobibliography*

\department {Dottorato di ricerca in Ingegneria dell'Informazione}

\author{Luca Ciampi}
\title{Deep Learning Techniques for Visual Counting}
\tutor{Prof. Hazım Kemal Ekenel, Prof. Roberto Caldelli}
\supervisor{Dr. Giuseppe Amato, Prof. Marco Avvenuti, Dr. Claudio Gennaro}
\titleimage{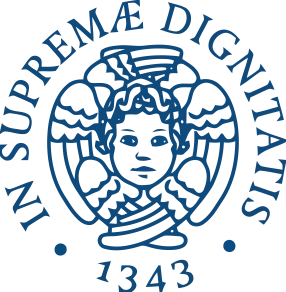} 
\phdyear{2022}
\phdmonth{April}
\phdcycle{Cycle XXXIV}

\definecolor{helmet}{RGB}{197, 17, 98}
\definecolor{hvv}{RGB}{0, 191, 165}
\definecolor{mask}{RGB}{255, 109, 0}

\usepackage{pifont}

\newacronym{ml}{ML}{Machine Learning}
\newacronym{dl}{DL}{Deep Learning}
\newacronym{dnn}{DNN}{Deep Neural Network}
\newacronym{cnn}{CNN}{Convolutional Neural Network}
\newacronym{dcnn}{DCNN}{Deep Convolutional Neural Network}
\newacronym{lstm}{LSTM}{Long Short-term Memory}
\newacronym{sgd}{SGD}{Stochastic Gradient Descent}
\newacronym{adam}{Adam}{Adaptive Moment Estimation}
\newacronym{relu}{ReLU}{Rectified Linear Unit}
\newacronym{crelu}{CReLU}{Concatenated Rectified Linear Unit}
\newacronym{resnet}{ResNet}{Residual Network}
\newacronym{bn}{BN}{Batch Normalization}
\newacronym{rpn}{RPN}{Region Proposal Network}
\newacronym{nms}{NMS}{Non-Maximum Suppression}
\newacronym{da}{DA}{Domain Adaptation}
\newacronym{uda}{UDA}{Unsupervised Domain Adaptation}
\newacronym{svm}{SVM}{Support Vector Machine}
\newacronym{fcn}{FCN}{Fully Convolutional Network}
\newacronym{tl}{TL}{Transfer Learning}
\newacronym{gan}{GAN}{Generative Adversarial Network}

\newacronym{yolo}{YOLO}{You Only Look Once}
\newacronym{ssd}{SSD}{Single Shot Detector}

\newacronym{ilsvrc}{ILSVRC}{ImageNet Large Scale Visual Recognition Challenge}
\newacronym{coco}{COCO}{Common Objects in Context}
\newacronym{trancos}{TRANCOS}{TRaffic ANd COngestionS}
\newacronym{webcamt}{WebCamT}{Webcam Traffic Video Dataset}
\newacronym{jta}{JTA}{Joint Track Auto}
\newacronym{gta}{GTA}{Grand Traffic Auto}
\newacronym{viped}{ViPeD}{Virtual Pedestrian Dataset}
\newacronym{mot17}{MOT17Det}{Multiple Object Tracking Detection Challenge 2017}
\newacronym{mot20}{MOT20Det}{Multiple Object Tracking Detection Challenge 2020}
\newacronym{carpk}{CARPK}{Car Parking Lot Dataset}
\newacronym{pucpr+}{PUCPR+}{PUCPR+}
\newacronym{pklot}{PKLot}{PKLot}
\newacronym{ndispark}{NDISPark}{Night and Day Instance Segmented Park} 
\newacronym{mbm}{MBM Cells}{Modified Bone Marrow Cells}
\newacronym{gcc}{GCC}{GTA5 Crowd Counting}
\newacronym{adi}{ADI Cells}{ADIpocyte Cells}
\newacronym{bcd}{BCData}{Breast Cancer Dataset}

\newacronym{gpu}{GPU}{Graphical Processing Unit}
\newacronym{vgg}{VGG}{Visual Geometry Group}
\newacronym{ai}{AI}{Artificial Intelligence}
\newacronym{uav}{UAV}{Unmanned Aerial Vehicle}
\newacronym{fps}{FPS}{Frames Per Second}
\newacronym{ppe}{PPE}{Personal Protective Equipment}
\newacronym{fov}{FOV}{Field Of View}
\newacronym{ransac}{RANSAC}{RANdom SAmple Consensus}

\newacronym{mae}{MAE}{Mean Absolute Error}
\newacronym{mse}{MSE}{Mean Squared Error}
\newacronym{rmse}{RMSE}{Root Mean Squared Error}
\newacronym{mare}{MARE}{Mean Absolute Relative Error}
\newacronym{game}{GAME}{Grid Average Mean absolute Error}
\newacronym{ssim}{SSIM}{Structural Similarity Index Measure}
\newacronym{ap}{AP}{Average Precision}
\newacronym{map}{mAP}{mean Average Precision}
\newacronym{iou}{IoU}{Intersection over Union}
\newacronym{tp}{TP}{True Positive}
\newacronym{fp}{FP}{False Positive}
\newacronym{fn}{FN}{False Negative}
\newacronym{tn}{TN}{True Negative}
\newacronym{auc}{AUC}{Area Under Curve}
\newacronym{mmd}{MMD}{Maximum Mean Discrepancy}

\makeglossaries

\begin{document}
\selectlanguage{english}

\maketitle

\pagestyle{empty}

\cleardoublepage
\newpage

\thispagestyle{empty}
    \null\vspace{\stretch {1}}
        \begin{flushright}
           \textit{To my family}
        \end{flushright}
\vspace{\stretch{2}}\null

\cleardoublepage
\newpage

\pagestyle{empty}

\thispagestyle{empty}
    \null\vspace{\stretch {1}}
        \begin{flushright}
                ``If you look at a question like, 'Would an elephant fit though a doorway?', \\
                while most people can answer that question almost instantaneously, \\
                machines will struggle. \\
                What's easy for one is hard for the other, and vice versa. \\
                That is what I call the AI paradox.''\\
            
                Oren Etzioni
        \end{flushright}
\vspace{\stretch{2}}\null

\cleardoublepage
\newpage

\pagestyle{empty}
\setcounter{page}{1}
\pagenumbering{Roman}

\chapter*{Acknowledgements}
\lettrine{I}{t} seems that the end of this unexpected journey has come. The COVID-19 pandemic made this path even weirder. However, I was lucky to be surrounded by great people during my Ph.D. career, both physically and virtually.

Foremost, I would like to express my deepest gratitude to Dr. Claudio Gennaro and Dr. Giuseppe Amato, who offered me this opportunity, advising me throughout the entire path. 

A special thank you also goes to Nicola Messina, the 'companion' of this journey, and to Fabio Carrara and Dr. Fabrizio Falchi, with whom I collaborated closely in many of the research activities.

I gratefully acknowledge Prof. Marco Avvenuti for supervising my Ph.D., and Prof. Hazim Kemal Ekenel and Prof. Roberto Caldelli, who evaluated my work and provided valuable feedback.

I want to thank all the Artificial Intelligence for Media and Humanities lab (the NeMIS lab when I started the Ph.D.) for allowing me to join its team. I really enjoyed my time at the CNR because of you. Special mentions go to Fabrizio Sebastiani, Alejandro Moreo Fernàndez, Paolo Bolettieri, Fabio Valerio Massoli, Alessandro Nardi, Lucia Vadicamo, Marco Di Benedetto, Claudio Vairo and Franca Debole.

A special mention goes to Prof. Joao Paulo Costeira and Carlos Santiago from the Instituto Superior Técnico. Lisbon has remained in my heart.

My most profound appreciation goes to all my friends, past and present. For sure, I forget someone, however, thank you to Pilleri, Sara, Michele, Grazia, Lara, Adolfo, Vittorio, Giacomo, Zuba, Franchini, Debora, my namesake Luca Ciampi, Sbocca, Tonio, Tuma, Pipotto, Moya.

I am deeply grateful to my family: my parents, Rosa and Nicola, my brother Fabio and his companion Elisa, and the newcomer Mauro, also known as Maurizio.

Most of all, incommensurable gratefulness is due to my girlfriend and soul mate Alessia, the best person I know, for her incredible heart and, as in all things, her invaluable support.
She stood next to me during the entire journey, and I owe you so much for all the time spent together that I will always treasure.
Without you, it wouldn't have been possible.

\selectlanguage{italian}

\selectlanguage{english}

\cleardoublepage
\newpage

\pagestyle{fancy}

\selectlanguage{english}
\chapter*{Summary}
\lettrine{T}{he} explosion of \acrfull{dl} added a boost to the already rapidly developing field of Computer Vision to such a point that vision-based tasks are now parts of our everyday lives. 
Applications such as image classification, photo stylization, or face recognition are nowadays pervasive, as evidenced by the advent of modern systems trivially integrated into mobile applications.

In this thesis, we investigated and enhanced the visual counting task, which automatically estimates the number of objects in still images or video frames. 
Recently, due to the growing interest in it, several \acrfull{cnn}-based solutions have been suggested by the scientific community. These artificial neural networks, inspired by the organization of the animal visual cortex, provide a way to automatically learn effective representations from raw visual data and can be successfully employed to address typical challenges characterizing this task, such as low-quality images, different illuminations, and object scale variations.
But apart from these difficulties, in this dissertation, we identified some other crucial limitations in the adoption of \acrshortpl{cnn}, proposing general solutions that we experimentally evaluated in the context of the counting task which turns out to be particularly affected by these shortcomings.

In particular, we tackled the problem related to the lack of data needed for training current \acrshort{dl}-based solutions. Given that the budget for labeling is limited, data scarcity still represents an open problem that prevents the scalability of existing solutions based on the supervised learning of neural networks and that is responsible for a significant drop in performance at inference time when new scenarios are presented to these algorithms. This concern is particularly evident in tasks such as the counting one, where the objects to be labeled are hundreds, or even thousands, per image, significantly increasing the human effort needed for the annotation procedure.
We proposed solutions addressing this issue from several complementary sides. We introduced synthetic datasets gathered from virtual environments resembling the real world, where the training labels are automatically collected, therefore drastically reducing the human effort for the annotation procedure. We proposed \acrfull{da} strategies, both supervised and unsupervised, aiming at mitigating the domain gap existing between the training and test data distributions. We presented a counting strategy in a weakly labeled data scenario, i.e., in the presence of non-negligible disagreement between multiple annotators, enhancing counting performance by taking advantage of the redundant information due to raters' judgment differences.
Moreover, we tackled the non-trivial engineering challenges coming out of the adoption of \acrshort{cnn}-based techniques in environments with limited power resources, mainly due to the high computational budget the \acrshort{ai}-based algorithms require. We introduced solutions for counting vehicles directly onboard embedded vision systems, i.e., devices equipped with constrained computational capabilities that can capture images and elaborate them. Finally, we designed an embedded modular Computer Vision-based and \acrshort{ai}-assisted system that can carry out several tasks to help monitor individual and collective human safety rules, such as estimating the number of people present in a region of interest.

\selectlanguage{english}


\selectlanguage{italian}
\chapter*{Sommario}
\lettrine{L}{a} recente diffusione del Deep Learning ha ulteriormente accelerato il già rapido sviluppo della Computer Vision, fino al punto che molte applicazioni riguardanti questa disciplina fanno ormai parte della nostra quotidianità. La classificazione di immagini, la stilizzazione di foto, o il riconoscimento facciale, sono applicazioni diventate pervasive, come dimostrato dal fatto che sono sempre più spesso integrate nei dispositivi mobili, quali ad esempio gli smartphone.

In questa tesi, è stato considerato il conteggio visivo, che ha lo scopo di stimare automaticamente il numero di oggetti afferenti ad una determinata categoria presenti in immagini statiche o frame estratti da video. Recentemente questo argomento ha ricevuto una notevole attenzione da parte della comunità scientifica, la quale ha proposto numerosi soluzioni principalmente basate sulle reti neurali convoluzionali. Queste ultime sono particolari reti neurali artificiali che, ispirandosi alla corteccia visiva celebrale degli animali, sono in grado di apprendere automaticamente delle rappresentazioni numeriche efficaci per le immagini, partendo dai dati visivi grezzi (pixel); esse sono state appunto impiegate con successo anche per contrastare le principali difficoltà caratterizzanti il conteggio visivo, come ad esempio la bassa qualità delle immagini analizzate, le differenti illuminazioni e la variazione di grandezza degli oggetti. Oltre a questi ostacoli, in questa tesi sono stati identificati ulteriori limiti nell'adozione di questi algoritmi, proponendo soluzioni generali che sono state valutate sperimentalmente nel contesto del conteggio visivo, particolarmente afflitto da queste problematiche.

In particolare, è stato affrontato il problema derivante dalla scarsità di dati necessari per la fase di addestramento supervisionato di questi approcci. Posto che il budget per l'annotazione dei dati è limitato, la loro carenza rimane tutt'ora un problema irrisolto che limita la scalabilità delle soluzioni esistenti, e che è responsabile di un significativo degrado delle prestazioni quando questi algoritmi vengono impiegati in nuovi scenari. Questa problematica è particolarmente riscontrabile nelle applicazioni quali il conteggio visivo, che richiede l'annotazione manuale di centinaia, se non di migliaia, di oggetti per ogni singola immagine, facendo aumentare in maniera significativa lo sforzo umano necessario per sopperire a questa procedura. In questa tesi sono state proposte varie strategie che contrastano questo problema da diverse direzioni complementari. Sono stati introdotti dataset sintetici acquisiti da mondi virtuali che simulano il mondo reale, e dove le annotazioni necessarie per la fase di addestramento degli algoritmi basati sull' Intelligenza Artificiale sono collezionate automaticamente. Sono state proposte delle tecniche di Domain Adaptation, sia supervisionate che non supervisionate, aventi lo scopo di mitigare il gap esistente tra le distribuzioni dei dati utilizzati per la fase di addestramento e quella di test. E' stata presentata una strategia di conteggio visivo in un contesto in cui le annotazioni presentavano errori, ovvero una notevole discrepanza fra molteplici annotatori, traendo vantaggio dalle informazioni derivanti dalle differenze di giudizio di questi ultimi. Inoltre, è stato anche affrontato il non banale problema ingegneristico dovuto all'utilizzo delle reti neurali convoluzionali in contesti caratterizzati da scarse capacità computazionali. A questo proposito, sono state introdotte soluzioni per il conteggio visivo di veicoli effettuato direttamente all'interno di sistemi aventi ridotte capacità di calcolo, ma in grado di catturare ed elaborare immagini. Infine, è stato progettato e presentato un sistema modulare basato sulla Intelligenza Artificiale capace di espletare diversi compiti aventi lo scopo di aiutare a controllare il rispetto di regole nella sfera della sicurezza umana individuale e collettiva, come ad esempio monitorare il numero di persone presenti in una determinata zona di interesse.  

\selectlanguage{english}


\selectlanguage{english}
\chapter*{List of publications}


\section*{International Journals}
\begin{enumerate}
    \item Ciampi, L., Messina, N., Falchi, F., Gennaro, C., and Amato, G. (2020, September). Virtual to real adaptation of pedestrian detectors. \emph{Sensors}. (Vol. 20(18), pp. 5250). MDPI.
    \item Di Benedetto, M., Carrara, F., Ciampi, L., Falchi, F., Gennaro, C., and Amato, G. (2022, March). An Embedded Toolset for Human Activity Monitoring in Critical Environments. \emph{Expert Systems with Applications}. (In Press). Elsevier.
\end{enumerate}

\section*{International Conferences/Workshops with Peer Review}
\begin{enumerate}
    \item Ciampi, L., Carrara, F., Amato, G., and Gennaro, C. (2022, February). Counting or Localizing? Evaluating Cell Counting and Detection in Microscopy Images. In \emph{2022 International Joint Conference on Computer Vision, Imaging and Computer Graphics Theory and Applications (VISIGRAPP)}. (Vol. 4: VISAPP, pp. 887--897). SCITEPRESS.
    \item Ciampi, L., Santiago, C., Costeira, J. P., Gennaro, C., and Amato, G. (2021, February). Domain Adaptation for Traffic Density Estimation. In \emph{2021 International Joint Conference on Computer Vision, Imaging and Computer Graphics Theory and Applications (VISIGRAPP)}. (Vol. 5: VISAPP, pp. 185--195). SCITEPRESS.
    \item Ciampi, L., Santiago, C., Costeira, J. P., Gennaro, C., and Amato, G. (2020, September). Unsupervised Vehicle Counting via Multiple Camera Domain Adaptation. In \emph{2020 International Workshop on New Foundations for Human-Centered AI (NeHuAI) at 2020 European Conference on Artificial Intelligence (ECAI)}. (Vol. 2659, pp. 82--85). CEUR-WS.
    \item Amato, G., Ciampi, L., Falchi, F., Gennaro, C., and Messina, N. (2019, September). Learning pedestrian detection from virtual worlds. In \emph{International Conference on Image Analysis and Processing}. (pp. 302--312). Springer, Cham.
    \item Amato, G., Ciampi, L., Falchi, F., and Gennaro, C. (2019, June). Counting vehicles with deep learning in onboard uav imagery. In \emph{2019 IEEE Symposium on Computers and Communications (ISCC)}. (pp. 1--6). IEEE.
    \item Amato, G., Bolettieri, P., Moroni, D., Carrara, F., Ciampi, L., Pieri, G., Gennaro, C., Leone, G. R, and Vairo, C. (2018, December). A wireless smart camera network for parking monitoring. In \emph{2018 IEEE Globecom Workshops (GC Wkshps)}. (pp. 1--6). IEEE.
    \item Ciampi, L., Amato, G., Falchi, F., Gennaro, C., and Rabitti, F. (2018, June). Counting Vehicles with Cameras. In \emph{2018 Italian Symposium on Advanced Database Systems (SEBD)}. (Vol. 2161). CEUR-WS.

\end{enumerate}


\selectlanguage{english}

\printglossary[type=\acronymtype,title=List of Abbreviations]
\let\cleardoublepage\clearpage
\cleardoublepage
\newpage
\tableofcontents
\cleardoublepage
\newpage

\setcounter{page}{1}
\pagenumbering{arabic}

\cleardoublepage
\let\ref\Cref 
\newcommand{\figfrom}[1]{Image by~\citet{#1}}

\graphicspath{{img/intro/}}

\chapter{Introduction}
\label{ch:introduction}
The interest in a machine capable of simulating the powerful capacities of human vision is old.
It appears simple because people, even very young children, trivially solve it. Nevertheless, it largely remains an unsolved task because of the limited understanding of biological vision and the complexity of visual perception in a dynamic and nearly infinitely varying physical world. Nowadays, the significant diffusion of cheap cameras and smartphones leads to an exponential daily production of digital visual data, such as images and videos. In this context, a constant increase of attention to the automatic understanding of the visual content is currently taking place. Consequently, Computer Vision has become one of the hottest sub-fields of \acrlong{ai} and \acrlong{ml}, to such a point that applications of Computer Vision techniques have become pervasive and are now parts of our everyday lives. These include face recognition, photo stylization, or machine vision in self-driving cars, to name a few. This dissertation investigates and enhances the visual counting task, which aims to automatically estimate the number of objects in still images or video frames.

\begin{figure}[htbp]
    \centering
  \subfloat{
       \includegraphics[width=0.325\linewidth, height=3.2cm]{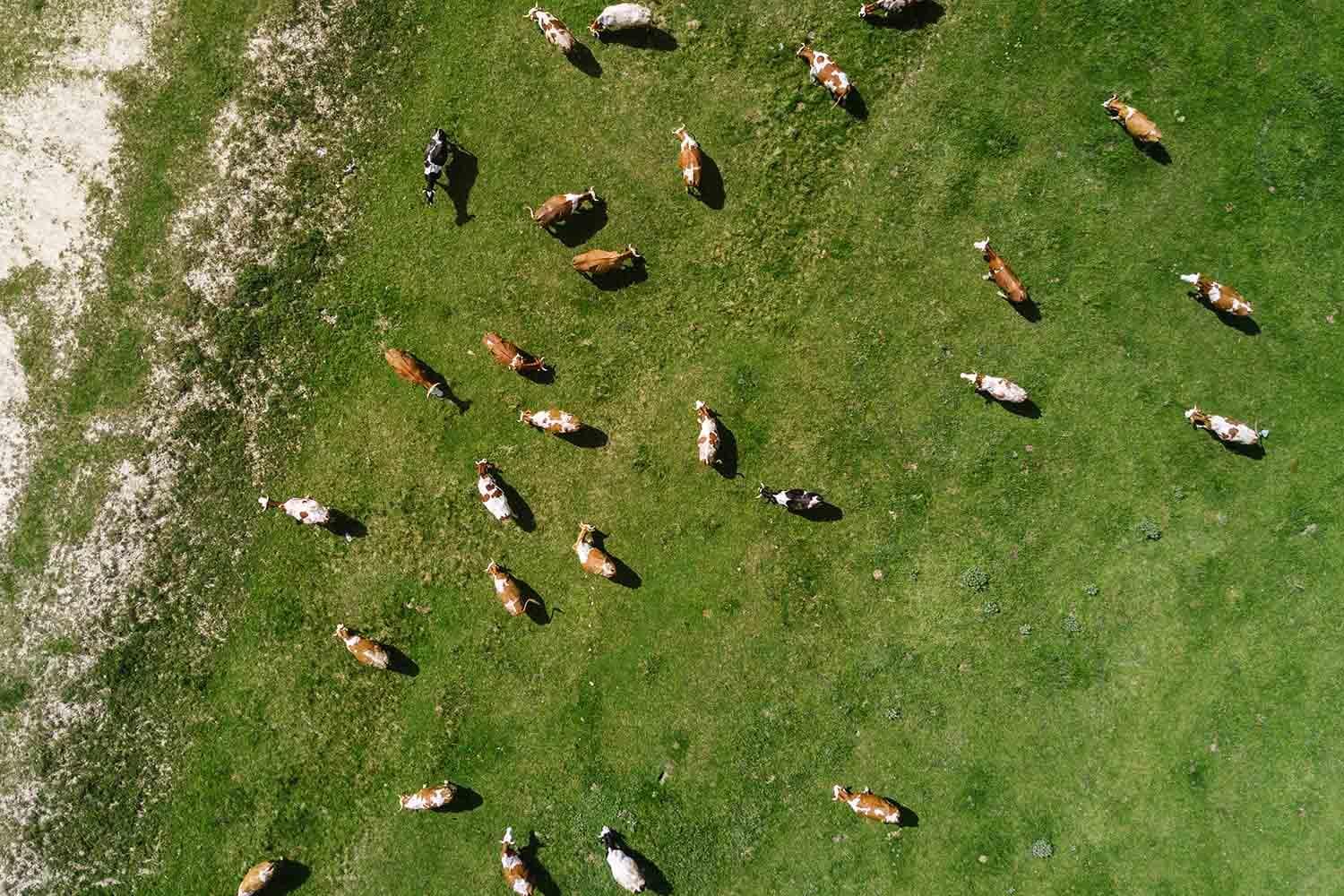}}
  \subfloat{
        \includegraphics[width=0.325\linewidth, height=3.2cm]{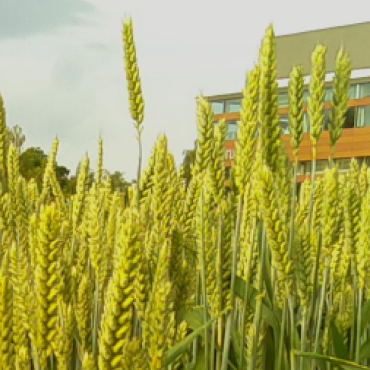}}
  \subfloat{
        \includegraphics[width=0.325\linewidth, height=3.2cm]{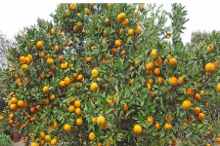}}
 \hfill
    \\ [1.1ex]
  \subfloat{
        \includegraphics[width=0.325\linewidth, height=3.2cm]{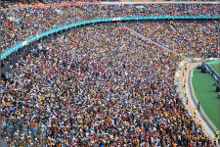}}
  \subfloat{
        \includegraphics[width=0.325\linewidth, height=3.2cm]{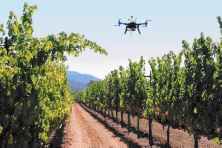}}
  \subfloat{
        \includegraphics[width=0.325\linewidth, height=3.2cm]{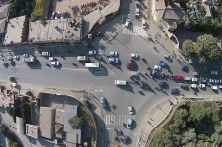}}
 \hfill
    \\ [1.1ex]
  \subfloat{
        \includegraphics[width=0.325\linewidth, height=3.2cm]{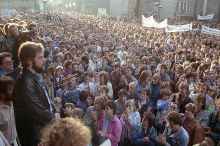}}
  \subfloat{
        \includegraphics[width=0.325\linewidth, height=3.2cm]{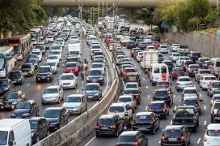}}
  \subfloat{
        \includegraphics[width=0.325\linewidth, height=3.2cm]{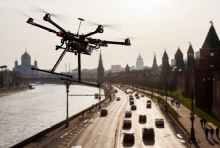}}
 \hfill
    \\ [1.1ex]
  \subfloat{
        \includegraphics[width=0.325\linewidth, height=3.2cm]{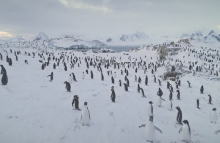}}
  \subfloat{
        \includegraphics[width=0.325\linewidth, height=3.2cm]{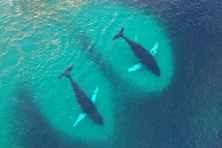}}
  \subfloat{
        \includegraphics[width=0.325\linewidth, height=3.2cm]{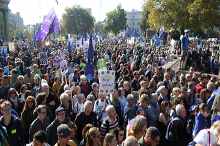}}
 \hfill
    \\ [1.1ex]
  \subfloat{
        \includegraphics[width=0.325\linewidth, height=3.2cm]{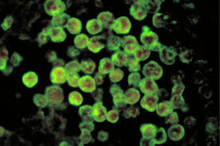}}
  \subfloat{
        \includegraphics[width=0.325\linewidth, height=3.2cm]{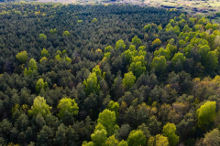}}
  \subfloat{
        \includegraphics[width=0.325\linewidth, height=3.2cm]{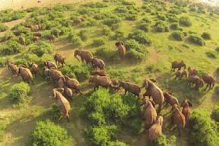}}
 \hfill

  \caption{\textbf{Some examples of real-world counting applications.} They range from estimating the number of people present at any event to counting vehicles in urban scenarios. But they also involve biology, counting cells from microscopic images, agriculture/farming helping to monitor the livestock or counting fruits on the trees, or even more wild contexts, counting animals in ecological surveys or leaves and tillers to assess plant health.}
  \label{visual_counting_applications} 
\end{figure}

Visual counting has attracted significant attention from many research communities due to its broad real-world applicability and its inherently interdisciplinary topic. A notable example in which it is involved is represented by crowd counting, where the goal is to estimate the number of people present at an event \cite{iterative_crowd_counting, crowd_counting_1}. The exponential growth in the world population and urbanization have led to increased sporting events, political rallies, or public demonstrations, resulting in more frequent crowd gatherings in recent years. Therefore, it is essential to analyze crowd behavior in such scenarios; thus, it is not surprising that crowd analysis is a research topic addressed by researchers from different communities (such as sociology \cite{socio_1, socio_2}, psychology \cite{psyco_1}, physics \cite{physics_1}, biology \cite{bio1, bio_2}, and public safety) since it is of paramount importance for a variety of critical standpoints and a multitude of perspectives, like the social, political or security-related ones. For example, according to their political positions, it is common for different sides (i.e., the organizers and the opposite parts) to claim different numbers for crowd gatherings. But apart from this, many scenarios involving crowd gatherings such as concerts and sports events address the risk of disasters related to the excessive number of present people. In such cases, the number of people can be exploited as an effective trigger for early overcrowding detection and an appropriate future strategy for managing the crowd \cite{abdelghany2014modeling, almeida11crowd}. Another prominent example of a real-world counting application is evaluating the number of vehicles in urban scenarios, like highways or parking areas \cite{cars_counting_aerial, costeira_lstm}. Building automated counting solutions to deal with this problem would allow the development of systems that precisely monitor the evolution of traffic jams. This information would be invaluable for the public authorities in charge of maintaining and planning road infrastructures. Other fields of applications comprise biology, where, for example, counting bacterial cells from microscopic images may be an indicator of the presence of some diseases \cite{count-ception, counting_cells_zisserman, cell_counting_2}. Also, in the agriculture/farming field, the counting task can help monitor the livestock, estimate the number of fruits on the trees \cite{fruits_1}, or count plants to assess the seedling emergence rate \cite{counting_plants_1, counting_plants_2, counting_plants_3}. More, counting objects is also important in more wild contexts, like counting animals in ecological surveys to monitor the population of a specific area \cite{counting_wild} or counting the number of trees in an aerial image of a forest. Counting flowers for estimating the start and duration of flowering is instead relevant for demarcating plant growth stages \cite{flower_1, flower_2}. In contrast, counting leaves and tillers is a trait pertinent to assessing plant health \cite{leaf_1, leaf_2}. We show some examples of real-world counting applications in \ref{visual_counting_applications}.

In humans, studies have demonstrated that as a consequence of the subitizing ability \cite{piazza2002subitizing}, the brain switches between two techniques to count objects \cite{hyde2011two}. The fast and accurate Parallel Individuation System (PIS) is employed when the observed objects are less than five. Otherwise, the inaccurate and error-prone Approximate Number System (ANS) is used. Thus, Computer Vision approaches offer a fast and helpful alternative for counting objects, at least for crowded scenes.

In principle, the key idea behind counting objects using Computer Vision-based techniques is elementary: density times area. However, objects are not localized regularly across the scene. Instead, they cluster in certain regions and are spread out in others. Another factor of complexity is represented by perspective distortions created by different camera viewpoints in various scenes, resulting in considerable variability of scales of objects. Other challenges to be considered are inter-object and intra-object occlusions, the high similarity of appearance between objects and background elements, different illuminations, and low image quality. Several machine learning-based approaches (especially supervised) have been suggested in the last years to overcome these challenges. In particular, significant improvements have been made through extensive use of \acrfull{dl}-based strategies \cite{deep_learning}, especially exploiting \acrlong{cnn}s (CNNs) \cite{lecun1998gradient}, a type of artificial neural network inspired by biological processes in that the connectivity pattern between neurons resembles the organization of the animal visual cortex.  \acrshort{cnn}s revolutionized feature engineering and visual understanding, outperforming handcrafted models on multiple vision tasks; they were recently made feasible, mainly thanks to the availability of powerful hardware and software infrastructure, as well as the increasing size of the datasets. 
This latter point is particularly crucial. The success of \acrshort{dl}-based methods often assumes the availability of a well-labeled and representative set of training images. However, the annotation process is an extremely costly, tedious, and error-prone procedure that requires a great human effort, especially in tasks such as the counting one, where the raters (i.e., the annotators) have to \textit{manually} label tens of thousands of objects, often characterized by non-trivial patterns. Furthermore, it is not only a matter of \textit{quantity} of labeled data to have in place but also of \textit{quality}. Training data should be error-free and cover the highest number of different scenarios and contexts to ensure that the algorithm can generalize to new data at inference time.
As a result, considering the limited labeling budget, data scarcity still represents an open problem that prevents the scalability of current \acrshort{dl}-based solutions.
Moreover, \acrshort{cnn}-based techniques pose non-trivial engineering challenges in their adoption mainly due to the high computational budget that drastically limits their applications in environments with limited power resources, which currently delegate complex data analysis to a centralized server.

In this thesis, we tackle these described critical limitations and challenges in the usage of \acrlong{dl}-based solutions in the counting task, proposing their adoption in novel approaches.

\section{Objectives and Contributions}
\label{sec:intro:contributions}
Excluding the first two chapters, which introduce our work and provide the reader with relevant background, we divide the dissertation into five parts covering different topics and aspects concerning the visual counting task. Specifically, we report below the significant contributions we propose in this study, highlighting the main challenges and the proposed solutions to tackle them.

\paragraph{Counting Vehicles Onboard Embedded Vision Systems.}
The ubiquity of video surveillance cameras in modern cities and the significant development of \acrfull{ai} provide new opportunities for the development of functional smart Computer Vision-based applications and services for citizens, mostly based on \acrshort{dl} solutions. 
However, this application is often hampered by the limited computational resources on disposable devices.
In this context, \ref{ch:counting-on-the-edge} explores the adoption of \acrshort{dl}-based solutions for counting vehicles directly onboard embedded vision systems, i.e., devices equipped with limited computational capabilities that can capture images and elaborate them. In particular, in our investigation, we propose a \acrshort{dl} solution to automatically detect and count vehicles in images taken from a \acrfull{uav}. We experiment over two real-world datasets showing that our approach results in state-of-the-art performances, running at a speed of 4 \acrfull{fps} on an NVIDIA Jetson TX2 board. Then, in the same chapter, we introduce a novel multi-camera system that combines a \acrshort{cnn}, which can locate and count vehicles present in images belonging to individual smart cameras, along with a decentralized geometry-based approach that is responsible for aggregating the data gathered from all the devices and estimating the number of cars present in the \textit{entire} parking lot. Again, all the computations are performed onboard the embedded vision systems. 

\paragraph{Virtual To Real Adaptation of Pedestrian Detectors.} A crucial task in many intelligent video surveillance systems is pedestrian detection since it is the main building block for a myriad of applications. Counting people is one of them. \acrshort{cnn}-based pedestrian detectors have demonstrated their superiority compared to the approaches relying on hand-crafted features. However, as mentioned above, the crux of \acrshort{cnn}s is that to generalize well at inference time, they require a massive amount of diverse labeled data during the training phase, covering the widest number of different scenarios. Since manually annotating new collections of images is expensive and requires a significant human effort, an appealing solution is to gather synthetic data from virtual environments resembling the real world, where the labels are \textit{automatically} collected by interacting with the graphical engine.
In this direction, in \ref{ch:virtual-to-real} we introduce \acrfull{viped}, a new synthetic dataset generated with the highly photo-realistic graphical engine of a video game that represents the first synthetic annotated collection of images suitable for the pedestrian detection task in the literature. We use it to train the \acrshort{cnn}-based detector.
However, data coming from virtual worlds cannot be fully exploited due to the Synthetic-to-Real Domain Shift, i.e., the image appearance difference between the synthetic training data and the real-world ones on which the pedestrian detector, in the end, shall be used. This domain gap between the two data distributions leads to performance degradation of the \acrshort{cnn} at test time, and so, intending to mitigate it, we propose two different \textit{Supervised \acrfull{da}} strategies.

\paragraph{\acrlong{uda} for Traffic Density Estimation.} Monitoring traffic flows in cities is crucial to improving urban mobility, and images are the best sensing modality to perceive and assess the flow of vehicles in large areas. However, as already stated, current machine learning-based technologies using images hinge on large quantities of annotated data, preventing their scalability to city-scale as new cameras are added to the system.
Scenarios that are never seen during the supervised training phase systematically lead to performance degradation of these approaches due to the existence of a \textit{Domain Shift} between the distributions of the training and test data. Gathering synthetic data is a promising solution since labels are automatically collected, as seen in the previous chapter. Still, data coming from virtual worlds cannot be fully exploited due to the Synthetic-to-Real Domain Shift.
To tackle these challenges, in \ref{ch:uda-counting} we propose a new methodology to design image-based vehicle density estimators and counting via an \textit{\acrfull{uda}} technique, differently from the previous chapter in which we instead rely on a \textit{supervised} adaptation of the network using real-world data.
In particular, during the training phase, we exploit the supervised learning provided by the synthetic automatically labeled data exploiting \acrfull{gta} dataset, the first collection of images with precise \textit{per-pixel} annotations gathered using the graphical engine of a video game,
and, at the same time, we infer some knowledge from the real-world \textit{unlabeled} images. In other words, we tackle the problem of data scarcity from two complementary sides: on the one hand, we exploit the significant variability of the synthetic data, while, on the other hand, we mitigate the domain gap existing between the synthetic and the real-world images in an \textit{unsupervised} fashion.
We base our unsupervised approach on adversarial learning performed directly on the generated density maps, i.e., in the output space, given that in this specific case, the output space contains valuable information such as scene layout and context. To the best of our knowledge, this is the first attempt to introduce a \acrshort{uda} scheme for the counting task. 
We conduct experiments considering the Synthetic-to-Real Domain Shift and accounting for other scenarios and datasets, showing the superiority of our approach compared to state-of-the-art techniques.

\paragraph{Counting Biological Structures with Raters’ Uncertainty.} 
\acrlong{dl} models have achieved astonishing results for counting biological structures in microscopy images. 
However, as already seen in the previous chapters concerning other tasks and applications, the success of these supervised methods assumes the availability of a representative set of \textit{well-labeled} images. In chapter \ref{ch:counting-with-uncertainty}, we tackle the problem of data scarcity in a different setting. In particular, we tackle the task of counting biological structures from microscopy images under the assumption of having training datasets characterized by \textit{weak labels}, that is, in the presence of non-negligible disagreement between multiple raters. This often occurs in medical images where non-trivial intrinsic patterns can produce weak annotations due to raters' judgment differences, even among experts.
More reliable labels can be obtained by aggregating and averaging the decisions given by several raters to the same data. Still, the scale of the counting task and the limited budget for labeling prohibit this. 
Consequently, raters prefer to label new data rather than label the same data more than once, resulting in large, single-rater weakly labeled datasets and very small multi-rater data, from which it is crucial to make the most. 
To this end, we propose a two-stage counting strategy in a weakly labeled data scenario.
In the first stage, we train state-of-the-art \acrshort{dl}-based methodologies to detect and count biological structures exploiting a large set of single-rater labeled data sure to contain errors; in the second stage, using a small set of multi-rater data, we refine the predictions, increasing the correlation between the scores assigned to the samples and the agreement of the raters on the annotations. 
We assess our methodology on a novel dataset comprising fluorescence microscopy images of mice brains containing extracellular matrix aggregates named perineuronal nets.
We demonstrate that we significantly enhance counting performance, improving confidence calibration by taking advantage of the redundant information characterizing the small sets of available multi-rater data.

\paragraph{Monitoring People in Critical Environments.} From a more practical perspective, in \ref{ch:counting-for-covid}, we present an embedded modular Computer Vision-based and \acrshort{ai}-assisted system that can carry out several tasks to help monitor individual and collective human safety rules, processing the captured images directly on an off-the-shelf commercial and low-cost device. Our solution put in practice some of the techniques described in the previous chapters and consists of multiple modules, each responsible for specific functionality that the user can easily enable and configure. In particular, by exploiting one of these modules or combining some of them, our framework makes available many capabilities. One of them aims at estimating the number of people present in a region of interest, a piece of crucial information to monitor the area occupancy. By measuring, and eventually limiting, the number of people who can visit a location at any one time, it is possible to reduce the likelihood of setting up people gatherings drastically.
To validate our solution, we test all the functionalities that our framework, deployed on an embedded device, makes available, exploiting some novel datasets that we collected and annotated on purpose. Experiments show that our system can effectively carry out all the functionalities that the user can set up, providing a valuable asset to monitor compliance with safety rules automatically.
\\

Finally, \ref{ch:conclusion} concludes the dissertation by discussing our contributions and defining new future research lines.

\graphicspath{{img/background/}}

\chapter{Background}
\label{ch:background}

In this chapter, we provide to the reader some primary notions, concepts, and related work about some topics addressed in this thesis, which will be helpful in a better understanding of them. First, we provide some notions about the main visual counting approaches, and we review some of the solutions introduced in the literature over the last few years, focusing on the \acrshort{cnn}-based ones. We also illustrate the main metrics exploited in the literature to measure the performance of the counting solutions. Then, we describe the more influential \acrshort{cnn}-based object detectors since they are the primary building block for the counting by detection approach. Next, we describe the Domain Adaptation task, a technique that tackles one of the cruces of the \acrshort{cnn}s and all the supervised learning methods, i.e., the need for a massive amount of labeled data to guarantee that they generalize well to diverse testing scenarios. This problem directly involves the counting task since the dataset annotation procedure is, in this case, particularly costly in terms of human effort. Finally, we give an overview of the datasets exploited in this dissertation.

\section{Visual Counting}
\label{sec:back:visual-counting}

Over the last few years, researchers have attempted to address the object counting task using a variety of approaches such as detection-based counting, clustering-based counting, regression-based counting, and counting by density estimation. The initial works mainly employed hand-crafted features while the more recent ones exploit \acrlong{cnn}s-based techniques that have demonstrated significant improvements. In this section, although this thesis focuses on \acrshort{cnn}-based approaches, we first briefly review methods using hand-crafted features for the sake of completeness. Thus, we illustrate the metrics commonly exploited in the literature to measure the performance of the counting solutions. Finally, we consider \acrshort{cnn}-based counting solutions. In particular, we review some of the more influential \acrshort{cnn}-based approaches performing the counting task by regression and, especially, by density estimation, which has been successfully applied in highly crowded scenarios.
On the other hand, we do not treat in this section \acrshort{cnn}-based detection approaches, but we prefer to illustrate the detectors upon they are built on in \ref{sec:back:cnn-based-detectors}.

\subsection{Review of Traditional Approaches}
\label{sec:back:visual-counting:traditional-approaches}

The authors in \cite{survey_traditional_approaches} and \cite{survey_sindagi} broadly classified traditional crowd counting methods based on hand-crafted features into the following categories: counting by detection, counting by clustering, counting by regression, and counting by density estimation. This section briefly reviews them, highlighting their strengths and weaknesses.

\subsubsection{Counting by Clustering}
The techniques belonging to the counting by clustering category tackle the counting task in an unsupervised way. A clear advantage is that such an approach does not need to be trained, and it is out of the box. These algorithms are left free to discover and present the structure in the data. However, the counting accuracy of such fully unsupervised methods is generally limited. These techniques fall into two categories: counting by self-similarities and counting by motion similarities.

\begin{figure}[htbp]
\centerline{\includegraphics[width=.80\textwidth]{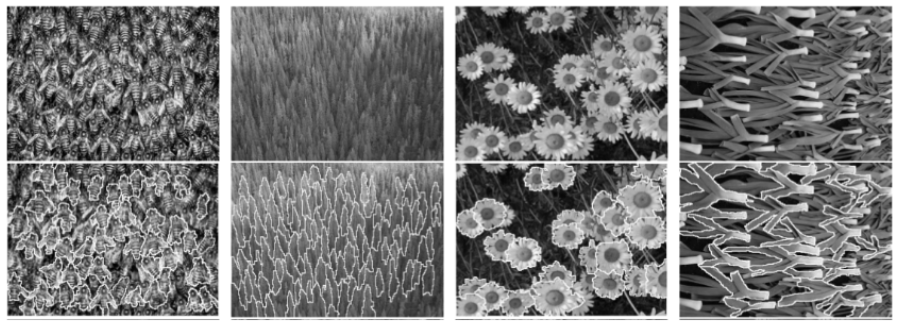}}
\caption{\textbf{Example of clustering by self similarities \cite{counting_clustering}.} Image features are probabilistically grouped into clusters.}
\label{fig:counting_clustering_self_similarities}
\end{figure}

The \textit{clustering by self-similarities} technique relies on tracking simple image features and probabilistically grouping them into clusters, like in \cite{counting_clustering}. Then, one can count clusters belonging to a certain category. An example of self-similarities clustering is shown in \ref{fig:counting_clustering_self_similarities}.

The \textit{clustering by motion similarities} approach relies instead on the assumption that a pair of points that appears to move together is likely to be part of the same individual. Hence coherent feature trajectories can be grouped to represent independently moving entities. Examples can be found in \cite{motion_clustering_1, motion_clustering_2}. In addition to a generally limited accuracy, the main drawback in such a method is that it only works with continuous image frames and not with static images. Furthermore, false estimations may arise when people remain static in a scene, exhibiting sustained articulations or sharing common feature trajectories over time. An example of motion similarities clustering is shown in \ref{fig:counting_clustering_self_similarities}.

\begin{figure}[htbp]
\centerline{\includegraphics[width=.30\textwidth]{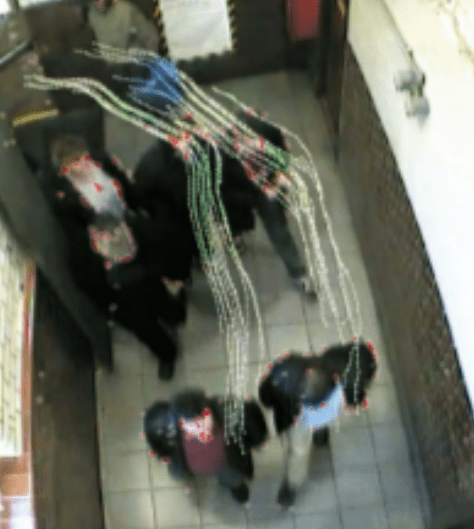}}
\caption{\textbf{Example of clustering by motion similarities \cite{motion_clustering_1}.} The assumption is that a pair of points that appears to move together is likely to be part of the same individual.}
\label{fig:counting_clustering_motion_similarities}
\end{figure}

\subsubsection{Counting by Detection}
The counting by detection approach is probably the most natural and intuitive counting method. It is a supervised approach where a sliding window detector previously trained is used to detect objects in the scene. This information is then used to count the number of objects. Even though this method is quite simple to understand, it suffers in scenes with occlusions. We can distinguish various categories of this counting technique: monolithic detection, part-based detection, and shape matching.

In the \textit{monolithic detection}, a classifier for the whole object appearance is trained using a set of training images. This approach typically employs features such as Haar wavelets, gradient-based features such as histograms of oriented gradient (HOG) feature \cite{hog}, edgelets \cite{edgelets} or shapelets \cite{shapelet}, to represent the whole object appearance. These methods, like \cite{monolithic_1, monolithic_2, monolithic_3}, are helpful in sparse scenes but suffer in settings having many objects with occlusions and clutters.

\begin{figure}[htbp]
\centerline{\includegraphics[width=.50\textwidth]{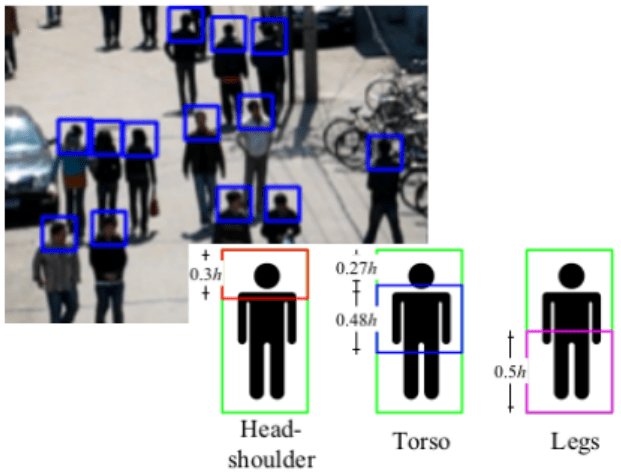}}
\caption{\textbf{Exmple of pedestrian detection using a part-based technique \cite{part_det_1, part_det_2}.} The classifier is trained for recognizing specific parts of a person (such as head and shoulders).}
\label{fig:counting_part_based_det}
\end{figure}

In the \textit{part-based detection}, a classifier for specific parts of an object (such as head and shoulders for people detection) is trained from a set of training images. This approach is a little more robust in crowded scenes than the previous one since we want to identify smaller but more distinct parts of an object. Examples in literature are \cite{part_det_1, part_det_2, part_det_3}. \ref{fig:counting_part_based_det} shows an example of part-based detection.

In the \textit{shape matching detection}, the trained classifier is about object shapes, for example, composed of ellipses. This classifier is then employed to search for the maximum a posteriori shape configuration of foreground objects, revealing the count and location and the pose of each object in a scene. Examples in literature are \cite{shape_det_1, shape_det_2}.

\subsubsection{Counting by Regression}
\begin{figure}[htbp]
\centerline{\includegraphics[width=.80\textwidth]{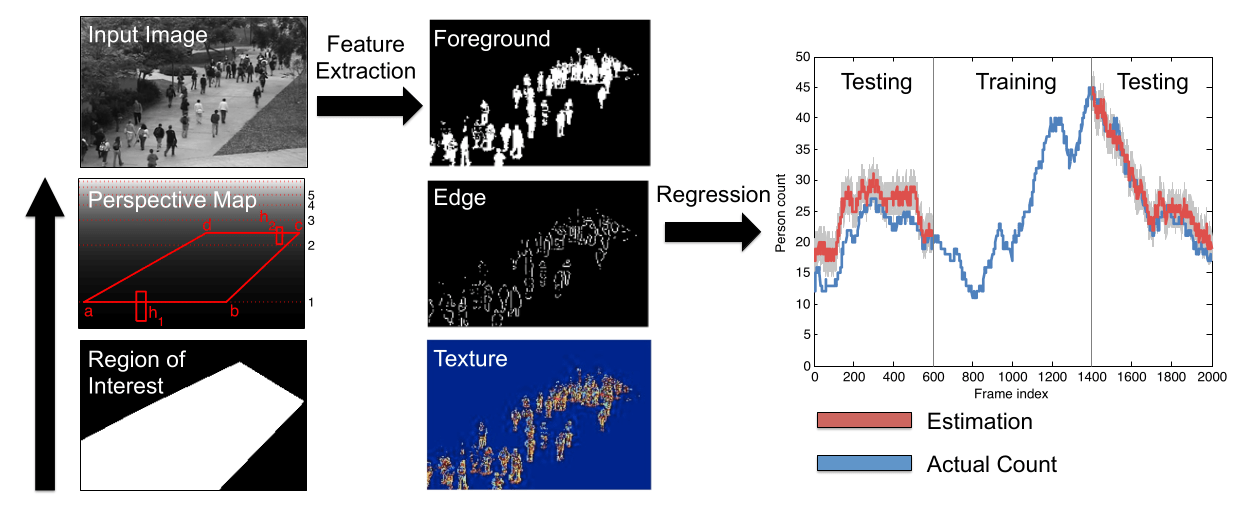}}
\caption{\textbf{A typical pipeline of counting by regression \cite{survey_traditional_approaches}.} In the first step, the so-called geometric correction is performed by defining a perspective map of the scene; then, holistic features are extracted and employed to train a regressor.}
\label{fig:regression_pipeline}
\end{figure}

As mentioned before, counting by detection and clustering approaches are unreliable when scenes present occluded objects. Counting by regression is a supervised method that avoids dependency on learning detectors (like counting by detection technique) and tracking features (like counting by clustering). Still, it estimates the count based on a holistic and collective description of objects patterns. In other words, this approach tries to establish a direct mapping (linear or not) from the image features to the number of objects present in an image without direct object detection or tracking. Since it does not rely on a specific classifier or model previously trained, it is more robust to occlusions and perspective distortions. A typical pipeline of counting by regression, shown in Figure \ref{fig:regression_pipeline}, consists of the following steps \cite{survey_traditional_approaches}: 
\begin{itemize}
    \item perform the so-called geometric correction, i.e., define a perspective normalization map of the scene;
    \item extract holistic features from the image, like edges or textures;
    \item train a regressor using the perspective normalized features.
\end{itemize}

\subsubsection{Counting by Density Estimation}
Counting by density estimation is a supervised technique, introduced in the seminal work \cite{learning_to_count}, that extends in some way the counting by regression approach previously described. Here, the (linear or not) mapping is between image features and the corresponding density map, i.e., a continuous-valued function. Then, it is possible to calculate the integral over any region in this density map obtaining the estimated number of objects within it.

In other words, given an input image, the high-level idea is to compute a density function $F$ as a real function of the pixels in this image. The notion of density function loosely corresponds to density's physical/mathematical idea. Then, given $F$ and a query about the number of objects in the entire image $I$ or in an image sub-region $S \subseteq I$, the number of objects is estimated by integrating (i.e., summing up the pixel values) $F$ over the region of interest.

This approach is robust to occlusions and perspective distortions because it does not rely on a previously trained classifier, just like the counting by regression approach. However, the crucial difference with the latter is that now we are exploiting a pixel-level mapping, where each pixel of the image is represented by a feature vector and mapped to a pixel of the corresponding density map. Therefore, unlike in the counting by regression technique, we incorporate spatial information in the learning process. Considering spatial information in the learning process also affects the number of training images needed and the ground truth labels generation. Indeed, in the counting by density estimation approach, we need, in general, a smaller number of training images than counting by regression technique, but, as a drawback, it is more complicated to generate the labels. Indeed, in the counting by regression approach, a training label corresponds to a raw number corresponding to the total number of objects present in the image. In contrast, the counting by density estimation technique also needs an (at least coarse) localization of the entities we want to count.
Furthermore, it is worth noting that building the ground truth density map is not a trivial problem. Due to perspective distortions, the objects are of different sizes, and labels should reflect this behavior. To summarize, in the regression-based approach, we need a large but easy-to-build training set, while in the density by estimation method, we need a training set that is more compact but more challenging to build.

In the context of density-based approaches, the most widely used labels needed for the supervised training are the dotted annotations, obtained by putting a single dot on each object instance in each image. Formally, we assume to have a set of $N$ training images ${I_1, I_2, \dots, I_N}$. We also assume that each image $I_i$ is labeled with a set of 2D points $\textbf{P}_i = {P_1, \dots, P_{K}}$, where $K$ is the total number of annotated objects. For a training image $I_i$, we define the ground truth density map as:
\begin{equation}
    \forall p \in I_i, \quad H_{i}(p) = \sum_{P \in \textbf{P}_i}\delta(p - P).
\end{equation}
Here, $p$ denotes a pixel, while a point identifying an object is represented as a delta function. Converting it into a continuous density function with Gaussian kernel $G_\sigma$ we obtain:
\begin{equation}
    \forall p \in I_i, \quad F_{i}(p) = \sum_{P \in \textbf{P}_i}\delta(p - P)*G_\sigma.
\end{equation}
The sum of the density map is equivalent to the total number of objects. It is worth noting that the Gaussian spread parameter $\sigma$ depends on the size of each object in the image, considering the perspective transformation. However, it is almost impossible to manually obtain the size of the occluded objects in a high-density environment. So this parameter is a dataset-specific quantity empirically estimated. Then, given a set of training images together with their ground truth densities, we aim to learn a transformation of the feature representation of the image that approximates the density function at each pixel so that it minimizes the sum of the mismatches between the ground truth and the estimated density functions.

\subsection{Evaluation Metrics}
\label{sec:back:visual-counting:metrics}
This section describes the metrics commonly exploited in the literature aiming to measure the performance of the counting solutions.

The most widely used counting metric is the \textit{\acrfull{mae}}, defined as:

\begin{equation}
\label{mae_def}
MAE = \frac{1}{N} \sum_{n=1}^{N} |c_{gt}^{n} - c_{pred}^{n}|,
\end{equation}

\noindent where $N$ is the total number of test images, $c_{gt}^{n}$ is the actual count (i.e., the ground truth), and $c_{pred}^{n}$ is the predicted count of the n-th image. 

Another frequently used metric is the \textit{\acrfull{mse}}, defined as follows:

\begin{equation}
\label{mse_def}
MSE = \frac{1}{N} \sum_{n=1}^{N} (c_{gt}^{n} - c_{pred}^{n})^{2},
\end{equation}

\noindent where, again, $N$ represents the total number of images, $c_{gt}^{n}$ is the ground truth, and $c_{pred}^{n}$ is the predicted count of the n-th image. It is worth noting that, as a result of the squaring of each difference, the \acrshort{mse} effectively penalizes large errors more heavily than small ones, and so it is more useful when large errors are particularly undesirable. Sometimes, the same error is expressed as the \textit{\acrfull{rmse}}, that is defined as the square root of the \acrshort{mse}:

\begin{equation}
\label{rmse_def}
RMSE = \sqrt{\frac{1}{N} \sum_{n=1}^{N} (c_{gt}^{n} - c_{pred}^{n})^{2}}.
\end{equation}

The above metrics are indicative of quantifying the mean error in the counting estimation for a set of images. However, as pointed out by \cite{survey_traditional_approaches}, these metrics do not contain information about the relation of the error and the total number of objects present in the \textit{single} images. A counting error of a certain magnitude perpetrated in a scenario containing thousands of objects can be considered not as bad compared to the same error in a scenario having sparse objects. To this end, another performance metric is taken into account, which is essentially a normalized \acrshort{mae}, named \textit{\acrfull{mare}}, defined as:

\begin{equation}
\label{mre_def}
MARE = \frac{1}{N} \sum_{n=1}^{N} \frac{|c_{gt}^{n} - c_{pred}^{n}|}{c_{gt}^{n}},
\end{equation}

\noindent where the notations are the same used before.

\begin{figure}[htbp]
\centerline{\includegraphics[width=.9\textwidth]{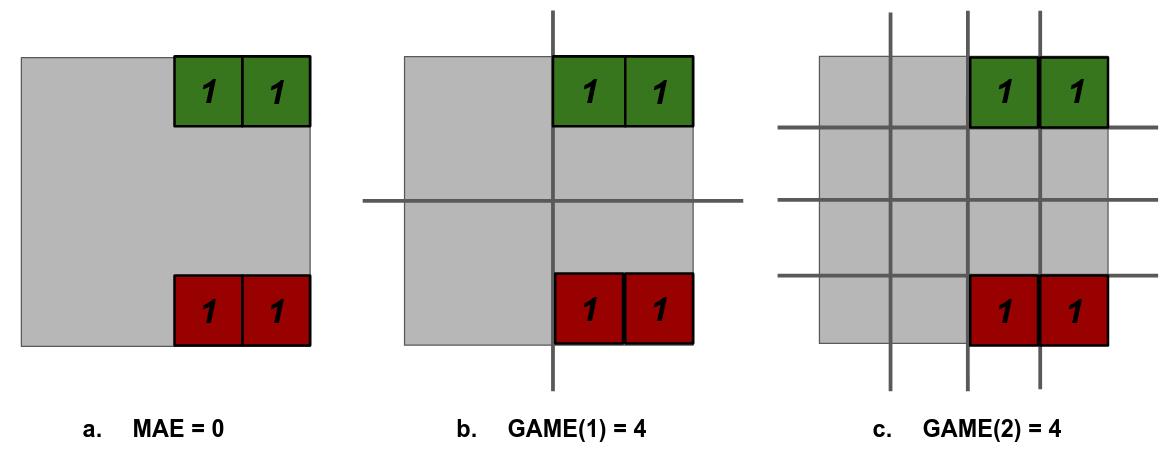}}
\caption{\textbf{An explanatory example for the \acrshort{game} and \acrshort{mae} metrics.} The 'ones' represent the count of the objects and their location in the image. We highlighted in green the estimation and in red the ground truth. The \acrshort{mae} in (a) is of 0, even though the objects have not been correctly located. In (b) and (c) we show that instead the \acrshort{game} is able to penalize the count when the localization is wrong.}
\label{fig:game}
\end{figure}

While these metrics seem fair for establishing a comparative, it is worth noting that they often lead to mask mistaken estimations. The reason is that the \acrshort{mae}, the \acrshort{mse} and the \acrshort{mare} do not take into account \textit{where} the estimations have been done in the images. Regarding the detection-based counting approaches, it is possible to exploit the standard metrics commonly used in the detection task that relies on the predicted bounding boxes localizing the objects. We describe more in detail these metrics in \ref{sec:back:cnn-based-detectors}. However, the problem is more remarkable in the density-based approaches, which provide only a weak localization of the objects.
In order to provide a more accurate evaluation, the authors of \cite{ExtremelyTrancos} introduced the \textit{\acrfull{game}} metric, suitable for the regression by density estimation counting solutions. Their goal is straightforward: to offer an evaluation metric that simultaneously considers the object count and the estimated locations for the objects. In particular, with the \acrshort{game} metric, they suggested to sub-divide the image in $4^{L}$ non-overlapping regions and computing the \acrshort{mae} in each of these sub-regions. The \acrshort{game} can be formulated as:

\begin{equation}
\label{game_def}
GAME(L) = \frac{1}{N} \sum_{n=1}^{N} (\sum_{l=1}^{4^{L}} |c_{gt}^{l} - c_{pred}^{l}|),
\end{equation}

\noindent where $N$ is the total number of test images, $c_{pred}^{l}$ is the estimated count in a region $l$ of the n-th image, and $c_{gt}^{l}$ is the ground truth for the same region in the same image. The higher $L$, the more restrictive the \acrshort{game} metric will be. Note that the \acrshort{mae} can be obtained as a particular case of the \acrshort{game} when $L=0$. \ref{fig:game} shows an example for the \acrshort{game} and the \acrshort{mae} metrics, highlighting that the \acrshort{game} metric is able to penalize those predictions having a good \acrshort{mae} but a wrong localization of the objects.

Another metric widely employed in the counting by density estimation task is the \textit{\acrfull{ssim}}. Even in this case, like for the \acrshort{game} metric, the goal is to assess the \textit{quality} of the counting results, taking into account also possible mistaken estimations. In particular, here, the goal is to assess the quality of the estimated density maps. \acrshort{ssim}, introduced in \cite{ssim}, is used in general as a metric to measure the \textit{similarity} between two images. In the case of the counting by density estimation, the images are represented by the predicted and the ground truth density maps, respectively. A common solution to assess the similarity between two images is to quantify the difference in the values of each of the corresponding pixels between the sample and the reference images by using, for example, Mean Squared Error. However, unlike most of the other methods that attempt to quantify the visibility of errors (differences) between the two images, inspired by the fact that human visual perception is highly adapted for extracting structural information from a scene, the authors of \cite{ssim} proposed to introduce an alternative, complementary framework for quality assessment based on the degradation of structural information. In particular, the \acrshort{ssim} extracts three key features from an image: i) luminance, ii) contrast, and iii) structure, and the comparison between the two images is performed exploiting and combining these three indices. The \acrshort{ssim} value can be a value between -1 and +1. A value of +1 indicates that the two given images are very similar or the same, while a value of -1 indicates the two given images are very different. Often these values are adjusted to be in the range $[0, 1]$, where the extremes hold the same meaning. It is worth noting that the authors suggested applying the above metrics regionally (i.e., in small sections of the image and taking the mean overall) instead of using them globally (i.e., all over the image at once). The reasons are that the statistical image features are usually highly spatially non-stationary and also because the image distortions, which may or may not depend on the local image statistics, may also be space-variant. The authors used an $11\times11$ circular-symmetric Gaussian Weighing function (basically, an $11\times11$ matrix whose values are derived from a Gaussian distribution) which moves pixel-by-pixel over the entire image. At each step, the local statistics and the \acrshort{ssim} index are calculated within the local window. Once computations are performed all over the image, the global \acrshort{ssim} is computed taking the mean of all the local \acrshort{ssim} values.

\subsection{CNN-based methods}
\label{sec:back:visual-counting:cnn-based}

The success of \acrshortpl{cnn} in numerous Computer Vision tasks has inspired researchers to exploit their abilities also for the counting task. In particular, \acrshortpl{cnn}-based detectors have been used to improve the performance of the counting by detection approaches. We review the main detectors based on \acrlongpl{cnn} in \ref{sec:back:cnn-based-detectors}. Moreover, \acrshortpl{cnn} have also been heavily exploited for learning non-linear functions from crowd images to their corresponding density maps or corresponding counts, and a multitude of methods have been proposed in the literature. Considering that reviewing all state-of-the-art methods is impractical, in this section, we sort out and describe some mainstream algorithms that were found to be more relevant and influential. In particular, we focus on the modern \acrshort{cnn}-based density estimation methods published in recent years, including also some early works based on regression for the sake of completeness.  

We divide the \acrshort{cnn}-based crowd counting models into three main categories because of the different types of network architectures they rely on:
\begin{itemize}
    \item \textit{Basic \acrshort{cnn}}: this network architecture adopts the standard \acrshort{cnn} layers, i.e., convolutional layers, pooling layers and fully connected layers. 
    These architectures are simple and easy to implement yet usually perform low accuracy.
    \item \textit{Multi-column}: this network architecture tries to ensure robustness to the large variation of the object scales present in the images. Usually, it adopts different branches (or columns) corresponding to different receptive fields aiming at capturing multi-scale information.
    \item \textit{Single-column}: the single-column network architectures usually deploy single and deeper \acrshort{cnn}s compared to the structure of multi-column network architectures. The premise is not to increase the complexity of the network.
\end{itemize}

\begin{figure}[htbp]
\centerline{\includegraphics[width=.9\textwidth]{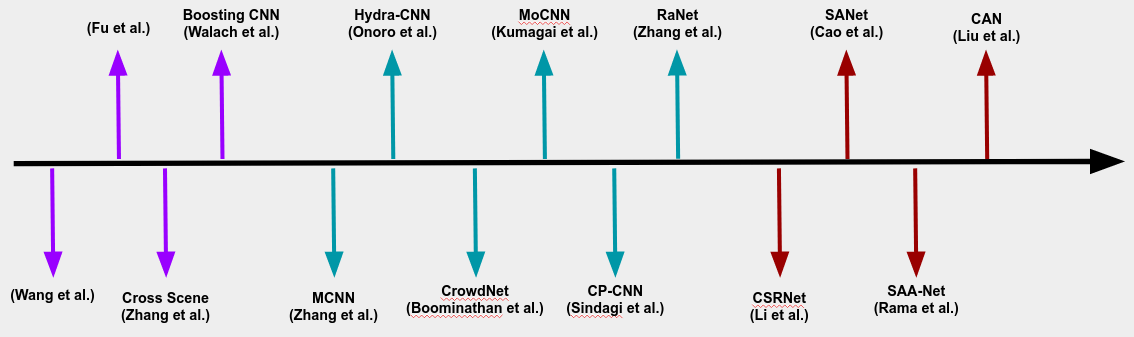}}
\caption{\textbf{A brief chronology of counting by regression techniques.} Milestone models mentioned in this figure: Wang et al. \cite{crowd_counting_1}, Fu et al. \cite{fast_crowd}, Cross-Scene \cite{cross_scene}, Boosting CNN \cite{count_boosting}, MCNN \cite{multi_column}, Hydra-CNN \cite{towards_perspective}, CrowdNet \cite{crowdnet}, MoCNN \cite{mixture_cnns}, CP-CNN \cite{cp_cnn}, RaNet \cite{ranet}, CSRNet \cite{csrnet}, SANet \cite{sanet}, SAA-Net \cite{DBLP:journals/corr/abs-1901-06026}, CAN \cite{DBLP:conf/cvpr/LiuSF19}. The trend in the past few years has been designing counting by regression models based on basic CNN (in purple), multi-column CNN (in blue) and single-column CNN (in red) architecture.}
\label{fig:trend_counting}
\end{figure}

A brief chronology is shown in \ref{fig:trend_counting}, which illustrates the main advancements and milestones of the \acrshortpl{cnn}-based counting by regression techniques, in particular the ones based on the density estimation.

\subsubsection{Basic \acrshort{cnn}-based Architectures}
Fu et al. \cite{fast_crowd}, and Wang et al. \cite{crowd_counting_1} have been among the first ones to exploit \acrshort{cnn}s tackling the counting task with the density estimation and regression techniques. In particular, the authors of \cite{fast_crowd} proposed a \acrshort{cnn} for crowd density estimation; they significantly improved the estimation speed by removing some network connections according to the observation of the existence of similar feature maps.
On the other hand, the authors in \cite{crowd_counting_1} proposed an end-to-end deep \acrshort{cnn} model able to estimate the number of people present in images of highly dense crowds, regressing from the image features to the count. They exploited the AlexNet network \cite{alexnet}, where the final fully-connected layers of 4096 neurons, initially responsible for classifying objects, have been replaced by a single neuron layer in charge of predicting the count. Besides, training data is augmented with additional negative samples whose ground truth count is set to zero to reduce false predictions like buildings and trees. 

Zhang et al. \cite{cross_scene} figured out that earlier methods drastically reduced their performance when applied to a new scene different from the ones present in the training set. So, they proposed to learn a mapping from the images to the counts and to adapt this mapping to new target scenes, performing cross-scene counting. To achieve this, they proposed a data-driven method where the network is fine-tuned using training samples that are considered similar to the target scene without using any extra-label information. Additionally, they introduced a new dataset, named WorldExpo’10, to evaluate cross-scene crowd counting. Inspired by this latter work, Walach and Wolf \cite{count_boosting} introduced a novel strategy based on layered boosting and selective sampling. Layered boosting involves iteratively adding \acrshort{cnn} layers to the model such that every new layer is trained to estimate the residual error of the earlier prediction. For instance, after the first \acrshort{cnn} layer is trained, the second \acrshort{cnn} layer is trained on the difference between the estimation and the ground truth. This layered boosting approach is based on the notion of Gradient Boosting Machines (GBM) \cite{GBM}. The other contribution made by the authors is the use of a sample selection algorithm to improve the training process by reducing the effect of low-quality samples such as trivial samples or outliers. According to the authors, the samples that are correctly classified early on are trivial samples. Presenting such samples for training even after the networks have learned to classify them introduces bias in the network for such samples, thereby affecting its generalization performance. Another source of training inefficiency is the presence of outliers such as mislabeled samples. Apart from affecting the network’s performance, these samples increase the training time. To overcome this issue, such samples are removed from the training process for some epochs.

\subsubsection{Multi-column \acrshort{cnn}-based Architectures}
\begin{figure}[htbp]
\centerline{\includegraphics[width=.85\textwidth]{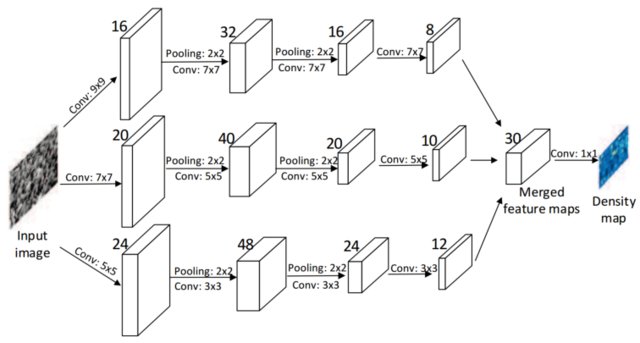}}
\caption{\textbf{Overview of the MCNN (Multi-Column Neural Network (image taken by Zhang et al. \cite{multi_column}).} This architecture is the first one that follows the multi-column paradigm in the counting task.}
\label{fig:overview_mcnn}
\end{figure}

Zhang et al. \cite{multi_column} have been the first ones to propose a multi-column-based architecture (\textit{Multi-Column Neural Network - MCNN}) for images having arbitrary crowd density and perspective. Inspired by the success of the multi-column networks for image recognition \cite{DBLP:conf/cvpr/CiresanMS12}, the proposed method claims to ensure robustness to the large variation of the object scales present in the different scenarios. The proposed architecture consists of three different branches (or columns) corresponding to filters with receptive fields of different sizes (large, medium, small). In the end, the features maps coming from the three branches are merged, and a final $1\times1$ convolutional layer is responsible for predicting the final density map. We report the overview of the architecture in \ref{fig:overview_mcnn}. Additionally, the authors introduced a new method for generating the ground truth crowd density maps. In contrast to existing methods that either use sum of Gaussian kernels with a fixed variance or perspective maps, they proposed to take into account the perspective distortion by estimating the spread parameters of the Gaussian kernels based on the size of the head of each person within the image. However, since it is impractical to estimate head sizes and their underlying relationship with density maps, they exploited an important property observed in high-density crowd images: the head size is related to the distance between the centers of two neighboring persons. In other words, the spread parameter for each person is data-adaptively determined based on its average distance to its neighbors. Note that the ground truth density maps created using this technique incorporate distortion information without using perspective maps. Finally, the authors introduced a new challenging dataset for the crowd counting task, the \textit{ShanghaiTech} dataset, described in \ref{sec:back:datasets}.

\begin{figure}[htbp]
\centerline{\includegraphics[width=.9\textwidth]{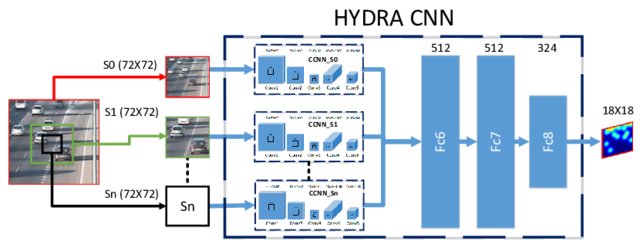}}
\caption{\textbf{Overview of the Hydra CNN architecture (image taken by Onoro and Sastre \cite{towards_perspective}).} Different heads extract image descriptors at different scales, while the body fuses them.}
\label{fig:overview_hydra}
\end{figure}

Similar to the above approach, Onoro and Sastre \cite{towards_perspective} developed a scale aware counting model called \textit{Hydra \acrshort{cnn}} that can estimate object densities in a variety of crowded scenarios without any explicit geometric information of the scene. Hydra-CNN consists of 3 heads and a body, with each head learning features for a particular scale. Each head of the Hydra-CNN is constructed using a deep, fully convolutional neural network. The outputs of the heads are then concatenated and fed to the body, which consists of two fully connected layers followed by a rectified linear unit (ReLu), a dropout layer, and a final fully connected layer to estimate the object density map. While the different heads extract image descriptors at different scales, the body learns a high-dimensional representation that fuses the multi-scale information provided by the heads. This network design of Hydra CNN is inspired by the work of Li et al. \cite{visual_saliency}. Finally, the network is trained with a pyramid of image patches extracted at multiple scales. The authors demonstrated through their experiments that the Hydra CNN performs successfully in scenarios and datasets with significant variations in the scene. The overview of the architecture is shown in \ref{fig:overview_hydra}.

\begin{figure}[htbp]
\centerline{\includegraphics[width=.9\textwidth]{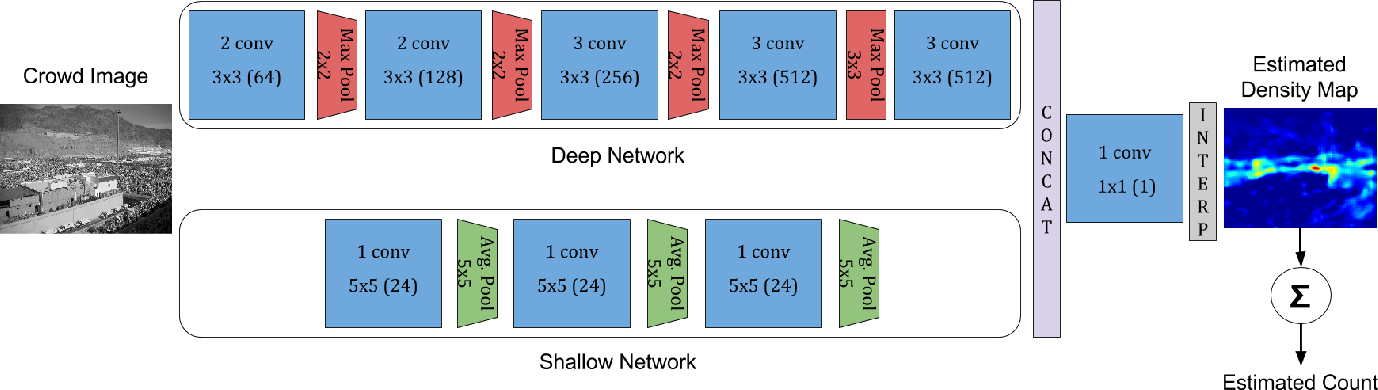}}
\caption{\textbf{Overview of the CrowdNet architecture (image taken by Boominathan et al. \cite{crowdnet}).} This architecture is a combination of deep and shallow fully convolutional networks to capture high-level and low-level semantic information, respectively.}
\label{fig:overview_crowdnet}
\end{figure}

In an effort to capture semantic information in the image, Boominathan et al. \cite{crowdnet} introduced \textit{CrowdNet}, a deep learning-based framework for estimating crowd density from static images of highly dense crowds. The authors exploited a combination of deep and shallow fully convolutional networks to predict the density map for a given crowd image. The deep network captures the high-level semantic information for objects near the camera captured in a significant level of detail (e.g., faces/body detectors). In contrast, the shallow one captures basic low-level patterns for objects away from the camera or when images are captured from an aerial viewpoint (e.g., representing as a head blob a person). Like in the MCNN architecture, in the end, the features maps produced by the two branches are merged, and a final $1\times1$ convolutional layer is responsible for predicting the final density map. The overview of the architecture is shown in \ref{fig:overview_crowdnet}.

\begin{figure}[htbp]
\centerline{\includegraphics[width=.8\textwidth]{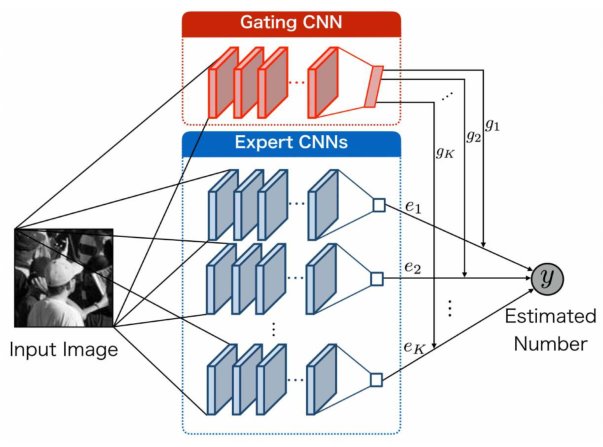}}
\caption{\textbf{Overview of the MoC (Mixture of CNN) architecture (image taken by Kumagai et al. \cite{mixture_cnns}).} A Gating CNN adaptively selects the appropriate CNN among the Experts, according to the appearance of the input image, predicting appropriate probabilities for each of them. These probabilities are then used as weighting factors to compute the weighted average of the counts predicted by all the experts.}
\label{fig:overview_mixture_cnn}
\end{figure}

In \cite{switching_cnn}, Sam et al. argue that can reach better performance by training the regressor of a multi-column network exploiting only a particular set of the available training patches by leveraging the variation of crowd density within an image. To this end, they proposed a switching \acrshort{cnn}-based architecture that selects the optimal regressor suited for a particular input patch. The authors described the network training phase as composed of multiple stages. First, the independent regressors are pre-trained on image patches to minimize the Euclidean distance between the estimated density map and ground truth. Then, the switch classifier, based on VGG-16 architecture \cite{vgg_16}, is trained to select the optimal regressor for accurate counting. Finally, the switch classifier and the \acrshort{cnn}-based regressors are co-adapted in the coupled training stage. 
More, Kumagai et al. \cite{mixture_cnns} proposed a \textit{Mixture of \acrshortpl{cnn} (MoCNN)}, consisting of a mixture of expert \acrshortpl{cnn} and a gating \acrshortpl{cnn} that adaptively selects the appropriate \acrshort{cnn} among the experts, according to the appearance of the input image. In particular, the gating \acrshort{cnn} predicts appropriate probabilities for each of the experts; these probabilities are further used as weighting factors to compute the weighted average of the counts predicted by all the experts. The overview of the architecture is shown in \ref{fig:overview_mixture_cnn}.

Sindagi et al. presented in \cite{cp_cnn} another method for crowd density and count estimation called \textit{Contextual Pyramid CNN (CP-CNN)}. The authors explicitly incorporate global and local contextual information of crowd images. In particular, the proposed architecture consists of four modules: i) a \acrshort{cnn}-based Global Context Estimator (GCE) to incorporate global context, trained to encode the context of an input image and to classify them into different density classes; ii) a Density Map Estimator (DME), which is a multi-column architecture-based \acrshort{cnn} with appropriate max-pooling layers, used to transform the image into high-dimensional feature maps; iii) a Local Context Estimator \acrshort{cnn} (LCE) trained on input image patches to encode local context information, aiming at guiding the DME to estimate better quality maps; iv) a Fusion-CNN (F-CNN) that fuses the contextual information obtained by LCE and GCE with the output of DME. 

In another work  \cite{drsan}, a unified neural network framework has been proposed, named \textit{Deep Recurrent Spatial-Aware Network (DRSAN)}. It incorporates a Recurrent Spatial-Aware Refinement (RSAR) module iteratively conducting two components: i) a Spatial Transformer Network \cite{spatial_transformer_nets} that dynamically locates an attentional region from the crowd density map and transforms it to the suitable scale and rotation for optimal crowd estimation; ii) a Local Refinement Network that refines the density map of the attended region with residual learning. 

Hossain et al. have been instead the first ones to exploit attention models. In \cite{saan}, they introduced a multi-column \acrshort{cnn}-based architecture with scale-aware attentions for crowd density estimation and counting. The attention in their model plays a similar role to the "switch" (i.e., density classifier) in \cite{switching_cnn}. However, in \cite{switching_cnn},  the switch makes a hard decision by selecting a particular scale based on the output of the density classifier and only uses the features corresponding to that scale for the final prediction. The problem is that if the density classifier is not completely accurate, it might select the wrong scale and, in the end, lead to incorrect density estimation. In contrast, the attention model proposed in \cite{saan} acts as a "soft switch." Instead of selecting a particular scale, the authors re-weight the features of a particular scale based on the attention score corresponding to that scale. 
On the other hand, the authors of \cite{ranet} proposed \textit{RANet (Relational Attention Network)}, characterized by a self-attention mechanism for capturing interdependence of pixels. RANet improves the self-attention mechanism by introducing two attention modules, i.e., local self-attention (LSA) and global self-attention (GSA). To be more specific, LSA applies the self-attention to the original feature, only operating on spatially local neighbors which are closely correlated to the central position. Moreover, correlated pixels from long distances should also be taken into account for capturing inter-dependencies, while, to be efficient, GSA selects distinctive pixels from the whole map by max pooling. 

Even though these multi-column networks have achieved significant progress, they still suffer from several significant disadvantages, as demonstrated by Li et al. \cite{csrnet}. First, it isn't easy to train the multi-column networks due to their complex structures. Next, using different branches but almost the same network structures inevitably leads to redundancy in information and waste in the consumption of parameters for density-level classifiers rather than the generation of final density maps. As with all the disadvantages, multi-column network architectures may be ineffective in a narrow sense. Thus it motivates many researchers to exploit simpler yet effective and efficient networks. Therefore, single-column network architectures are come out to cater to the demands of more challenging situations in crowd counting.

\subsubsection{Single-column \acrshort{cnn}-based Architectures}
The authors of CSRNet \cite{csrnet}, as mentioned above, have been the first ones to demonstrate that multi-column networks suffer from many significant drawbacks. But furthermore, they also introduced a novel single-column \acrshort{cnn}-based approach that can perform accurate density estimation and that represents a new milestone for the counting task. In particular, they presented \textit{CSRNet - Congested Scene Recognition Network}, which can understand highly congested scenes and perform accurate count estimation as well as produce high-quality density maps. In particular, the network is composed of two major components. Following other previous work, like \cite{crowdnet}, the authors use a modified version of the well-known VGG-16 network \cite{vgg_16} as the front-end for 2D feature extraction because of its strong transfer learning ability, carving the first 13 layers and adding a $1\times1$ convolutional layer as the output layer. The output size of this front-end network is $1/8$ of the original input size. The authors argue that continuing to stack more convolutional layers and pooling layers (i.e., the basic components in the VGG-16 network) would have led to an output size further shrunken, with the consequence of low-quality density maps. On the other hand, inspired by \cite{dilated_1, dilated_2, deeplab}, they proposed a back-end consisting of dilated kernels. The basic concept of dilated convolutions is to deliver larger reception fields replacing the pooling operations, extracting deeper information of saliency, and maintaining the output resolution. Formally, a 2-D dilated convolution can be defined as follow:

\begin{equation}
    y(m, n) = \sum_{i=1}^{M} \sum_{j=1}^{N} x(m + r \times i,n + r \times j) w(i,j),
\end{equation}

where $y(m,n)$ is the output of dilated convolution associate to the input $x(m,n)$ and a filter $w(i,j)$ with the length and the width of $M$ and $N$, respectively. The parameter $r$ is the dilation rate. If $r = 1$, a dilated convolution turns into a normal convolution. 

Cao et al. proposed in \cite{sanet} a novel encoder-decoder network, named \textit{Scale Aggregation Network (SANet)}. Inspired by the success of the Inception architecture \cite{inception} in the image recognition domain, the authors decided to employ scale aggregation modules in the encoder to improve the representation ability and scale diversity of the features. The decoder is instead composed of a set of convolutions and transposed convolutions. The latter is exploited to generate high-resolution and high-quality density maps, of which the sizes are the same as input images. Inspired by \cite{DBLP:journals/corr/ZhaoGFK15}, they employed a combination of Euclidean loss and local pattern consistency loss to exploit the local correlation in density maps. In particular, the local pattern consistency loss is computed by the SSIM \cite{ssim} index to measure the structural similarity between the estimated density map and the corresponding ground truth. Finally, they used Instance Normalization (IN) \cite{DBLP:conf/iccv/HuangB17} layers to alleviate the vanishing gradient problem and a simple but effective patch-based training and testing scheme to diminish the impact of statistical shifts between training and test images.

Zhang et al. proposed in \cite{DBLP:conf/wacv/ZhangSC18} the \textit{Scale-adaptive CNN (SaCNN)} architecture, another deep single column \acrshort{cnn} for density estimation. In this case, the spatial resolution has been preserved by small fixed-sized filters. Furthermore, the authors introduced a multi-task loss by adding a relative head-count loss to the density map loss that significantly improves the network generalization in crowd scenes with few pedestrians, where most representative works perform poorly. More, similarly to \cite{csrnet}, Chen et al. in \cite{DBLP:conf/wacv/ChenBSG19} proposed the \textit{Scale Pyramid Network (SPN)} which adopts a shared single deep column structure and extracts multi-scale information in high layers by Scale Pyramid Module. In particular, in the Scale Pyramid Module, they specifically employed different rates of dilated convolutions in parallel instead of traditional convolutions with different sizes.

Rama et al. in \cite{DBLP:journals/corr/abs-1901-06026} introduced \textit{SAA-Net - Scale-Aware Attention Network}, proposing a novel multi-branch scale-aware attention network that exploits the hierarchical structure of convolutional neural networks. This hierarchical structure progressively expands the receptive field of the network feature maps, implicitly capturing information at different scales. Inspired by the skip branches in FCN \cite{fcn} and SSD \cite{ssd}, they proposed to generate multiple density maps from these intermediate feature maps directly. As the feature map generated by the last convolutional layer has the largest receptive field, it carries high-level semantic information that can be used to localize large heads; on the other hand, feature maps generated by the intermediate layers are more accurate and robust in counting extremely small heads (i.e., the crowds), and they contain important details about the spatial layout of the people and low-level texture patterns. To aggregate these maps into the final prediction, they proposed a novel soft attention mechanism that learns a set of gating masks, one for each map. These masks learn to attend to large heads from the density map predicted by the last convolutional layer and smaller ones from earlier layers. Finally, they proposed a new scale-aware loss function to regularize the multi-scale estimates further and guide them to specialize in specific head sizes.

Liu et al. \cite{DBLP:conf/cvpr/LiuSF19} argued that previous methods indiscriminately fused the information at all scales gathered by the \acrshort{cnn}-based models (combining either density maps extracted from image patches at different resolutions or feature maps obtained with convolutional filters of different sizes), ignoring the fact that the suitable scale varies smoothly over the image and should be handled adaptively. To this end, they introduced Context-Aware Network (CAN), where the features have been extracted at multiple scales by a deep model that also learns how to combine them adaptively. In other words, this novel deep network
architecture adaptively encodes multi-level contextual information into the features it produces, without explicitly requiring defining patches, but rather by learning how to weigh these features for each pixel. By leveraging multi-scale pooling operations, this framework can cover an arbitrarily large range of receptive fields.

Due to their architectural simplicity and training efficiency, single-column network architecture has received more and more attention in recent years.

\section{CNN-based Objects Detectors}
\label{sec:back:cnn-based-detectors}

Object detection is one of the most important and challenging branches of Computer Vision. It deals with detecting instances of semantic objects of a certain class (such as humans, buildings, or cars) in digital images and videos \cite{DBLP:journals/tcsv/DasiopoulouMKPS05}. This task has attracted increasing attention due to its wide range of applications and recent technological breakthroughs. At present, most of the state-of-the-art object detectors employ \acrlong{cnn}s as their backbones and detection networks to extract features from images, classification, and localization, respectively. Existing object detectors can usually be divided into two categories: \textit{two-stage} detectors and \textit{one-stage} detectors. Two-stage detectors have high localization and object recognition accuracy, whereas the one-stage detectors achieve higher inference speed. Here, we prefer to consider a slightly different taxonomy, also considering a third category comprising the so-called \textit{anchor-less} detectors. The main difference between the other detectors is that the anchor-less ones do not use \textit{anchors}, i.e., pre-defined boxes that the network exploits as priors to find objects in the images. This section briefly describes the main characteristics of these detector classes, reporting the most significant and influential methods.

\subsection{Metrics}
\label{sec:back:cnn-based-detectors:metrics}
Object detection metrics assess how well a model performs on an object detection task. They also enable us to compare multiple detection solutions objectively or compare them to a benchmark. In literature, many different metrics are proposed, leading to some confusion. Therefore, in this section, we explain the main object detection metrics.

Since the classification task only evaluates the probability of the class object appearing in the image, it is straightforward for a classifier to identify correct predictions from incorrect ones. However, the object detection task localizes the object with a bounding box associated with its corresponding confidence score. Therefore, the first metric that is needed is one that determines how many objects were detected correctly and how many false positives were generated. This metric is called \textit{\acrfull{iou}}. In particular, it measures the overlap between two boundaries. In the case of the object detection task, \acrshort{iou} is useful to measure how much the predicted bounding box overlaps with the ground truth bounding box. Is it is defined as:

\begin{equation}
    \label{iou}
    IoU = \frac{\text{Area of Intersection of two boxes}}{\text{Area of Union of two boxes}}.
\end{equation}

By computing the \acrshort{iou} score for each detection, we set a threshold for converting those real-valued scores into classifications, where \acrshort{iou} values above this threshold are considered positive predictions, and those below are considered to be false predictions. More precisely, the predictions are classified into \acrfull{tp}, \acrfull{fn}, and \acrfull{fp}. It is worth noting that \acrfull{tn} predictions are not considered since they describe the situation where empty boxes are correctly detected as "non-object." The model would identify thousands of empty boxes in this scenario, which adds little to no value to the algorithm. \acrshort{tp}s, \acrshort{fp}s and \acrshort{tp}s are exploit to compute Precision and Recall.

\textit{Precision} is the ratio of the number of \acrfull{tp} to the total number of positive predictions. On the other hand, \textit{Recall} is the ratio of the number of \acrshort{tp} to the total number of actual (relevant) objects. Formally:

\begin{equation}
    \label{precision_eq}
    Precision = \frac{\text{\acrshort{tp}}}{\text{\acrshort{tp}} + \text{\acrshort{fp}}},
\end{equation}

\begin{equation}
    \label{recall_eq}
    Recall = \frac{\text{\acrshort{tp}}}{\text{\acrshort{tp}} + \text{\acrshort{fn}}}.
\end{equation}

\noindent High Recall but low Precision implies that all ground-truth objects have been detected, but most detections are incorrect (i.e., there are many \acrshort{fp}s). Low Recall but high Precision implies that all predicted boxes are correct, but most ground truth objects have been missed (i.e., there are many \acrshort{fn}s). The ideal detector has high Precision and high Recall so that most ground truth objects have been detected correctly.

Sometimes, these two metrics are employed to compute the \textit{F1 score}, a weighted average of the precision and recall, ranging from 0 to 1, where 1 means highest accuracy, and defined as:

\begin{equation}
    \label{f1_eq}
    \text{\textit{F1 score}} = 2 \times \frac{\text{Precision} \times \text{Recall}}{\text{Precision} + \text{Recall}}.
\end{equation}

Instead, it is worth noting that \textit{Accuracy}, i.e., the percentage of correctly predicted examples out of all predictions, formally known as

\begin{equation}
    \label{accuracy_eq}
    Accuracy = \frac{\text{\acrshort{tp}} + \text{\acrshort{tn}}}{\text{\acrshort{tp}} + \text{\acrshort{fn}} + \text{\acrshort{tn}} + \text{\acrshort{fn}}},
\end{equation}

\noindent can be very misleading when dealing with imbalanced class data, where the number of instances is not the same for each class. Object detection datasets fall into this category since the class distribution is considerably non-uniform. This is another reason indicating that \acrshort{tn}s are not essential in the object detection task. 

Another aspect to consider when dealing with object detection that influences the Precision and the Recall metrics is the confidence score, i.e., the probability that a bounding box contains an object. Usually, the detector outputs the confidence score, which can be used to filter out the predictions. When choosing a high confidence threshold, the model becomes robust to positive examples (i.e., boxes containing an object). Hence there will be fewer positive predictions. As a result, false negatives increase, and false positives decrease, thus reducing the recall (the denominator increases in the Recall formula) and improving the Precision (the denominator decreases in the precision formula). Similarly, further lowering the threshold causes the Precision to decrease and Recall to increase. Therefore, the confidence threshold is a tunable parameter that can determine the performance of the model.

\begin{figure}[htbp]
\centerline{\includegraphics[width=.8\textwidth]{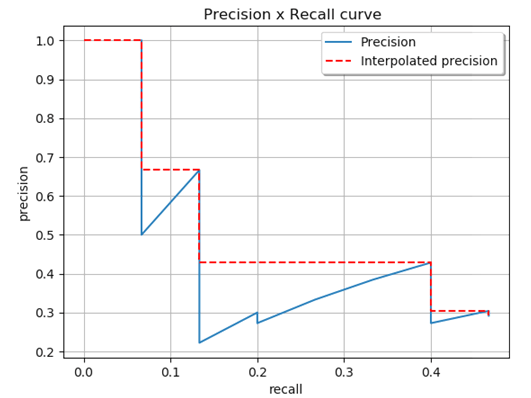}}
\caption{\textbf{An example of the Precision-Recall Curve (in blue).} The red dashed curve represents instead the smoothed PR curve, useful for the computation of the \acrshort{ap}.}
\label{fig:pr_curve_example}
\end{figure}

A metric summarizing both Recall and Precision and providing a model-wide evaluation is the \textit{Precision-Recall curve}. Since both metrics do not use true negatives, the Precision-Recall curve is suitable for assessing the performance of the model on imbalanced datasets. The Precision-Recall curve plots Recall on the x-axis and Precision on the y-axis, where each point in the curve represents Recall and Precision values for a particular confidence value. See the blue plot in \ref{fig:pr_curve_example} for an example of a Precision-Recall curve.

Recall values increase as we go down the prediction ranking. Consequently, the curve can be noisy and have a particular saw-tooth shape that makes it difficult to estimate the performance of the model and similarly tricky to compare different models with their Precision-Recall curves crossing each other. Therefore the idea is to assess the area under the curve using a numerical value called \textit{\acrfull{ap}}. More in detail, \acrshort{ap} is a single number metric that encapsulates both Precision and Recall and summarizes the Precision-Recall curve by averaging Precision across Recall values from 0 to 1. More in detail, there are some techniques aimed at smoothing out the saw-tooth pattern of the curve. One of the most popular solutions is to sample the curve at all unique recall values $r_i$, whenever the maximum precision value drops, and compute $p(r_i)$. \acrshort{ap} is then defined as the sum of the rectangular blocks.

The \textit{\acrfull{map}} is the average \acrshort{ap} over the $N$ class categories of the dataset. In addition, some of the more popular detection challenges, that represent the \textit{de facto} standard, consider also the \acrshort{iou}. In particular, the PASCAL VOC challenge uses the \acrshort{map} as a metric with an \acrshort{iou} threshold of 0.5, while MS COCO averages the \acrshort{map} over different \acrshort{iou} thresholds (0.5, 0.55, 0.6, 0.65, 0.7, 0.75, 0.8, 0.85, 0.9, 0.95) with a step of 0.05, denoting this metric by mAP@[.5,.95]. Therefore, COCO not only averages AP over all classes but also on the defined \acrshort{iou} thresholds.

\subsection{Two-stage Detectors}
\label{sec:back:cnn-based-detectors:two-stages-detectors}
Two-stage frameworks divide the detection pipeline into two steps: the region proposal and the classification stages. These architectures first propose several object candidates, known as regions of interest (RoI) or region proposals, operating in the so-called ’recognition using regions’ paradigm \cite{recognition_using_regions}. In the second stage, these proposals are classified, and their localization is refined. In the following, we describe the more influential two-stage detectors presented in the literature in the last few years.

\begin{figure}[htbp]
\centerline{\includegraphics[width=.9\textwidth]{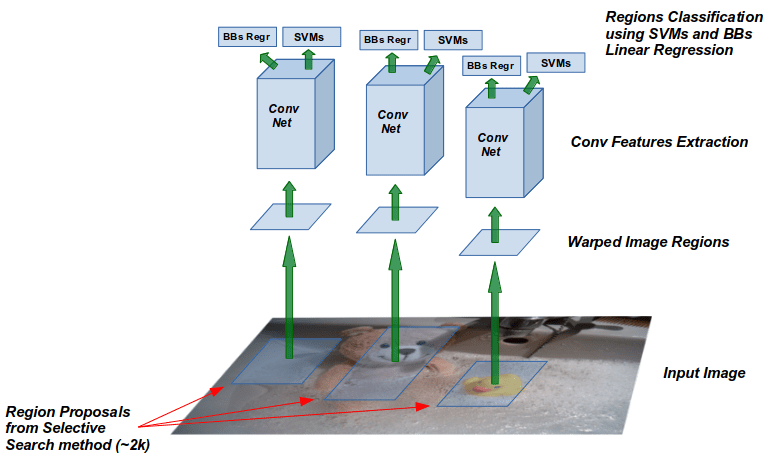}}
\caption{\textbf{Overview of the R-CNN architecture.} Region proposals are created using the selective-search algorithm \cite{selective_search} and then are passed through a \acrshort{cnn}-based backbone to perform the features extraction operation. The final layers are responsible of classify the produced feature vectors and to predict bounding boxes localizing the objects.}
\label{fig:rcnn_architecture}
\end{figure}

\paragraph{R-CNN.} \textit{Regions with \acrshort{cnn} features (R-CNN)} \cite{rcnn_det} is probably the first detector showing that \acrshortpl{cnn} can lead to dramatically higher object detection performance compared to traditional methods relying on hand-crafted features. Operating in the two-stage framework, R-CNN takes an image as input and outputs bounding boxes localizing the detected objects together with labels classifying them into some pre-defined object categories. The first stage aiming at producing a bunch of region proposals is carried out by the Selective Search algorithm \cite{selective_search}; at a high level, the selective-search algorithm looks at the image through windows of different sizes, trying to group adjacent pixels by texture, color, or intensity to identify objects. Then, these proposals are warped to a fixed square size and passed through a \acrshort{cnn}-based backbone, typically a modified version of AlexNet \cite{alexnet} or of VGG-16 \cite{vgg_16}, in charge of performing the features extraction operation. Thus, the corresponding outputs are fixed-length feature vectors describing each region proposal. On the final layer of the detection pipeline, R-CNN adds a set of class-specific linear \acrfull{svm} \cite{svm} trained using these fixed-length features and having the goal of classifying them into pre-defined object categories. Furthermore, R-CNN runs a simple linear regression (inspired by \cite{DBLP:journals/pami/FelzenszwalbGMR10}) on the same fixed-length feature vectors to generate the final result, i.e., bounding box coordinates to fit the dimensions of the objects. An overview of this detection pipeline is shown in \ref{fig:rcnn_architecture}.

\begin{figure}[htbp]
\centerline{\includegraphics[width=.9\textwidth]{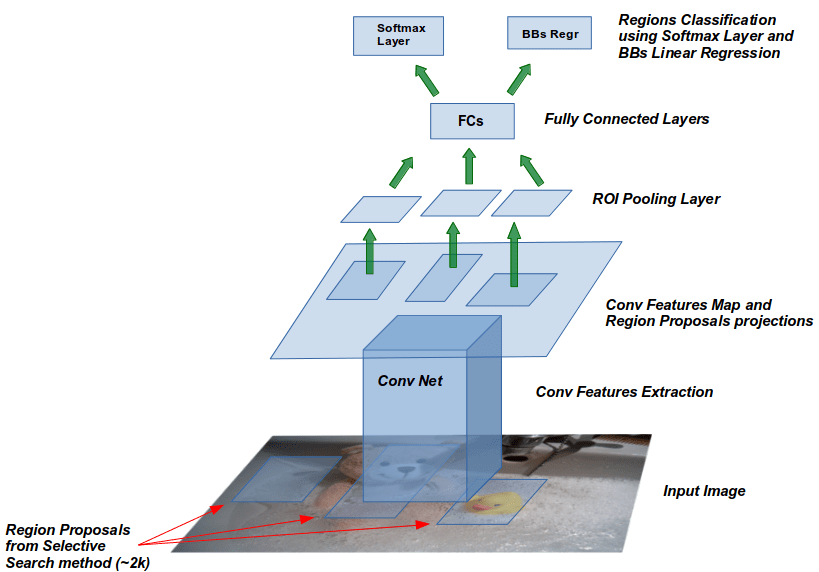}}
\caption{\textbf{Overview of the Fast R-CNN architecture.} The main difference compared to R-CNN is that Fast R-CNN exploits the backbone just once per image using a network layer known as RoIPool (Region of Interest Pooling) and sharing that computation across all the proposals.}
\label{fig:fastrcnn_architecture}
\end{figure}

\paragraph{Fast R-CNN.} One year after the release of R-CNN, the same authors proposed \textit{Fast R-CNN} \cite{fast_rcnn}, an evolution of their first object detection framework. Indeed, despite achieving high object detection quality, RCNN has notable drawbacks, the main of which concerns the training of the \acrshort{svm} classifiers and bounding box regression: this process is expensive in both disk space and time since \acrshort{cnn} features are extracted independently from each region proposal in each image, without sharing computation, and thus posing great challenges for large-scale detection. On the other hand, Fast R-CNN exploits the backbone just once per image using a network layer known as RoIPool (Region of Interest Pooling) and sharing that computation across all the proposals. In particular, the ROI pooling operation produces fixed-size feature maps from a list of non-uniform inputs defining and localizing the region proposals. From every region of interest from this input list, the RoIPoll layer takes a section of the input feature map that corresponds to it and scales it to some pre-defined size. The scaling is done by dividing the region proposal into equal-sized sections and finding the largest value in them. The result is that from a list of rectangles with different sizes, it gets a list of corresponding fixed-length feature vectors from the feature map. Furthermore, other improvements done in Fast R-CNN concerns the training procedure. In particular, the authors of Fast R-CNN proposed an end-to-end training procedure (when ignoring the process of region proposal generation) that can update all the network layers using a multi-task loss, unlike the multistage training pipeline exploited in R-CNN, which is slow and hard to optimize because each individual stage must be trained separately. To this end, each feature vector is fed into a sequence of fully connected layers that finally branch into two sibling output layers. One is a softmax layer that replaces the \acrshort{svm}s and is responsible for the classification task. The other is instead in charge of regressing the bounding box coordinates localizing the objects. We show in \ref{fig:fastrcnn_architecture} the architecture of this detection pipeline.

\paragraph{Faster R-CNN.}
Even with all the advancements observed in Fast R-CNN, there was still one remaining bottleneck in the process: the region proposer. As explained above, the first step of the detection pipeline is to generate a bunch of potential bounding boxes or Regions of Interest likely to contain objects. As we saw, in R-CNN and Fast R-CNN, these proposals were created using the Selective Search algorithm \cite{selective_search}, a fairly slow process that was found to be the bottleneck of the overall process. In 2015, a team of Microsoft research proposed in \cite{faster_rcnn} a further evolution of R-CNN and Fast R-CNN, named \textit{Faster R-CNN}, finding a way to make the region proposal step more efficient. In particular, they proposed to compute the region proposals exploiting the features of the images extracted by the \acrshort{cnn} that acts as the backbone in Fast R-CNN. In other words, they suggested using the extracted features of the images not only for the classification and bounding boxes regression steps of the detection pipeline but also for the region proposals computation instead of running the slow and costly selective search algorithm. More in detail, Faster R-CNN introduces a new network, the \acrfull{rpn}, that analyzes the features of the images, ranking region boxes (called \textit{anchors}), and proposes the ones most likely containing objects. Basically, the authors proposed a single network comprising \acrshort{rpn} for region proposal and Fast RCNN for region classification. \acrshort{rpn} and Fast RCNN share a large number of convolutional layers and, in particular, the features from the last shared convolutional layer are used for region proposal and region classification from separate branches, thus enabling highly efficient region proposal computation. \acrshort{rpn} is, in fact, a \acrlong{fcn}; Faster RCNN is thus a purely \acrshort{cnn} based framework without using handcrafted features.

\begin{figure}[htbp]
\centerline{\includegraphics[width=.8\textwidth, height=9cm]{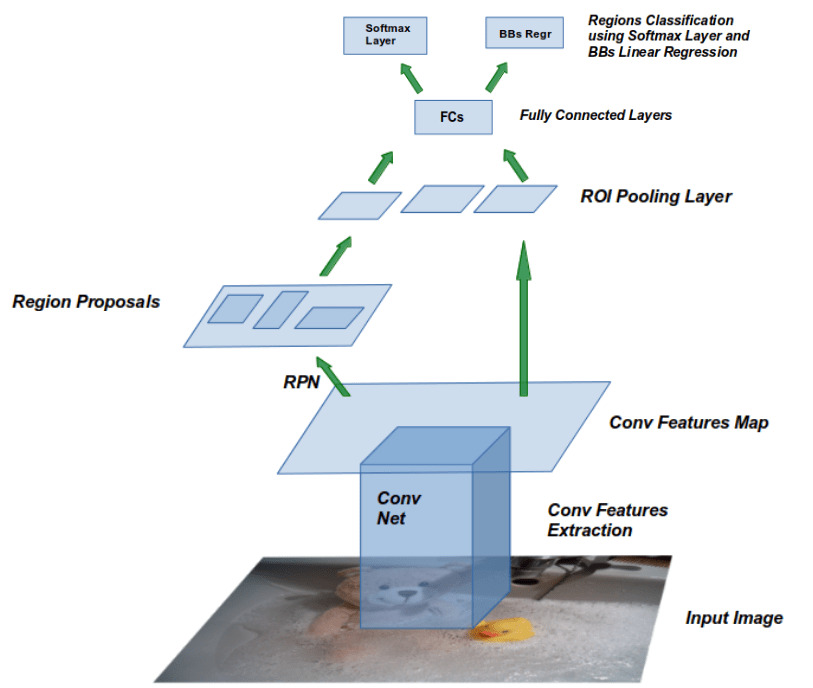}}
\caption{\textbf{Overview of the Faster R-CNN architecture.} The authors in \cite{faster_rcnn} find a way to make the region proposal step more efficient exploiting a new network, the \acrlong{rpn}, that analyzes the features of the images, ranking region boxes (called \textit{anchors}), and proposing the ones most likely containing objects.}
\label{fig:fasterrcnn_architecture}
\end{figure}

\textit{Anchors} play a crucial role in Faster R-CNN. Basically, an anchor is a box. Intuitively, we know that objects in an image should fit specific common aspect ratios and sizes. For instance, to detect people, rectangular boxes that resemble the shapes of humans are required. In such a way, Faster R-CNN proposed to use $k$ boxes having fixed sizes and aspect ratios, called indeed anchor boxes, that are slid over the image looking for objects. In the default configuration of Faster R-CNN, the authors suggested using nine anchors at a position of an image. 
The architecture of Faster R-CNN is shown in \ref{fig:fasterrcnn_architecture}. The standard choice of anchors proposed by the authors of \cite{faster_rcnn} work well for many datasets. However, one has the freedom to design different kinds of anchors/boxes. For example, if the network is designed to detect pedestrians, it is better to consider short, big, or square boxes. A neat set of anchors may increase the speed as well as the accuracy. 

\paragraph{Mask R-CNN.}
So far, we have analyzed networks able to use \acrshort{cnn} features to effectively locate and categorize different objects in an image with bounding boxes. \textit{Mask R-CNN} \cite{mask_rcnn} extends such techniques to go one step further and locate exact pixels of each object instead of just bounding boxes (i.e. performing \textit{instance segmentation}). Mask R-CNN does this by adding a branch to Faster R-CNN that outputs a binary mask or each RoI that determines whether or not a given pixel is part of an object. This branch, as in Faster R-CNN for the \acrshort{rpn}, is a \acrlong{fcn} on top of a \acrshort{cnn} feature map. To avoid the misalignments caused by the original RoIPool layer, an additional RoIAlign layer was proposed to preserve the pixel-level spatial correspondence.

\subsection{Single-stage Detectors}
\label{sec:back:cnn-based-detectors:single-stage-detectors}
In contrast to the two-stage detectors, one-stage detectors predict bounding boxes in a single step without using region proposals, and they rely on a fixed number of "predictions on a grid." In other words, there is no intermediate task (as we have already seen in the previous \ref{sec:back:cnn-based-detectors:two-stages-detectors} with region proposals) that must be performed to produce the output, leading to simpler and faster model architectures, although it can sometimes struggle to be flexible enough to adapt to arbitrary tasks (such as mask prediction). Furthermore, these detectors generally reach lower accuracy than the two-stages methods. 
In the following, we illustrate the more influential one-stage detectors introduced in the literature in the last few years. 

\paragraph{YOLO.} \cite{yolo_v1} \acrfull{yolo} is a popular one-stage object detector proposed by Redmon et al. in 2015, casting object detection as a regression problem from image pixels to spatially separated bounding boxes and associated class probabilities.
Since the region proposal generation stage is completely drained, \acrshort{yolo} directly predicts detections using a small set of candidate regions. In particular, \acrshort{yolo} divides an input image into a $S \times S$ regular grid. Each grid cell is responsible for predicting B objects encoded in 5 terms: x, y, w, h, and a confidence score. The (x, y) coordinates represent the center of the box enclosing the object relative to the bounds of the grid cell. The width and height are instead predicted relatively to the whole image. Finally, the confidence score tells us how certain the predicted bounding box encloses some object. This score does not say anything about what kind of object is in the box, just if the shape of the box is any good. Each grid cell also predicts C conditional class probabilities. These probabilities are conditioned on the grid cell containing an object. \acrshort{yolo} only predicts one set of class probabilities per grid cell, regardless of the number of boxes B, and builds a so-called class probability map. At test time, the confidence score for the bounding box and the class prediction are combined into one final score that tells us the probability that this bounding box contains a specific type of object. Since, in the end, \acrshort{yolo} predicts $S \times S \times B$ bounding boxes, additional filtering using \acrlong{nms} is needed.

It is worth noting that \acrshort{yolo} imposes strong spatial constraints on bounding boxes predictions since each grid cell only predicts B boxes (in the paper, B is set to two) and can only have one class. This spatial constraint has two effects: it limits the number of nearby objects that the model can predict and, as a result, \acrshort{yolo} struggles with small objects that appear in groups. Still, it also helps mitigate multiple detections of the same object. The authors of \acrshort{yolo} implement this model as a \acrshort{cnn}: the initial convolutional layers of the network extract features from the image. In contrast, the final fully-connected layers predict the output probabilities and coordinates. 

\paragraph{YOLOv2 and YOLO9000.} \acrshort{yolo}v2 \cite{yolo9000} is the second version of \acrshort{yolo}, which adopts a series of design decisions from past works with novel concepts to improve speed and precision. In particular: i) it adds a \acrshort{bn} layer ahead of each convolutional layer which accelerates the network to get convergence and helps regularize the model; ii) it removes the final fully-connected layers; iii) it adopts an anchor-based prediction mechanism where the anchors are learned with k-means; iv) for localizing smaller objects, it concatenates the higher resolution features with the low-resolution features by stacking adjacent features into different channels; v) it performs a multi-scale training; vi) it proposes a new classification backbone namely Darknet-19, with 19 convolutional layers and five max-pooling layers, which requires fewer operations to process an image yet achieves high accuracy. Furthermore, in the same work, the authors also introduced YOLO9000, which can detect over 9000 object categories in real-time by proposing a joint optimization method to train simultaneously on ImageNet \cite{imagenet_dataset} and COCO \cite{lin2014microsoft} with WordTree, combining data from multiple sources.

\paragraph{SSD.} \acrfull{ssd} \cite{ssd} is a single-shot detector for multiple categories within the one-stage paradigm that effectively combines ideas from \acrshortpl{rpn} in Faster R-CNN \cite{faster_rcnn}, \acrshortpl{yolo} \cite{yolo_v1} and multiscale convolutional features \cite{DBLP:journals/pami/HariharanAGM17}. Like \acrshort{yolo}, SSD predicts a fixed number of bounding boxes and objectness scores, followed by \acrshort{nms} to filter out the final detections. The backbone is based on standard classification \acrshortpl{cnn}, like VGG16 \cite{vgg_16}, truncated before any classification layers. Then, several auxiliary convolutional layers, progressively decreasing in size, are added on top of it. Since the information in the last layer with low resolution may be too coarse spatially to allow precise localization, SSD uses shallower layers with higher resolution for detecting small objects. More, for objects of different sizes, SSD performs detection over multiple scales by operating on multiple convolutional feature maps, each of which predicts category scores and box offsets for bounding boxes of appropriate sizes.

\paragraph{YOLOv3.} \acrshort{yolo}v3 \cite{yolo_v3} is an improved version of \acrshort{yolo}v2.  The most salient feature of YOLOv3 is that it makes detections at three different scales, which are precisely given by downsampling the dimensions of the input image by 32, 16, and 8, respectively. Detections at different scales help address the issue of detecting small objects, a frequent complaint with YOLOv2. More, YOLOv3 proposes a deeper and robust backbone for features extraction, called Darknet-53 (in contrast with the Darknet-19 characterizing YOLOv2).

\paragraph{RetinaNet.}
\cite{retinanet} RetinaNet is another one-stage detector that introduced a new loss function to be exploited during the training phase, named \textit{Focal loss}. The authors of this network demonstrated that one-stage detectors did not reach the accuracy of two-stage detectors because of the extreme class imbalance encountered during training. In particular, in two-stage detectors such as Faster R-CNN, the first stage (i.e., the \acrshort{rpn}) narrows down the number of candidate object locations to a small number (e.g., 1–2k), filtering out most background samples. Then, in the second classification stage, sampling heuristics, such as a fixed foreground-to-background ratio (1:3) or online hard example mining (OHEM) \cite{ohem} are performed to maintain a manageable balance between foreground and background. On the other hand, a one-stage detector must process a much larger set of candidate object locations regularly sampled across an image. In practice, this often amounts to enumerating about 100k locations that densely cover spatial positions, scales, and aspect ratios. This large number prevents the adoption of the above-mentioned sampling heuristics, and, consequently, the training procedure is dominated by easily classified background examples. To contrast this issue, the authors of \cite{retinanet} introduced the focal loss, which, intuitively, can automatically down-weight the contribution of easy examples during training and rapidly focus the model on hard examples. In particular, the focal loss is the reshaping of cross-entropy loss such that it down-weights the loss assigned to well-classified examples. In other words, this novel loss focuses training on a sparse set of hard examples and prevents the vast number of easy negatives from overwhelming the detector during training.

\subsection{Anchor-less Detectors}
\label{sec:back:cnn-based-detectors:anchor-less-detectors}
As already stated, most \acrshort{cnn}-based object detectors use multiple (typically 3 or 5) prior boxes, or anchors, to encode their predictions. These anchors have different sizes and aspect ratios, and they are slid over the image to find candidates for the objects to localize. This method has two major drawbacks. First, it slows down training since, typically, there is the need for a very large set of anchors because the detector is trained to classify whether each anchor box sufficiently overlaps with a ground truth box, and a large number of anchor boxes is needed to ensure sufficient overlap with most ground truth boxes; as a result, only a tiny fraction of anchor boxes will overlap with ground truth, creating a huge imbalance between positive and negative anchor boxes \cite{retinanet}. Second, it introduces many hyperparameters and design choices, including how many anchors, sizes, and aspect ratios should be employed. Such decisions are made mainly via ad-hoc heuristics and can become even more complicated when combined with multiscale architectures \cite{ssd, retinanet, DBLP:conf/icra/ArakiOHYF20}. Hence, recently, some works have explored anchorless detectors.

\paragraph{CornerNet.}
CornerNet \cite{cornernet} is a one-stage object detector that get over anchor boxes. It localizes object bounding boxes as a pair of key points, i.e., the top-left corner and the bottom-right corner, using a single \acrshort{cnn}, predicting a heatmap for the top-left corners of all instances of the same object category, a heatmap for all bottom-right corners, and an embedding vector for each detected corner. This embedding vector, inspired by the associate embedding in \cite{DBLP:conf/nips/NewellHD17}, is such that if a top-left corner and a bottom-right corner belong to the same bounding box, the distance between their embeddings should be small. The actual values of the embeddings are unimportant. Only the distances between the embeddings are used to group the corners. Furthermore, the authors also introduced a novel component named corner pooling, a new type of pooling layer that helps a convolutional network better localize corners of bounding boxes. 

\paragraph{CenterNet.}
Published by Zhou et al., \cite{centernet} introduced another deep detection architecture that removes the need for anchors, inspired by CornerNet \cite{cornernet}. During the training-set preparation, the authors proposed to draw a map with delta functions located at the center of the ground-truth bounding boxes. Then, they used a Gaussian filter to smear these centers, generating a smooth distribution that reaches a peak at each object center. In the end, for each image, they generated an associated ground truth heatmap. The model then uses two prediction heads: one is trained to predict the confidence heatmap, whereas the other head is trained, like in other objects detectors, to predict regression values for box dimensions and offsets, which refers to the box center predicted by the first head. The more interesting part is perhaps the fact that the confidence heatmap can be directly exploited to remove irrelevant predictions without decoding the boxes and without breaking the deep \acrshort{cnn} flow. In particular, the authors proposed running a max-pooling operation on the heatmap, aiming to elevate each cell a little bit, except at the very peak. Then, a small flat region is created very near the peak. Next, they suggested running a boolean element-wise comparison between the max-pool input and output that return the values as FP values (either 0. or 1.), creating a new heatmap where everywhere is 0 (since the max-pool operation elevated each value) except local maxima, where the value is 1. Finally, they proposed to run an element-wise multiplication between the output of this latter stage and the input to the max-pool to return each local maximum to its confidence level and leave the non-maximum values as 0. Like other detectors, a confidence threshold has been applied as the last step, decoding only the boxes corresponding to surviving maxima.

\section{Domain Adaptation}
\label{sec:back:domain_adaptation}
One of the most severe problems that \acrlong{dl} users are facing is related to data scarcity to be used during the training phase. Indeed, collecting and annotating datasets for every new task is a highly costly and time-consuming process. Consequently, sufficient training data is not always available, and data dependency is still a severe issue in specific domains. 
\acrshort{ml} models assume that training data (be it large or small) is representative of the underlying distribution. However, if the inputs at test time differ significantly from the training data, the model might experience performance degradation. As an illustrative example, let us assume that we want to automatically segment images captured by a car's camera to know what lies ahead (buildings, trees, other vehicles, pedestrians, traffic lights, etc.). Specifically, let us suppose to train the model in a supervised fashion by exploiting a well-labeled dataset comprising images taken in New York City. If we test this application on the streets of New York City, the model will work fine.
On the other hand, if we test the same model in Rome, we will experience remarkable performance degradation. The model did not perform well in this new scenario because the problem domain is changed; for instance, the cars look very different in Rome (there are no yellow cabs like in NYC), and the streets are not as straight anymore. In this case, \acrfull{da} represents a possible solution to mitigate the problem. \acrshort{da} is a field of \acrshort{ml} that deals with scenarios where a model trained on a source distribution is used in the context of a different (but related) target distribution. In general, \acrshort{da} uses labeled data in one or more source domains to solve new tasks in a target domain. The level of relatedness between the source and target domains usually determines how successful the adaptation will be. 

Formally, following the notation introduced in \cite{da_survey_1, da_survey_2}, we define a \textit{domain} $\mathcal{D}$ consisting of two components: a \textit{d}-dimensional feature space $\mathcal{X} \subset \mathbb{R}^d$ and a marginal probability distribution $P(X)$, where $X = \{x_1, \dots, x_n\} \subset \mathcal{X}$. Given a specific domain, $\mathcal{D} = \{\mathcal{X}, P(X)\}$,  we formulate a \textit{task $\mathcal{T}$} defined by a label space $\mathcal{Y}$ and the conditional probability distribution $P(Y|X)$, where $Y = \{y_1, \dots, y_n\}$ is the set of the corresponding labels for $X$. In general, $P(Y|X)$ can be learned in a supervised manner from these feature-label pairs $\langle x_i, y_i \rangle$.

Let us assume to have two domains with their related tasks; a \textit{source} domain $\mathcal{D}_S = \{\mathcal{X}_S, P(X_S)\}$ with $\mathcal{T}_S = \{\mathcal{Y}_S, P(Y_S|X_S)\}$ and a \textit{target} domain $\mathcal{D}_T = \{\mathcal{X}_T, P(X_T)\}$ with $\mathcal{T}_T = \{\mathcal{Y}_T, P(Y_T|X_T)\}$. Traditional \acrshort{ml} methods can be used to solve the problem, i.e., to learn $P(Y_T|X_T)$, if the two domains and their learning tasks are the same, i.e., $\mathcal{D}_S = \mathcal{D}_T$ and $\mathcal{T}_S = \mathcal{T}_T$. $\mathcal{D}_S$ and $\mathcal{D}_T$ become the \textit{training} and the \textit{test} set, respectively. 

However, when this assumption does not hold, and the domains are different, then either (i) the feature spaces between the domains are different, i.e., $\mathcal{X}_S \neq \mathcal{X}_T$, or (ii) the feature spaces between the domains are the same, but the marginal probability distributions between domain data are different, i.e., $P(X_S) \neq P(X_T)$ where $X_{S_i} \in \mathcal{X}_S$ and $X_{T_i} \in \mathcal{X_T}$, the models trained on $\mathcal{D}_S$ might perform poorly on $\mathcal{D}_T$, or they are not directly applicable to $\mathcal{T}_T$. When there exists some relationship, explicit or implicit, between the feature spaces of the two domains, we say that the source and target domains are \textit{related}; in this case, it is possible to exploit the information from $\{\mathcal{D}_S, \mathcal{T}_S\}$ to learn $P(Y_T|X_T)$, in a process known as \textit{\acrfull{tl}}. 

It is possible to distinguish between \textit{homogeneous \acrshort{tl}} and \textit{heterogeneous \acrshort{tl}}. In the first case, the feature spaces of the source and target domains are identical ($\mathcal{X}_S = \mathcal{X}_T$). Hence, the source and target datasets are generally different in terms of data distributions ($P(X_S) \neq P(X_T)$) due to distribution shift. On the other hand, in the second case, the feature spaces of the source and target domains can have different representations ($\mathcal{X}_S \neq \mathcal{X}_T$), or they can even be of different modalities such as image \textit{vs.} text.

Based on these definitions, the authors of \cite{da_survey_1} further classify the \acrshort{tl} approaches into three main categories, considering not only the different situations concerning the source and target domains but also the corresponding tasks. In the \textit{inductive \acrshort{tl}}, the source and the target tasks are different ($\mathcal{T}_S \neq \mathcal{T}_T$), no matter whether the source and target domains are the same or not; at least a few labeled data in the target domain are required as the training data to \textit{induce} the target predictive function. In the case of \textit{transductive \acrshort{tl}}, instead, the source and the target distributions are different due to selection bias or distribution mismatch ($\mathcal{X}_S \neq \mathcal{X}_T$). Finally, in the \textit{unsupervised \acrshort{tl}}, both the domains and the tasks are different ($\mathcal{X}_S \neq \mathcal{X}_T$ and $\mathcal{T}_S \neq \mathcal{T}_T$) and, in general, labels are not available neither for the source nor for the target. The goal is to exploit the (unlabeled) information in the source domain to solve unsupervised learning tasks in the target domains. Some tasks belonging to the latter category are clustering, dimensional reduction and density estimation \cite{DBLP:conf/icml/DaiYXY08, DBLP:conf/pkdd/WangSZ08}. According to this classification, \acrshort{da} methods are transductive \acrshort{tl} solutions with the assumption that the tasks are the same, i.e., $\mathcal{T}_S = \mathcal{T}_T$ and the differences are only caused by domain divergence, i.e., $\mathcal{D}_S \neq \mathcal{D}_T$.

In addition, \acrshort{da} solutions, both homogeneous and heterogeneous, are further divided into three categories. In the \textit{supervised \acrshort{da}}, a small amount of labeled target data, $\mathcal{D}_{TL}$, are present but commonly not sufficient for tasks. In the \textit{semi-supervised \acrshort{da}} both limited labeled data $\mathcal{D}_{TL}$ and redundant unlabeled data $\mathcal{D}_{TU}$ are available in the training stage, which allows the networks to learn the structure information of the target domain. Finally, in the \textit{unsupervised \acrshort{da}} \footnote{Note that the unsupervised DA is not related to the unsupervised TL, for which no source labels are available, and, in general, the task to be solved is unsupervised.} no labeled but sufficient unlabeled target domain data, $\mathcal{D}_{TU}$, are observable when training the network.

\paragraph{Deep Domain Adaptation.} Although some studies have shown that deep networks can learn more transferable representations that disentangle the exploratory factors of variations underlying the data samples and group features hierarchically by their relatedness to invariant factors, Donahue et al. \cite{DBLP:conf/icml/DonahueJVHZTD14} showed that a domain shift still affects their performance. The deep features would eventually transition from general to specific, and the transferability of the representation sharply decreases in higher layers. Therefore, recent work has addressed this problem by deep \acrshort{da}, which combines \acrshort{dl} and \acrshort{da}. Although shallow methods where deep networks are employed only to extract features and do not help transfer knowledge directly can be considered deep \acrshort{da} (such as \cite{DBLP:conf/iccv/LuZCWXSH17}), here we focus on a more strict definition. We discuss deep \acrshort{da} techniques where the \acrshort{da} is in some way embedded in the learning process (i.e., back-propagated) and where the goal is to produce deep domain invariant feature representations. Deep \acrshort{da} techniques generally fall into three categories, namely \textit{discrepancy-based}, \textit{adversarial-based} and \textit{reconstruction-based}, that we analyze in the following. It is worth noting that most of these techniques are proposed in the context of the classification task. At the end of this section, we discuss different application examples using deep \acrshort{da}, a field which, however, remains more unexplored.

\subsection{Discrepancy-based Domain Adaptation}
These approaches aim to reduce the shift between the two domains, fine-tuning a deep network model by minimizing some divergence criteria between the source and the target data distributions. Indeed, the authors of \cite{DBLP:conf/nips/YosinskiCBL14} demonstrated that fine-tuning can enhance the generalization capabilities of deep networks, mitigating the limitations characterizing the learned transferable features due to representation specificity. In particular, the fine-tuning operation aims to train a base network exploiting the labeled source data (labeled or unlabeled), and then directly re-use the first $n$ layers to carry out a target network using the target data, where the remaining layers are randomly initialized and trained using a discrepancy-based loss function. Depending on the size of the target dataset and on its similarity with the source one \cite{best_practices_fine_tuning}, the first $n$ layers of the target network can be frozen or fine-tuned during the training. In the end, the minimization of this divergence criterion between the source and the target data distributions aims to achieve a domain-invariant feature representation.

Discrepancy-based \acrshort{da} can be further classified into four categories: \textit{Class Criterion}, \textit{Statistic Criterion}, \textit{Architecture Criterion} and \textit{Geometric Criterion}.

\paragraph{Class Criterion.}
It is the most basic training loss in deep \acrshort{da}. Here, the assumption to be made is to have a small \textit{labeled} target dataset to be used as a guide for transferring knowledge between different domains. After pre-training the network using the source data, the remaining layers of the target model are trained to exploit the target class label information. Ideally, the class label information is given directly in the supervised process, commonly using as the training loss $L$ the negative log-likelihood of the ground-truth class $y_i$ and the softmax predictions of the model $\hat{y_i}$, which represent class probabilities \cite{class_crierion_3, class_crierion_1, class_crierion_2}:

\begin{equation*}
    L = - \sum_{i=0}^{N} y_i \log \hat{y_i}.
\end{equation*}

\noindent To extend this, Hinton et al. \cite{soft_label} proposed a modified version of the softmax function to \textit{soft} label loss:

\begin{equation*}
    q_i = \frac{\exp{(\frac{z_i}{T}})}{\exp{(\sum_j \frac{z_j}{T})}},
\end{equation*}

\noindent where $z_i$ is the logit output computed for each class, while $T$ is a temperature that is normally set to 1 in standard softmax, but it takes a higher value to produce a softer probability distribution over classes, aiming at obtaining much of the information about the learned function that resides in the ratios of very small probabilities. 

Relaxing the assumption of having a \textit{labeled} target dataset, so when there is no class label information in the target domain, some works tried to imitate the ability of humans to identify unseen classes given only a high-level description. For example, the authors in \cite{class_crierion_4} introduced high-level semantic attributes per class, and Gebru et al. \cite{class_crierion_5} leveraged these attributes to improve performance in the \acrshort{da} of fine-grained recognition. Occasionally, when fine-tuning the network in the setting of unsupervised \acrshort{da}, it is possible to preliminarily obtain a label of the target data, which is called a \textit{pseudo label}, based on the maximum \textit{a posteriori} probability. For example, Yan et al. \cite{class_crierion_6}, after the initialization of the target model exploiting the source data, defined the class \textit{a posteriori} probability $p(y^{t}_{j} = c|x^{t}_{j})$ using the output of the target model and assigning pseudo-label $\hat{y^{t}_{j}}$ to $x^{t}_{j}$ by $\hat{y^{t}_{j}} = \argmax_c p(y^{t}_{j} = c|x^{t}_{j})$.

\paragraph{Statistic Criterion.}
This criterion aims to try to align the statistical distribution shift between the source and the target domains. 

The most commonly method for comparing and reducing this distribution shift is the \acrfull{mmd} \cite{DBLP:journals/jmlr/GrettonBRSS12}. In general, \acrshort{mmd} is defined by the idea of representing distances between distributions as distances between mean embeddings of features.




In subsequent works, Tzeng et al. \cite{deep_mmd_1} and Long et al. \cite{deep_mmd_2} involved also deep \acrshort{cnn}, extending \acrshort{mmd}. In particular, the authors of \cite{deep_mmd_1} introduced the Deep Domain Confusion network (DDC), comprising two \acrshort{cnn}s for the source and the target domains with shared weights; DDC is optimized in the source domain using a classification loss, while the domain difference is assessed by an adaptation layer exploiting the \acrshort{mmd} metric. 
%
%
Since DDC adapts only one layer of the network, limiting the transferability of multiple layers, rather than using a single layer and a linear \acrshort{mmd}, the authors in \cite{deep_mmd_2} introduced a Deep Adaptation Network (DAN), aiming at matching the shift across domains in marginal distributions, by adding multiple adaptation layers and exploring multiple kernels. However, since, in practice, the source classifier cannot be directly used in the target domain, Long et al. \cite{deep_mmd_3} proposed a joint adaptation network (JAN) to align the shift in the joint distributions of input features and output labels in multiple domain-specific layers based on a Joint \acrshort{mmd} (JMMD) criterion. Another remarkable work is \cite{class_crierion_6}, where the authors introduced a weighted \acrshort{mmd} model characterized by an extra weight for each class belonging to the source domain when the target domain class weights are not the same as those in the source domain.


\paragraph{Architecture Criterion.}
The goal of these methods is to improve the ability to learn more transferable features by optimizing and modifying the architecture of the adopted deep networks. 

In \cite{archi_criterion_1}, the authors introduced a two-stream architecture with related (and not shared) weights. Specifically, they introduced a weight regularizer $r_w(\cdot)$ between the weights of the source and the target models, accounting for the differences between the two domains. The weight regularizer $r_w(\cdot)$ is expressed as the exponential loss function below:

\begin{equation*}
    r_w(\theta^{s}_j, \theta^{t}_j) = \exp (\|\theta^{s}_j - \theta^{t}_j\|^2) -1,
\end{equation*}

\noindent where $\theta^{s}_j$ and $\theta^{t}_j$ represent the parameters of the $j^{th}$ layer of the source and target models, respectively. In \cite{archi_criterion_2}, the idea is that domain-related knowledge is represented by the statistics of the \acrfull{bn} layer \cite{batch_norm} (while the class-related knowledge is represented in the weight matrix). Basically, \acrshort{bn} normalizes the mean and standard deviation for each individual feature channel such that each layer receives data from a similar distribution, regardless if it comes from the source or the target domain. Thus, \acrshort{bn} is exploited to align the distribution, recomputing the mean and standard deviation in the target domain. Formally:

\begin{equation*}
    BN(X^t) = \lambda \left( \frac{x - \mu(X^t)}{\sigma(X^t)}\right)  + \beta,
\end{equation*}

\noindent where $\lambda$ and $\beta$ are parameters learned from the target data, and $\mu(x)$ and $\sigma(x)$ are the mean and standard deviation computed independently for each feature channel. Instead, Xiao et al. \cite{archi_criterion_3} proposed a domain-guided dropout to solve the problem of multi-\acrshort{da} muting neurons that are not effective and related for each domain. Instead of exploiting a specific dropout rate, they assigned dropout depending on the gain of the loss function of each neuron on the domain sample when the specific neuron is removed.

\paragraph{Geometric Criterion.}
This criterion assumes that the relationship of geometric structures can reduce the domain shift, and thus it bridges the source and target domains according to their geometrical properties. In particular, it mitigates the domain shift by integrating intermediate subspaces (a fixed \cite{geometric_criterion_2} or infinite number \cite{geometric_criterion_3}) on a geodesic path between the two domains; both source and target data are projected to the obtained intermediate subspaces to align the distribution. For example, Chopra et al. \cite{geometric_criterion_1} proposed a model called deep learning
for \acrshort{da} by interpolating between domains (DLID). In particular, starting with all the source data samples and gradually replacing them with the target data samples, DLID creates intermediate datasets; then, a deep feature extractor is trained in an unsupervised fashion, exploiting the sparse predictive decomposition.

\subsection{Adversarial-based Domain Adaptation}
These techniques aim to encourage domain confusion through an adversarial objective, minimizing the distance between the empirical source and target mapping distributions; here, a domain discriminator classifies whether a data point is drawn from the source or the target domain. The main idea derives from the method adopted by the \acrfull{gan} \cite{gan}.
In particular, \acrshort{gan} comprises two modules: a Generative model $G$ responsible for extracting the data distribution, and a Discriminative model $D$ (also called Discriminator) in charge of distinguishing whether a sample comes from $G$ or training datasets, by predicting a binary label. The two networks are trained using the label prediction loss in a min-max fashion: $G$ is optimized to minimize the loss, while, at the same time, $D$ is trained to maximize the probability of assigning the correct label. In \acrshort{da}, this principle has been employed to ensure that the network cannot distinguish between the source and target domains.

Following \cite{deep_da_survey}, we classify Adversarial-based into two categories: \textit{Generative Models} and \textit{Non-generative Models}

\paragraph{Generative Models.} These \acrshort{da} strategies combine the discriminator with a generative component, often based on \acrshort{gan}s. One of the typical cases is to use source images, noise vectors, or both, to generate simulated samples that are similar to the target ones, preserving the annotation information in the source domain and consequently tackling the problem of the lack of training data. A notable example is the generation of synthetic data with ground-truth annotations, which is helpful for training the target model.
For example, in \cite{da_gan_1} a domain discriminator is employed to guarantee the invariance of content to the source domain, and a real/fake discriminator is paired with the generator to produce similar images to the target domain. More, the authors of \cite{da_gan_2} introduced a method to improve the realism of synthetic images using unlabeled real data, combining an adversarial loss and a self-regularization loss. Instead, Bousmalis et al. \cite{da_gan_3} introduced a novel model exploiting a \acrshort{gan} conditioned by both a noise vector and source images; in this case, the classifier is trained to predict labels for both source and synthetic images, while the discriminator is trained to predict the domain labels belonging to the target and the synthetic images. 

\paragraph{Non-generative Models.}
The goal of these models is to learn a discriminative feature representation using the labels in the source domain and to map the target data to the same space through a domain-confusion loss. Thus, the feature extractor learns domain-invariant representations rather than generating models with input image distributions. With these representations, the distribution of both domains can be similar enough such that the classifier is fooled and can be directly used in the target domain even though it is trained on source samples. Therefore, whether the representations are domain-confused or not is crucial to transferring knowledge. Inspired by \acrshort{gan}, domain confusion loss, which the discriminator produces, is introduced to improve the performance of deep \acrshort{da} without generators. Ganin et al. \cite{ganin2015unsupervised} introduced DANN (Domain-Adversarial Neural Network), integrating a Gradient Reversal Layer (GRL) into the standard architecture to make similar the two domains. In particular, DANN minimizes the domain confusion loss for all samples and label prediction loss only for source samples while maximizing domain confusion loss via the use of the GRL. The authors of \cite{class_crierion_1} proposed to add a domain classification layer performing binary domain classification and introduced a domain confusion loss aiming at making its outputs to be as close as possible to a uniform distribution over binary labels. Volpi et al. \cite{DBLP:conf/cvpr/VolpiMSM18} performed data augmentation in the source feature space exploiting a feature generator S. They obtained domain invariant features with a min-max game against generated features by S.

\subsection{Reconstruction-based Domain Adaptation}
This third category of deep \acrshort{da} approaches has the goal to reconstruct the source or target samples, assuming that it can help improve the performance of \acrshort{da}. In particular, the reconstructor can ensure both specificities of intra-domain representations and indistinguishability of inter-domain representations. 

It is possible to divide this approaches in two categories: \textit{Encoder-Decoder Reconstruction} and \textit{Adversarial Reconstruction}.

\paragraph{Encoder-Decoder Reconstruction.}
The \acrshort{da} approaches based on encoder-decoder reconstruction typically learn the domain-invariant representation by a shared encoder and maintain the domain-specific representation by a reconstruction loss in the source and target domains. Xavier and Bengio \cite{DBLP:conf/icml/GlorotBB11} proposed to use as the encoder the Stacked Denoising Autoencoder (SDA) \cite{DBLP:journals/jmlr/VincentLLBM10}. SDA extracts a high-level representation of the inputs (coming from all the domains), which is then passed to a linear classifier - trained on the source domain labeled data - to perform predictions on the target data. A more efficient evolution of this approach is presented in \cite{DBLP:conf/icml/ChenXWS12}. Ghifary et al. \cite{DBLP:conf/eccv/GhifaryKZBL16} introduced the Deep Reconstruction Classification Network (DRCN), a \acrshort{cnn} comprising a shared encoder and two branches. The encoder provides the feature representation; the first branch learns to classify source data by exploiting supervision from source labels; the second is a deconvolutional network optimized for unsupervised reconstruction with target data. On the other hand, the authors of \cite{DBLP:conf/nips/BousmalisTSKE16} proposed the Domain Separation Network (DSN), comprising two encoders. One is a shared-weight encoder aiming at learning to capture shared representations; the second is a private encoder used for domain-specific components in each domain. More, a shared decoder learns to reconstruct the input samples by exploiting both the private and the shared representation.

\paragraph{Adversarial Reconstruction.}
Adversarial reconstruction is inspired by dual learning \cite{DBLP:conf/nips/HeXQWYLM16}, and it is adopted in deep \acrshort{da} with the help of dual \acrshortpl{gan}. Zhu et al. \cite{DBLP:conf/iccv/ZhuPIE17} proposed Cycle \acrshort{gan}, a technique for image-to-image translation without paired training examples. This algorithm employs two generators, which learn a mapping $G: X \to Y$ and the inverse mapping $F: Y \to X$. It exploits two discriminators $D_X$ and $D_Y$ and two losses; $D_X$ uses an adversarial loss to measure how realistic the generated image is ($G(X) \thickapprox Y$ or $G(Y) \thickapprox X$); $D_Y$ utilizes a cycle consistency loss (or reconstruction loss) to measure how well the original input is reconstructed after a sequence of two generations ($F(G(X)) \thickapprox X$ or $G(F(Y)) \thickapprox Y$). Similarly, dual \acrshort{gan} \cite{DBLP:conf/iccv/YiZTG17} and disco \acrshort{gan} \cite{DBLP:conf/icml/KimCKLK17} have been proposed. The former one, employed skip connections between mirrored downsampling and upsampling layers \cite{unet, DBLP:conf/cvpr/IsolaZZE17} for the generator, and the Markovian patch-\acrshort{gan} architecture \cite{DBLP:conf/eccv/LiW16} for the discriminator. The second one exploited as the reconstruction loss various distance functions such as \acrfull{mse}, cosine distance and hinge loss.

Deep \acrshort{da} strategies have been recently applied to other fields other than the classification task. In particular, we discuss deep \acrshort{da} techniques suited for face recognition, person re-identification, object detection, and semantic segmentation.

\paragraph{Face recognition.}
In face recognition, the domain shift may be represented by expressions, poses, resolution, and others. For instance, since current benchmarks do not perform well on young children due to a lack of training data, Xia et al. \cite{DBLP:conf/iccvw/Xia0W17} proposed to transfer the knowledge from adults to infants. More, the authors of \cite{DBLP:conf/icip/HongIRY17} exploited synthetic data having varying poses, increasing the number of images in the source domain, and proposed an adversarial training strategy to fill the gap between the synthetic data and the real-world ones. 

\paragraph{Person re-identification.} 
Person re-identification (re-ID) aims at recognizing the identities of the people between different cameras. Here, deep \acrshort{da} strategies can be helpful when trained models on one dataset are directly employed in another dataset. Deng et al. \cite{DBLP:conf/cvpr/Deng0YK0J18} proposed Similarity Preserving Generative Adversarial Network (SPGAN). Inspired by Cycle \acrshort{gan} \cite{DBLP:conf/iccv/ZhuPIE17} and Siamese network, this solution translates in an unsupervised fashion the labeled source image into the target domain. Then it trains re-ID models with supervised learning that exploits the translated images. On the other hand, Xiao et al. \cite{archi_criterion_3} introduced a dropout algorithm able to discard useless neurons under a guide provided by the characteristics of the domain.

\paragraph{Object detection.}
Most deep \acrshort{da} techniques applied to the object detection task rely on two-stage detectors (see also \ref{sec:back:cnn-based-detectors} for further details) because they produce, as the first step, a set of regions that have therefore been classified in the second stage. Considering this region selection mechanism being domain-independent, deep \acrshort{da} strategies can be helpful in the second-stage classification process to adapt to the target domain. In other words, deep \acrshort{da} methods act in the second stage of the detection pipeline, where R-CNNs train classifiers over the proposed regions. To this end, most works exploit the limited images annotated with bounding boxes to learn to detect and, in parallel, use weakly labeled data (such as image-level class labels) during the second stage to adapt the detector to the target domain. For instance, Hoffman et al. \cite{DBLP:conf/nips/HoffmanGTHDGDS14} proposed the Large-Scale Detection through Adaptation (LSDA) network where, firstly, a classification layer is trained for the target domain; then, a pre-trained source detection model is employed, together with output layer adaptation strategies aiming at update the target classification parameters. Instead, Chen et al. \cite{DBLP:conf/cvpr/Chen0SDG18} considered the Faster R-CNN detector incorporating an image-level and an instance-level adaptation component in the detection pipeline; they minimized the domain discrepancy by exploiting an adversarial strategy during the training phase. 

\paragraph{Semantic segmentation.}
\acrfullpl{fcn} demonstrated to be effective for dense per-pixel prediction; however, also their performance degraded because of domain shift. Therefore, some works have been proposed to tackle it. Most of them rely on deep unsupervised \acrshort{da}. Zhang et al. \cite{DBLP:journals/pami/ZhangDFG20} exploited synthetic data to enhance performance in real-world images using the global label distribution loss of the images and local label distribution loss of the landmark super-pixels in the target domain as a regularizer for the fine-tuning of the network. On the other hand, Chen et al. \cite{DBLP:conf/iccv/ChenCCTWS17} proposed a framework for cross-city semantic segmentation. \cite{hong2018conditional} employed a residual network and adversarial training to make the source feature maps closer to the target ones. The authors of \cite{chen2019learning} combine semantic segmentation and depth estimation to boost the adaptation performance, providing to the discriminator the segmentation and the depth prediction maps jointly.

\section{Datasets and Annotation Types}
\label{sec:back:datasets}
\begin{figure}
    \centering
  \subfloat{
       \includegraphics[width=0.325\linewidth, height=2.8cm]{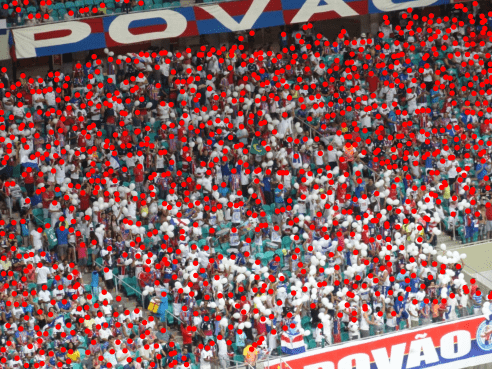}}
  \subfloat{
        \includegraphics[width=0.325\linewidth, height=2.8cm]{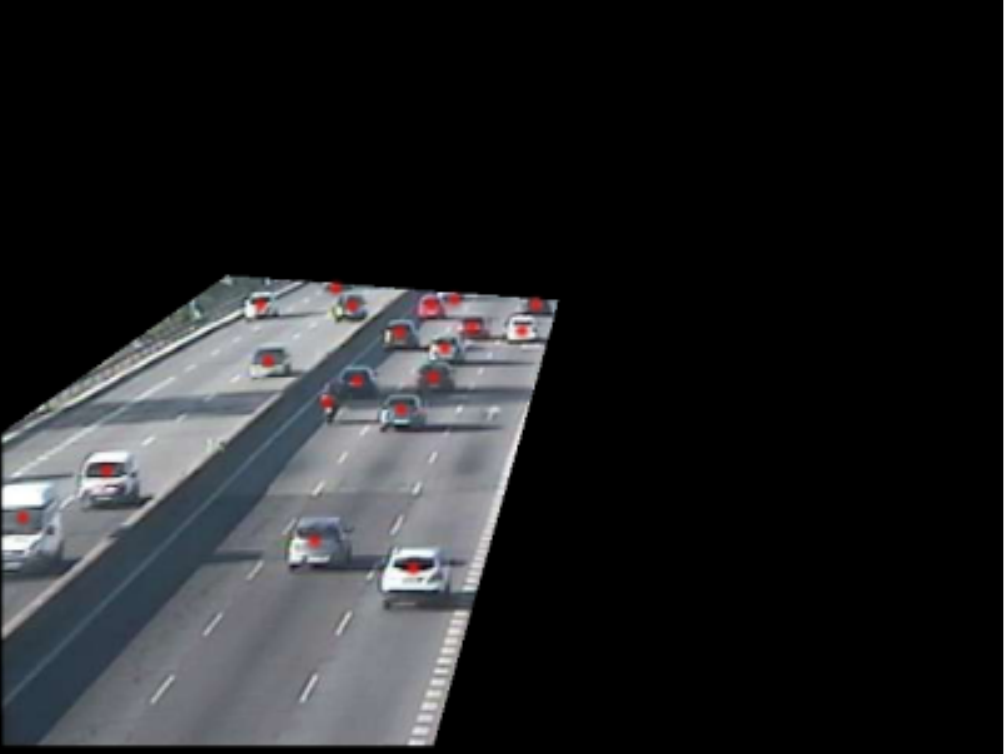}}
  \subfloat{
        \includegraphics[width=0.325\linewidth, height=2.8cm]{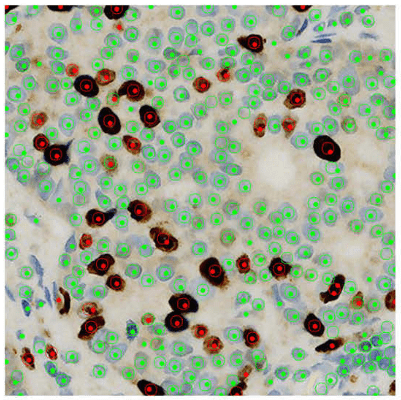}}
 \hfill
    \\ [0.5ex]
  \subfloat{
        \includegraphics[width=0.325\linewidth, height=2.8cm]{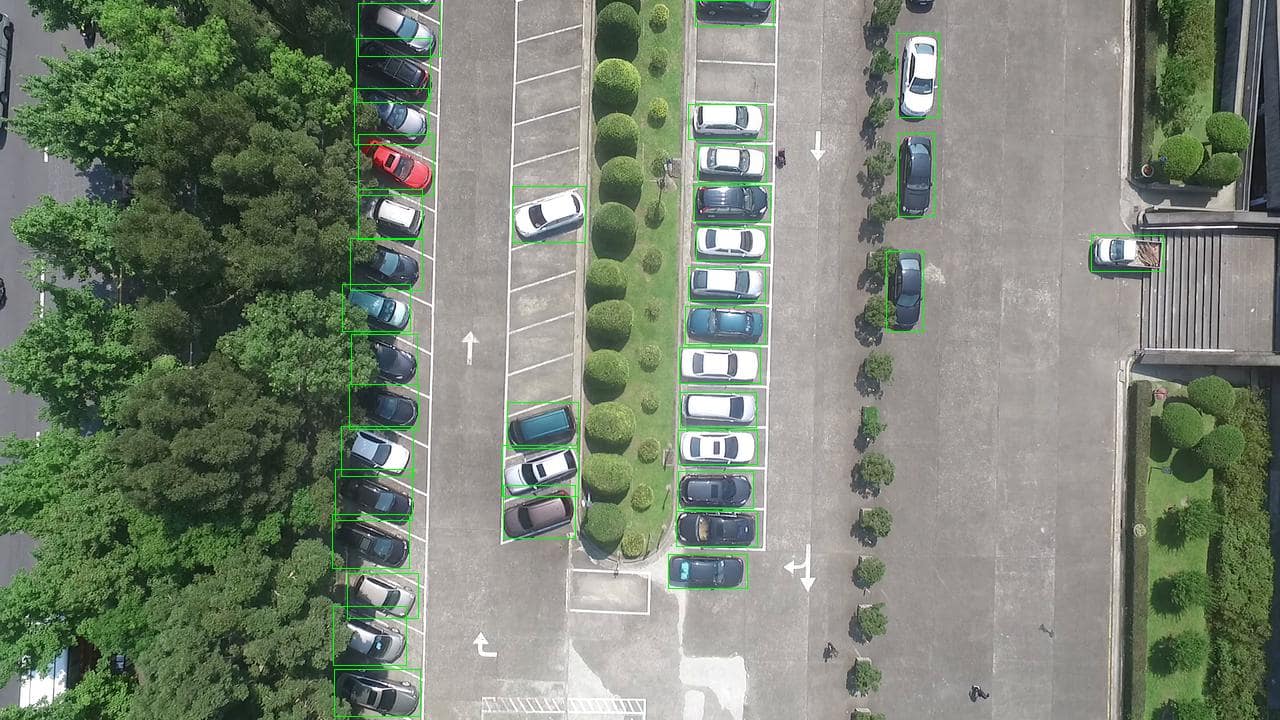}}
  \subfloat{
        \includegraphics[width=0.325\linewidth, height=2.8cm]{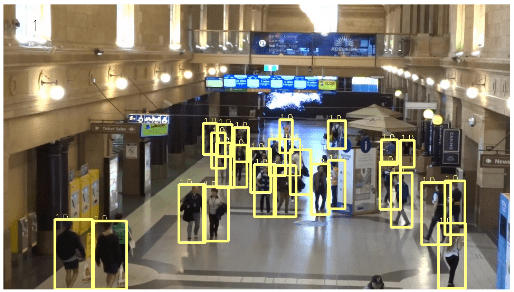}}
  \subfloat{
        \includegraphics[width=0.325\linewidth, height=2.8cm]{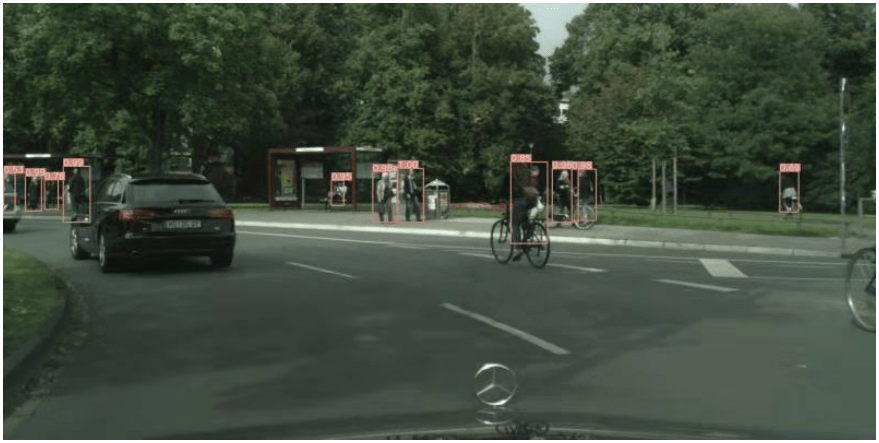}}
 \hfill
    \\ [0.5ex]
  \subfloat{
        \includegraphics[width=0.325\linewidth, height=2.8cm]{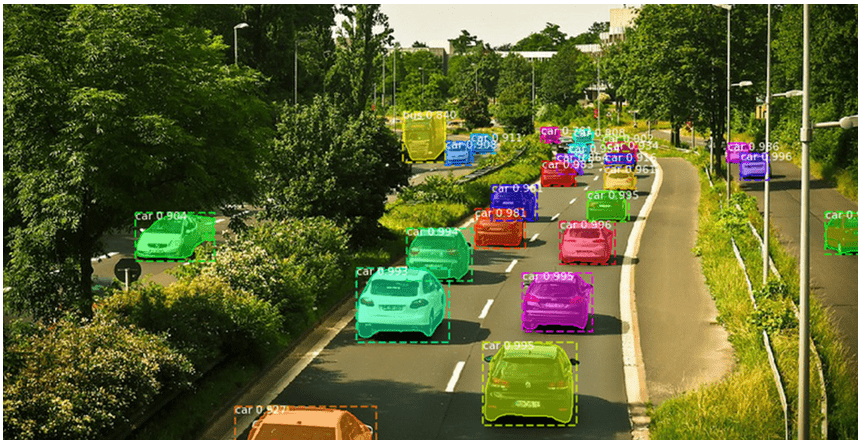}}
  \subfloat{
        \includegraphics[width=0.325\linewidth, height=2.8cm]{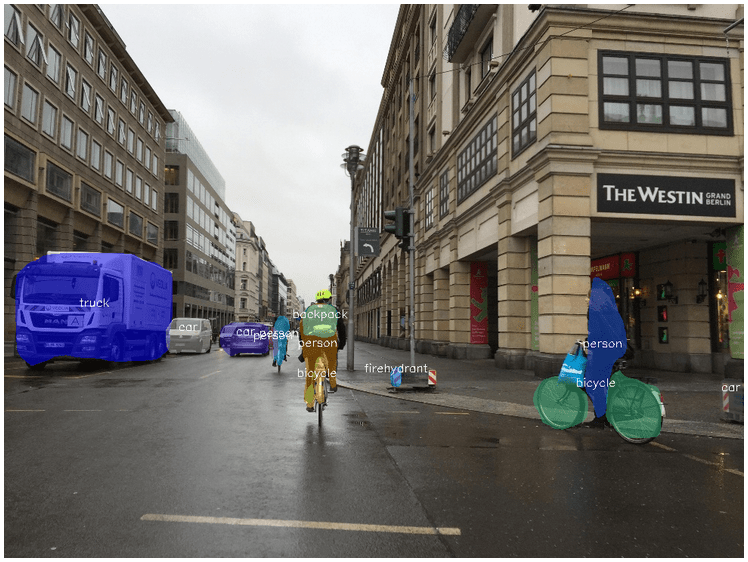}}
  \subfloat{
        \includegraphics[width=0.325\linewidth, height=2.8cm]{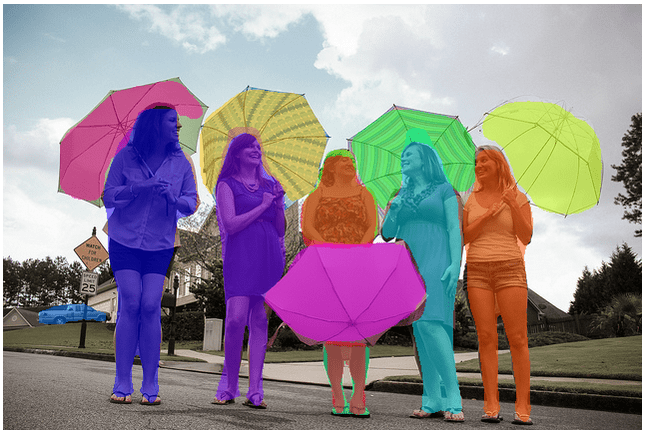}}
 \hfill

  \caption{\textbf{Samples of annotations used for the counting task.} Top row: \textit{dot annotations}, where dots indicate a coarse localization of the object instances. Middle row: \textit{bounding box annotations}, where rectangles localize the objects also giving an estimate of their size. Bottom row: \textit{per-pixel annotations}, where per-pixel masks precisely identify the objects. More informative labels lead to a higher human effort for the annotation procedure.}
  \label{ann_types} 
\end{figure}

The availability of training data for the supervised learning process is one of the backbones for Computer Vision \acrshort{ai}-based model development, and it is not surprising that advances have been driven by the constant growth of available well-labeled data and benchmarks. This section describes the datasets employed directly or indirectly in the experiments presented in this work. Some of these have been collected during this thesis and represent a concrete scientific contribution to the Computer Vision community. Others were already publicly available and have been exploited mainly for state-of-the-art comparisons. We decide to group the datasets depending on their annotations. In fact, it is worth noting that, in some cases, the annotations delineate the task for which the datasets can be exploited and represent an indicator of the human effort employed for their generation. 
Concerning the counting task, bare minimum annotations correspond to provide the overall count of objects in each training image. However, in this way, all spatial information is lost. This lack of knowledge makes hard the training phase of the supervised models. The next level of annotation is to specify the object position by putting a single dot on each object instance in each image. Although the provided information consists of weak localization of the objects, this knowledge can be profitably placed in the training loop. More precise labels, like bounding boxes or per-pixel annotations, lead to a considerable higher human effort for the annotation procedure; on the other hand, these last labels carry more helpful information, such as the dimension of the objects. Figure \ref{ann_types} shows some samples of annotations used for the counting task.

\subsection{Dot-annotated Datasets}
These datasets are characterized by labels consisting of dots placed on the object centroids or over significant parts of the objects, for example, the heads of the people. Dotting (e.g., pointing) is the natural way to count objects for humans, at least when their number is high. Arguably, providing dotted annotations for the training images is no harder for humans than giving just the raw counts. Furthermore, dots provide additional information, e.g., a spatial arrangement of the objects, that can be exploited in the supervised training process. Overall, it should be noted that dotted annotation is less labor-intensive than the bounding-box annotation, let alone pixel-accurate annotation, commonly used by supervised techniques in many computer vision tasks. On the other hand, these labels provide just a weak localization of the objects but do not give additional crucial information, like their size. Therefore, they represent a good trade-off between the information usable during the training phase and the human effort in terms of labeling. In the following, we briefly outline the dot-annotated datasets exploited in this dissertation.

\paragraph{\acrfull{trancos}~\cite{ExtremelyTrancos}.}
The \acrshort{trancos} dataset is a public dataset containing 1244 dot-annotated images of different congested traffic scenes captured by surveillance cameras, mainly exploited for evaluating counting by density estimation approaches. The ground truth density maps are generated by putting one Normal Gaussian kernel for each dot present in the scene, having a value of \(\sigma\) empirically decided by the authors, who also provided the regions of interest (ROIs) for each image.

\paragraph{VGG Cells~\cite{learning_to_count}.}
This public dataset contains 200 RGB synthetic images simulating bacterial cells from fluorescence-light microscopy at various focal distances. The size of each image is fixed to 256×256×3 pixels, and the cells are designed to be clustered and occluded with each other. The authors annotated the images by putting dots in the cell centroids.

\paragraph{\acrfull{mbm}~\cite{mbm_original}.}
This dataset contains RGB real microscopy images of human bone marrow with various cell types stained blue, annotated with dots in their centroids. The original image size of each image was 1200×1200×3 pixels, but the authors in \cite{count-ception} split each of them into four images with the size of 600×600 pixels. The images in this dataset have non-homogeneous tissue background and significant cell shape variance. 

\paragraph{\acrfull{gcc}~\cite{gcc_dataset}.}
The \acrshort{gcc} dataset is a large-scale and diverse synthetic crowd counting dataset gathered from the video-game Grand Theft Auto V (GTA5) and automatically annotated. It consists of 15,212 images, with a resolution of $1080\times1920$, containing 7,625,843 automatically dot-annotated persons in 400 different scenarios with various locations, weather conditions, and crowd densities. Compared with the existing datasets, \acrshort{gcc} is a more large-scale crowd counting dataset in both the number of images and persons.

\paragraph{ShanghaiTech~\cite{multi_column}.} The ShanghaiTech dataset is a large-scale crowd dataset of nearly 1,200 manually dot-annotated images with a total of 330,165 people with centers of their heads. This dataset consists of two parts: part A, containing 482 images crawled randomly from the Internet, and part B, composed of 716 images taken from the busy streets of metropolitan areas in Shanghai. The crowd density varies significantly between the two subsets, making this dataset more challenging. 

\paragraph{UCF-QNRF~\cite{composition_loss}.} The UCF-QNRF dataset is a collection of images gathered from three sources: Flickr, Web Search, and the Hajj footage. The authors performed the entire annotation process in two stages, the first one for the labeling and the second one for the verification, for a total of 2,000 human-hours spent through to its completion. This dataset comprises 1,535 images with more than 1 million dot annotations on the centers of the heads of the pedestrian, divided into training and test subsets.

\paragraph{NWPU-Crowd~\cite{nwpu_dataset}.} The NWPU dataset is a large-scale congested crowd counting and localization dataset consisting of 5,109 images, in a total of 2,133,375 annotated heads with points and boxes. Compared with other real-world datasets, it has the most extensive density range. Another peculiarity of this dataset is that it also comprises negative samples like high-density crowd images to assess the robustness of models.

\paragraph{PNN.} This dataset consists of a collection of fluorescence microscopy images of mice brain slices containing Perineuronal Nets (PNNs), extracellular matrix aggregates surrounding the cell body of a large number of neurons throughout the nervous system. It is worth noting that the non-trivial appearance of PNNs causes difficulty in localizing them accurately, even for human experts.
For this reason, a small part of the dataset has been labeled by multiple raters; nonetheless, the maximum agreement between raters is roughly 70\%, highlighting the need for an automated counting technique that accounts for uncertain patterns. This dataset has been introduced on purpose in this dissertation, specifically in \ref{ch:counting-with-uncertainty}, and it has been made available for the scientific community at \cite{luca_ciampi_2021_5567032}.

\paragraph{\acrfull{adi}~\cite{count-ception}.}
The \acrshort{adi} dataset is a human subcutaneous adipose tissue dot-annotated collection of microscopy images. It consists of 200 Regions Of Interest (ROI) of 150×150×3 pixels in size sampled from high-resolution histology slides representing adipocyte cells. The average cell count across all images is $165 \pm 44.2$, and the size of the biological structures can vary dramatically, representing a challenging test case for automated cell counting procedures.

\paragraph{\acrfull{bcd}~\cite{Huang_2020}.}
This dataset is a recent collection of 1,338 images based on Ki-67 staining with 181,074 dot-annotated cells divided into two classes (positive and negative tumor cells, i.e., malignant and not malignant, respectively). Unlike other datasets, BCData is not only large in scale concerning the labeled objects but also considering the number of different unique patient cases (that are 394). The size of each image is fixed to 640×640×3 pixels, and the authors divided the dataset into training, validation, and testing split at a ratio of approximately 6:1:3 (803, 133, and 402 images, respectively).

\subsection{Bounding Box-annotated Datasets}
These datasets are characterized by labels consisting of bounding boxes localizing the objects.
These annotations provide a more precise localization than dots, although the considered regions often contain background portions near the objects. On the other hand, the human effort for labeling is higher than dot annotations. This category of datasets is commonly exploited for detection and tracking tasks. In the following, we briefly describe the ones used in this dissertation.

\paragraph{\acrfull{webcamt}~\cite{understandingCosteira}.} The \acrshort{webcamt} dataset is a collection of traffic scenes recorded using city-cameras. 
It is particularly challenging to analyze due to the low-resolution \((352\times240)\), high occlusion, and large perspective. It comprises about 40,000 images belonging to 10 different cameras and consequently having different views. 

\paragraph{\acrfull{viped}~\cite{ciampi_iciap, ciampi_viped}.} \acrshort{viped} is a new synthetic dataset introduced in this thesis (specifically, in \ref{ch:virtual-to-real}), generated with the highly photo-realistic graphical engine of the video game GTA V (Grand Theft Auto V) by Rockstar North and \textit{automatically} annotated with precise bounding boxes. It extends the \acrfull{jta} dataset, presented in \cite{fabbri_jta}. The dataset includes a total of about 500K images extracted from 512 full-HD videos of different urban scenarios. These videos are organized into a training set (256 videos) and a test set (the remaining 256 videos). The dataset is made freely available for the scientific community at \href{https://ciampluca.github.io/viped}{https://ciampluca.github.io/viped}.

\paragraph{MOT17Det~\cite{mot17_dataset} and MOT20Det~\cite{mot20_dataset}.} The MOT17Det and MOT20Det datasets are two collections of images (5,316 and 8,931, respectively), annotated with bounding boxes, taken from multiple sequences describing crowded scenarios having different characteristics, like viewpoints, weather conditions, and camera motions. They are suitable for detection and tracking tasks. The authors provided training and test subsets, but they released only the ground-truth labels for the former. The main difference between MOT20Det compared to MOT17Det is that the first contains more crowded scenarios.

\paragraph{CityPersons~\cite{citypersons_dataset}.} The CityPersons dataset consists of a set of stereo video sequences recorded from a moving car in the streets of different cities in Germany and neighboring countries. In particular, the authors provide 5,000 frames from 27 cities labeled with bounding boxes and split across train/validation/test subsets.

\paragraph{CrowdHuman~\cite{crowdhuman_dataset}.} CrowdHuman is a benchmark dataset for pedestrian detection. It comprises 15,000, 4,370, and 5,000 images for training, validation, and testing, respectively, describing diverse, crowded scenarios, with an average number of persons in an image of 22.6. The authors annotated each human instance with a head bounding box, a human visible-region bounding box, and a full-body bounding box. 

\paragraph{PRW~\cite{prw_dataset}.} The PRW dataset contains 11,816 frames where 932 different pedestrian identities are annotated with their bounding boxes. The authors provide the training and the test splits.

\paragraph{CUHK-SYSU~\cite{cuhk-sysu_dataset}.} The CUHK-SYSU is a large-scale benchmark dataset containing 18,184 images, 8,432 different identities, and 96,143 pedestrian bounding boxes. It is divided into training and test subsets.

\paragraph{CrowdVisorPisa.} The CrowdVisorPisa dataset is a novel collection of images that we collected and annotated on purpose for this dissertation. In particular, we stored 15 different sequences gathered from a webcam located in a public square of the city of Pisa, Italy, each of which comprises ten images captured with a time interval of 1 second. We manually labeled all frames, localizing the pedestrian instances with bounding boxes. Furthermore, we also annotated each sequence taking track of the different pedestrian entities entering or exiting the scene. We divided the dataset into train and test splits, considering 10 and 5 different sequences. The former split is exploited to train the object detector module, while the latter is to evaluate the performance of some modules of our framework. It is worth noting that, due to camera positioning not modifiable for local restrictions, this dataset represents a particularly challenging scenario as people instances are small and sometimes difficult to localize. A more detailed description is present in \ref{ch:counting-for-covid}.

\paragraph{CrowdVisorPPE.} The CrowdVisorPPE is a new dataset introduced in this thesis comprising 54,017 images representing pedestrians with and without wearable \acrlong{ppe}. Roughly half of the dataset are synthetic images procedurally generated using the GTA V video game engine as in~\cite{ppe_detection}, whereas the other half comprises real-world photographic images taken from the Web and manually annotated. The \acrshort{ppe} classes of interest, i.e., helmets, high-visibility vests (HVVs), and face masks, are annotated with bounding boxes. A more detailed description is present in \ref{ch:counting-for-covid}.

\paragraph{PKLot~\cite{pklot_dataset}.} The PKLot dataset contains 12,416 images of parking lots extracted from surveillance camera frames. There are images of sunny, cloudy, and rainy days and the parking spaces are labeled as occupied or empty. All images were acquired at the parking lots of the Federal University of Parana (UFPR) and the Pontical Catholic University of Parana (PUCPR), both located in Curitiba, Brazil.

\paragraph{\acrfull{carpk}~\cite{HsiehLH17}.} The \acrshort{carpk} dataset is the first large-scale aerial dataset for counting cars in parking lots. It includes 989 and 459 training and test samples, respectively, each of resolution $720 \times 1,280$. The training images are taken from three different parking lot scenes, while the test set is taken from a fourth scene. The total number of car instances is 42,274 in the range [1,87] (i.e., from a minimum of one car up to a maximum of 87 cars in the whole lot) and in the test dataset is 47,500 in the range [2,188].

\paragraph{\acrshort{pucpr+}~\cite{HsiehLH17}.} The \acrshort{pucpr+} dataset is published in the same paper as the \acrshort{carpk} dataset. It is a subset of the PUCPR dataset \cite{pklot_dataset}, adapted by the authors for the counting task. It contains images captured using a fixed camera from a height of the 10\textsuperscript{th} floor of a building, which provides a slanted view of the parking lot. This dataset has 100 and 25 training and test samples, respectively. Images are taken under three different weather conditions (sunny, rainy, and cloudy), resulting in different illuminations of the scene. The total number of car instances in the training dataset is 1,299 in the range [0,331] and in the test dataset is 3,920 in the range [1,328].

\paragraph{CNRPark+EXT~\cite{AmatoCarOccupancy2016,AmatoCarOccupancy2017}.}
This dataset is a collection of images collected from the parking lots of the National Research Council (CNR) campus in Pisa. It is comprised of roughly 150,000 images of parking spaces taken from nine smart cameras with different viewpoints, on different days with different weather and light conditions, and includes occlusion and shadow situations that make the occupancy detection task challenging. It is employed for the lot occupancy detection task, and each image represents a parking space that can be vacant or occupied.

\subsection{Per-pixel Annotated Datasets}
This class of datasets is characterized by labels involving the pixels of the images. In particular, \textit{instance segmentation} represents the more informative annotation type, combining pixel-wise classification of the entire image and the localization of the single detected object instances. However, this great amount of information does not come for free. Indeed, instance segmentation is the more costly and time-consuming labeling procedure in terms of human effort. In the following, we briefly illustrate the datasets belonging to this category used in this dissertation.

\paragraph{\acrfull{coco}~\cite{lin2014microsoft}.}
The Microsoft \acrfull{coco} dataset is a collection of images designed to represent a vast array of objects that we regularly encounter in everyday life. It provides suitable data for several computer vision tasks, like large-scale object detection, segmentation, and captioning, comprising 883,331 labeled object instances belonging to 80 common object categories in 121,408 images. In particular, each image is associated with an instance-level pixel-wise segmentation of the object instances it depicts, together with captions describing the whole scene. This dataset is considered the gold standard benchmark for evaluating the performance of state-of-the-art computer vision models. Still, it also provides a solid base dataset to train models that can be then fine-tuned by exploiting custom datasets, to learn other tasks.  

\paragraph{\acrfull{ndispark}.} This dataset was created on purpose in this thesis and made publicly available at \href{https://ciampluca.github.io/unsupervised\_counting/}{https://ciampluca.github.io/unsupervised\\\_counting}. It is a small, manually annotated dataset for counting cars in parking lots, consisting of about 250 images. This dataset is challenging and describes most of the problematic situations we can find in a real scenario: seven different cameras capture the images under various weather conditions and viewing angles. Another challenging aspect is the presence of partial occlusion patterns in many scenes, such as obstacles (trees, lampposts, other cars) and shadowed cars. Furthermore, it is worth noting that images are taken during the day and the night, showing utterly different lighting conditions and that, unlike most counting datasets, the \acrshort{ndispark} dataset is precisely annotated with instance segmentation labels, allowing us to generate accurate ground truth density maps for the counting task since the size of the vehicles is well-known. A more detailed description is provided in \ref{ch:uda-counting}.

\paragraph{\acrfull{gta}.} The \acrshort{gta} dataset was created on purpose in this dissertation and made publicly available at \href{https://ciampluca.github.io/unsupervised\_counting/}{https://ciampluca.github.io/unsupervised\_counting}.
It is a vast collection of about 15,000 synthetic images of urban traffic scenes collected using the highly photo-realistic graphical engine of the GTA V - Grand Theft Auto V video game. About half of them concern urban city areas, while the remaining involve sub-urban areas and highways. To generate this dataset, we designed a framework that \textit{automatically} and precisely annotates the vehicles present in the scene with per-pixel annotations. As in the \acrshort{ndispark} dataset, the instance segmentation labels allow us to produce accurate ground truth density maps for the counting task since the size of the vehicles is well-known. A more detailed description is provided in \ref{ch:uda-counting}.

\graphicspath{{img/counting-on-the-edge/}}

\chapter{Counting Vehicles in Embedded Vision Systems}
\label{ch:counting-on-the-edge}

Traffic-related issues are constantly increasing, and tomorrow's cities cannot be considered intelligent if they do not enable smart mobility. 
This concept is becoming more critical since road congestion is growing, caused by the increasing number of people traveling anywhere. 
Smart mobility applications, such as smart parking and road traffic management, are widely employed worldwide, making our cities more livable, benefiting our lives, reducing costs, and improving energy usage.
The ubiquity of video surveillance cameras in modern cities and the significant development of \acrlong{ai} provide new opportunities for the development of such helpful applications and services for citizens.

Indeed, images are perhaps the best sensing modality to perceive and assess the flow of vehicles in large areas. Like no other sensing mechanism, city camera networks can monitor large areas while simultaneously providing visual data to \acrshort{ai} systems to extract relevant information from this deluge of data. However, this application is often hampered by the limited computational resources on disposable devices. Indeed, \acrlong{dl}-based solutions, including \acrlong{cnn}s, often require a considerable amount of computational resources. To automatically learn a functional hierarchy of features, these models are defined as \textit{deep}, i.e., need to stack many parametric transformations (called layers). Consequently, the model requires a considerable computational and memory budget for the training and the evaluation, often limiting the applications of this kind of solution in restricted environments with limited power resources, such as IoT devices, delegating complex data analysis to a centralized server.


In this chapter, we explore the adoption of \acrfull{dl}-based solutions for counting vehicles directly onboard embedded vision systems, i.e., devices equipped with limited computational capabilities that can capture images, process them extracting some knowledge, and eventually communicate with other devices sending the elaborated information. In particular, we firstly propose a \acrshort{dl} solution to automatically detect and count vehicles in images taken from a \acrfull{uav}. We experimented over two real-world datasets showing that our approach results in state-of-the-art performances, running at a speed of 4 \acrfull{fps} on an NVIDIA Jetson TX2 board. Then, in another research activity, we introduce a novel vehicle counting solution considering a different scenario, consisting of a parking area monitored by multiple smart cameras. Indeed, in many real-world scenarios, one can benefit from using multiple cameras to monitor the same parking lot from different perspectives and viewpoints. Furthermore, multiple neighboring cameras can also be helpful to cover a wider area. At the same time, such a setting introduces issues related to merging the knowledge extracted from the single cameras with partially overlapping \acrlong{fov}s. To this end, we propose a multi-camera system that combines a \acrshort{cnn} running onboard the smart cameras, which can locate and count vehicles present in images belonging to the individual \acrshort{fov}s, along with a decentralized approach, again running directly on the disposable devices, that is responsible for analyzing and merging these results, exploiting the geometry of the captured images to estimate the number of vehicles present in the overlapped \acrlong{fov}s, and automatically outputs the number of cars present in the \textit{entire} parking area.

We organize the chapter into two distinct parts. In the first one, corresponding to \ref{sec:counting-on-the-edge:counting-drones}, we describe the vehicle counting solution that analyzes images captured by a \acrfull{uav}. In particular, we introduce the topic, we review some works related to our, we describe the proposed method, and we show the experiments and the obtained results. In the second part of the chapter, corresponding to \ref{sec:counting-on-the-edge:multi-camera-counting}, we instead introduce the decentralized multi-camera counting solution, we review some related works, we explain our proposed method, and finally, we show the experimental results.

The research presented in this chapter was published in \cite{ciampi_sebd, ciampi_globecom, ciampi_iscc, ciampi_multi_camera}.

\section{Counting Vehicles in Onboard UAV Imagery}
\label{sec:counting-on-the-edge:counting-drones}

\subsection{Introduction}
\label{sec:counting-on-the-edge:counting-drones:intro}
\begin{figure}
\centerline{\includegraphics[width=.90\textwidth]{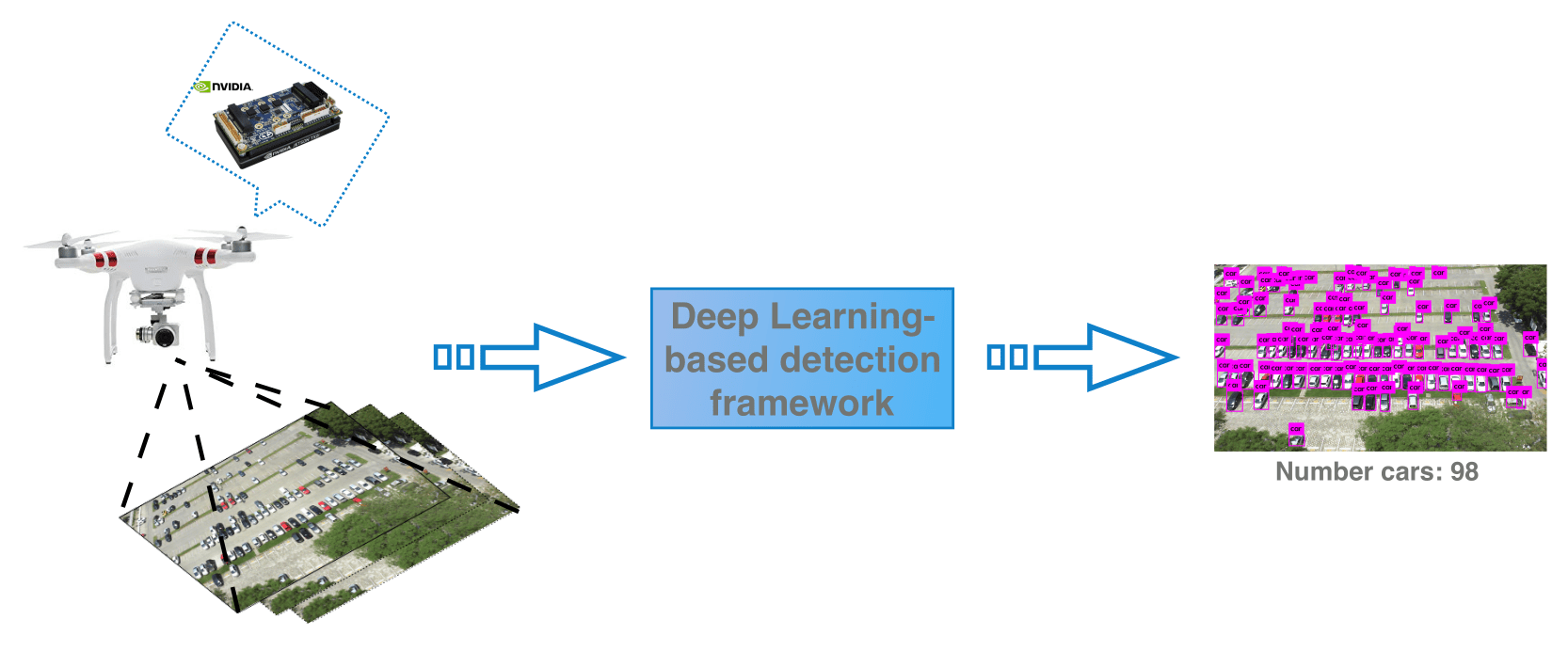}}
\caption{\textbf{Overview of the system architecture}. We localize and counting vehicles from \acrshort{uav} imagery, directly onboard drones equipped with low-power vision embedded systems.
}
\label{fig:drone_architecture}
\end{figure}

With the advent of \acrlong{uav}s, new potential applications emerge for unconstrained images and videos analysis for aerial view cameras, ranging from agriculture to security. Visual tasks like object detection, classification, or segmentation are the essential building blocks for many of them.
In this section, we address the counting problem for evaluating the number of vehicles present in parking lots from \acrshort{uav} imagery, i.e., images of parking areas taken from a drone view. In this scenario, the cars appear in various orientations, often within the same scene, and can be partially occluded by trees or bridges. Other challenges are represented by clutters caused by motorbikes, buildings, or different light conditions. To this end, we propose a real-time counting solution to be used directly onboard drones equipped with low-power vision embedded systems. Specifically, we relied on an existing CNN-based detection framework that we re-trained, specializing it for this specific task and making it more suitable for working in scenarios constrained by limited computational capabilities. We evaluated the effectiveness and the reliability of our solution, testing it on two publicly available car counting datasets: the \textit{CARPK} dataset and the \textit{PUCPR+} dataset \cite{HsiehLH17}. In both cases, we achieved state-of-the-art counting errors. For our tests, we used an NVIDIA Jetson TX2 Developer Kit, a power-efficient embedded \acrshort{ai} computing device running at 4 Frames Per Second (FPS). An overview of the system architecture is shown in \ref{fig:drone_architecture}

\subsection{Related Works}
\label{sec:counting-on-the-edge:counting-drones:related}
In this section, we briefly review some works related to our and with which we have compared the obtained results.
Prior methods for monitoring the parking lot often assume that the locations of the monitored objects of a scene, e.g., the lots, are already known in advance, and the cameras are fixed. In this setting, the counting task is cast as a classification problem in which the network analyzes the single regions of images corresponding to the lots \cite{DBLP:conf/cvpr/AhrnbomAN16, AmatoCarOccupancy2016, AmatoCarOccupancy2017, pklot_dataset}. However, these car counting methods are not directly applicable in unconstrained drone videos. As stated in \ref{sec:back:visual-counting}, most of the existing counting approaches can be classified into two supervised learning-based categories: i) \textit{counting by regression} that tries to establish a direct mapping (linear or not) from the image features to the number of objects present in the scene or a corresponding density map (i.e., a continuous-valued function), and ii) \textit{counting by detection} where the goal is to localize instances of the objects and then count them.

In \cite{HsiehLH17} the authors proposed a detection-based solution, introducing a novel Layout Proposal Network (LPN) that counts and localizes vehicles in drone images, leveraging the spatial layout information (e.g., cars often park regularly). Furthermore, they also presented the CARPK and the PUCPR+ datasets, two collections of images suitable for the vehicle counting task that we exploited to evaluate the performance of our solution. In \cite{large_contextual_dataset} the authors presented instead ResCeption, a network that combines residual learning with Inception-style layers \cite{inception} and is able to count vehicles in one look. The authors of \cite{class_agnostic} proposed an approach that claims to be able to count objects belonging to a generic class not decided a priori, i.e., a class-agnostic counting network. To this end, the network must be adapted using very few annotated data. They achieved this by recasting the counting problem as an image matching problem, where counting instances is performed by matching (self-similar patches) within the same image. This solution is tested by counting bacterial cells, people and vehicles. Aich et al. in \cite{improving_with_heatmap} proposed a simple and effective way to improve one-look regression models for object counting from images, enhancing them by regulating activation maps from the final convolution layer of the network with coarse ground-truth activation maps generated from simple dot annotations. They called this strategy Heatmap Regulation (HR). These mentioned solutions are not designed to work on embedded real-time devices due to time constraints and the limited available computing resources. One prior method able to perform onboard computations is described in \cite{shuffledet}, where the authors presented a detection-based solution computationally inexpensive for vehicle detection in \acrshort{uav} imagery.

\subsection{Method}
\label{sec:counting-on-the-edge:counting-drones:method}
We propose a vehicle counting solution from UAV imagery based on detections, i.e., first, we localize the cars, and then we count the found instances. We relied on \textit{YOLOv3} (You Only Look Once) \cite{yolo_v3}, a popular deep \acrlong{fcn}, employed in many detection framework, that we described in \ref{sec:back:cnn-based-detectors:single-stage-detectors}. In a nutshell, unlike previous detectors based on a sliding-window classifier or region proposals, \acrshort{yolo} addresses the detection problem by exploiting the \textit{one-look} paradigm, achieving higher inference speed, and making it more suitable for working in limited computation capabilities scenarios.

\begin{figure}
\centerline{\includegraphics[width=.90\textwidth]{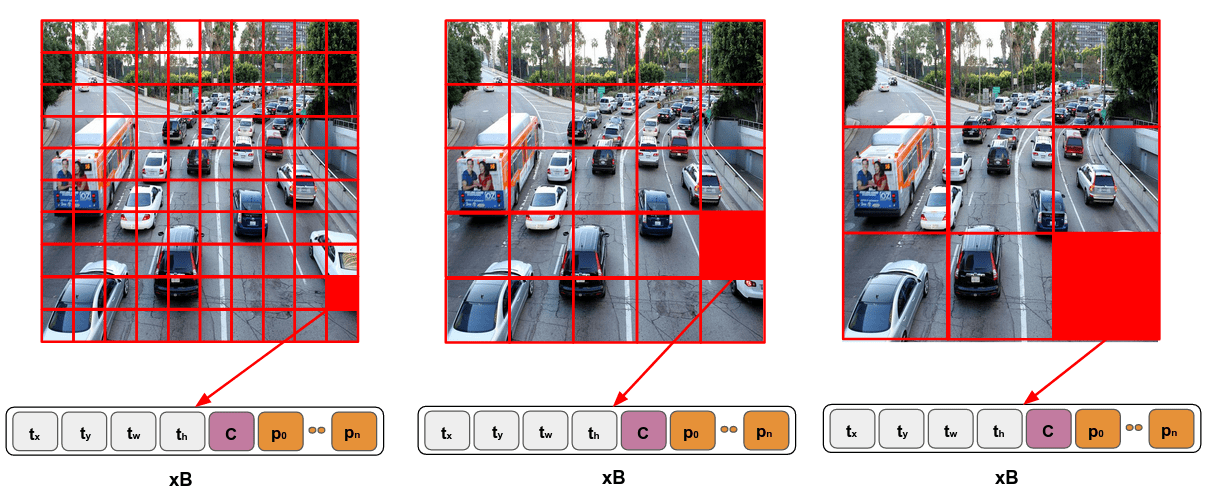}}
\caption{\textbf{Output vectors produced by the \acrshort{yolo}v3 network.}\acrshort{yolo}v3 divides the input images in $S=3$ different regular grids, making detections at different scales. For each cell of a grid it predicts $B=3$ bounding boxes each of them encoded in a vector of $(c, t_x, t_y, t_w, t_h) \times (p_i, ..., p_{n})$ features, where $c$ is the objecteness of the associated bounding box, $t_x, t_y$ are the features related to the offsets of the bounding box center w.r.t. a corner of the grid cell, $t_w, t_h$ are the features related to the dimensions of the bounding box and $(p_i, ..., p_{n})$ are conditional probabilities, $P(Class|Object)$, for each of the considered $n$ object classes. At the end, YOLO outputs a tensor of dimensions $S_i \times S_i \times (B \times (n + 5))$, where $n$ is the number of object classes and $S_i$ are the three different grids.}
\label{fig:yolo_scales}
\end{figure}

\acrshort{yolo}v3, as well as its predecessors, divides the input images using a regular $S \times S$ grid. For each cell of the grid, it predicts $B=3$ bounding boxes, each of them encoded in a vector of $(c, t_x, t_y, t_w, t_h) \times (p_i, \dots, p_{num\_classes})$ features, where $c$ is the objectness of the associated bounding box, i.e., a sort of a confidence measure of how likely a box is containing, or not, an object, $t_x, t_y$ are the features related to the offsets of the bounding box center w.r.t. a corner of the grid cell, $t_w, t_h$ are the features related to the dimensions of the bounding box and $(p_i, \dots, p_{num\_classes})$ are conditional probabilities, $P(Class|Object)$, for each of the considered object classes. So, in the end, the output of the network is a prediction consisting of a tensor of dimensions $S \times S \times (B \times (5 + num\_classes))$. Moreover, \acrshort{yolo}v3, differently from the previous versions, uses $3$ of these regular grids as outputs, making detections at different scales and so predicting three of these tensors at three different depths of the architecture. \ref{fig:yolo_scales} shows an example of these three grids applied to a sample image. In particular, to perform these computations, \acrshort{yolo}v3 uses a deep custom architecture that acts as the backbone for the features extraction, called \textit{darknet-53}, which is made up of 53 layers. The authors stacked onto it 53 additional layers, building an architecture of 106 layers able to perform the detection task. The three detections at the three different scales are made by the 8{2\textsuperscript{nd}}, the 9{4\textsuperscript{th}} and the 10{6\textsuperscript{th}} layer, respectively, by applying $1 \times 1$ detection kernels on the feature maps. 

In this work, we considered as a starting point a model of \acrshort{yolo} pre-trained on the COCO dataset \cite{lin2014microsoft}, a large collection of images describing complex everyday scenes of common objects in their natural context, categorized in 80 different categories. Since this network is a generic objects detector, we specialized it to localize and recognize object instances belonging to only a specific category - i.e., the car category in our case. Specifically, we extracted the weights of the first 81 layers of this pre-trained model since these layers capture universal features (like curves and edges), which are also relevant to our task. Then, we re-trained the detector by initialing the first 81 layers with the previously extracted weights and the weights associated with the remaining layers at random, exploiting annotated data describing our specific scenario for the supervised learning. Thus, we got the network to focus on learning dataset-specific features in the last layers. Furthermore, in such a way, we reduced the dimension of the predicted output tensors, from  $S_i \times S_i \times (B \times (80 + 5)) = S_i \times S_i \times 255$ to $S_i \times S_i \times (B \times (1 + 5)) = S_i \times S_i \times 18)$, where $S_i$ are the three different grids exploited by \acrshort{yolo}, saving memory usage and making our solution more suitable for the limited computational capabilities of the embedded vision system employed in our testing scenario.

\subsection{Experimental Evaluation}
\label{sec:counting-on-the-edge:counting-drones:esperiments}
This section illustrates the employed experimental setup, the performed experiments, and, finally, the obtained results. We start by describing the exploited datasets and the testing scenario. Then we discuss the results compared with the state-of-the-art ones. Following other counting benchmarks, we used the \textit{\acrfull{mae}} and the \textit{\acrfull{rmse}} as the performance metrics (see also \ref{sec:back:visual-counting:metrics} for more details). Furthermore, since we rely on a detection-based approach, we also exploit the \textit{Precision} and \textit{Recall} metrics to take into account the quality of the localization (see also \ref{sec:back:cnn-based-detectors:metrics}).

\subsubsection{Datasets}
\begin{figure}[htbp]
    \centering
  \subfloat{
       \includegraphics[width=0.45\linewidth]{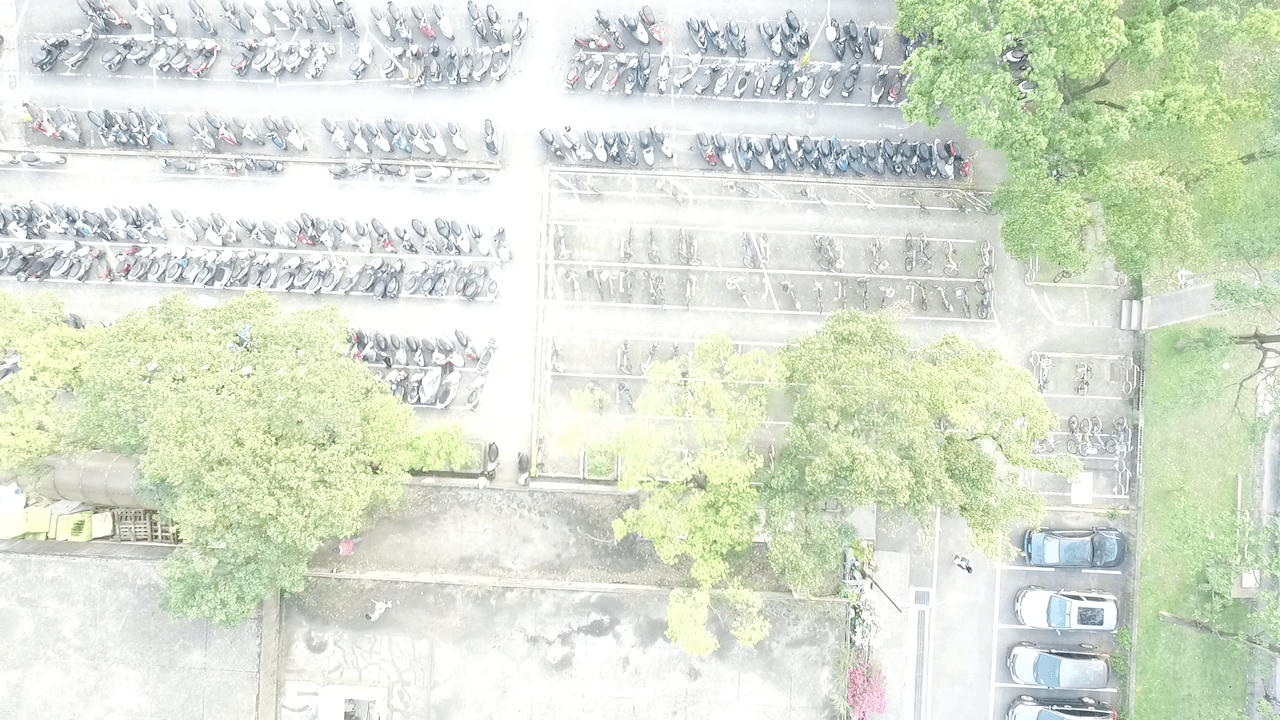}}
  \subfloat{
        \includegraphics[width=0.45\linewidth]{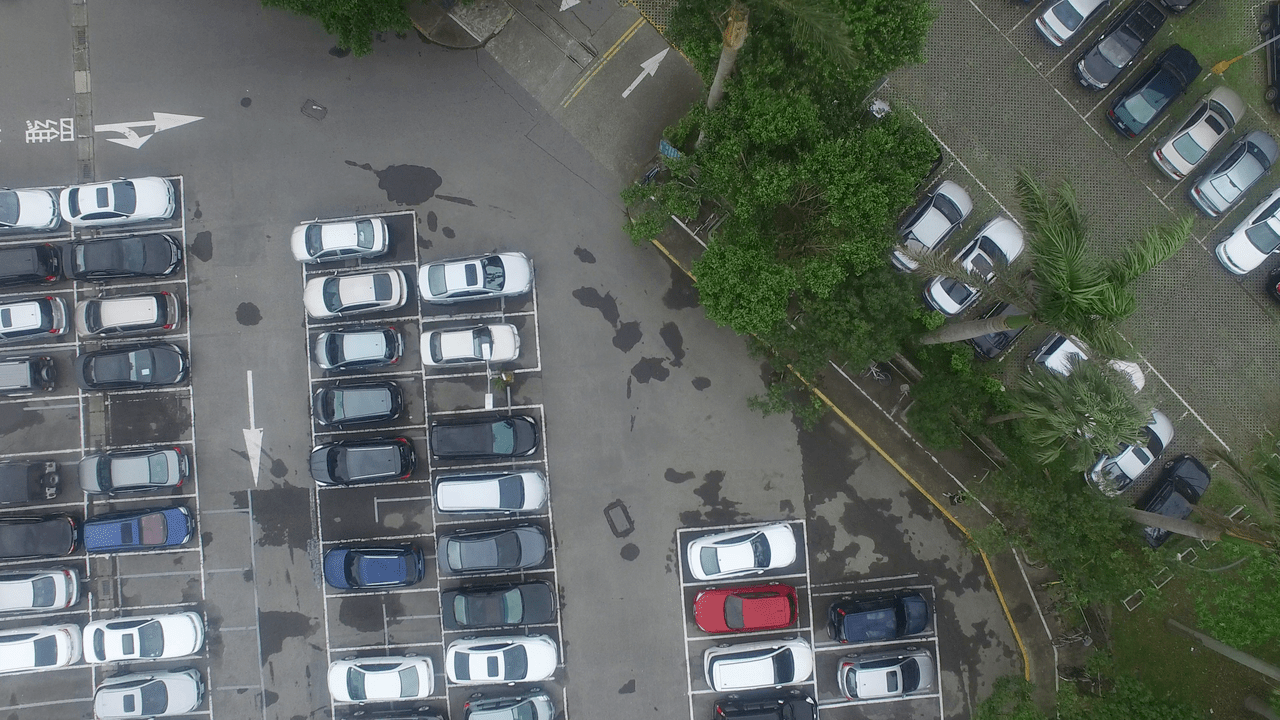}}
 \hfill
    \\ [1.1ex]
  \subfloat{
        \includegraphics[width=0.45\linewidth]{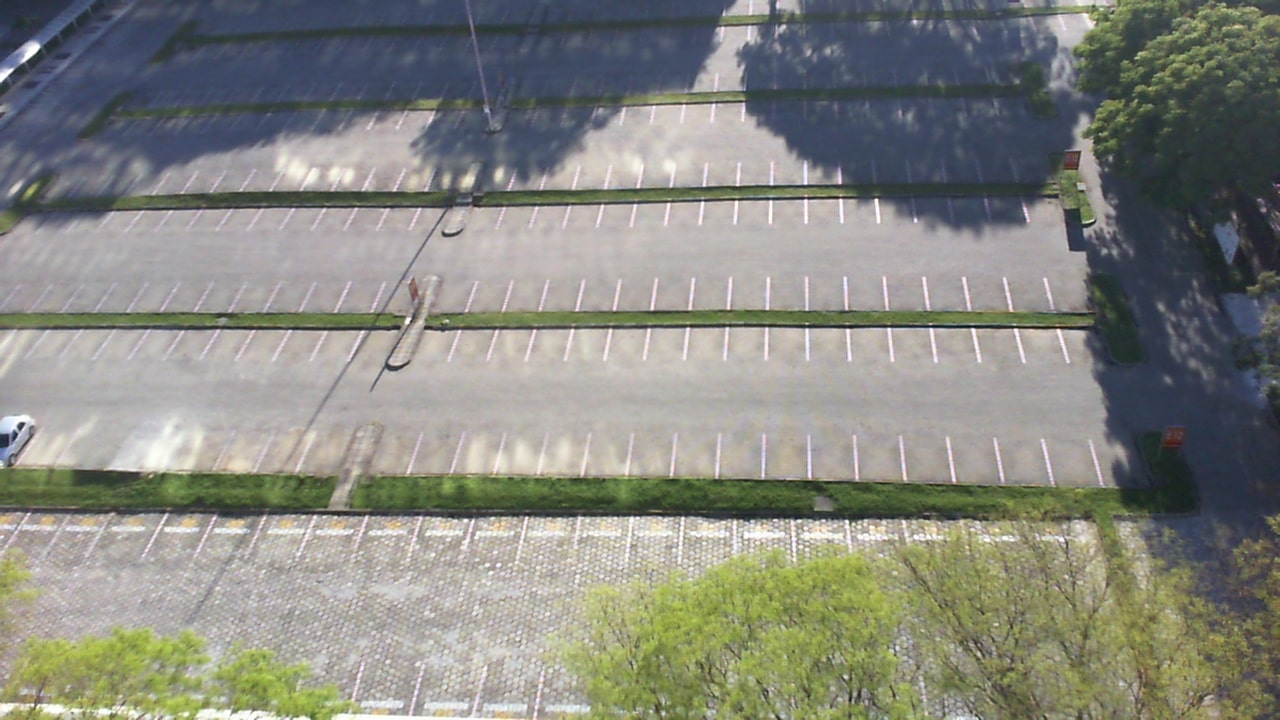}}
  \subfloat{
        \includegraphics[width=0.45\linewidth]{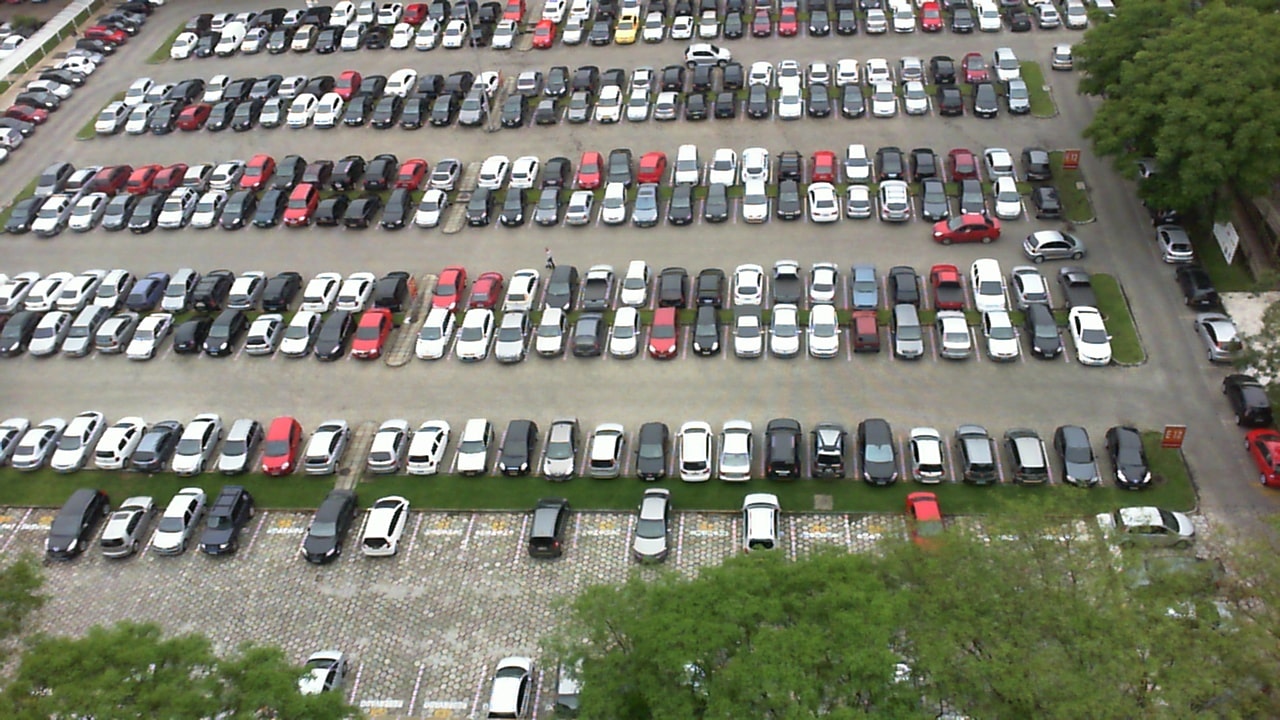}}
 \hfill
\caption{\textbf{Examples from the CARPK (top row) and PUCPR+ (bottom row) datasets.} The CARPK dataset is an aerial dataset taken from a drone, while the PUCPR+ dataset records a parking lot from a fixed camera located at the 10\textsuperscript{th} floor of a building. We exploit them to train and evaluate our vehicle counting solution.}
\label{fig:carpk_pucprplus_example} 
\end{figure}

To train and evaluate our vehicle counting solution, we exploited two public collections of images widely employed in the literature: the \textit{CARPK} and the \textit{PUCPR+} datasets, introduced in \cite{HsiehLH17}. The CARPK dataset is the first large-scale aerial dataset for counting cars in parking lots. It includes 989 and 459 training and test samples, respectively, each having resolution of $720 \times 1,280$. The training images are taken from three different parking lot scenes, while the test set is taken from a fourth scene. The total number of car instances is 42,274 in the range [1,87] (i.e., from a minimum of one car up to a maximum of 87 vehicles in the whole lot) and in the test dataset is 47,500 in the range [2,188]. The PUCPR+ dataset is published in the same paper as the CARPK dataset. It is a subset of the PUCPR dataset introduced in \cite{pklot_dataset}, adapted by the authors for the counting task. It contains images captured using a fixed camera from a height of the 10\textsuperscript{th} floor of a building, which provides a slanted view of the parking lot. This dataset has 100 and 25 training and test samples, respectively. Images are taken under three different weather conditions (sunny, rainy, and cloudy), resulting in different scene illuminations. The total number of car instances in the training dataset is 1,299 in the range [0,331] and in the test dataset is 3,920 in the range [1,328]. Some sample images from the dataset are shown in \ref{fig:carpk_pucprplus_example}.

\subsubsection{Experimental Scenario}
We used the darknet implementation of YOLO \footnote{https://pjreddie.com/darknet/}. We trained the network by exploiting the training subsets of the CARPK and PUCPR+ datasets, obtaining two different models specific for the two considered scenarios. Specifically, we picked up the best models in terms of \acrshort{mae} over the validation splits. Concerning the CARPK dataset, we obtained the validation subset by further dividing the training set, considering images belonging to one of the three different scenes. On the other hand, we adopted a 5-fold cross-validation for the PUCPR+ dataset since it is very small.
For the training phase, we exploited an NVIDIA GeForce RTX 2080 Ti GPU for 150 epochs, using \acrfull{sgd} for the optimization with a base learning rate of 0.001, momentum 0.9, and a weight decay of 0.0005. We employed a data augmentation strategy over the training images to prevent overfitting, randomly changing saturation, exposure (brightness), and hue (color), applying a blurring filter, and changing the size and aspect ratio.

We evaluated our final solutions by testing them over the test subsets of the CARPK and PUCPR+ datasets, also performing cross-dataset experiments, i.e., testing the model trained on CARPK over PUCPR+, and vice-versa. For this phase, we exploited an NVIDIA Jetson TX2 Developer Kit as an embedded vision device, simulating an autonomous system located onboard a \acrshort{uav}. This vision board is equipped with two 64bit CPUs having two and four cores each, an NVIDIA Pascal GPU with 256 CUDA cores, 4 GB of RAM shared between the system and the graphics accelerator and a 32 GB solid-state storage volume, together with a Camera Module able to capture images. The operating system is the NVIDIA’s Linux4Tegra(L4T) distribution based on Ubuntu. The operational autonomy can be guaranteed in a real scenario using, for example, a 3S 6000 mAh LiPo battery that provides a six-hour of autonomy, as stated in \cite{blanco2018deep}. 

\subsubsection{Results and Discussion}
\begin{figure*}
    \centering
  \subfloat{
       \includegraphics[width=0.45\linewidth, height=3.4cm]{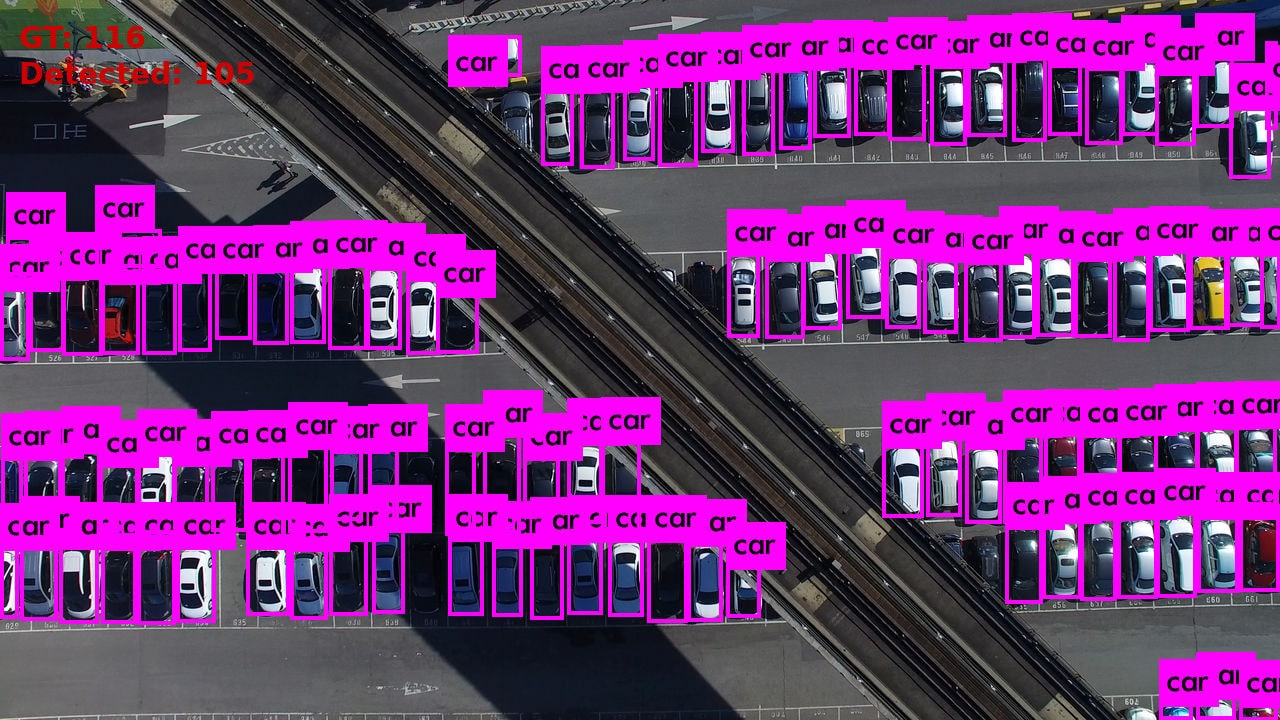}}
  \subfloat{
        \includegraphics[width=0.45\linewidth, height=3.4cm]{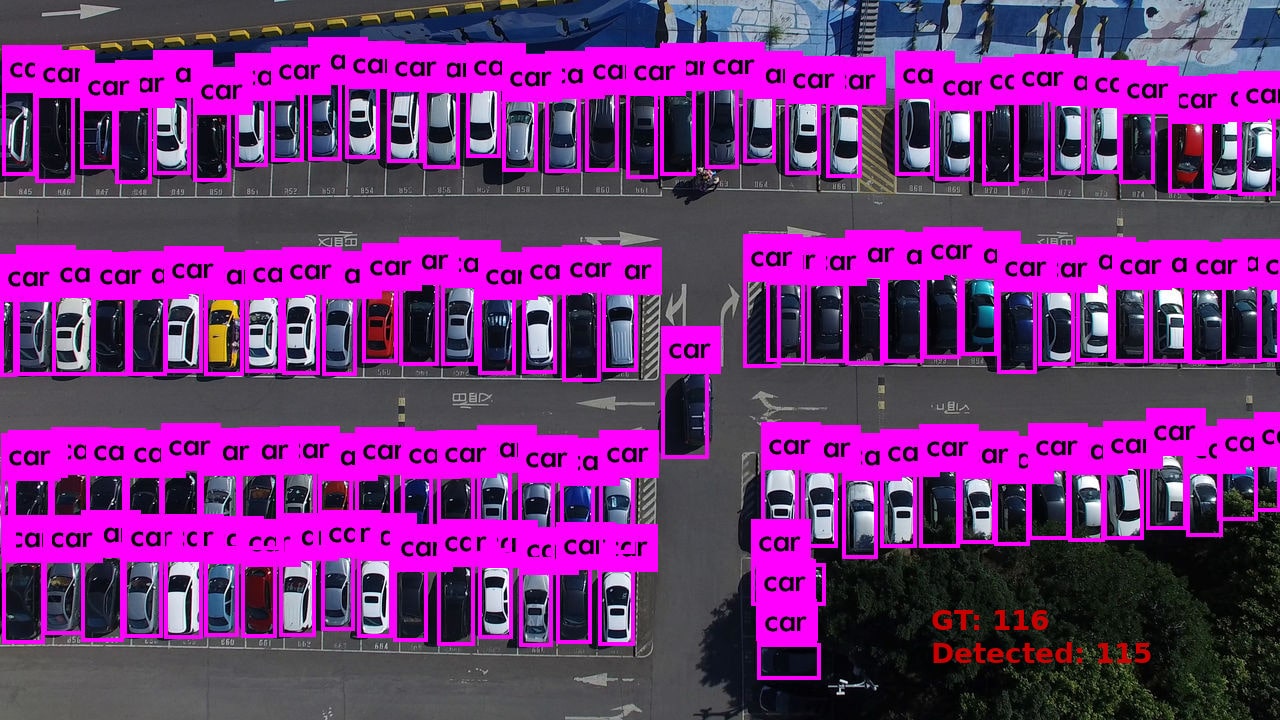}}
 \hfill
    \\ [0.5ex]
  \subfloat{
        \includegraphics[width=0.45\linewidth, height=3.4cm]{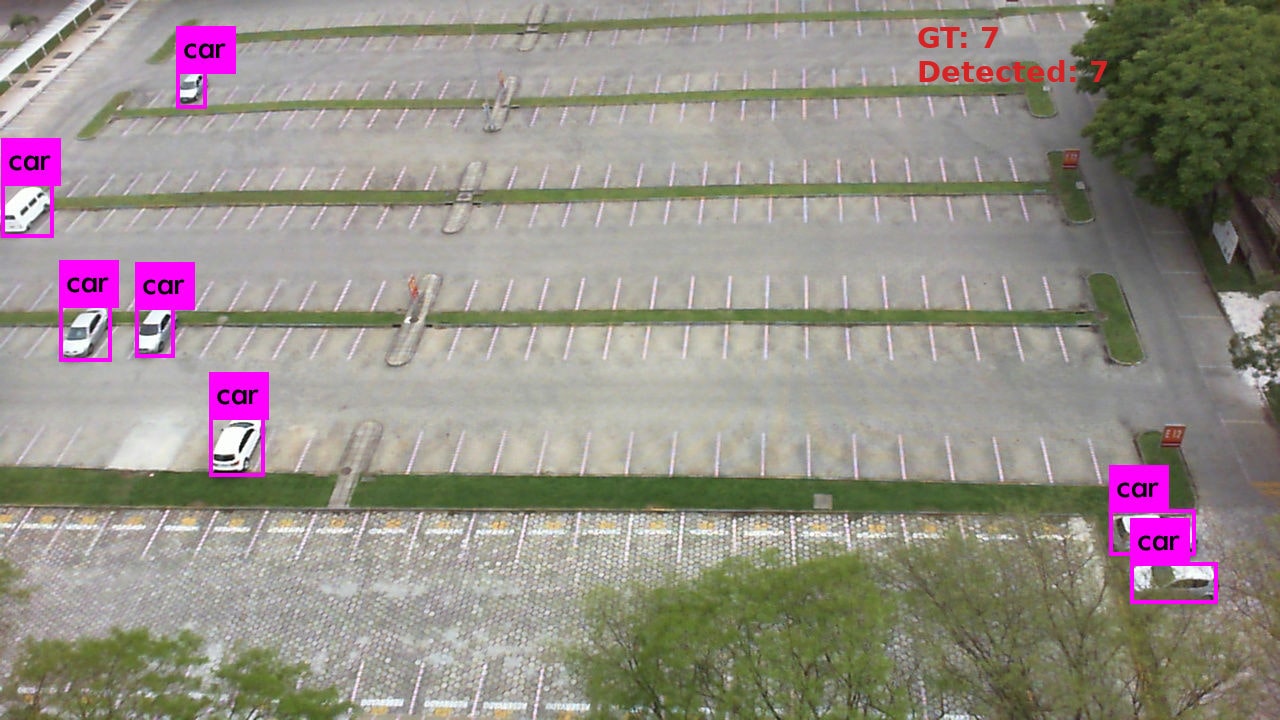}}
  \subfloat{
        \includegraphics[width=0.45\linewidth, height=3.4cm]{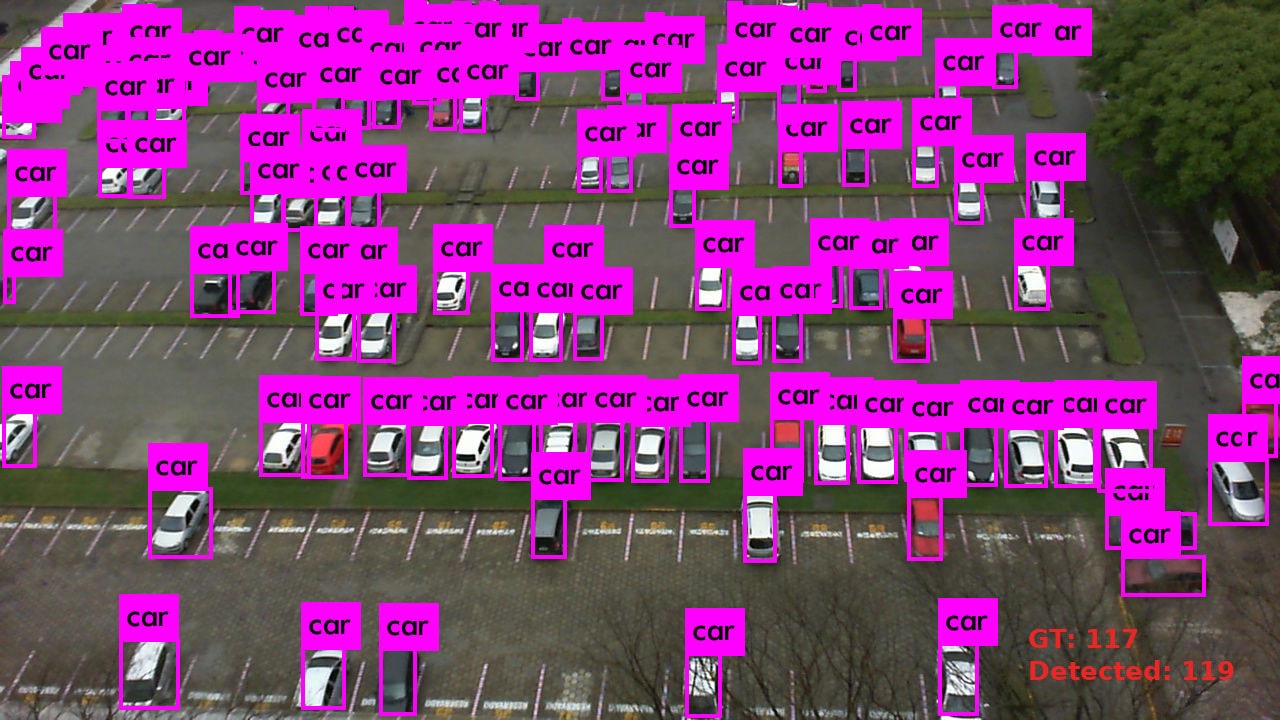}}
  \caption{\textbf{Some examples of the obtained results on the CARPK dataset (top row) and the PUCPR+ dataset (bottom row).} The detections are marked with a bounding box and a label indicating the object class. In addition, we report the total number of instances found together with the Ground Truth value.}
  \label{fig2} 
\end{figure*}

We report the obtained results in terms of \acrshort{mae}, \acrshort{rmse}, Precision, and Recall in \ref{tab1} and \ref{tab2}, concerning the CARPK and the PUCPR+ datasets, respectively, comparing the gained performance against other state-of-the-art approaches. On the other hand, \ref{fig2} shows some detection results.

\begin{table*}[t]
\caption{\textbf{State-of-the-art comparison on the CARPK test set (459 images and 47,500 total vehicles)}}
\begin{center}
\begin{tabular}{|c|c|c|c|c|}
\hline
\textbf{Method} & \textbf{MAE} & \textbf{RMSE} & \textbf{Recall} & \textbf{Precision} \\
\hline
One-Look Regression \cite{large_contextual_dataset, HsiehLH17} & 59.46 & 66.84 & - & - \\
\hline
Faster R-CNN \cite{faster_rcnn, HsiehLH17} & 47.45 & 57.55 & - & - \\
\hline
ShuffleDet \cite{shuffledet} & 26.75 & 38.46 & - & - \\ 
\hline
Faster R-CNN (RPN-small) \cite{faster_rcnn, HsiehLH17} & 24.32 & 37.62 & - & - \\
\hline
Spatially Regularized RPN \cite{HsiehLH17} & 23.80 & 36.79 & 57.5\% & - \\
\hline
Our solution $^*$ & 16.63 & 21.18 & - & - \\
\hline
GMN (3 images) \cite{class_agnostic} & 13.38 & 18.03 & 76.1\% & 85.1\% \\
\hline
VGG-GAP-HR \cite{improving_with_heatmap} & 7.88 & 9.30 & - & - \\
\hline
GMN (full dataset) \cite{class_agnostic} & 7.48 & 9.90 & 88.4\% & \textbf{91.8\%} \\
\hline
\hline
Our solution & \textbf{4.94} & \textbf{6.76} & \textbf{89.7\%} & 90.5\% \\
\hline
\end{tabular} \\[1ex]
* Trained using the PUCPR+ dataset
\label{tab1}
\end{center}
\end{table*}

\begin{table*}[t]
\caption{\textbf{State-of-the-art comparison on the PUCPR+ test set (25 images and 3,920 total vehicles)}}
\begin{center}
\begin{tabular}{|c|c|c|c|c|}
\hline
\textbf{Method} & \textbf{MAE} & \textbf{RMSE} & \textbf{Recall} & \textbf{Precision}\\
\hline
Faster R-CNN \cite{faster_rcnn, HsiehLH17} & 156.76 & 200.59 & - & -\\
\hline
ShuffleDet \cite{shuffledet} & 41.58 & 49.68 & - & -\\ 
\hline
Faster R-CNN (RPN-small) \cite{faster_rcnn, HsiehLH17} & 39.88 & 47.67 & - & -\\
\hline
Our solution $^*$ & 28.16 & 41.20 & - & - \\
\hline
One-Look Regression \cite{large_contextual_dataset, HsiehLH17} & 21.88 & 36.73 & - & -\\
\hline
Spatially Regularized RPN \cite{HsiehLH17} & 22.76 & 34.46 & 62.5\% & -\\
\hline
VGG-GAP-HR \cite{improving_with_heatmap} & 5.24 & 6.67 & - & -\\
\hline
\hline
Our solution & \textbf{2.80} & \textbf{4.53} & \textbf{95.3\%} & \textbf{95.2\%} \\
\hline
\end{tabular}
\\[1ex]
* Trained using the CARPK dataset
\label{tab2}
\end{center}
\end{table*}

We achieved state-of-the-art results in terms of counting errors in both datasets, also ensuring high Precision and Recall values. The Generic Matching Network (GMN) presented in \cite{class_agnostic} and tested on the CARPK dataset has the advantage of being class agnostic. However, the error obtained using only three training images is about three times larger than the one obtained with our approach. The error using the entire training dataset is instead about two times larger than the one we got, and in this case, the benefits of using a class agnostic solution are less self-evident. The VGG-GAP-HR solution \cite{improving_with_heatmap} has instead the advantage of needing a dot-annotated training dataset, while our solution needs bounding box labels. Dot-annotation is a less time-consuming operation than the bounding box one, but the error of our solution is about half for both datasets.
Finally, comparing our approach with the one proposed by \cite{shuffledet} in terms of images processed per second, it is notable that the algorithm in \cite{shuffledet} is faster than our (14 \acrshort{fps} instead of 4 \acrshort{fps}). Still, the error in \cite{shuffledet} is pretty larger in both the exploited benchmarks compared to ours.
Finally, concerning the cross-dataset experiments, we experienced a performance degradation due to the overfitting over a specific scenario; however, counting errors remain comparable with the ones obtained using most of the other approaches, thus showing capabilities by our solution to generalize to new contexts.  

\subsection{Summary}
\label{sec:counting-on-the-edge:counting-drones:conclusion}
In this first part of the chapter, we presented a useful tool based on a state-of-the-art object detector for evaluating the number of vehicles present in parking lots from \acrshort{uav} imagery. We tested our solution over two public datasets, exploiting an NVIDIA Jetson TX2 as an embedded vision device having limited computational capabilities that simulated an autonomous system located onboard a drone. We compared the obtained results against other approaches present in the literature, getting state-of-the-art performances.

\section{Multi-Camera Vehicle Counting in Parking Lots}
\label{sec:counting-on-the-edge:multi-camera-counting}

\subsection{Introduction}
\label{sec:counting-on-the-edge:multi-camera-counting:intro}
In this second part of the chapter, we tackle the problem of estimating the number of vehicles present in a parking lot using images captured by smart cameras. Whereas classic solutions for car counting or parking lot occupancy estimation are sensor-based (e.g., entrance-level photocells, per-space ground sensors), camera-based solutions provide several advantages, such as a) flexibility, as cameras can adapt to more challenging configurations of parking spaces (e.g., undelimited parking lots with non-fixed spaces), b) lower hardware and maintenance cost, as smart cameras can cost few tens of euros while each monitoring multiple parking spaces, and c) multi-purpose, as the same hardware can be used to perform additional tasks (e.g., surveillance). However, this vision-based counting task is challenging as the process of understanding the captured images faces many problems, such as shadows, light variation, weather conditions, and inter-object occlusions. 

\begin{figure}
\centerline{\includegraphics[width=.70\textwidth]{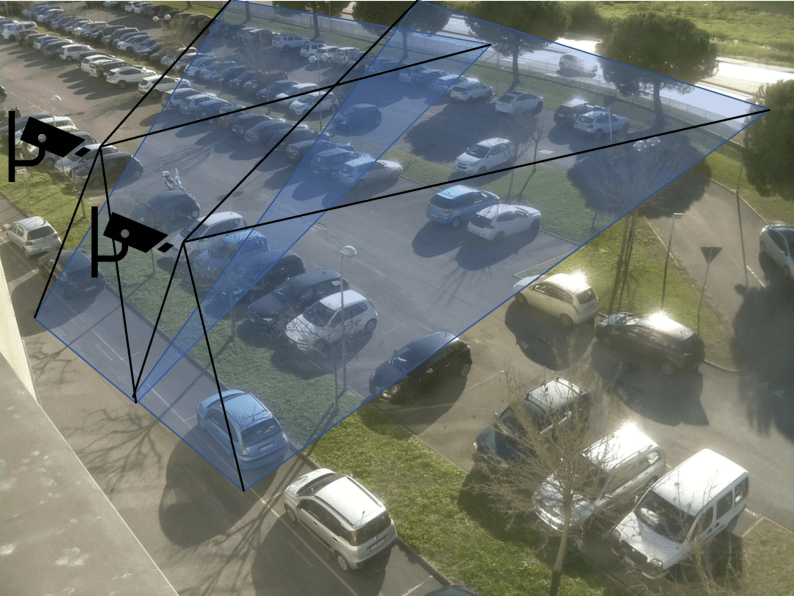}}
\caption{\textbf{An example of two cameras monitoring the same parking area with partially overlapping fields of view.}
This redundancy provides robustness and fault-tolerance, but it also raises the problem of aggregating knowledge extracted from the individual cameras.
}
\label{fig:multi_camera_example}
\end{figure}

Although most of the existing solutions concerning visual vehicle counting focus on the analysis of \textit{single} images, in many real-world scenarios, one can benefit from using multiple cameras to monitor the same parking lot from different perspectives and viewpoints. Furthermore, multiple neighboring cameras can also be helpful to cover a wider area. At the same time, such an approach introduces issues related to merging the knowledge extracted from the single cameras with partially overlapping \acrlong{fov}s (FOVs), as shown in \ref{fig:multi_camera_example}. To this end, we propose a novel solution to improve car counting when scaled up with multi-camera setups. We introduce a multi-camera system that combines a state-of-the-art \acrlong{cnn}, which can locate and count vehicles present in images belonging to individual cameras, along with a decentralized geometry-based approach that is responsible for aggregating the data gathered from all the devices and estimating the number of cars present in the \textit{entire} parking lot. A remarkable peculiarity of our solution is that it performs the task directly on the disposable devices, i.e., the smart cameras --- vision systems with limited computational capabilities able to capture images, extract information from them, make decisions, and communicate with other devices.  

We modeled our system as a graph where the nodes represent the smart cameras.
The total count is built exploiting the partial results computed in parallel by the single cameras and propagated through messages. Hence, our system scales better when the number of monitored parking spaces increases. Moreover, our solution does not require any extra information about the monitored parking area, such as the location of the parking spaces, nor any geometric information about the camera positions in the parking lot. In short, it is a flexible and ready-to-use solution that provides a simple ``plug-and-play'' insertion of new cameras into the system.

To validate our multi-camera solution, we employed and extended the \textit{CNRPark-EXT} dataset \cite{AmatoCarOccupancy2017}, a collection of images taken from the parking lot on the campus of the National Research Council (CNR) in Pisa, Italy. Images have been acquired by multiple cameras having partially overlapping fields of view and describing challenging scenarios with different perspectives, illuminations, weather conditions, and many occlusions. Since the annotations of this dataset concerned single images, we extended it by re-labeling a part of it to be consistent with our algorithm that instead considers the entire parking area. We conducted extensive experiments testing the generalization capabilities of the \acrshort{cnn} responsible for detecting vehicles in single images and the effectiveness of our multi-camera algorithm, demonstrating that our system is robust and benefits from the redundant information deriving from the different cameras improving the overall performance.

To summarize, the main contributions of this research are the following:
\begin{itemize}
\item we introduce a novel multi-camera system able to automatically estimate the number of cars present in the \textit{entire} monitored parking area. It runs directly on the disposable devices and combines a deep learning-based detector with a decentralized technique that exploits the geometry of the captured images;
\item we extend the \textit{CNRPark-EXT} dataset \cite{AmatoCarOccupancy2017}, a collection of images acquired by multiple cameras having partially overlapping fields of view and describing various parking lots, making it suitable with our scenario in which we consider the whole parking area;
\item we show through experiments that our system is robust, flexible, and can benefit from redundant information coming from different cameras while improving overall performance.
\end{itemize}

\subsection{Related Work}
\label{sec:counting-on-the-edge:multi-camera-counting:related}
Parking lot monitor using visual data is not new, and other works have already tackled it in the literature. In \cite{AmatoCarOccupancy2016,AmatoCarOccupancy2017}, the same authors of the CNRPark-EXT dataset presented a deep learning-based system for parking lot occupancy detection that can run in a Raspberry Pi directly on board a smart camera. In \cite{multi_camera_1}, the authors directly dealt with the issues deriving from the adoption of a multi-camera system. In particular, they applied a homography to project the detected vehicles from the plane of each camera to a common plane, where they performed a perspective correction to correct matching between the vehicle detections and the parking spots. Also, the authors in \cite{multi_camera_2} proposed a multi-camera system to classify parking spaces as vacant or occupied. In this solution, the acquired images are processed onboard Raspberry Pi devices. The extracted information about the status of parking spaces is then transmitted to a central server, which evaluates the parking spaces in the overlapping areas. Their algorithm is based on the histogram of oriented gradients (HOG)\cite{hog} feature descriptor and \acrfull{svm} classifiers. Since the HOG feature descriptor cannot adequately describe rotated vehicles, the authors have provided a descriptor with additional information about rotation to increase the system accuracy. However, these solutions rely on prior knowledge of the monitored scene, like the position of the parking spaces or some geometric information about the scene. In essence, a preliminary annotation of the new areas and a new training phase of the algorithm are often mandatory operations.
As a consequence, these techniques are not very flexible. On the other hand, we propose a simple yet effective solution that does not need extra information about the monitored scene. The smart cameras can automatically localize and count the vehicles present in their field of view, propagating the single results to the other devices through messages. A decentralized technique, again running directly onboard the devices, is instead in charge of analyzing and merging these results, exploiting the captured images geometry, and automatically outputs the number of cars present in the entire parking area.

\subsection{Method}
\label{sec:counting-on-the-edge:multi-camera-counting:method}

\subsubsection{Overview}

\begin{figure}[htbp]
\centering
  \begin{subfigure}{0.44\textwidth}
    \includegraphics[width=\textwidth]{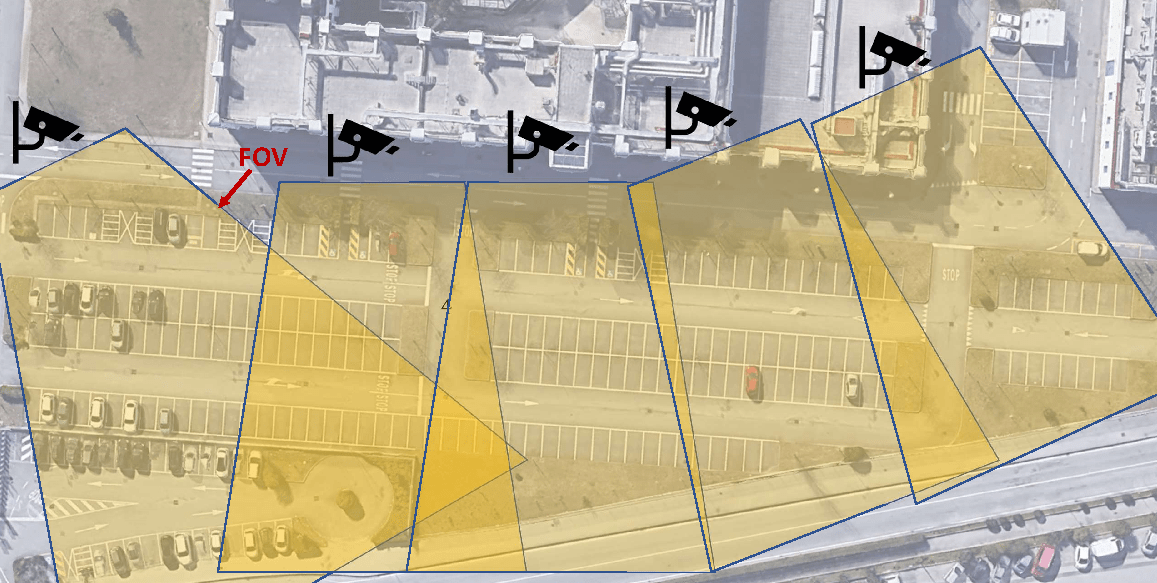}
    \label{system_overview_example}
  \end{subfigure} \hfill
  \begin{subfigure}{0.55\textwidth}
    \includegraphics[width=\textwidth]{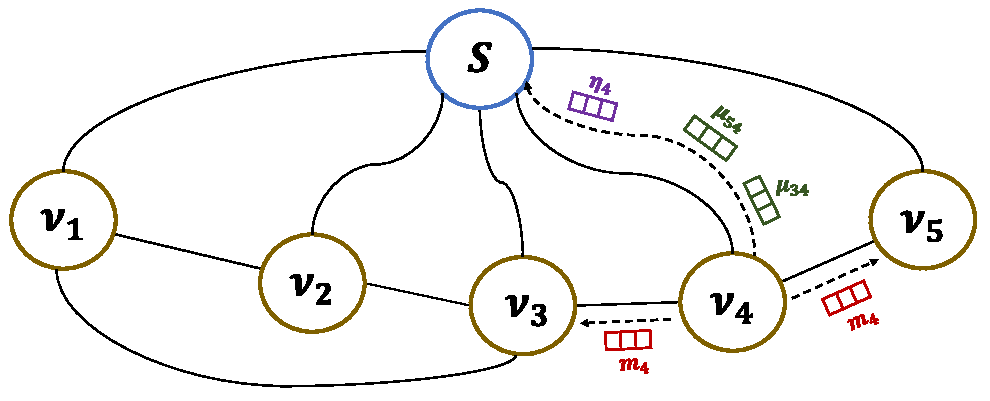}
    \label{system_model_example}
  \end{subfigure}
\caption{\textbf{An example of our multi-camera counting system, with $n=5$ smart cameras.} We model it as a graph $G$, comprised of $n$ nodes $\nu_i$ (one for each camera) and one Sink node $S$, $V = \{\nu_1, \nu_2, \cdots, \nu_n, S\}$. Each node $\nu_i$ can capture images, localize and count the vehicles present in its FOV, and communicate with its neighboring nodes through messages $m_i$ containing these detections. Moreover, each node $\nu_i$ can run a local counting algorithm in charge of computing partial counting results concerning the overlapped areas between its FOV and the ones belonging to its neighbors, exploiting images geometry. These partial results are sent through messages to the Sink node $S$, which is responsible for their fusion and provides the final result. Messages to $S$ can be of two types: i) $\eta_i$ containing the number of cars captured by the node $\nu_i$ in its FOV, and ii) $\mu_{j, i}$ representing the partial counting estimation related to the overlapping area between two neighboring nodes $\nu_i$ and $\nu_j$.}
\label{fig:system_overview}
\end{figure}

We base our multi-camera counting algorithm on the parallel processing of each of the smart cameras followed by the fusion of their results to estimate the number of vehicles present in the \textit{entire} parking area. \ref{fig:system_overview} shows an example of our multi-camera counting system, together with its graphical representation. We modeled our system as a graph $G$, comprised of $n$ nodes $\nu_i$ and one Sink node $S$, $V = \{\nu_1, \nu_2, \cdots, \nu_n, S\}$. Each node $\nu_i$ represents an independent device, i.e., a smart camera in our case. Two nodes $\nu_i$ and $\nu_j$ are considered neighbors if their \acrshort{fov}s overlap, and in this case, a directed edge of the graph connects them. Each device $\nu_i$ can capture images, localize and count the vehicles present in its \acrshort{fov} exploiting a deep learning-based detector, and communicate with its neighboring nodes through messages $m_i$ containing the cars detections. Furthermore, each node $\nu_i$ can also run a local counting algorithm in charge of computing partial counting results concerning the estimation of the number of vehicles present in overlapped areas between its \acrshort{fov} and the ones belonging to its neighbors. The fusion of the partial results is performed by the Sink node $S$, which is also in charge of providing the final result and synchronizing all the algorithm steps through synchronization signals headed towards the other nodes $\nu_i$. On the other hand, the nodes $\nu_i$ can also communicate through messages with the Sink node. Messages can be of two types: i) messages $\eta_i$ containing the number of cars captured by the node $\nu_i$ in its \acrshort{fov}, and ii) messages $\mu_{j, i}$ representing the partial counting estimation related to the overlapping area between two neighboring nodes $\nu_i$ and $\nu_j$.

\subsubsection{Initialization}

This step is aimed at \textit{automatically} initializing the system, estimating the geometric relationship between each node (i.e., each scene monitored by a smart camera) and its neighbors. The only hypotheses we impose are i) each smart camera is aware of the IP addresses of its neighbors, i.e., the cameras having the field of view overlapped with its own; ii) the Sink node $S$ is aware of the IP addresses of all the smart cameras belonging to the system.

The Sink node $S$ starts the initialization phase, sending a synchronization signal to the other nodes. Once received, each smart camera captures an image of the scene it monitors and sends it to all its neighbors. Once a smart camera $i$ receives an image from a neighboring camera $j$, it computes a homographic transformation $H_{j, i}$ between the image $j$ and the image $i$ describing its monitored scene. This allows us to establish a correspondence between the points belonging to the pair of images taken by the two cameras, which will be used subsequently in the algorithm. We formalized the system initialization for a generic node $\nu_i$ in the \ref{alg:system_init}.

\begin{figure}[htbp]
\centerline{\includegraphics[width=.80\textwidth]{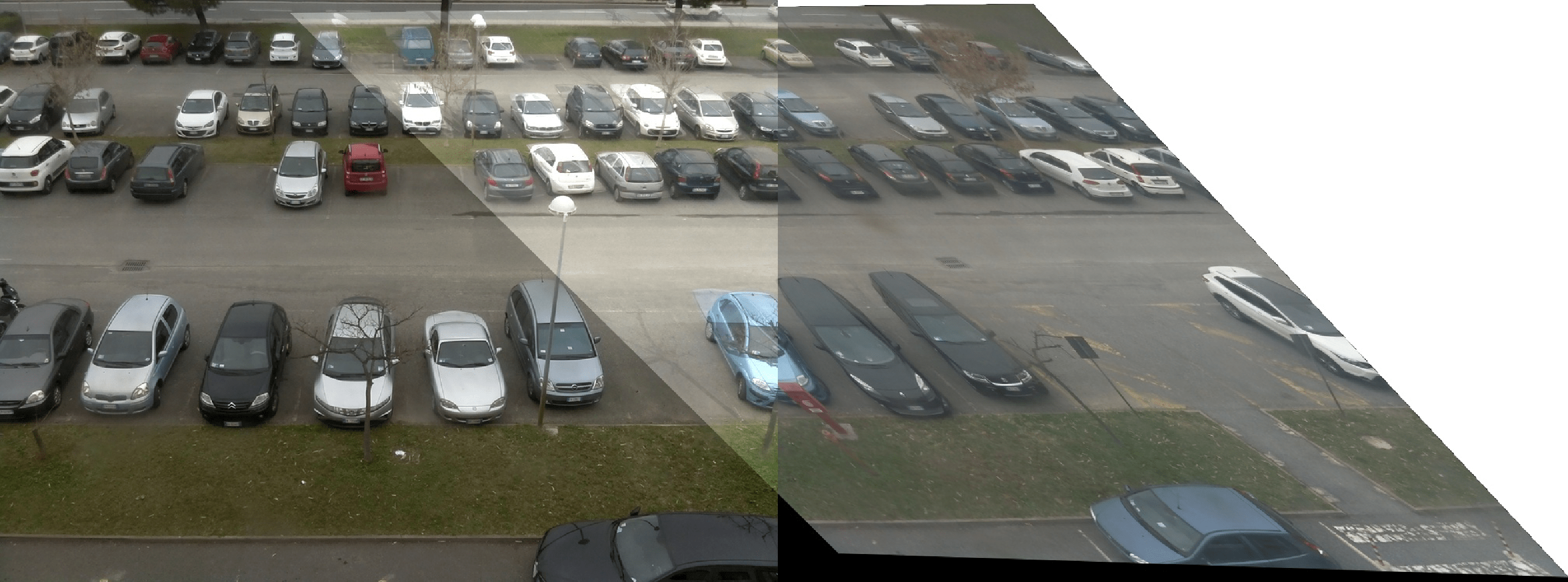}}
\caption{\textbf{Example of concatenation of two images using a homographic transformation.} It is also visible the overlapping area between them.}
\label{fig:stiching_example}
\end{figure}

However, finding this homography can be challenging because neighboring cameras can have different angles of view, leading to a perspective distortion between the images captured by them. Given a pair of neighboring nodes $\nu_i, \nu_j$, we employ a procedure that starts with finding the SIFT \cite{sift} key-points and feature descriptors of the images $i, j$ captured by the two nodes. Then, we match the two sets of feature descriptors performing the David Lowe’s ratio test \cite{sift}, and we further filter the matched feature descriptors by keeping only the pairs whose euclidean distance is below a given threshold. Finally, we obtain the homography transformation by applying the \acrfull{ransac} \cite{ransac} algorithm to the filtered feature descriptors. \ref{fig:stiching_example} shows the concatenation of two neighboring images $i$ and $j$ in which we apply the found homographic matrix to the image $i$, to have the same perspective as the image $j$. 

\begin{algorithm}[htbp]
\caption{\textbf{: Initialization} \\ At each Initialization Signal by $S$, each node $\nu_i$ performs the following steps:}

\begin{algorithmic}[1]
\State {\Call{ReceiveInitSignal()}{}} \Comment{waits the initialization signal from $S$}
\State {image$_i \gets$ \Call{CameraCapture()}{}}
\ForEach {$j \in J $} \Comment{$J$ is the set of neighboring nodes of node $\nu_i$}
\State {\Call{SendImage}{image$_i$,$\nu_j$}} \Comment{sends image$_i$ to node $\nu_j$}
\State {image$_j \gets$ \Call{ReceiveImage()}{}} \Comment{receives image$_j$ from node $\nu_j$}

\State {$H_{j,i} =$ \Call{ComputeHomography}{image$_j$, image$_i$}}
\EndFor
\end{algorithmic}

\label{alg:system_init}
\end{algorithm}

\subsubsection{Local Counting Algorithm}

The local counting algorithm runs directly and independently onboard the disposable devices, i.e., the smart cameras. It combines a \acrshort{cnn} in charge of the localization and the estimation of the number of vehicles present in the acquired single images, i.e., the contents of the messages $m_{i}$ and the quantities $\eta_i$ shown in \ref{fig:system_overview}, together with a geometric-based approach responsible of estimating the number of vehicles present in the overlapping areas between the nodes and their neighbors, i.e., the quantities $\mu_{j, i}$ in \ref{fig:system_overview}.

To count the vehicles, we exploited a detection-based approach since regression-based techniques cannot precisely localize the objects present in the scene and can eventually provide only a coarse position of the area in which they are distributed (see also \ref{sec:back:visual-counting} for further details), therefore resulting not suitable in our scenario where we need to know the precise localization (with boundaries) of the detected vehicles. As in the previous \ref{sec:counting-on-the-edge:counting-drones} where we proposed a solution able to estimate the number of vehicles in \acrshort{uav} imagery, also in this research, we faced challenges due to the imitated computational resources available in smart cameras. Here, we relied on \textit{Mask R-CNN} \cite{mask_rcnn}, a popular deep CNN for instance segmentation that operates within the `recognition using regions paradigm \cite{recognition_using_regions}, described in \ref{sec:back:cnn-based-detectors:two-stages-detectors}. In particular, it extends the \textit{Faster R-CNN} detector \cite{faster_rcnn} by adding a branch that outputs a binary mask indicating whether or not a given pixel is part of an object. Briefly, in the first stage, a \acrshort{cnn} acts as a backbone, extracting the input image features. Starting from this feature space, another \acrshort{cnn} named \acrfull{rpn} generates region proposals that might contain objects. \acrshort{rpn} slices pre-defined region boxes (called anchors) over this space and ranks them, suggesting those most likely containing objects. Once \acrshort{rpn} produces the Regions of Interested (ROIs), they might be of different sizes. Since it is hard to work on features having different sizes, \acrshort{rpn} reduces them into the same dimension using the Region of Interest Pooling algorithm. Finally, these fixed-size proposals are processed by two parallel CNN-based branches: one is responsible for classifying and localizing the objects inside them with bounding boxes; the second produces a binary mask that says whether or not a given pixel is part of an object. In the end, given an input image, the network produces per-pixel masks localizing the detected objects together with the associated labels classifying them. To make our counting solution able to run efficiently directly on the devices, we employed, as a backbone, the \textit{ResNet50} architecture, a lighter version of the popular \textit{ResNet101} \cite{resnet}. This simplification is also justified because the more powerful version of Mask R-CNN based on the ResNet101 model was designed for more complicated visual detection tasks than ours. Originally, Mask R-CNN was trained on the \textit{COCO} dataset \cite{lin2014microsoft} to detect and recognize 80 different classes of everyday objects. In our case, we have to localize and identify objects belonging to just one category (i.e., the \textit{vehicle} category). To this end, we further simplified the model by reducing the number of the final fully convolutional layers responsible for the classification of the detected objects, making the model lighter. Once we have localized the instances of the objects, we count them estimating the number of vehicles present in the scene.

The local counting is triggered by the Sink node $S$, sending a synchronization signal to all the smart cameras belonging to the system. Once received the synchronization signal, each node $\nu_i$ captures an image belonging to its underlying \acrshort{fov}, feeds it to the previously described CNN-based counting technique, and obtains as output a set of masks masks$_i$ localizing the vehicles present in the scene. The cardinality of this set of masks corresponds to the number of cars present in the image, i.e., the quantity $\eta_i$, that is sent through a message to the Sink node $S$. Then, the node $\nu_i$ packs this set of masks masks$_i$ in a message $m_i$, sends it to all its neighboring nodes $\nu_j$, and receives from them their corresponding set of masks masks$_j$ packed in a message $m_j$. Once received a message $m_j$, the node $\nu_i$ is responsible for analyzing the potential vehicles present in the overlapped area between its \acrshort{fov} and the ones of the nodes $\nu_j$. To this end, it employs the homographic transformation $H_{j, i}$ computed during the system initialization, projecting the masks belonging to the set masks$_j$ into its image plane, filtering them and discarding the ones that overlap with the masks belonging to the set masks$_i$ with a value of \acrfull{iou} greater than a threshold that we empirically found to be optimal at 0.2. These masks indeed localize vehicles already detected, and that should not be considered a second time. On the other hand, the cars left after this filtering are vehicles that were not detected in the \acrshort{fov} underlying the node $\nu_i$, but instead found by the node $\nu_j$, probably because of having a better view of this object. Referring to our graph modeling the system and reported in \ref{fig:system_overview}, the number of the cars detected by $\nu_j$ and already localized by $\nu_i$ corresponds to the message $\mu_{j, i}$, that is sent to the Sink node $S$. We detail all the described steps in the \ref{alg:local_counting} and in the \ref{alg:compute_num_overlaps}.

\begin{algorithm}[htbp]
\caption{\textbf{: Local Counting} \\ At each Computational Signal by $S$, each node $\nu_i$ performs the following steps:}

\begin{algorithmic}[1]
\State {\Call{ReceiveComputSignal()}{}} \Comment{waits the computational signal from $S$}
\State {image$_i \gets$ \Call{CameraCapture()}{}}

\State {masks$_i \gets$ \Call{MaskRCNN}{image$_i$}}
\State {$\eta_i \gets \left| \text{masks}_i\right |$}
\State {\Call{SendMessage}{$\eta_i, S$}} \Comment{sends $\eta_i$ to Sink node $S$}
\State {$m_i \gets$ \Call{PackMessage}{masks$_i$}} \Comment{builds message $m_i$ containing masks$_i$}
\ForEach {$j \in J $} \Comment{$J$ is the set of neighboring nodes of node $\nu_i$}
\State {\Call{SendMessage}{$m_i, \nu_j$}} \Comment{sends $m_i$ to node $\nu_j$}
\State {$m_j \gets$ \Call{ReceiveMessage()}{}} \Comment{receives message $m_j$ from node $\nu_j$}
 \State {masks$_j \gets$ \Call{UnpackMessage}{$m_j$}} \Comment{unpacks $m_j$ containing masks$_j$}
\State {$\mu_{j, i} \gets$ \Call{compute\_$\mu$}{masks$_i$, masks$_j$, $H_{j, i}$}}
\State {\Call{SendMessage}{$\mu_{j, i}, S$}} \Comment{sends $\mu_{j, i}$ to Sink node $S$}
\EndFor
\end{algorithmic}

\label{alg:local_counting}
\end{algorithm}  

\begin{algorithm}[htbp]
\caption{: \textbf{Computation of $\mu$} \\ $\mu$ represents the num of cars detected by $\nu_j$ and already detected by $\nu_i$ \\ Each node $\nu_i$ performs the following procedure:}

\begin{algorithmic}[1]
\Procedure{compute\_$\mu$}{masks$_i$, masks$_j$, $H_{j,i}$}

\State {n\_cars\_already\_detected $\gets 0$}
\ForEach {mask $\in$ masks$_j$}
\State  {mask$_h \gets $} \Call{Project}{$H_{j,i}$, mask} \Comment{projects mask points on plane $i$}
\If {mask$_h$ falls within image$_i$}
\State{mask$_\text{max} \gets \argmax_{m \in \text{masks}_i} \text{IoU}(\text{mask}_h, m)$}
 \If {IoU$(\text{mask}_h, \text{mask}_\text{max}) > \tau$ }
 \State {n\_cars\_already\_detected ++}
 \EndIf
\EndIf
\EndFor
\State {\textbf{return} n\_cars\_already\_detected}
\EndProcedure
\end{algorithmic}
\label{alg:compute_num_overlaps}
\end{algorithm}

\subsubsection{Global Counting Algorithm}
The global counting algorithm runs on the Sink node $S$, and it is responsible for the fusion of the partial results coming from all the other nodes and for finally outputting the number of cars present in the \textit{entire} monitored parking area.

This phase starts when $S$ receives all the $\eta_i$ and the $\mu_{j, i}$, i.e., the number of vehicles estimated in the single \acrshort{fov}s and the estimation of the number of cars already considered in the overlapping areas between neighboring cameras, from all the nodes belonging to the system. In particular, for each overlapped area shared between a pair of nodes $\nu_i, \nu_j$, the node $S$ receives two messages $\mu_{j, i}$ and $\mu_{i, j}$, the contents of which are computed by the two nodes employing two homographic transformations $H_{j, i}$ and $H_{i, j}$, respectively. These two quantities can be potentially different. We choose the best value by aggregating them, choosing between three different functions - max, min, and mean, finding that the latter is the best one. Finally, the node $S$ builds the final result, i.e., the estimation of the number of vehicles present in the \textit{entire} parking lot, by summing up all the $\eta_i$, and subtracting the aggregated values. We detail all these steps in the \ref{algo:global_counting}.

\begin{algorithm}[htbp]
\caption{\textbf{: Global Counting} \newline The Sink node $S$ performs the following steps:}

\begin{algorithmic}[1]
\ForEach {$(\mu_{i, j}, \mu_{j, i})$}
\State{$\overline{\mu_k} \gets$ \Call{Aggregate}{$\mu_{i, j}, \mu_{j, i}$}}
\EndFor
\State{global\_cars\_count $\gets \sum_{n=1}^{N} \eta_n - \sum_{k=1}^{K} \overline{\mu_k}$ \newline \Comment{$N$ is the set of nodes, $K$ is the set of aggregations}}
\end{algorithmic}

\label{algo:global_counting}
\end{algorithm}

\subsection{Experimental Setup}
\label{sec:counting-on-the-edge:multi-camera-counting:exp_setup}
This section describes the simulated scenario that we exploited for our experiments. In particular, we extend the \textit{CNRPark-EXT} dataset \cite{AmatoCarOccupancy2017}, adapting it to be suitable for the counting task and so that it was usable for training the vehicles counting \acrshort{cnn} running on the smart cameras and applicable to validate our multi-camera algorithm.
Furthermore, we briefly describe the \textit{PKLot} dataset \cite{pklot_dataset}, a public dataset comprising parking lot scenes that we exploit for further assessing the generalization capabilities of the local vehicle counting network. 
Finally, we report some implementation details.

\subsubsection{The CNRPark-EXT Dataset}
We exploited the \textit{CNRPark-EXT} public dataset introduced in \cite{AmatoCarOccupancy2017}, a collection of annotated images of vacant and occupied parking spaces on the campus of the National Research Council (CNR) in Pisa, Italy. This challenging dataset represents most of the problematic situations that can be found in a real scenario: nine different cameras capture the images under various weather conditions, angles of view, light conditions, and many occlusions. Furthermore, the cameras have their fields of view partially overlapped. Since this dataset is specifically designed for parking lot occupancy detection, it is not directly usable for the counting task. Indeed, each image, called \textit{patch}, contains one parking space labeled according to its occupancy status - 0 for vacant and 1 for occupied. Since this work aims at counting the cars present in the parking area, we extended it by considering the full images and adapting the ground truth to our purposes. 

To train and evaluate the local vehicle counting based on Mask R-CNN, we created a suitable label set. In this case, labels correspond to \textit{binary masks}, i.e., binary images identifying the polygons surrounding the vehicles we want to detect. Since mask creation is a very time-consuming operation, differently from \cite{ciampi_sebd}, we considered the \textit{raw} masks obtained directly from the bounding boxes localizing the occupied parking spaces. The idea is that we do not need precise polygons that identify the vehicles we want to detect. Still, we can use the region within the delimiters that identify the occupied parking spaces and the underlying part of the car.

On the other hand, to validate our multi-camera algorithm, we built a simulated scenario considering some sequences of images belonging to different cameras captured simultaneously. In other words, a sequence is defined as the set of images captured by the different smart cameras that are monitoring the parking area at the same moment. Hence, a sequence represents a snapshot of the \textit{entire} parking lot at a given timestamp, and it takes into account all the spaces from the available different views. We manually annotated these sequences to obtain the ground truth car counts. Specifically, we considered the single images composing a sequence, counting the vehicles present in the scenes, but taking care of accounting for them just once if they appear in more than one view. We labeled six different sequences, two for each weather condition, considering the images from camera$_2$ to camera$_9$. We did not consider camera$_1$ since it has small and particularly skewed field-of-view overlaps with the others cameras, hindering the automatic homography estimation and the subsequent projections.

\subsubsection{The PKLot Dataset}
To further validate the generalization capabilities of the CNN-based local vehicles counting algorithm, we exploited an additional public dataset, named \textit{PKLot} \cite{pklot_dataset}. In particular, this dataset is composed by three different scenarios describing three different parking lot scenes - \textit{UFPR04}, \textit{UFPR05} and \textit{PUC}. We considered only the first two subsets since the third one contains images captured from a fixed camera located at the height of the 10th floor of a building, which provides a slanted view of the parking lot and results in a different setting without intra-vehicle occlusions. Like the CNRPark dataset, also PKLot is specifically designed for the parking lot occupancy detection task. Thus, we manually re-labeled the ground truth to our purposes, obtaining a simulated scenario suitable for measuring the performance of our solution concerning the counting task.

\subsubsection{Implementation Details}
We report in this section some implementation details concerning the Mask R-CNN-based algorithm responsible for the prediction of the number of vehicles in the single images. In particular, we trained the modified Mask R-CNN initializing the weights of the ResNet50 backbone with the ones of a pre-trained model on \textit{ImageNet} \cite{imagenet_dataset}, a popular dataset for classification tasks, and the remaining ones at random. We freeze the backbone for the first ten epochs, and then we trained the whole network for 20 additional epochs. To prevent overfitting, we applied some standard augmentation techniques to the training data: images are horizontally flipped with a 0.5 probability, then their pixels are multiplied by a random value between 0.8 and 1.5, and finally, they are blurred using a Gaussian kernel with a standard deviation of a random value between 0 and 5. Then, to support training multiple images per batch, we resized all pictures to the same size. If an image is not square, we pad it with zeros to preserve the aspect ratio. In the end, we obtained images of size $1024\times1024$. At inference time, images are resized and padded with zeros to get a square picture of size $1024\times1024$, and no other augmentations occur.

\subsection{Experiments and Results}
\label{sec:counting-on-the-edge:multi-camera-counting:experiments}
In this section, we report the experiments and the obtained results. First, we evaluate the performance against other state-of-the-art solutions of the CNN-based technique responsible for estimating the vehicle number in images concerning the single-camera scenario, i.e., considering a single \acrshort{fov}. Then, we validate the effectiveness of our multi-camera algorithm, testing it in the simulated scenario previously described. Following other counting benchmarks, for all the experiments we exploited the \acrfull{mae}, the \acrfull{mse}, and the \acrfull{mare} as performance metrics, defined in \ref{sec:back:visual-counting:metrics}.

\subsubsection{Experiments on the Single-Camera Scenario}

\begin{figure}[htbp]
\centering
  \begin{subfigure}[b]{0.48\textwidth}
    \includegraphics[width=\textwidth]{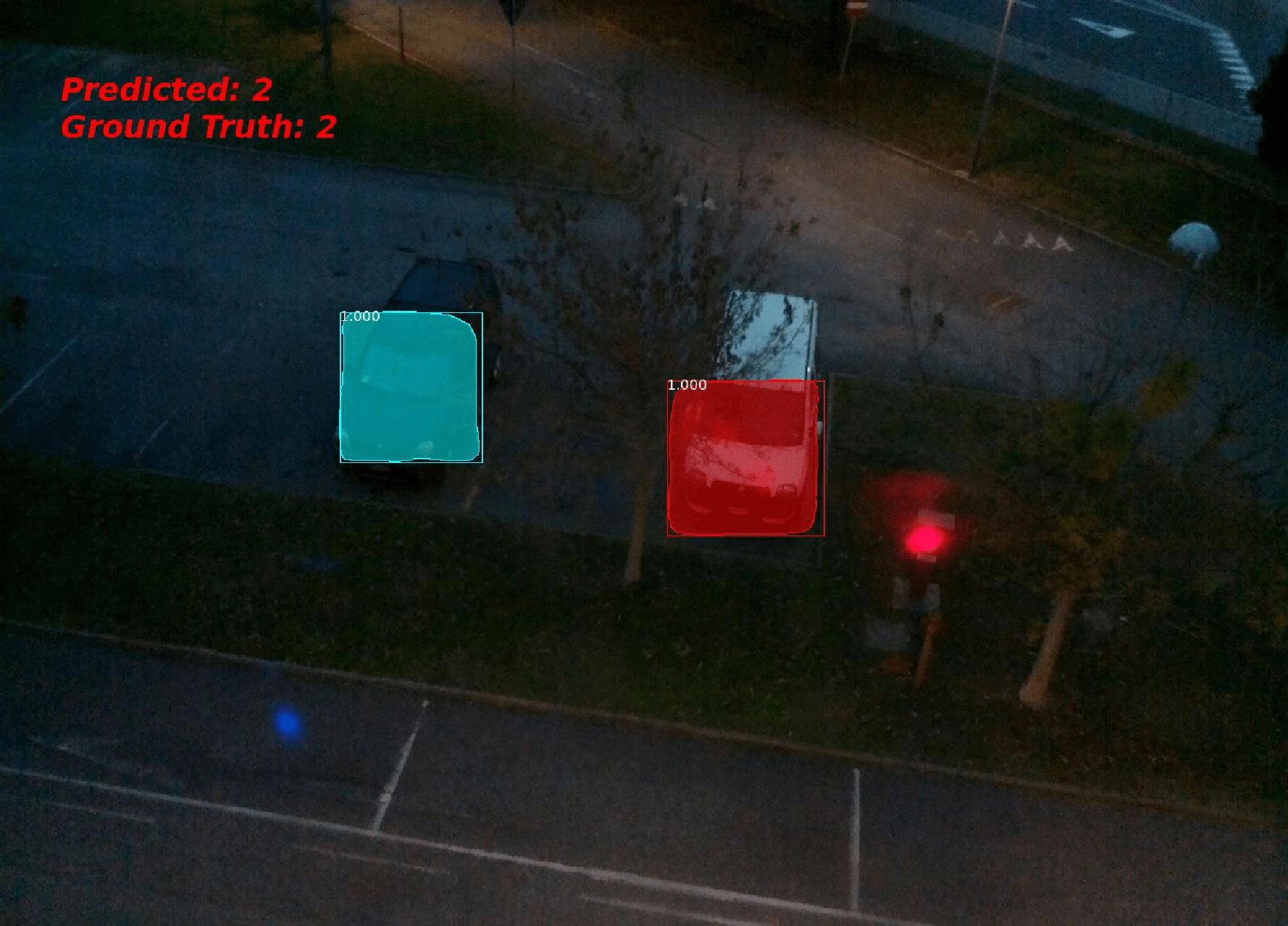}
    \caption{Image from Camera$_2$}
    \label{example_counting_output_a}
  \end{subfigure} \hfill
  \begin{subfigure}[b]{0.48\textwidth}
    \includegraphics[width=\textwidth]{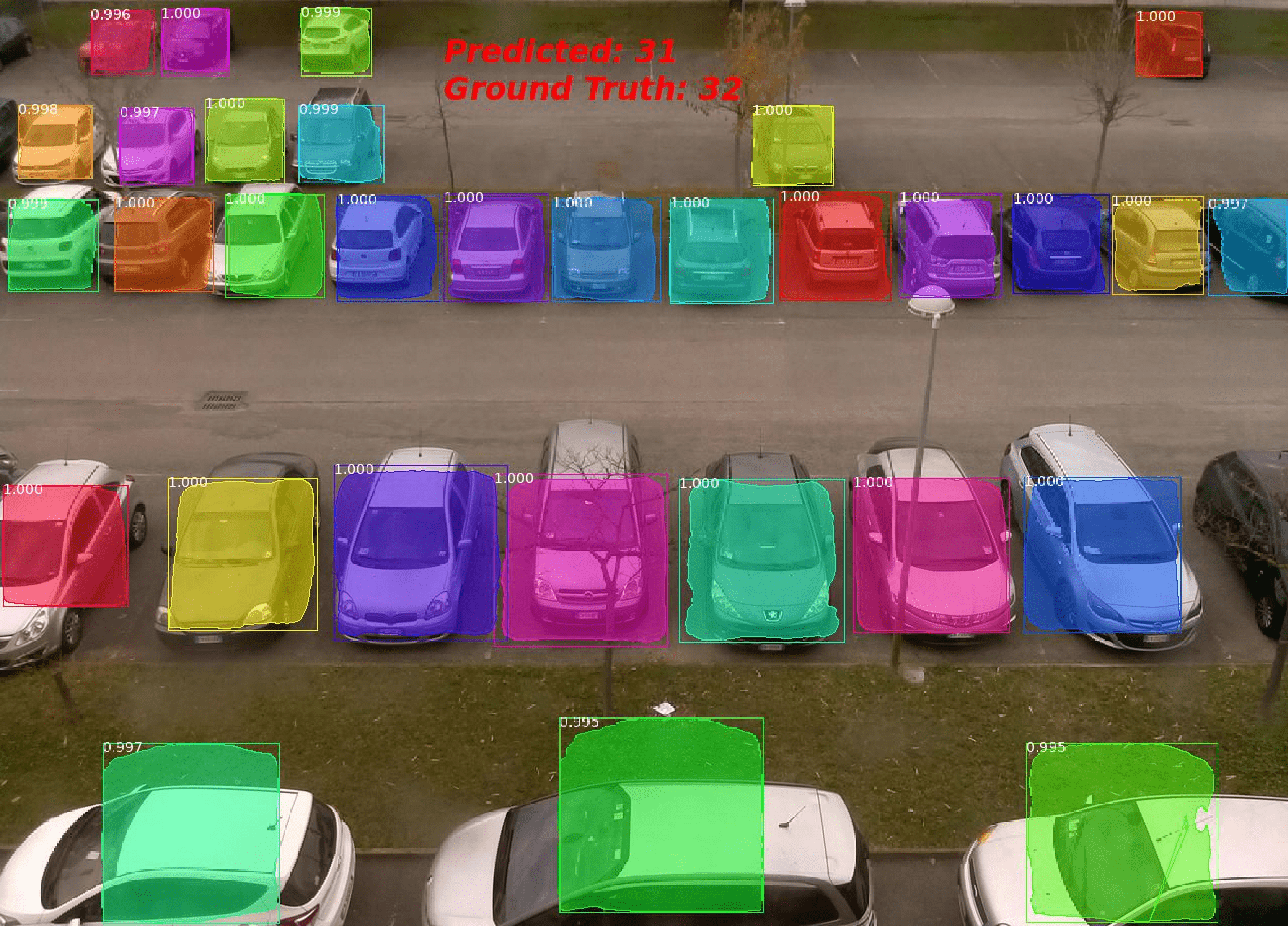}
    \caption{Image from Camera$_8$}
    \label{example_counting_output_b}
  \end{subfigure}
  \caption{\textbf{Examples of vehicle localization and counting in the Single-Camera Scenario.} Images are taken from the CNRPark-EXT dataset. We report the masks localizing detected cars and the estimate of their number.}
  \label{fig:example_counting_output}
\end{figure}

\textit{State-of-the-art comparison.} We compared our solution with the results obtained in \cite{ciampi_sebd}, where a centralized detection-based counting approach has been presented. It is based on the original version of Mask R-CNN having the ResNet101 model as features extractor, which has been fine-tuned on a very small manually annotated subset of the CNRPark-EXT dataset, starting from the model pre-trained on the COCO dataset \cite{lin2014microsoft}. Although this solution is computationally expensive and unsuitable for edge devices, it represents a direct comparison in terms of counting on the same dataset. We also compared our technique against the method proposed in \cite{AmatoCarOccupancy2017}, an approach for car parking occupancy detection based on \textit{mAlexNet}, a deep \acrshort{cnn} designed explicitly for smart cameras. This work represents an indirect method for counting cars in a parking lot, as the counting problem is cast as a classification problem: if a parking space is occupied, we increment the total number of cars. We illustrate the results in \ref{tab:results_edge_counting}, where we also report the performance obtained using the Mask R-CNN network without a preliminary fine-tuning on the CNRPark-EXT dataset. Our solution performed better than the other state-of-the-art considered methods, considering all three counting metrics. In particular, our approach outperformed the solution introduced in \cite{ciampi_sebd}, despite the latter employing a more deep and powerful \acrshort{cnn}, and it is designed to be used as a centralized-server solution. This is explained by the fact that in \cite{ciampi_sebd}, the authors fine-tuned the \acrshort{cnn} using a tiny dataset. Consequently, the algorithm overfits over the training data, and it cannot generalize over the test subset. It is also worthy of notice that our \acrshort{cnn} also outperformed the mAlexNet network, even though the latter knows the exact location of the parking spaces. \ref{fig:example_counting_output} shows some examples of images belonging to different cameras and different weather conditions together with the masks localizing the found vehicles. 

\begin{table}[t]
\caption{\textbf{Experiments on the single-camera scenario: comparison against SOTA.} Left-side: results obtained using our counting solution compared with other state-of-the-art approaches; we get the best results on all the three considered counting metrics. Right-side: evaluation of the generalization capabilities on the PKLot dataset \cite{pklot_dataset}, using the model trained on the CNRPark-EXT dataset; we achieved an error that is approximately four times lower than the one obtained with the COCO pre-trained model.}
\centering
\small
\setlength{\tabcolsep}{3pt}
\newcolumntype{R}{>{\raggedleft\arraybackslash}X}
\begin{tabularx}{\linewidth}{l*{3}{R}*{3}{R}}
\toprule
       & \multicolumn{3}{c}{CNRPark-EXT} & \multicolumn{3}{c}{PKLot} \\
         \cmidrule(l){2-4}                 \cmidrule(l){5-7}
Method & 
MAE & 
MSE & 
MARE &
MAE & 
MSE & 
MARE \\
\midrule
\cite{AmatoCarOccupancy2017}                             & 1.34 & 8.00 & 0.04 & \multicolumn{3}{c}{-} \\
\cite{ciampi_sebd} & 1.05 & 4.41 & 0.03 & \multicolumn{3}{c}{-} \\
ResNet50 Mask R-CNN                                       & 11.20 & 247.40 & 0.30 & 16.90 & 522.40 & 0.48 \\ 
Our solution                                              & \textbf{0.49} & \textbf{1.04} & \textbf{0.01} & \textbf{4.56} & \textbf{33.88} & \textbf{0.13}\\
\bottomrule
\end{tabularx}
\label{tab:results_edge_counting}
\end{table}

\noindent \textit{Generalization capabilities}.
Errors in vehicle detection and counting are due to many reasons, but critical points are different light conditions and diverse perspectives. Weather conditions might produce significant illumination changes since puddles and wet floors create a textural pattern that may lead to an error, and sunbeams can create reflections on the car windscreen, covering the majority of the images with saturated patterns. When a \acrshort{cnn} does not generalize well, it works well only in the conditions where it was trained. To measure the robustness of our approach to these scenarios, we performed two types of experiments exploiting the CNRPark-EXT dataset: i) \textit{inter-weather} and ii) \textit{inter-camera} experiments. In the former, we trained our \acrshort{cnn} with images taken in one particular weather condition, and we computed the performance metrics obtained on images having different weather conditions. In particular, we performed three experiments, training respectively on the \textit{Sunny}, \textit{Overcast} and \textit{Rainy} subsets of the CNRPark-EXT dataset. In the latter, we trained our algorithm employing images from one camera, and then we computed the performance metrics on pictures captured by another camera. In particular, we performed two experiments, training with images coming respectively from camera$_1$ and camera$_8$. We chose these two cameras because they are particularly representative since one has a side view of the parking lot while the other has a pure front view. We report the results of the two experiments in \ref{inter_weather_results} and \ref{inter_camera_results}, respectively. We achieved a good generalization in both the considered scenarios. We experienced a larger amount of error when the \acrshort{cnn} was trained and tested on two opposite weather conditions, for instance, \textit{Sunny} and \textit{Rainy}, while the more accurate model was the one trained on \textit{Overcast} weather conditions. However, the performance difference is quite small. On the other hand, in \textit{inter-camera} experiments, the model trained on camera$_8$ is the best, and it has a slight drop in performance only when tested on the camera$_1$ subset. The model trained on the camera$_1$ dataset performed in general worse. This is probably due to a bias in the CNRPark-EXT dataset, where the majority of the images are captured from a frontal viewpoint.

\begin{table}
\caption{\textbf{Experiments on the single-camera scenario: generalization capabilities.}}
\newcolumntype{C}{>{\centering\arraybackslash}X}
\newcolumntype{R}{>{\raggedleft\arraybackslash}X}
\small
\begin{subtable}{\linewidth}
\caption{Results of inter-weather experiments obtained when training on sunny, overcast, or rainy weather.}
\centering
\begin{tabularx}{.95\linewidth}{lC*{8}{R}}
\toprule
                     & \multicolumn{3}{c}{Sunny} & \multicolumn{3}{c}{Overcast} & \multicolumn{3}{c}{Rainy} \\
                       \cmidrule(lr){2-4}          \cmidrule(lr){5-7}             \cmidrule(lr){8-10}
Train Set            & MAE  & MSE  & MARE         & MAE  & MSE  & MARE            & MAE  & MSE  & MARE  \\
\midrule
Sunny    & -    & -    & -           & 0.29 & 0.34 & 0.009          & 0.96 & 2.78 & 0.02 \\
Overcast & 0.62 & 1.09 & 0.02        & -    & -    & -              & 0.56 & 1.26 & 0.01 \\
Rainy    & 0.84 & 1.65 & 0.02        & 0.49 & 0.65 & 0.01           & -    & -    & -    \\
\bottomrule
\end{tabularx}
\label{inter_weather_results}
\end{subtable}\\[6ex]
\begin{subtable}{\linewidth}
\caption{Results of inter-camera experiments obtained when training on camera 1 and camera 8.}
\centering
\begin{tabularx}{.95\linewidth}{cc*{9}{R}}
\toprule
&& \multicolumn{9}{c}{Test Set} \\
   \cmidrule(lr){3-11}
Metric & Train Set &          C1 &    C2 &    C3 &     C4 &     C5 &     C6 &     C7 &     C8 &    C9 \\
\midrule
\multirow{2}{*}{MAE} & C1 &     - &  0.77 &  1.21 &   2.53 &   3.26 &   2.57 &   2.88 &   2.88 &  1.54 \\
    & C8 &   3.87 &  0.85 &  0.76 &   0.45 &   0.48 &   0.71 &   1.07 &     - &  0.41 \\
\midrule
\multirow{2}{*}{MARE} & C1 &     - &  0.08 &  0.05 &   0.06 &   0.07 &   0.05 &   0.06 &   0.05 &  0.05 \\
    & C8 &   0.11 &  0.09 &  0.03 &   0.01 &   0.01 &   0.01 &   0.02 &     - &  0.01 \\
\midrule
\multirow{2}{*}{MSE} & C1 &     - &  1.48 &  2.91 &  10.61 &  20.24 &  13.50 &  19.82 &  17.30 &  7.19 \\
    & C8 &  22.60 &  1.78 &  1.36 &   0.57 &   0.74 &   0.95 &   4.97 &     - &  2.13 \\
\bottomrule
\end{tabularx}
\label{inter_camera_results}
\end{subtable}
\label{generalization_capabilities}
\end{table}

\noindent Moreover, to further validate the generalization capabilities of our approach, we considered our counting network trained on the entire training set of the CNRPark-EXT dataset, and we tested it over a different dataset, the PKLot dataset \cite{pklot_dataset}. Results are shown in \ref{tab:results_edge_counting} where we also report the performance obtained using the Mask R-CNN network without a preliminary fine-tuning on the CNRPark-EXT dataset. As we can see, using our solution, we achieved an error that is approximately four times lower than the one obtained with the COCO pre-trained model.

\subsubsection{Experiments on the Multi-Camera Scenario}
To the best of our knowledge, there are no annotated datasets in the literature suitable for evaluating counting algorithms operating on multiple \acrshort{fov}-overlapping cameras. Hence, we performed our experiments on the extended version of the CNRPark-EXT dataset, which we created on purpose in this dissertation. To demonstrate that our algorithm can benefit from the redundant information from the different cameras, we compared the obtained results against a baseline and a simplified version of our algorithm.

\begin{figure}[htbp]
\centering
  \begin{subfigure}[b]{0.48\textwidth}
    \includegraphics[width=\textwidth]{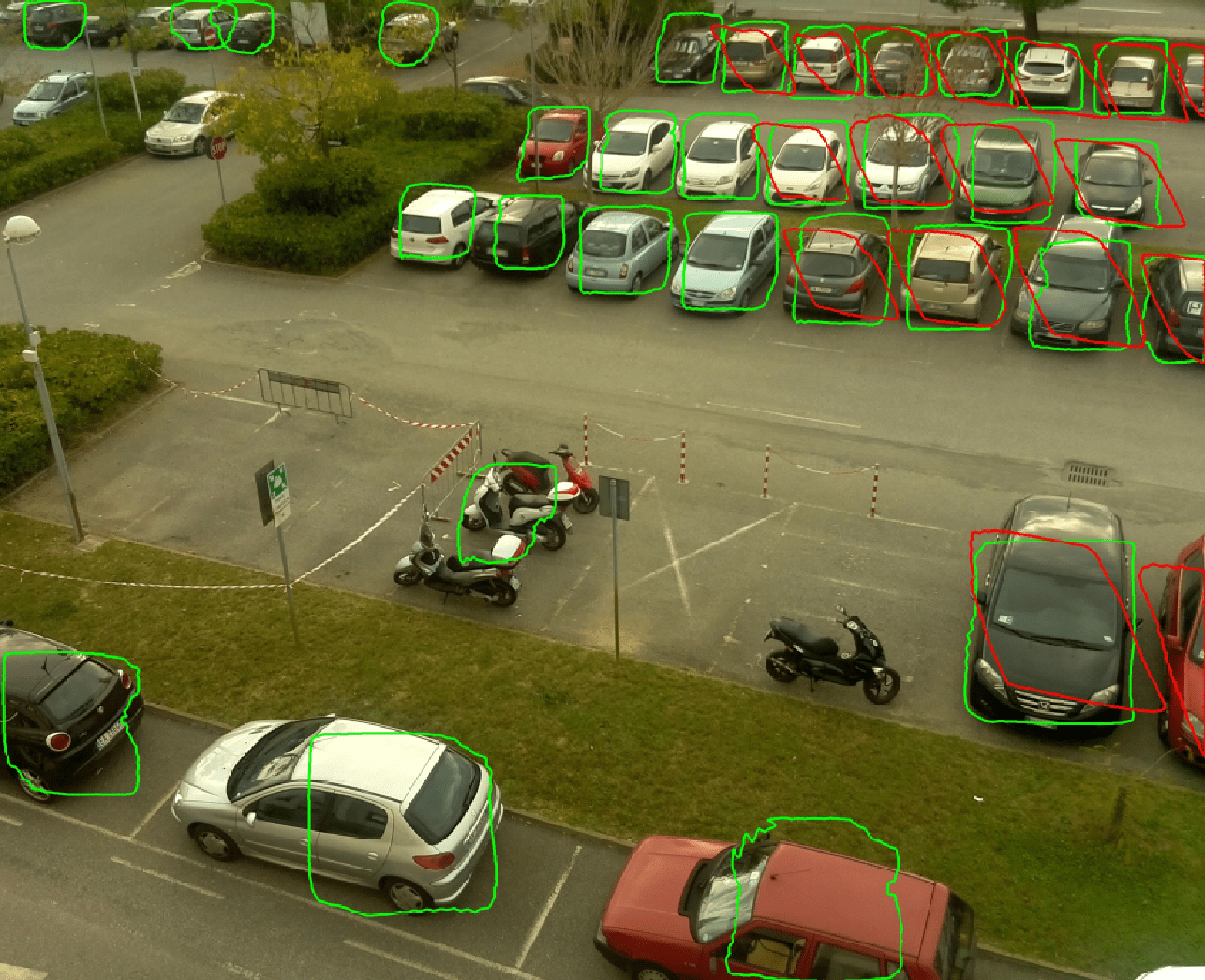}
    \caption{Image from Camera$_9$}
    \label{example_multi_camera_counting_output_a}
  \end{subfigure} \hfill
  \begin{subfigure}[b]{0.48\textwidth}
    \includegraphics[width=\textwidth]{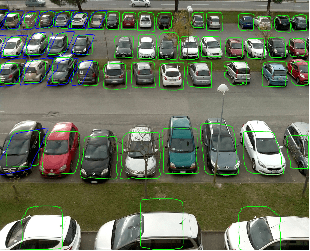}
    \caption{Image from Camera$_8$}
    \label{example_multi_camera_counting_output_b}
  \end{subfigure}
  \caption{\textbf{Example of vehicle localization and counting in the Multi-Camera Scenario.} We consider a pair of images belonging to two different cameras $i, j$ having a shared area in their FOV. We report in green the masks localizing the vehicles detected by a camera in its own FOV, while in red and blue, the masks projected from camera j to camera i and vice-versa, employing the homographic transformations pre-computed during the system initialization.}
  \label{fig:example_multi_camera_counting_output}
\end{figure}

In particular, we compared our solution against a system that is not aware of the other overlapped areas, and so it just sums up all the vehicles detected by all the cameras belonging to a sequence (Na\"ive Counting \textbf{N}). Then, we considered a more conservative approach, where the nodes employed the homographic transformations only with the purpose of black-masking the overlapped areas (Overlap Masking \textbf{M}). This latter baseline then loses the ability to take advantage of monitoring the same lots from different views. However, it is still aware of the locations of the overlapping areas, and it considers the vehicles inside them only once. Results are shown in \ref{tab:results_multi_camera_counting}. Our solution obtained the best results compared to the considered baselines in the three counting metrics and all the employed scenarios. We report the errors concerning the considered six sequences of the CNRPark-EXT dataset, together with the \acrshort{mae}, \acrshort{mse}, and \acrshort{mare}, which summarize the mean results regarding all the scenarios. As an example, in \ref{fig:example_multi_camera_counting_output} we also report the output of our multi-camera algorithm for a pair of images belonging to two different cameras having a shared area in their field of view, where we highlight in red and blue the masks projected from one camera to the other, using the previously computed homographic transformations.

\begin{table}[t]
\caption{\textbf{Experiments on the multi-camera scenario.} We consider the entire parking lot, comparing our solution against a baseline and a simplified version of our algorithm. We report the errors obtained on the six considered sequences (two for each weather condition) of the CNRPark-EXT dataset that we extend on purpose.}
\centering

\footnotesize
\newcolumntype{C}{>{\centering\arraybackslash}X}
\begin{tabularx}{\linewidth}{X *{3}{c} *{3}{c} *{3}{c} *{3}{c}}
\toprule
 & \multicolumn{3}{c}{Error} & \multicolumn{3}{c}{Absolute Err.} & \multicolumn{3}{c}{Squared Err.} & \multicolumn{3}{c}{Relative Err. (\%)} \\
 \cmidrule(l){2-4} \cmidrule(l){5-7} \cmidrule(l){8-10} \cmidrule{11-13}
  & \textbf{N} & \textbf{M} & \textbf{O} & \textbf{N} & \textbf{M} & \textbf{O} & \textbf{N} & \textbf{M} & \textbf{O} & \textbf{N} & \textbf{M} & \textbf{O} \\

\midrule
Overcast-1 &        124 &        -33 &    \textbf{2} &      124   &       33   &  \textbf{2}   &    15,376   &     1,089   &   \textbf{4}   &      71.6 &       19.0 &  \textbf{1.2} \\
Overcast-2 &        131 &        -26 &    \textbf{1} &      131   &       26   &  \textbf{1}   &    17,161   &       676   &   \textbf{1}   &      76.1 &      15.1 &  \textbf{0.6} \\
Rainy-1    &         80 &        -39 &   \textbf{-5} &       80   &       39   &  \textbf{5}   &     6,400   &     1,521   &  \textbf{25}   &      47.6 &      23.2 &  \textbf{2.9} \\
Rainy-2    &        105 &        -44 &    \textbf{-5} &      105   &       44   &   \textbf{5}   &    11,025   &     1,936   &   \textbf{25}   &      54.4 &      22.8 &   \textbf{2.6} \\
Sunny-1    &        117 &        -38 &     \textbf{2} &      117   &       38   &   \textbf{2}   &    13,689   &     1,444   &    \textbf{4}   &       68.0 &      22.1 &   \textbf{1.2} \\
Sunny-2    &        113 &          -37 &     \textbf{2} &      113   &       38   &   \textbf{2}   &    12,769   &     1,444   &    \textbf{4}   &      66.1 &      22.2 &   \textbf{1.2} \\
\midrule
Mean       &          111.6 &      -36.1 &     \textbf{-0.5} &      111.6 &       36.3 &   \textbf{2.8} &    12,736.6 &     1,351.6 &   \textbf{10.5} &      63.9 &      20.7 &   \textbf{1.6} \\
\bottomrule
\end{tabularx} \\[1ex]
\textbf{N}: Na\"ive Counting; \textbf{M}: Overlap Masking; \textbf{O}: Ours (mean aggr., IoU Threshold $\tau = 0.2$)
\label{tab:results_multi_camera_counting}
\end{table}

\subsection{Conclusion}
\label{sec:counting-on-the-edge:multi-camera-counting:conclusion}
In the second part of this chapter, we presented a distributed \acrlong{ai}-based system that automatically counts the vehicles present in a parking lot using images taken by multiple smart cameras. Unlike most of the works in literature, we introduced a multi-camera approach that can estimate the number of cars present in the \textit{entire} parking area and not only in the single captured images. The main peculiarities of this approach are that all the computation is performed in a distributed manner at the edge of the network and that there is no need for any extra information about the monitored parking area, such as the location of the parking spaces, nor any geometric information about the position of the cameras in the parking lot. We modeled our system as a graph. The nodes, i.e., the smart cameras, are responsible for estimating the number of cars present in their view and merging data from nearby devices with an overlapping field of view. Our solution is simple but effective, combining a \acrlong{dl} technique with a distributed geometry-based approach. We evaluated our algorithm on the CNRPark-EXT dataset, which we specifically extended to show how we benefit from redundant information from different cameras while improving overall performance.

\graphicspath{{img/virtual-to-real/}}

\chapter{Virtual To Real Adaptation of Pedestrian Detectors}
\label{ch:virtual-to-real}
A key task in many intelligent video surveillance systems is pedestrian detection, as it provides essential information for semantic understanding of video. Accurate detection of individual instances of pedestrians in images plays a vital role in a myriad of applications that can positively impact the quality of human life. They range from video surveillance \cite{bilal2016low, varga2017robust}, robotics, automotive \cite{gavrila2007multi, shashua2004pedestrian} and assistive technologies to people with visual disabilities \cite{tian2014rgb}, just to name a few. Moreover, pedestrian detection also provides the main building block for people counting using detection-based approaches. Once the pedestrian instances are localized, it is possible to estimate the number of people present in the scene or a region of interest inside it.

\acrlong{cnn}s-based methods \cite{lecun1998gradient} have recently demonstrated their superiority compared to the approaches relying on hand-crafted features. However, despite the recent advances, the pedestrian detection task remains a challenging active research area in Computer Vision. One of the main issues still affecting \acrshort{cnn}-based approaches for pedestrian detection is related to \textit{data scarcity}. Indeed, the crux of \acrshort{cnn}s is that to generalize well at inference time, they require a massive amount of diverse well-labeled data during the training phase, covering the widest number of different scenarios.
While there exist some large annotated generic datasets suitable for training these supervised learning networks, such as {ImageNet} \cite{imagenet_dataset} and {MS COCO} \cite{lin2014microsoft}, in many real-world situations they are not enough. Hence, as a consequence, a model that learned from these data usually experiences a drastic drop in performance at inference time when applied to other scenarios never seen during the training phase.  
This performance gap is due to the fact that the network learns from data belonging to one domain (named the \textit{training} or \textit{source} domain) and is then applied on another different domain (the \textit{test} or \textit{target} domain), and is commonly referred as \textit{Domain Shift} \cite{torralba2011unbiased}.

Since manually annotating new collections of images is expensive and requires great human effort, it is unfeasible to label new target data every time to supervise the model to acquire knowledge from the test domain. 
A recently promising approach is to gather data from virtual world environments that mimic as much as possible all the characteristics of the real-world scenarios and where the annotations can be acquired with an \textit{automated} process. To this end, in this chapter, we introduce \textit{\acrfull{viped}}, a new \textit{synthetic} dataset generated with the highly photo-realistic graphical engine of the video game {GTA V (Grand Theft Auto V)} by {Rockstar  North}, that extends the \acrfull{jta} dataset presented in \cite{fabbri_jta}. \acrshort{viped} is the first synthetic dataset suitable for the pedestrian detection task, annotated with bounding boxes locating the instances of the people present in the scenes.

While synthetic data collections are very appealing, usually, when training solely on a synthetic dataset, the model does not generalize well to real-world data on which the pedestrian detector, in the end, shall be used. In other words, the potentials inherent in synthetic data can only be partially exploited.
This is due to the \textit{Synthetic2Real} Domain Shift. In this case, the source and the target domains are the synthetic and the real-world, respectively. The gap between the two data distributions is caused by the fact that synthetic images' appearance is still significantly different from that observed in real-world images, even using current rendering techniques.
To mitigate this domain gap, in this chapter, we also propose two different supervised \textit{\acrfull{da}} methods, suitable for the pedestrian detection task but possibly applicable to general object detection. The first one consists of training the model by exploiting the synthetic data and then, in a second step, fine-tuning it using real-world images. Instead, the second one consists of an end-to-end training procedure in which we employed mixed batches containing both synthetic and real-world data.

We conducted extensive experimentation to validate our approaches. First, we tested the generalization capabilities of the detector over unseen scenarios. We showed that we obtained better or comparable results when training with the synthetic data compared to using the same model trained using only real-world images, just taking advantage of the variety of \acrshort{viped}. Secondly, we experimented with the two proposed supervised \acrlong{da} techniques to boost the performance over specific real-world scenarios. We demonstrated that we reduced the Synthetic2Real Domain Shift by bringing the two domains closer together, thus achieving better results. 

To summarize, the main contributions presented in this chapter are the following:
\begin{itemize}
    \item we present \acrfull{viped}, a new vast synthetic dataset collected using the photo-realistic video game GTA V (Grand Theft Auto V) and automatically annotated by the graphical engine. To the best of our knowledge, it was the first synthetic dataset suitable for pedestrian detection;
    \item we introduce two supervised \acrlong{da} techniques to mitigate the Synthetic2Real Domain Shift existing between the synthetic and the real-world images; 
    \item we demonstrate through extensive experimentation on various real-world pedestrian detection datasets the generalization capabilities of the detector trained using synthetic data, and that the two proposed \acrshort{da} solutions boost the performance over specific real-world scenarios, bringing the synthetic and the real-world domains closer.
\end{itemize}

The chapter is organized as follows. In \ref{sec:virtual-to-real:related-work}, we review some works related to the pedestrian detection task and the \acrshort{da}, focusing on the ones more significant for our proposal. In \ref{sec:virtual-to-real:viped}, we describe \acrshort{viped}, our new synthetic dataset. In \ref{sec:virtual-to-real:method}, we present the \acrshort{da} strategies we employed to contrast the Synthetic2Real Domain Shift. \ref{sec:virtual-to-real:experiments} shows the experimental evaluation of our approaches. 

The research presented in this chapter was published in \cite{ciampi_iciap, ciampi_viped}. The code, the models, and the dataset are made freely available at \href{https://ciampluca.github.io/viped}{https://ciampluca.github.io/viped}.

\section{Related Work}
\label{sec:virtual-to-real:related-work}
This section reviews some relevant works about the object and pedestrian detection task. We also analyze some previous studies on \acrshort{da}, focusing on the Synthetic2Real domain shift.

\paragraph{Pedestrian Detection}
Pedestrian detection is highly related to object detection. It deals with locating and recognizing instances of pedestrians' specific class, usually in images of urban environments, without considering group dynamics. We can subdivide approaches for the pedestrian detection task into two main research areas. The first class of detectors is based on handcrafted features, such as ICF (Integral Channel Features) \cite{ped_det_1, ped_det_2, ped_det_3, ped_det_4, ped_det_5}. These methods can usually rely on higher computational efficiency at the cost of lower accuracy. On the other hand, more recently, Deep Neural Network approaches have been explored. For example, \cite{ped_det_6, ped_det_7, ped_det_8} proposed some solutions based on the \acrshort{cnn} networks to detect pedestrians, even accounting for different scales.

\paragraph{Synthetic2Real Domain Adaptation}
With the need for huge amounts of labeled data, synthetic datasets have recently gained considerable interest. Some notable examples are GTA5 \cite{gta_5_semantic} and SYNTHIA \cite{synthia} for semantic segmentation. However, there is a non-negligible domain gap between the synthetic and the real worlds, as mentioned above. Many techniques try to fill this gap, using both supervised and unsupervised approaches. A more exhaustive survey about deep learning \acrshort{da} techniques is provided in \ref{sec:back:domain_adaptation} and in \cite{deep_da_survey}. For example, authors in \cite{learning_to_adapt} proposed an unsupervised domain adaptation solution for the segmentation task. Authors in \cite{fabbri_jta} created \acrfull{jta}, a synthetic dataset gathered from the highly photo-realistic video game GTA V. They demonstrated that it is possible to reach excellent results on tasks such as people tracking and pose estimation when validating on real data. In this work, we extend this dataset, making it suitable for the pedestrian detection task. Authors in \cite{ped_synth_1, ped_synth_2} have also focused on learning features from synthetic data for the pedestrian detection task. Still, they did not consider deep learning approaches, exploring only traditional detection techniques. In \cite{ped_synth_3}, instead, the authors employed a synthetic dataset to train a CNN able to detect objects belonging to different classes in a video. This \acrshort{cnn} is responsible only for the classification of the objects, while their detection relies on a background subtraction algorithm based on Gaussian Mixture Models (GMMs). The performance of this approach over real-world scenarios was evaluated employing two pedestrian detection datasets, one of which, the 2D MOT 2015 \cite{mot15}, is an older version of the dataset we used to carry out our experiments. To the best of our knowledge, \cite{driving_matrix} and \cite{ped_synth_3} are the closest works to ours. In particular, they also used {GTA V} as the source for the acquisition of the synthetic data, but they focused their efforts on the vehicle detection task.

\section{ViPeD}
\label{sec:virtual-to-real:viped}
\acrfull{viped} is a new {synthetic} dataset generated with the highly photo-realistic graphical engine of the video game {GTA V (Grand Theft Auto V)} by {Rockstar North}. It extends \acrfull{jta} dataset, presented in \cite{fabbri_jta}. The dataset includes a total of about 500K images, extracted from 512 full-HD videos of different urban scenarios. These videos are organized into a training set (256 videos) and a test set (the remaining 256 videos).

While we could reuse the already existing \acrshort{jta} images, we needed to generate suitable annotations for the pedestrian detection task. Indeed, the \acrshort{jta} dataset provides only skeletal information, which is helpful in the pose estimation and tracking tasks. Instead, in our scenario, we are required to annotate each pedestrian with the four coordinates (x, y, w, h) delimiting its minimum enclosing bounding box. Hence, we employed the already available \acrshort{jta} images producing a new set of labels suitable for our task.

Estimating the precise bounding box surrounding each pedestrian instance can be tricky, as we did not have access to the underlying GTA game engine. Other works tried to overcome this problem by using some interesting work-around. For example, \cite{driving_matrix} extracted the semantic masks around each object in the scene and separated the instances by exploiting the depth information available through the depth buffer.  

Our solution relied instead on the skeletal information already provided by the \acrshort{jta} annotations. Indeed, differently from \cite{driving_matrix}, we dealt with multiple instances of pedestrians in possibly highly crowded scenarios. In these cases, the depth information may be insufficient for distinguishing two different pedestrians, leading to possible severe bounding box estimation errors.

\begin{figure}
    \centering
  \subfloat[\label{dataset_skeleton}]{%
       \includegraphics[width=0.46\linewidth]{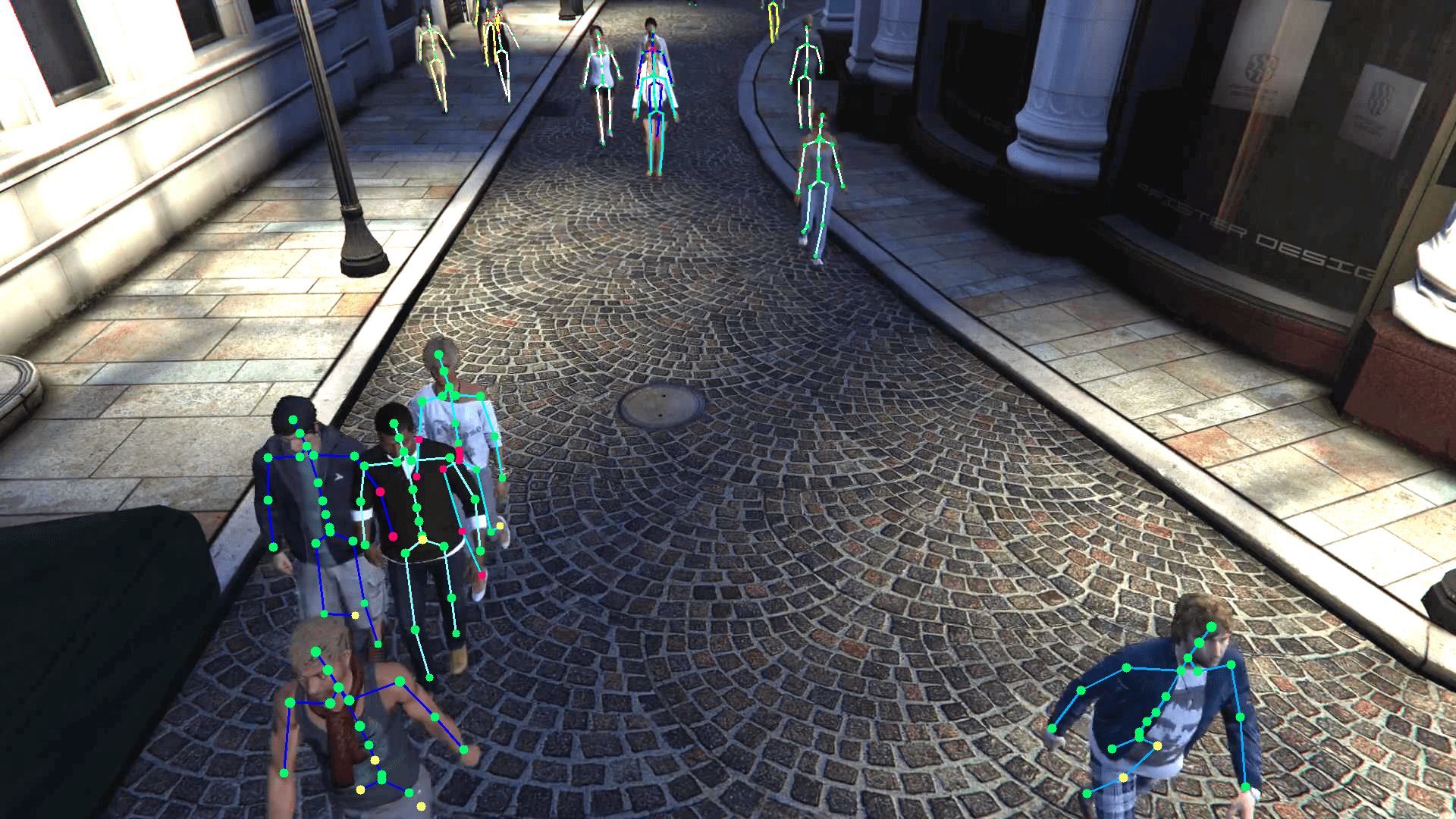}}
    \hfill
  \subfloat[\label{dataset_bbs}]{%
        \includegraphics[width=0.46\linewidth]{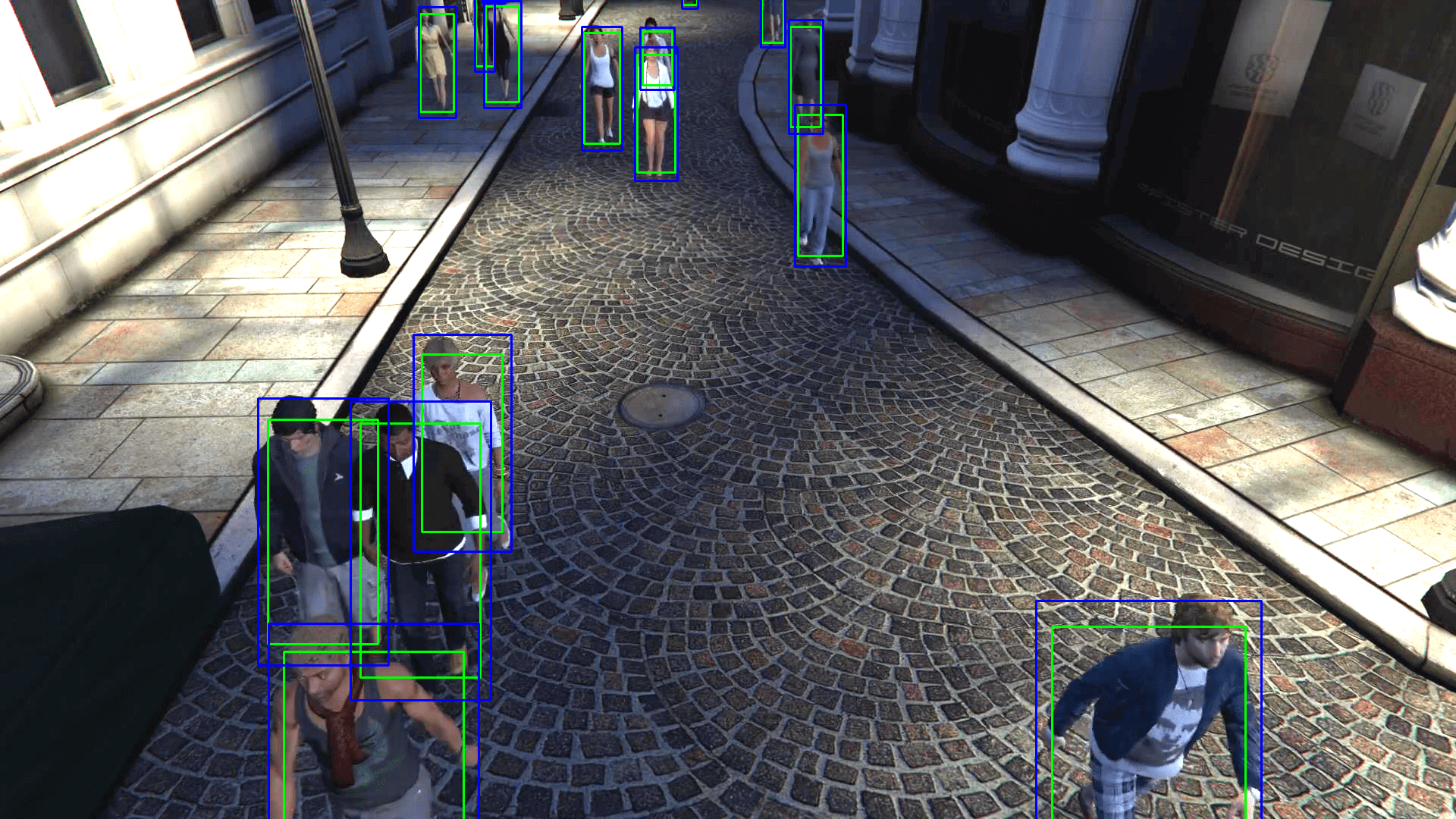}}
  \label{dataset} 
  \caption{(\textbf{a}) Pedestrians in the {JTA (Joint Track Auto)} dataset with their skeletons. (b) Examples of annotations in the \acrfull{viped} dataset; original bounding boxes are in green, while the sanitized ones are in blue.}
\end{figure}

As a first approximation, we exploited the skeleton joints' positions in screen coordinates, directly available from the \acrshort{jta} annotations, for drawing the minimum bounding box enclosing all the skeleton joints (green boxes in \ref{dataset_bbs}). However, it can be noticed that the bounding boxes produced using this simple procedure are undersized compared to the full-sized pedestrian instance, as the skeleton always lays below the skin surface. We solved this issue by constructing a more precise bounding box (blue boxes in \ref{dataset_bbs}), obtained by estimating an amount of padding through a simple heuristic.

In particular, we estimated the height of a pedestrian mesh, denoted as $h_m$, from the height $h_s$ of its skeleton, through the formula:
\begin{equation}
\label{mat:find_alpha}
    h_m = h_s + \frac{\alpha}{z},
\end{equation}
\noindent where $z$ is the distance of the pedestrian center of mass from the camera, and $\alpha$ is a parameter that depends on the camera projection matrix.

The $z$ value for each pedestrian was already part of the \acrshort{jta} annotations, while $\alpha$ is unknown since we could not access the camera parameters. Then, we evaluated $\alpha$ from \ref{mat:find_alpha}, estimating $h_m$ for a small representative population of pedestrians. To this end, we isolated 50 random pedestrians from different scenarios, and we manually annotated them with their height in pixels units. At this point, it has been possible to recover the value of $\alpha$ from \ref{mat:find_alpha} performing a regression to find the best fit.

The height padding depends basically only on the distance of the pedestrian from the camera. Instead, the width is also linked to the specific pedestrian pose. However, we found that we can ignore these pose-dependent effects while still obtaining an excellent estimate by deriving the pedestrian width $w_m$ assuming no changes in the original bounding box aspect ratio. For this reason, we derived $w_m$ from the computed $h_m$ as follows:

\begin{equation}
    w_m = h_m \frac{w_s}{h_s} = h_m r,
\end{equation}
\noindent where $r$ is the aspect ratio of the bounding box enclosing the skeleton. Some examples of final estimated bounding boxes are shown in blue in \ref{dataset_bbs}.

\begin{figure}
    \centering
    \includegraphics[width=0.62\linewidth]{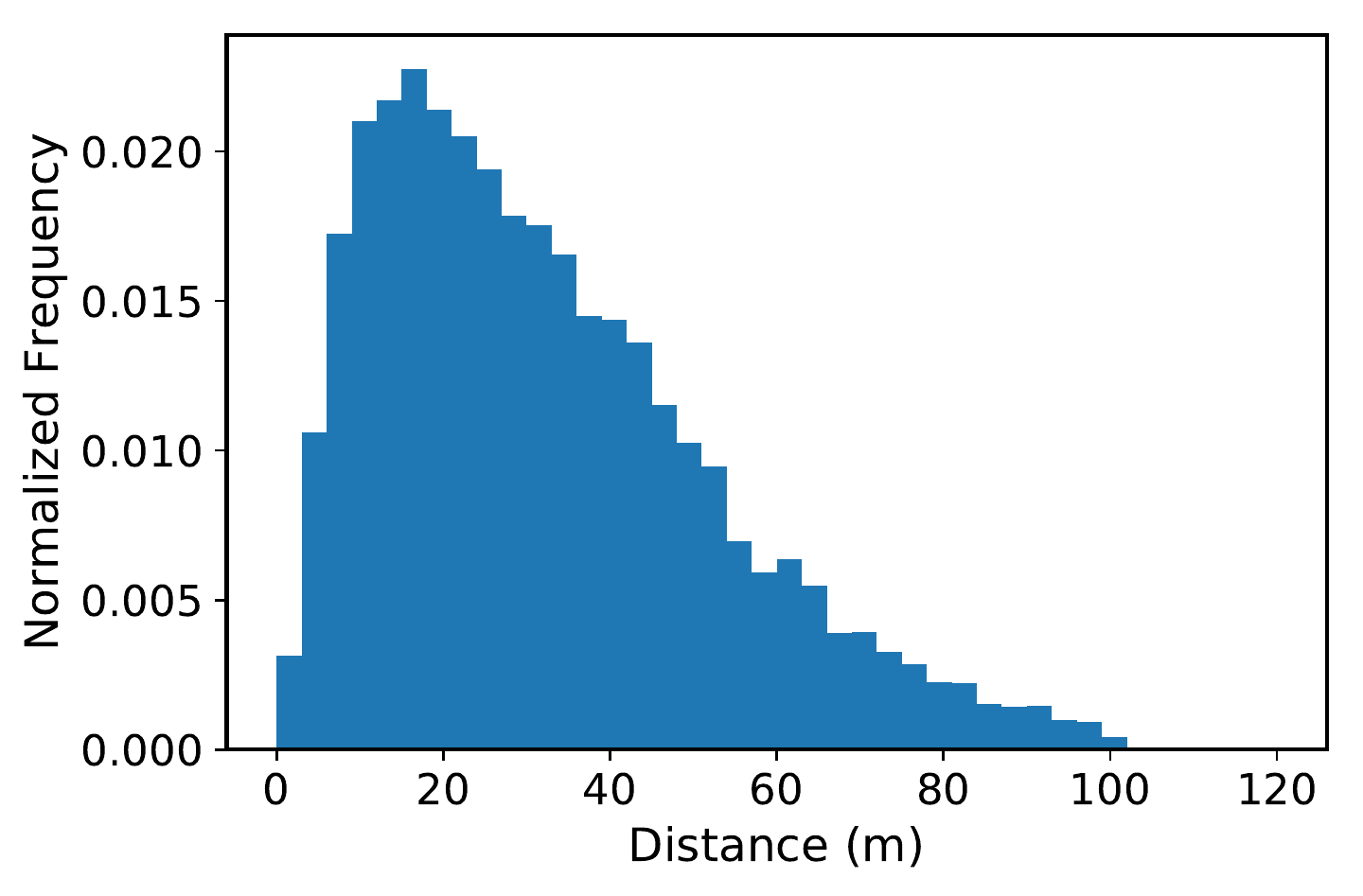}
  \caption{\textbf{Histogram of distances between pedestrians and cameras.} We show the distribution of the distances of the pedestrians from the camera characterizing our \acrfull{viped}.}
  \label{distance_hist} 
\end{figure}

We then assessed the quality of the produced bounding boxes. In \ref{distance_hist}, we report
a histogram depicting the distribution of the distances of the pedestrians from the camera. We observed that human annotators tend not to annotate pedestrians far more than a certain amount from the camera in real-world datasets. We computed this distance limit by finding the minimum bounding box height, in pixels, occurring in human annotations of the MOT17Det \cite{mot17_dataset} dataset, and seeing at what distance from the camera we reached this bounding box limit size on the \acrshort{jta} annotations. We concluded that human annotators did not include bounding boxes for pedestrians farther than 30--40 meters from the camera. Then, to be consistent with real-world datasets on which we will validate our approach, we cleaned the produced bounding boxes by pruning all the ones enclosing pedestrians farther than 40 meters.

\begin{figure}
    \centering
  \subfloat{
       \includegraphics[width=0.48\linewidth]{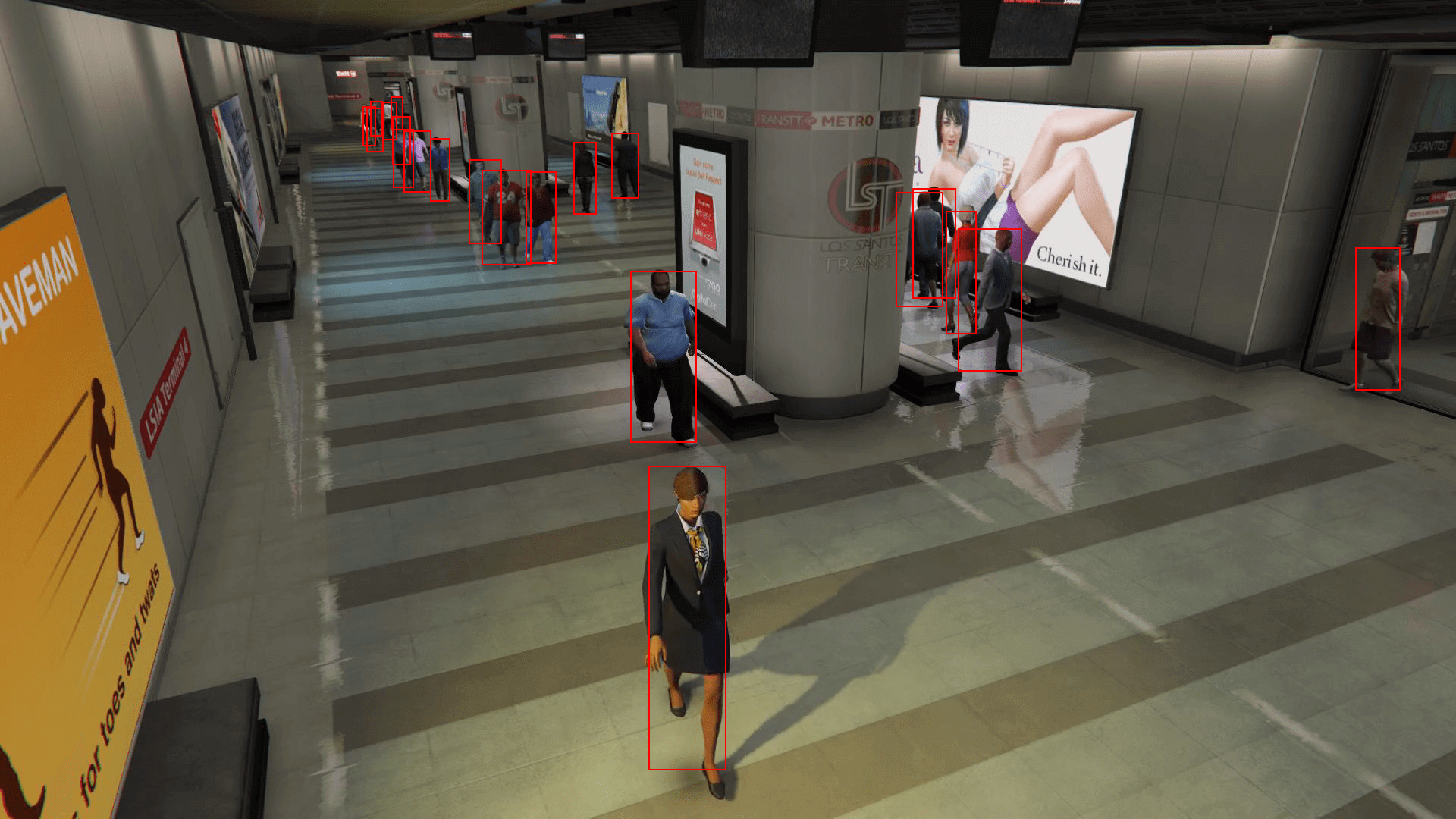}}
    \hfill
  \subfloat{
        \includegraphics[width=0.48\linewidth]{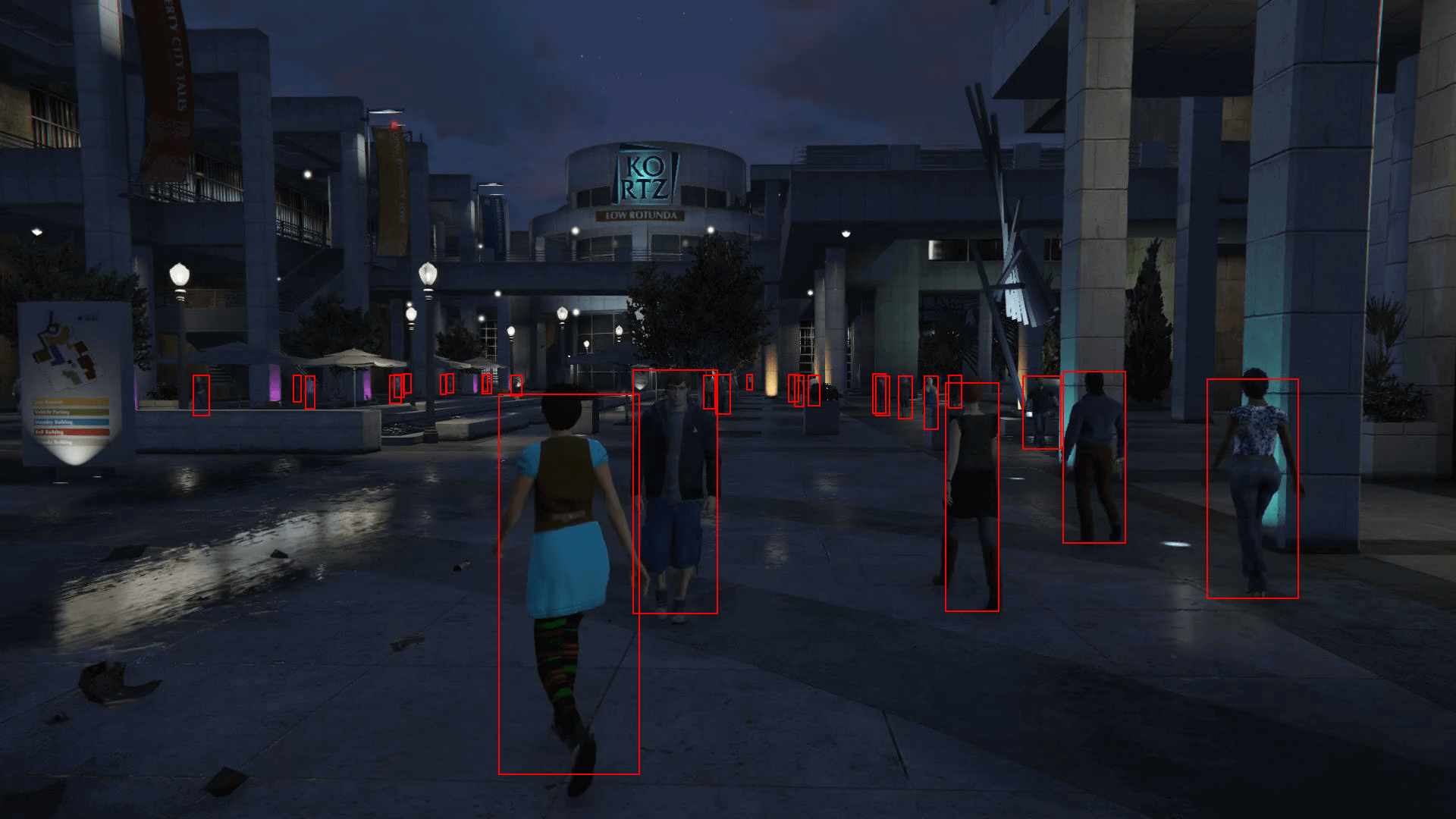}}
    \\ [2ex]
  \subfloat{
        \includegraphics[width=0.48\linewidth]{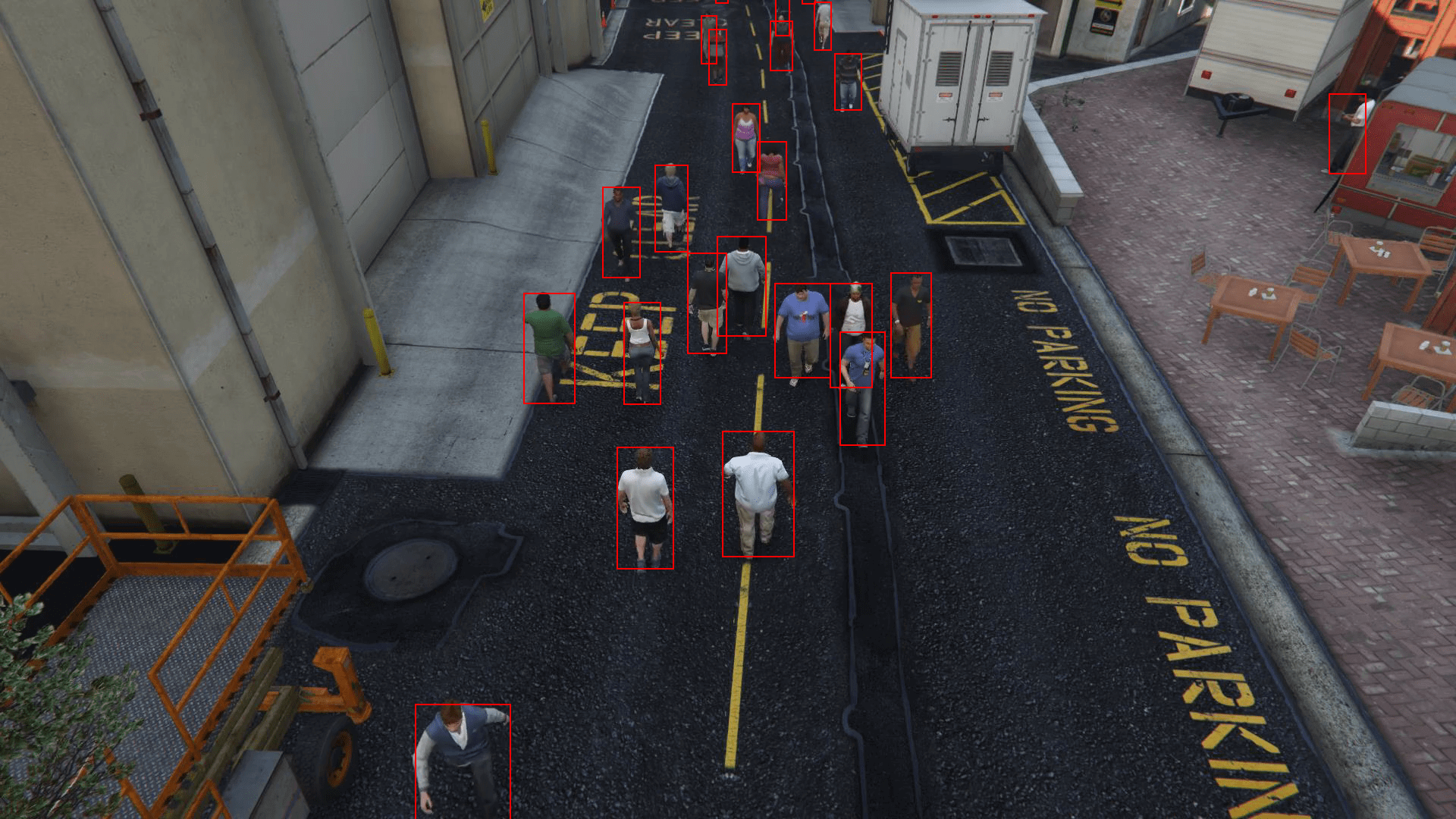}}
    \hfill
  \subfloat{
        \includegraphics[width=0.48\linewidth]{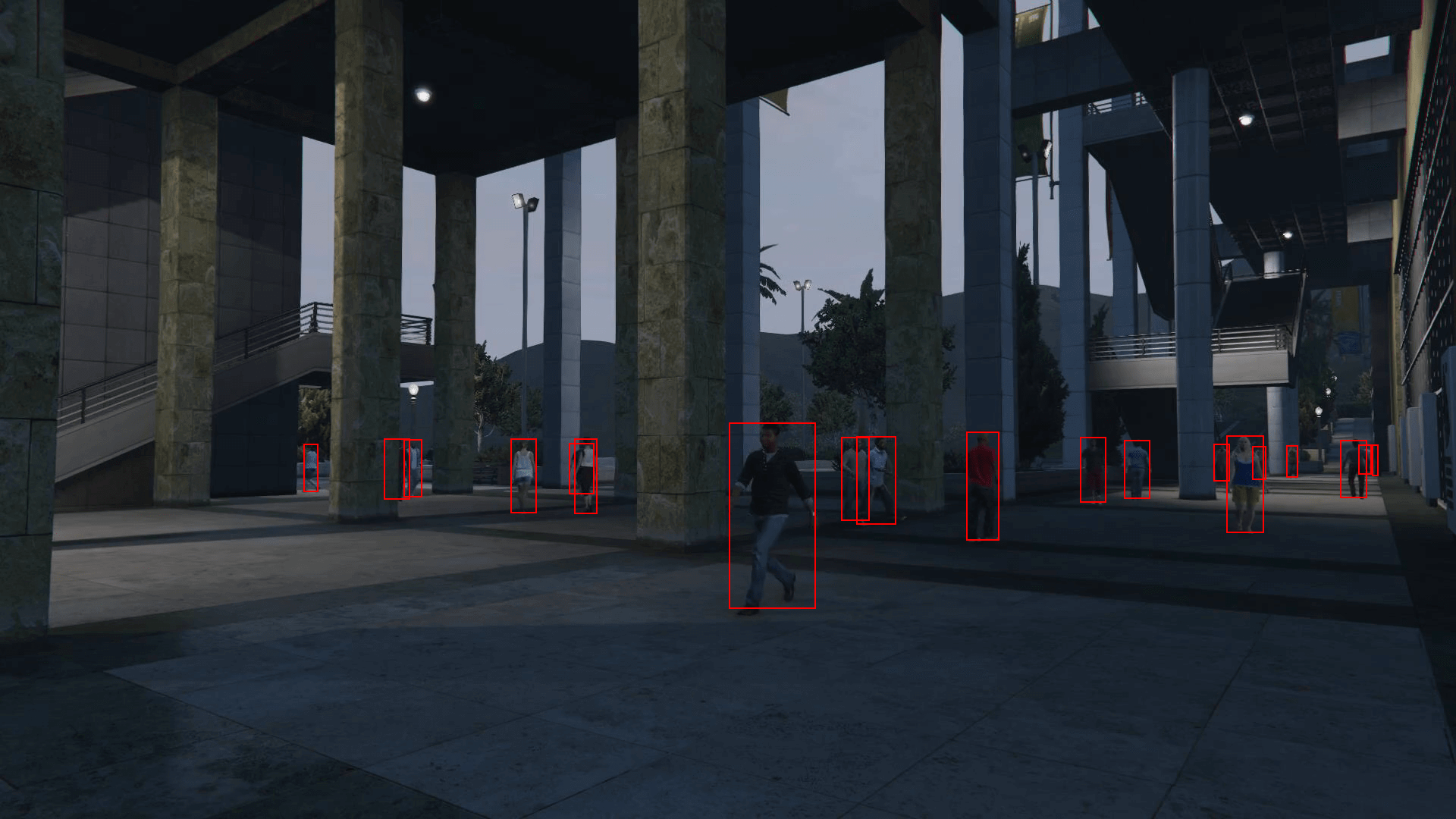}}
    \\ [2ex]
  \subfloat{
        \includegraphics[width=0.48\linewidth]{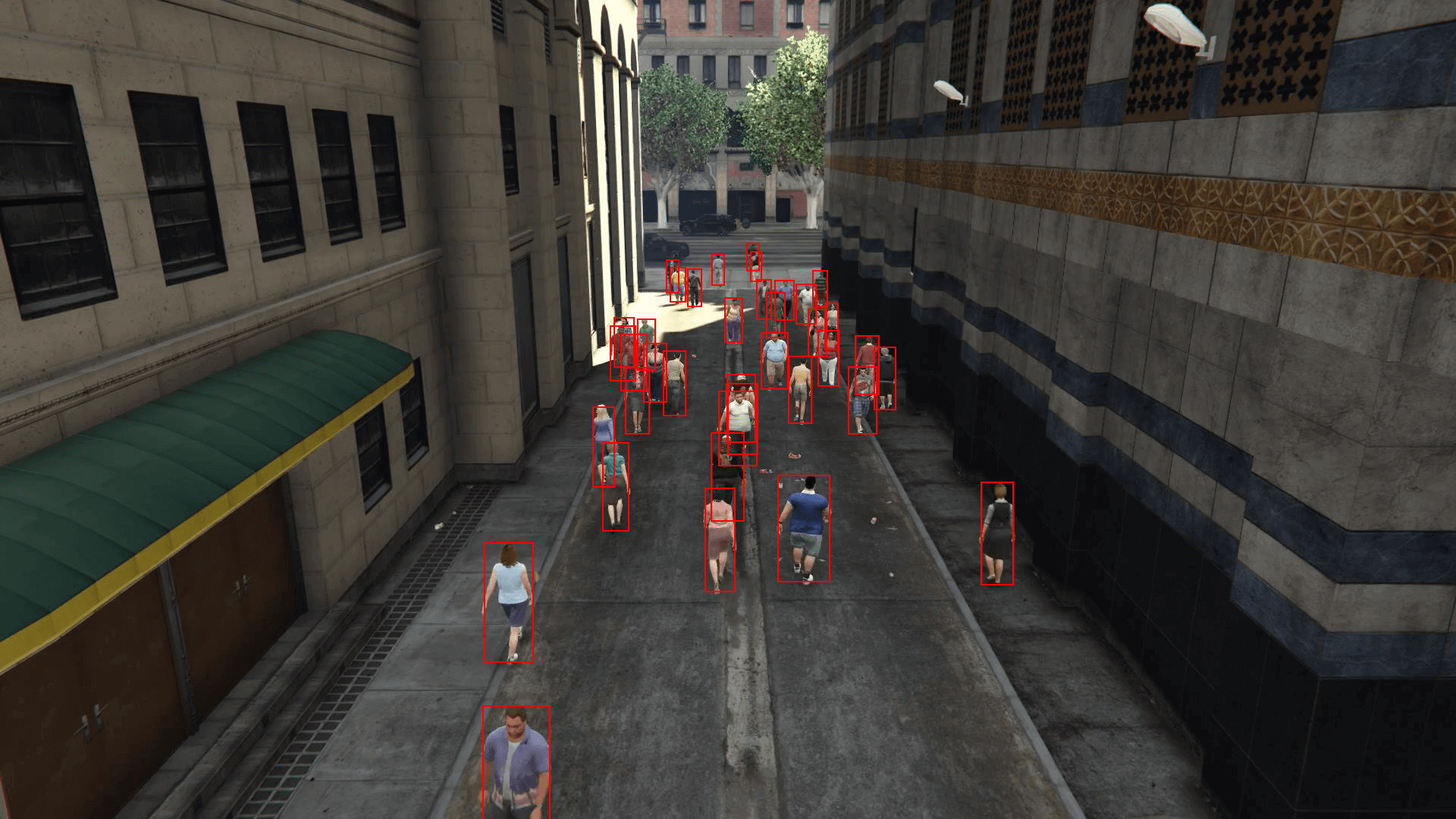}}
    \hfill
  \subfloat{
        \includegraphics[width=0.48\linewidth]{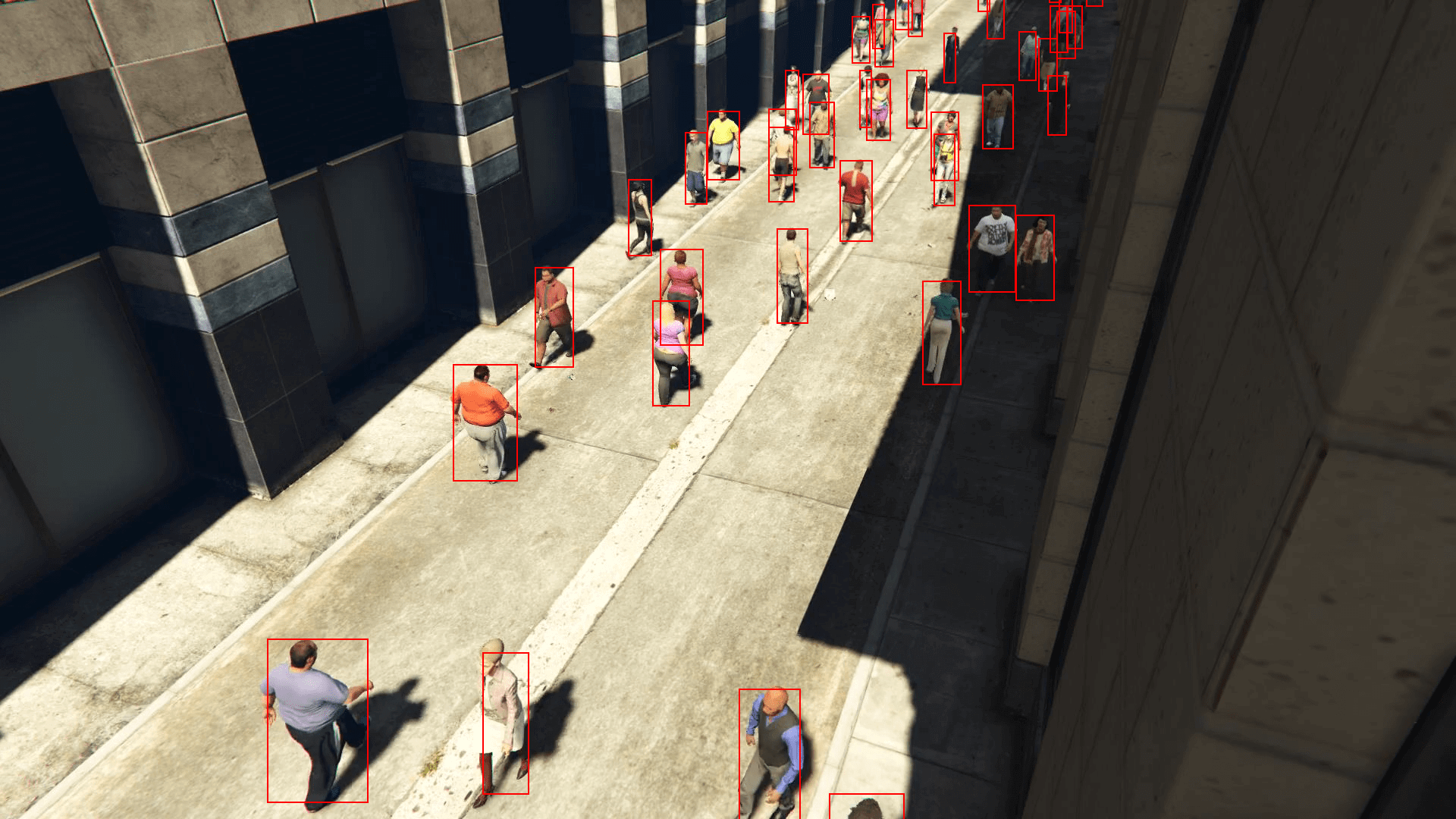}}
    \\ [2ex]
  \subfloat{
        \includegraphics[width=0.48\linewidth]{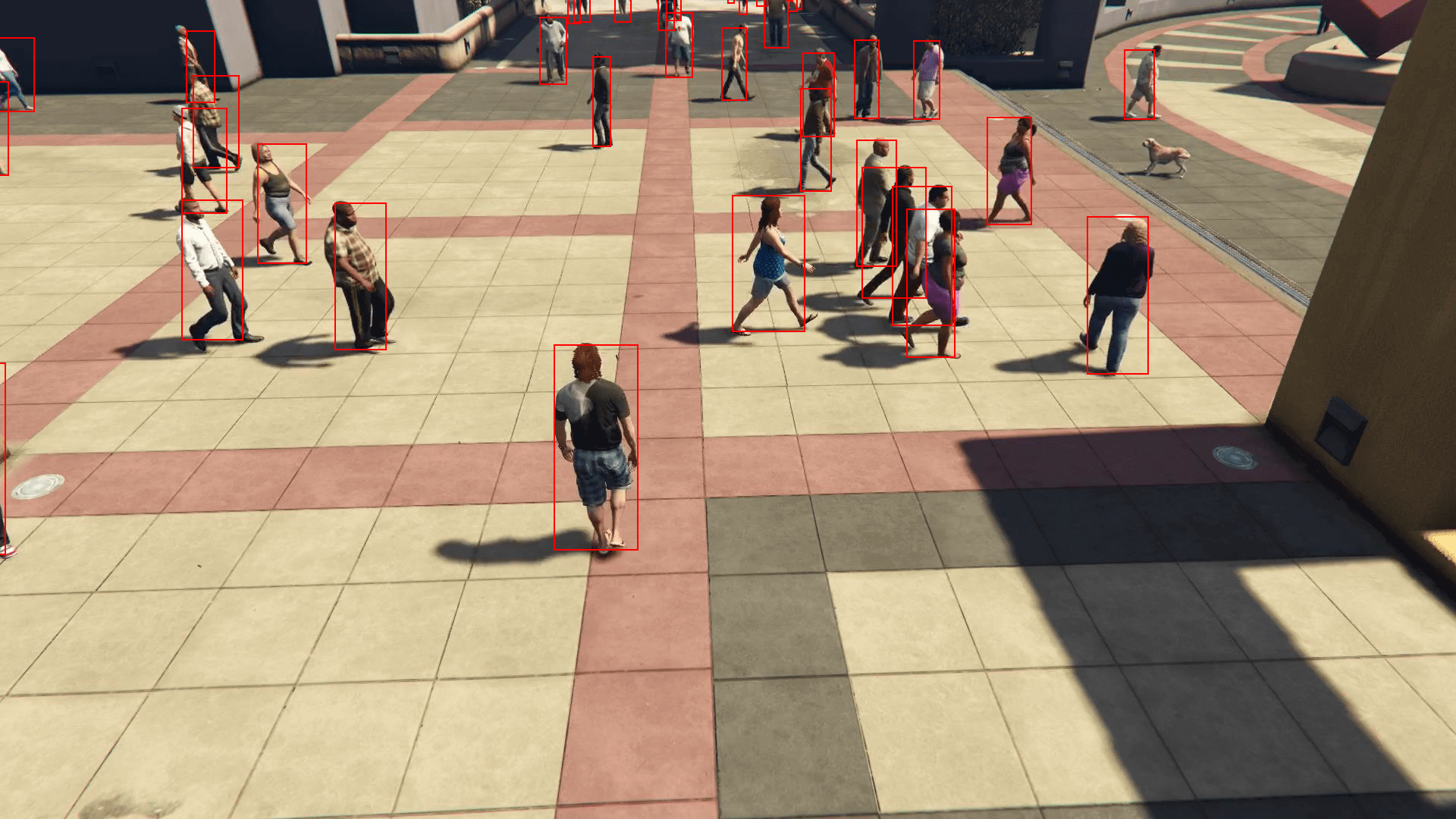}}
    \hfill
  \subfloat{
        \includegraphics[width=0.48\linewidth]{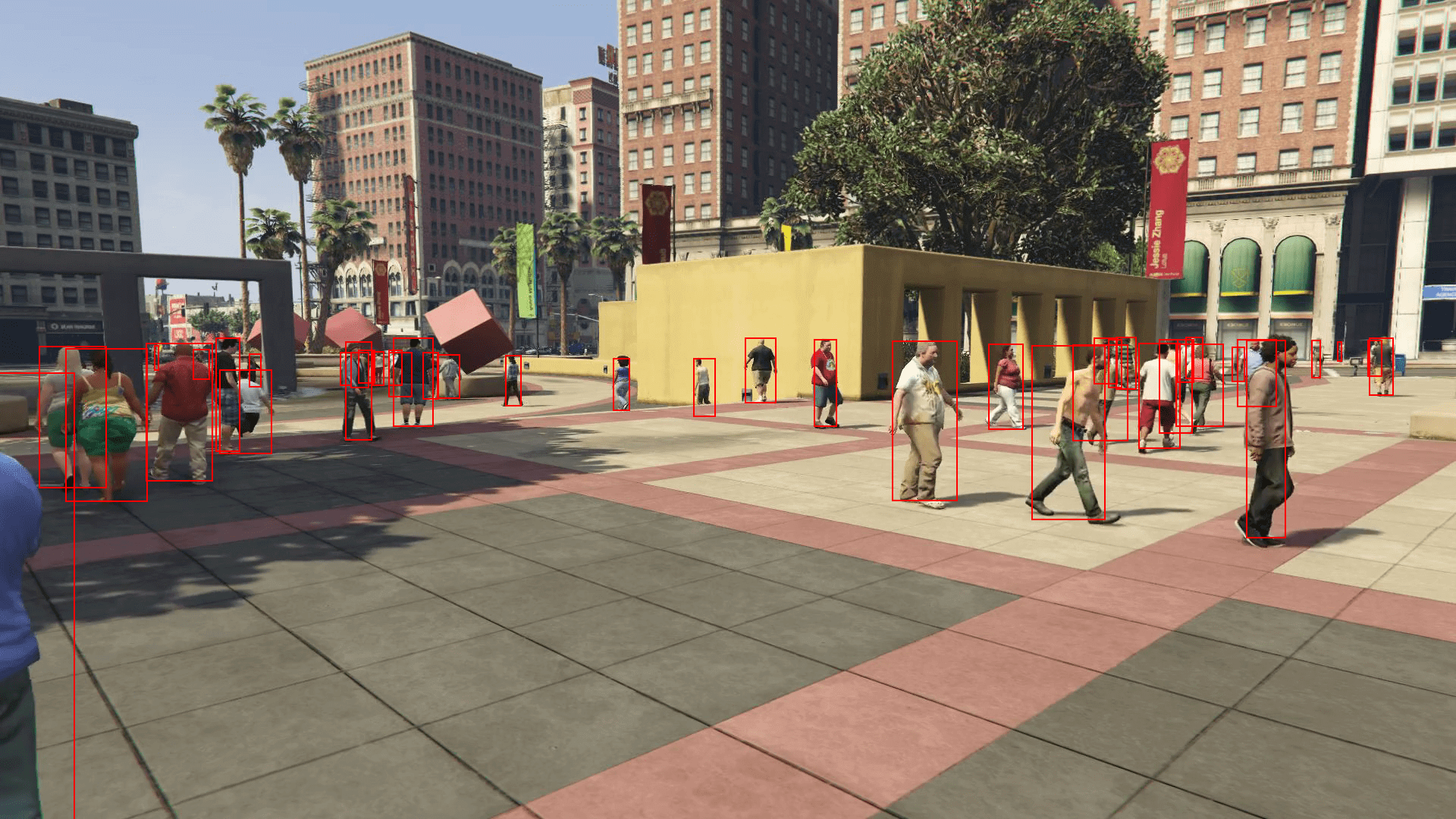}}
    \\ [2ex]
  \subfloat{
        \includegraphics[width=0.48\linewidth]{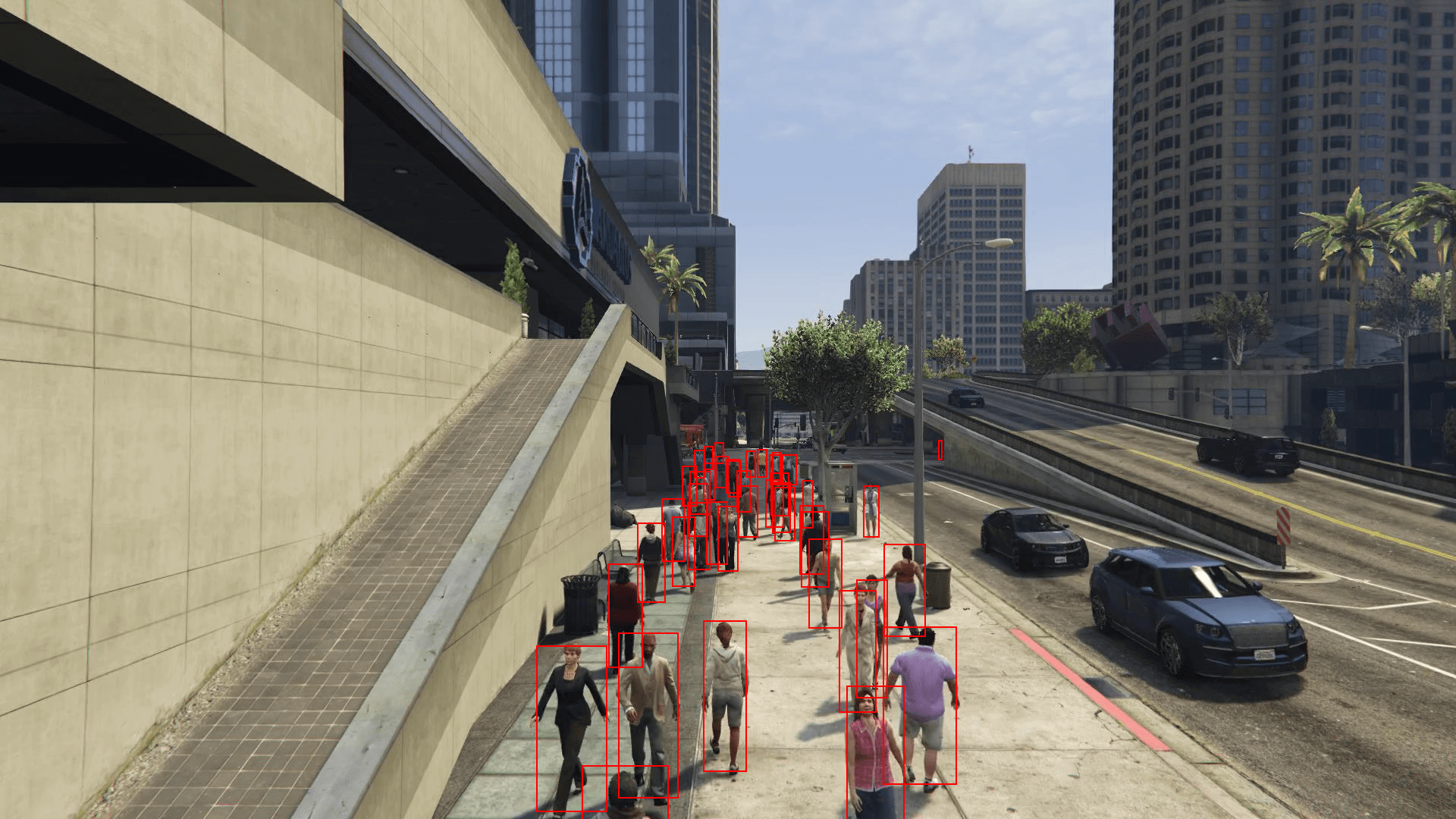}}
    \hfill
  \subfloat{
        \includegraphics[width=0.48\linewidth]{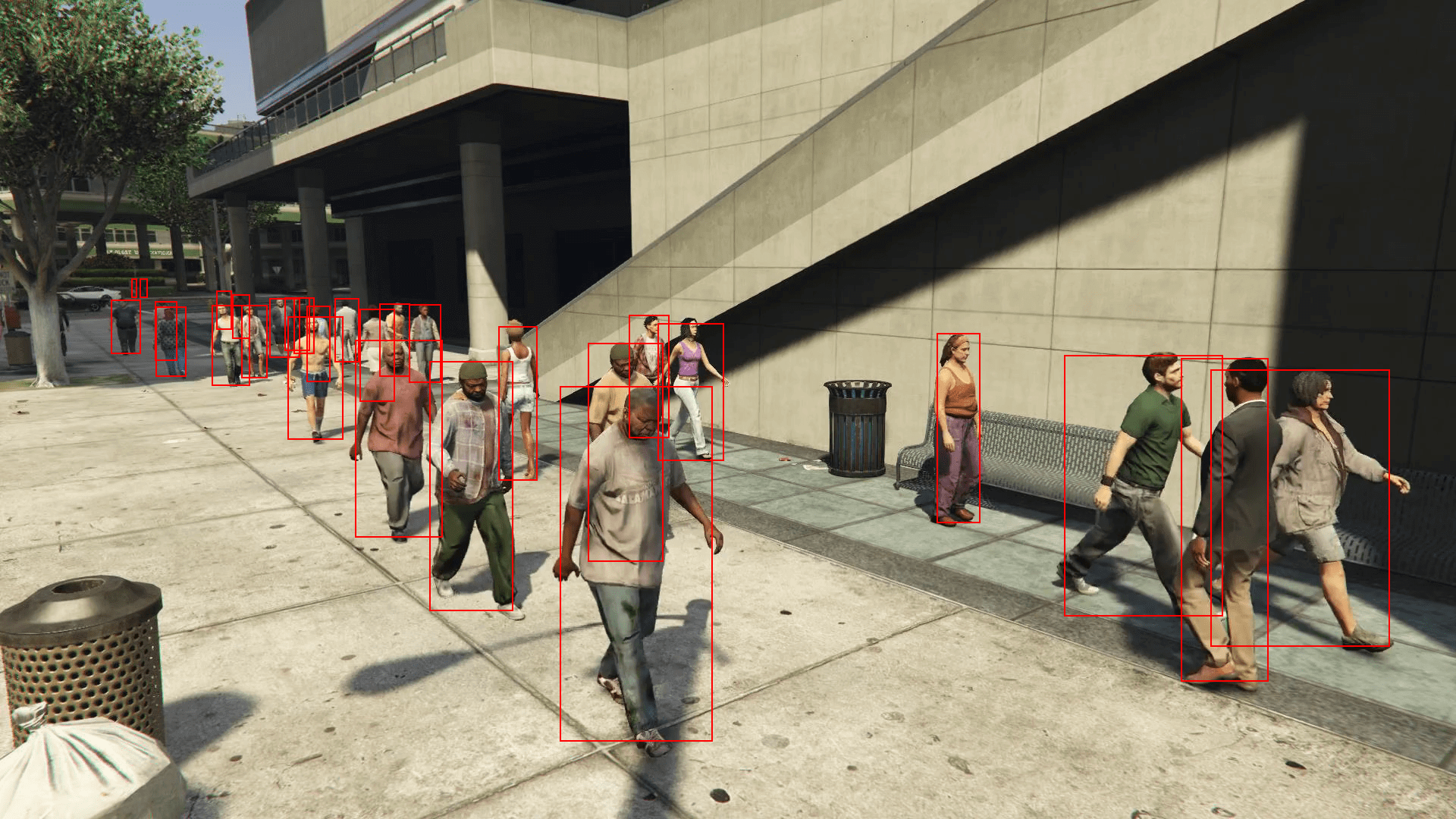}}

  \caption{\textbf{Samples of \acrfull{viped}.} We show some images of our synthetic dataset together with the sanitized bounding boxes localizing pedestrians..}
  \label{viped_examples} 
\end{figure}

In \ref{viped_examples}, we report some examples of images of the \acrshort{viped} dataset together with the sanitized bounding boxes.

\section{Domain Adaptation for Synthetic2Real Pedestrian Detection}
\label{sec:virtual-to-real:method}
In this section, we describe the object detector and the domain adaptation strategies we employed in this research. We exploited Faster R-CNN \cite{faster_rcnn}, a widely used state-of-the-art object detector that we briefly review in \ref{sec:virtual-to-real:method:Faster_RCNN} and that it is also described more in detail in \ref{sec:back:cnn-based-detectors:two-stages-detectors}.  
We trained this \acrshort{cnn} using \acrshort{viped}, our collection of synthetic images automatically annotated, already outlined in the previous section. To mitigate the existing domain shift between these data and the real-world ones, we proposed two supervised \acrlong{da} techniques (for a better explanation of \acrshort{da}, see also \ref{sec:back:domain_adaptation}). The first one, described in \ref{sec:virtual-to-real:method:domain_adaptation_1}, consists of training the detector with the synthetic data and then fine-tuning it by exploiting the real-world images. In the second approach, described in \ref{sec:virtual-to-real:method:domain_adaptation_2}, we employed instead another supervised technique, called Balanced Gradient Contribution (BGC) \cite{synthia,training_constrained}, where we mixed the synthetic and the real-world data during the training phase. \ref{finetuning_arch} and \ref{mixedbatch_arch} show an overview of the two solutions.

\subsection{Faster R-CNN Object Detector}
\label{sec:virtual-to-real:method:Faster_RCNN}
We exploited Faster-RCNN \cite{faster_rcnn} as object detector architecture. Our choice fell on Faster R-CNN since it provides state-of-the-art performance. Furthermore, we did not consider pedestrian detection-specific solutions since the two proposed \acrlong{da} techniques can also be applied to other tasks, accounting for another class of objects different from the pedestrian one.

We described in detail Faster R-CNN in \ref{sec:back:cnn-based-detectors:two-stages-detectors}. It is a two-stage \acrshort{cnn}-based algorithm composed of different networks: the backbone, the \acrfull{rpn}, and the Evaluation Network (EN). In the first stage, a \acrshort{cnn} acts as a backbone, extracting the input image features. Starting from these features space, the \acrshort{rpn} is in charge of generating region proposals that might contain objects. Briefly, \acrshort{rpn} slices pre-defined region boxes (called anchors) over this space and ranks them, suggesting the ones most likely containing objects. Once \acrshort{rpn} produces the Regions Of Interests (ROIs), they might be of different sizes. Since it is hard to work on features having different sizes, \acrshort{rpn} reduces them into the same dimension using the Region of Interest Pooling algorithm. The EN finally processes these fixed-size proposals, responsible for classifying and locating the objects inside them. Then, given an input image, the EN network's final outputs are class scores and bounding box coordinates.

Faster R-CNN is then a versatile and modular network in which it is possible to change the building blocks. Regarding the backbone, our choice fell on the ResNet-50 network, a lighter version of the very popular ResNet-101 network \cite{resnet}. Indeed, Faster R-CNN with ResNet-50 can produce satisfactory detection results compared to the low computational resources and the time required during the training and test phases. 

\subsection{Domain Adaptation using Real-World Fine-Tuning}
\label{sec:virtual-to-real:method:domain_adaptation_1}
The first proposed \acrshort{da} solution relies on a \acrfull{tl} strategy. As pointed out in \cite{deep_da_survey} and in \ref{sec:back:domain_adaptation}, \acrshort{da} is a particular \acrfull{tl} case that employs labeled data in one or more relevant source domains to execute the task in a target domain. In particular, the crucial point in this methodology consists of fine-tuning a previously trained model with the target-domain data. 

We divided our fine-tuning methodology into two different steps. In the first step, we considered as the baseline the Faster R-CNN detector described in the above section, having a ResNet-50 backbone pre-trained on the COCO dataset \cite{lin2014microsoft}, a large collection of images depicting complex everyday scenes of ordinary objects in their natural context, divided into 80 different categories. Since this network is a generic object detector that can distinguish between many different classes of objects, we modified the EN building block to adapt the model to our purposes. In particular, we reduced the last fully connected layers of the detector to recognize and locate object instances belonging only to a specific category, i.e., the pedestrian category. Then, we trained this modified Faster R-CNN-based network exploiting our synthetic images of \acrshort{viped}, leaving all the model weights unfrozen during this phase so that the back-propagation algorithm can tune them.

\begin{figure}
    \centering
    \includegraphics[width=0.98\linewidth]{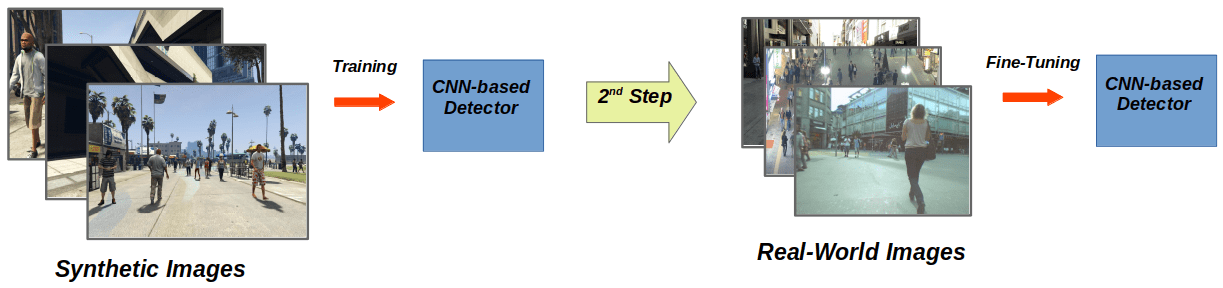}
  \caption{\textbf{Overview of the first domain adaptation technique.} In a first step, we train the detector using \acrshort{viped}, our synthetic collection of images. Then, in a second step, we fine-tune the network using real-world images.}
  \label{finetuning_arch} 
\end{figure}

In the second step, we fine-tuned this pre-trained model using real-world images as the target domain in the second step. So, in the end, the network will have processed both source and target images, memorizing in its weights information from both the domains. \ref{finetuning_arch} shows an overview of this approach. This method, looking at real-world images in this last step, is beneficial for boosting the performance of the detector in a specific real-world target scenario.

\subsection{Domain Adaptation using Balanced Gradient Contribution}
\label{sec:virtual-to-real:method:domain_adaptation_2}
The second \acrshort{da} approach is an end-to-end training, so it benefits from not relying on a two-step process like the previous one. 

\begin{figure}
    \centering
    \includegraphics[width=0.95\linewidth]{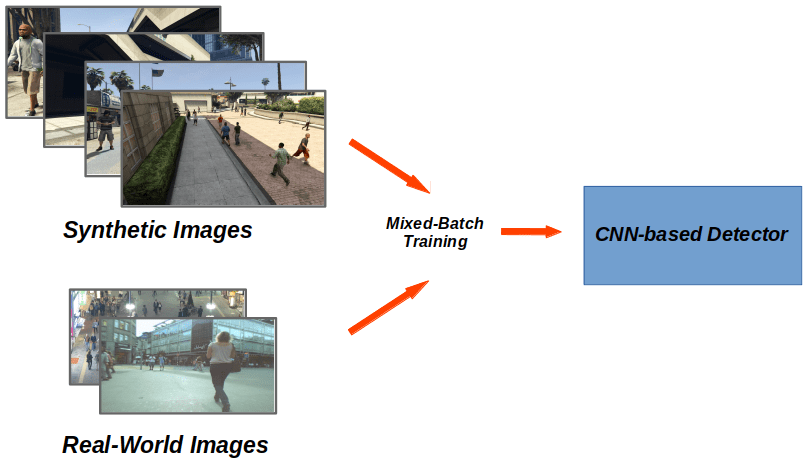}
  \caption{\textbf{Overview of the second domain adaptation technique.} We mitigate the {Synthetic2Real} domain shift in a single-step training procedure, employing mixed batches containing both synthetic and real images at the same time.}
  \label{mixedbatch_arch} 
\end{figure}

As in the previous solution, we started with the modified Faster R-CNN detector having the ResNet-50 backbone pre-trained on the COCO dataset. This time, we trained the network using mixed batches, i.e., we employed batches containing synthetic and real-world images simultaneously, given a fixed mixing ratio. As explained in \cite{training_constrained}, the real-world data acts as a regularization term over the synthetic data training loss. In particular, we exploited batches composed of $2/3$ of synthetic images and $1/3$ of real-world data. Thus, statistics from both domains are considered throughout the entire procedure, creating a more accurate model for both. Again, during this phase, we left all the network weights unfrozen so that the back-propagation algorithm could modify the network parameters accordingly. Consequently, we mitigated the Synthetic2Real Domain Shift straight in a single-step training. \ref{mixedbatch_arch} shows an overview of this approach.

In the experiments, we employed this technique for boosting the performance of the detector on a precise target scenario, using batches composed of the synthetic data and the real-world images specific for the particular considered scenes. Besides, we exploited this solution also for achieving wider generalization capabilities, considering batches containing synthetic images and generic real-world images containing pedestrians.

\section{Experimental Evaluation}
\label{sec:virtual-to-real:experiments}
In this section, we briefly report some details about the real-world datasets exploited for the experiments. Then, we show and discuss the results concerning the generalization capabilities of our detector trained using \acrshort{viped}. Finally, we illustrate the performance of the two \acrlong{da} techniques over specific real-world scenarios.

\subsection{Real-World Datasets}
\label{sec:virtual-to-real:experiments:mot_datasets}
\textit{MOT17Det} \cite{mot17_dataset} and \textit{MOT19Det} \cite{mot20_dataset} are datasets for pedestrian detection that are particularly suitable for surveillance applications. They comprise a collection of challenging images (5,316 and 8,931, respectively) taken from multiple sequences with various crowded scenarios having different viewpoints, weather conditions, and camera motions. The authors provided training and test subsets, but they released only the ground-truth annotations belonging to the former. The performance metrics concerning the test subsets are instead available submitting results to their \textit{MOT Challenge} website ({\url{https://motchallenge.net}}). The main peculiarity of MOT19Det compared to MOT17Det is the massive crowding of the collected scenarios.

\textit{CityPersons} dataset \cite{citypersons_dataset} consists of a large and diverse set of stereo video sequences recorded in streets of different cities in Germany and neighboring countries. In particular, the authors provided 5,000 images from 27 cities labeled with bounding boxes and divided across train/validation/test subsets. This dataset is more focused on self-driving applications, and images are collected from a moving car.

\textit{COCOPersons} dataset is a split of the popular COCO dataset \cite{lin2014microsoft} comprising images collected in general contexts belonging to 80 categories. We filtered these images considering only the ones belonging to the \textit{persons} category. Hence, we obtained a new dataset of about 66,000 images containing at least one pedestrian instance.

\subsection{Experiments}
We evaluated the detection performances using the standard \acrfull{map} metric (see \ref{sec:back:cnn-based-detectors:metrics} for a more detailed explanation). In particular, we considered the detection proposals having a score confidence greater than 0.05. Then, we employed the COCO \acrshort{map} \cite{lin2014microsoft} and the MOT AP metrics \cite{mot17_dataset}, fixing the \acrshort{iou} threshold to 0.5 and varying only the detection confidence threshold.

\subsubsection{Testing Generalization Capabilities}
To test the generalization capabilities, we trained the detector on a source domain, and then we validated it on a different target domain. In particular, we trained the model using a dataset, and then we tested it on another one. In this way, we guaranteed that the two distributions were different and not related.

In particular, we trained the modified Faster R-CNN-based detector described in \ref{sec:virtual-to-real:method:Faster_RCNN} using \acrshort{viped}. This procedure corresponds to the first step of the previously described domain adaptation solution (see \ref{sec:virtual-to-real:method:domain_adaptation_1}). We evaluated this model using the real-world datasets MOT17Det, MOT19Det, and CityPersons, defining three subsets containing images not present in the training subset.
To form a solid baseline for this experiment, we trained the same detector using every one of the three real-world datasets, and then we tested them over the remaining two. We also reported a further baseline considering the detector trained only on the real-world general-purpose COCO dataset, considering only the detections belonging to the \textit{person} category.

We also experimented with the mixed-batch \acrshort{da} approach explained in \ref{sec:virtual-to-real:method:domain_adaptation_2}, using the same evaluation protocol as before. We exploited batches composed of $2/3$ of \acrshort{viped} and by the remaining $1/3$ of COCOPersons. We chose the latter as the real-world dataset since it depicts humans in highly heterogeneous scenarios, and it is not biased towards a specific application (e.g., autonomous driving). Again, we evaluated this model testing on all the three remaining real-world datasets.

We report the results in \ref{virtual-to-real:tab1}. Note that we omit results concerning a specific dataset if employed during the training phase for a fair evaluation of the overall generalization capabilities. In most cases, as we can see, our network performed better than those trained using only the manually annotated real-world datasets, taking advantage of the high variability and size of the \acrshort{viped} dataset. In particular, concerning the MOT17Det dataset, all our solutions trained with synthetic data outperformed those trained with real ones. We obtained the best results using the mixed-batch approach. Considering the MOT19Det dataset, we achieved the best result in training the detector with \acrshort{viped}. CityPersons is the only dataset on which the algorithm maintained higher performances when trained with real-world data. In particular, the highest \acrshort{map} on CityPersons is obtained when the detector is trained with the MOT17Det dataset. However, in this case, the mixed-batch approach achieved results comparable with the baselines.

\begin{table}[t]
\caption{\textbf{Evaluation of the generalization capabilities.} The first section of the table reports results obtained training the detector with real-world data, while the latter is related to the model trained over synthetic images. \acrshort{viped} + Real refer to the mixed batch experiments with $2/3$ \acrshort{viped} and $1/3$ of \textit{COCOPersons}. Results are evaluated using the COCO mAP. We report in bold the best results.}
\begin{center}
\begin{tabular}{lccc}
\toprule
& \multicolumn{3}{c}{\textbf{Test Dataset}} \\
\cmidrule(lr){2-4}
\textbf{Training Dataset} & MOT17Det & MOT19Det & CityPersons \\
\midrule
COCO & 0.636 & 0.466 & 0.546 \\
\midrule
MOT17Det & - & 0.605 & \textbf{0.571} \\
\midrule
MOT19Det & 0.618 & - & 0.419 \\
\midrule
CityPersons & 0.710 & 0.488 & - \\
\midrule
\acrshort{viped} & 0.721 & \textbf{0.629} & 0.516 \\
\midrule
\acrshort{viped} + Real & \textbf{0.733} & 0.582 & 0.546 \\
\bottomrule
\end{tabular}
\label{virtual-to-real:tab1}
\end{center}
\end{table}

\subsubsection{Testing Domain Adaptation Techniques over Specific Real-world Scenarios}
To test how the two proposed domain adaptation techniques behave when considering specific target real-world scenarios, we considered the MOT17Det and MOT19Det real-world datasets.

Regarding the fine-tuning \acrshort{da} approach, we considered as training sets those proposed by the authors of \cite{mot17_dataset, mot20_dataset}, and we obtained the evaluation of our results over the test sets by submitting them to the {Mot Challenge} website. For the mixed-batch \acrshort{da} solution, during the training phase, we injected in the same batch $2/3$ of synthetic images from the \acrshort{viped} dataset and $1/3$ of real-world images from the training subsets of the {MOTDet17} or the {MOT19Det} dataset. Again, we validated our results by submitting them to the MOT Challenge website.

\ref{virtual-to-real:tab2} and \ref{virtual-to-real:tab3} report the results for the two considered scenarios. We report our results together with the state-of-the-art approaches publicly released in the MOT Challenges (at the time of writing). As we can see, the two \acrshort{da} approaches can mitigate the Synthetic2Real Domain Shift. In both datasets, we obtained an improvement in performance compared to the results in \ref{virtual-to-real:tab1}. It is also worth noting that we achieved competitive results in both scenarios compared to the state-of-the-art, reaching (at the time of writing) the first and the second places in the leader boards of the MOT17Det and MOT19Det challenges, respectively.

\begin{table}[t]
\caption{\textbf{Evaluation of the two \acrshort{da} techniques on the \textit{MOT17Det} dataset.} FT-DA (Fine Tuning DA) is the first proposed solution, while MB-DA (Mixed Batch DA) is the second one. Results are evaluated using the MOT mean average precision (mAP).}
\begin{center}
\begin{tabular}{ccc}
\toprule
\textbf{Method} & \textbf{MOT AP} \\
\midrule
YTLAB \cite{ped_det_8} & 0.89 \\
\midrule
KDNT \cite{poi} & 0.89 \\
\midrule
\acrshort{viped} FT-DA (our) & \textbf{0.89} \\
\midrule
\acrshort{viped} MB-DA (our) & 0.87\\
\midrule
ZIZOM \cite{Lin2018} & 0.81 \\
\midrule
SDP \cite{ped_det_7} & 0.81 \\
\midrule
FRCNN \cite{faster_rcnn} & 0.72 \\
\bottomrule
\end{tabular}
\label{virtual-to-real:tab2}
\end{center}
\end{table}

\begin{table}[t]
\caption{\textbf{Evaluation of the two \acrshort{da} techniques on the \textit{MOT19Det} dataset.} FT-DA (Fine Tuning DA) is the first proposed solution, while MB-DA (Mixed Batch DA) is the second one. Results are evaluated using the MOT mAP.}
\begin{center}
\begin{tabular}{ccc}
\toprule
\textbf{Method} & \textbf{MOT AP} \\
\midrule
SRK\_ODESA & \textbf{0.81} \\
\midrule
CVPR19\_det & 0.80 \\
\midrule
Aaron & 0.79 \\
\midrule
PSdetect19 & 0.74 \\
\midrule
ViPeD FT-DA (our) & 0.80 \\ 
\midrule
ViPeD MB-DA (our) & 0.80 \\
\bottomrule
\end{tabular}
\label{virtual-to-real:tab3}
\end{center}
\end{table}

\section{Summary}
\label{sec:virtual-to-real:summary}
In this chapter, we addressed the pedestrian detection task by proposing a \acrshort{cnn}-based solution trained using synthetically generated data. The choice of training a \acrshort{cnn} using synthetic data is motivated by the fact that the network, to generalize well, requires a considerable amount of manually annotated images representing different scenarios. This procedure usually requires a significant human effort, and it is error-prone. Consequently, often, training of \acrshort{cnn}s has to deal with \textit{data scarcity}, resulting in performance degradation when applied to new scenarios at inference time.

To this end, we introduced a synthetic dataset named \acrfull{viped}, containing a massive collection of images rendered from the highly photo-realistic video game GTA V developed by Rockstar North and a full set of precise bounding boxes annotations around all the visible pedestrians. It is worth noting that, in this case, the labels have been \textit{automatically} collected without human intervention. To the best of our knowledge, it was the first synthetic dataset suitable for training \acrshort{cnn}-based pedestrian detectors. 
However, the potential inherent in synthetic data can only be partially exploited. Indeed, there is a domain gap between the synthetic and the real-world data because synthetic images’ appearance is still significantly different from that observed in real-world images, even using current rendering techniques.
To this end, we proposed two different supervised \acrlong{da} techniques to mitigate this Synthetic2Real Domain Shift, which are suitable for pedestrian detection and possibly applicable to more general object detection tasks.

The experiments showed that, in most cases, the detector trained with the synthetic data generalizes better on unseen scenarios than the same algorithm trained using only the manually annotated real-world datasets. Moreover, the two proposed \acrshort{da} approaches were able to mitigate the underlying differences between the two worlds, obtaining a performance improvement in specific real-world scenarios. In our opinion, the result of this research opens new perspectives to address the scalability of pedestrian and object detection methods for large physical systems with limited supervisory resources. Using our freely available model trained using \acrshort{viped}, future researchers will have at their disposal a detector able to localize instances of people over images belonging to a multitude of different scenarios and, therefore, a system robust to newly added sources of data. On the other hand, they will also have the possibility of further specializing the detector to work over newly added real-world scenarios using our two domain adaptation techniques, obtaining an additional performance boost. However, the adoption of supervised domain adaptation methods to mitigate the gap between the two data distributions has a price since it still employs labeled data from the real-world domain. In the next chapter, we will propose an \textit{unsupervised} \acrshort{da} strategy, which overcomes this problem, relying only on supervision in the synthetic domain and inferring some knowledge from test data without using the labels.

\graphicspath{{img/uda-counting/}}

\chapter{Unsupervised Domain Adaptation for Traffic Density Estimation and Counting}
\label{ch:uda-counting}

As seen in the previous chapters, with the advent of Convolutional Neural Networks (CNNs), supervised learning has reached excellent results across many Computer Vision application areas, such as object detection and counting. However, as pointed out in \ref{ch:virtual-to-real}, most CNN-based methods require a large amount of well-labeled data for supervised learning and make a common assumption: the training and testing data are drawn from the same distribution. The direct transfer of the learned features between different domains does not work very well because the distributions are different. Thus, a model trained on one domain, named \textit{source}, usually experiences a drastic drop in performance when applied on another domain, named \textit{target}. This problem is commonly referred to as the \textit{Domain Shift}. As we have seen, \acrfull{da} is a common technique to address this problem. It adapts a trained neural network by fine-tuning it with a new set of labeled data belonging to the new distribution. However, in many real cases, gathering a further collection of labeled data is expensive, especially for tasks that imply per-pixel annotations, like semantic or instance segmentation. A promising solution is to collect synthetic data where annotations are gathered automatically by the graphical engine. Still, data coming from virtual worlds cannot be fully exploited due to the Synthetic-to-Real Domain Shift, and supervised adaptations using real-world data are often needed.

\textit{\acrfull{uda}} addresses the domain shift problem differently. It does not use labeled data from the target domain and relies only on supervision in the source domain. Specifically, \acrshort{uda} takes a source labeled dataset and a target \textit{unlabeled} one. The challenge here is to automatically infer some knowledge from the target data to reduce the gap between the two domains.

This chapter proposes an end-to-end CNN-based \acrshort{uda} algorithm for traffic density estimation and counting based on adversarial learning. Adversarial learning is performed directly on the generated density maps, i.e., in the \textit{output space}, given that in this specific case, the output space contains valuable information such as scene layout and context. We focused on vehicle counting, but the approach is suitable for counting any other types of objects. To the best of our knowledge, this was the first attempt to introduce a \acrshort{uda} scheme for counting to reduce the gap between the source and the target domain without using additional labels.

\begin{figure*}[!htbp]
\centering
\begin{tabular}{cccc}
\includegraphics[width=.22\textwidth,height=2.9cm]{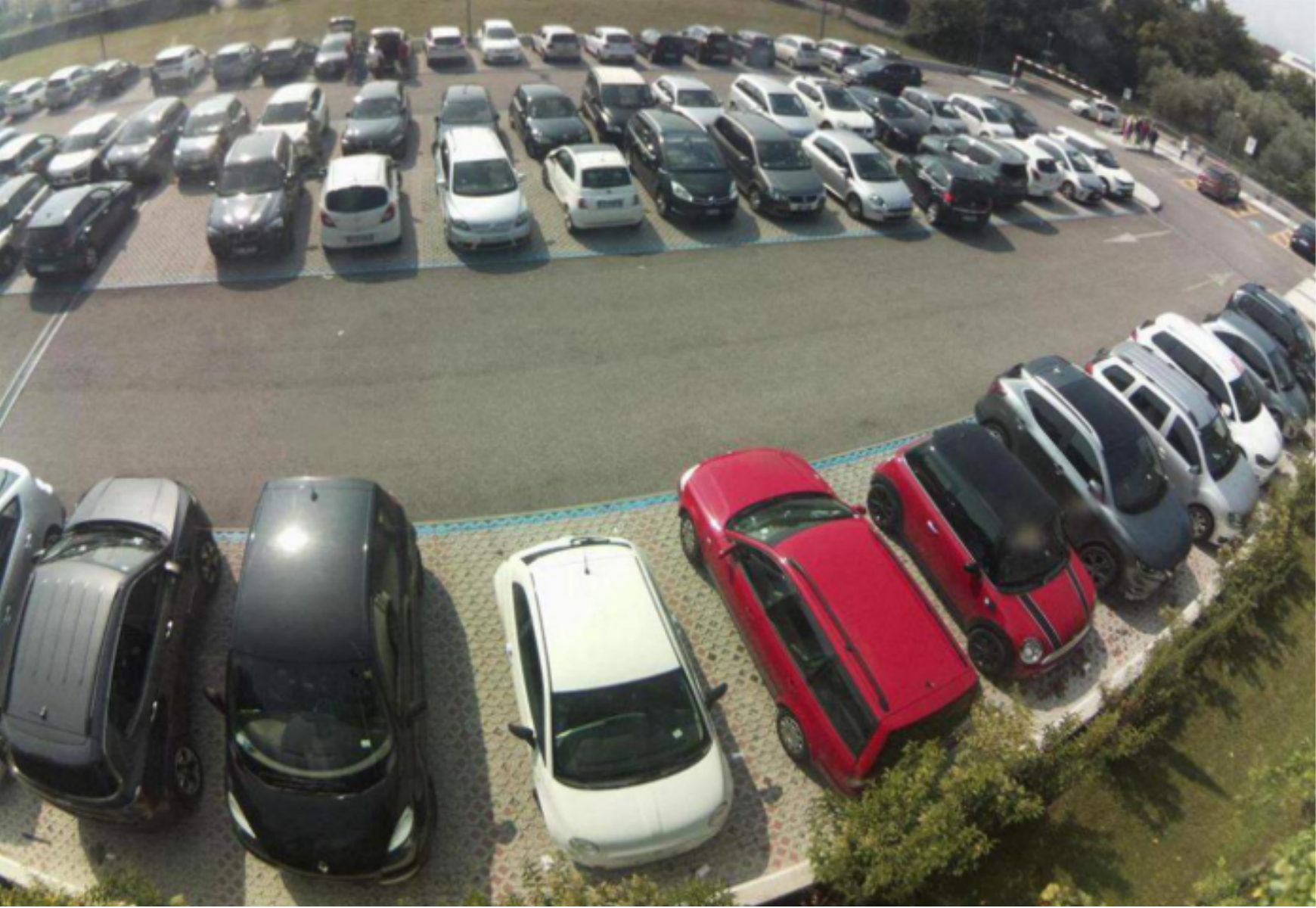} & 
\includegraphics[width=.22\textwidth,height=2.9cm]{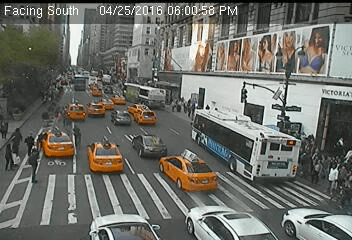} &
\includegraphics[width=.22\textwidth,height=2.9cm]{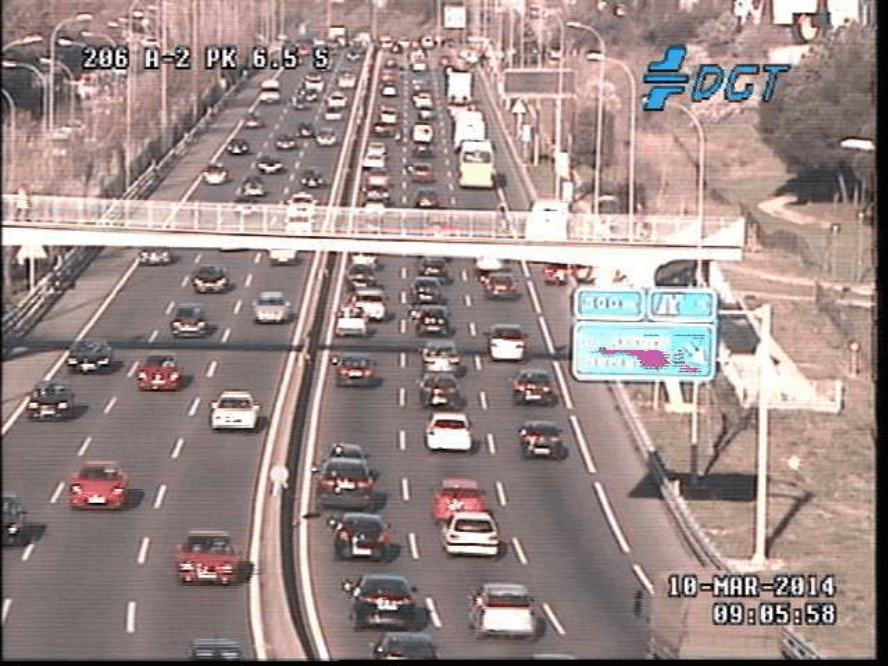} &
\includegraphics[width=.22\textwidth,height=2.9cm]{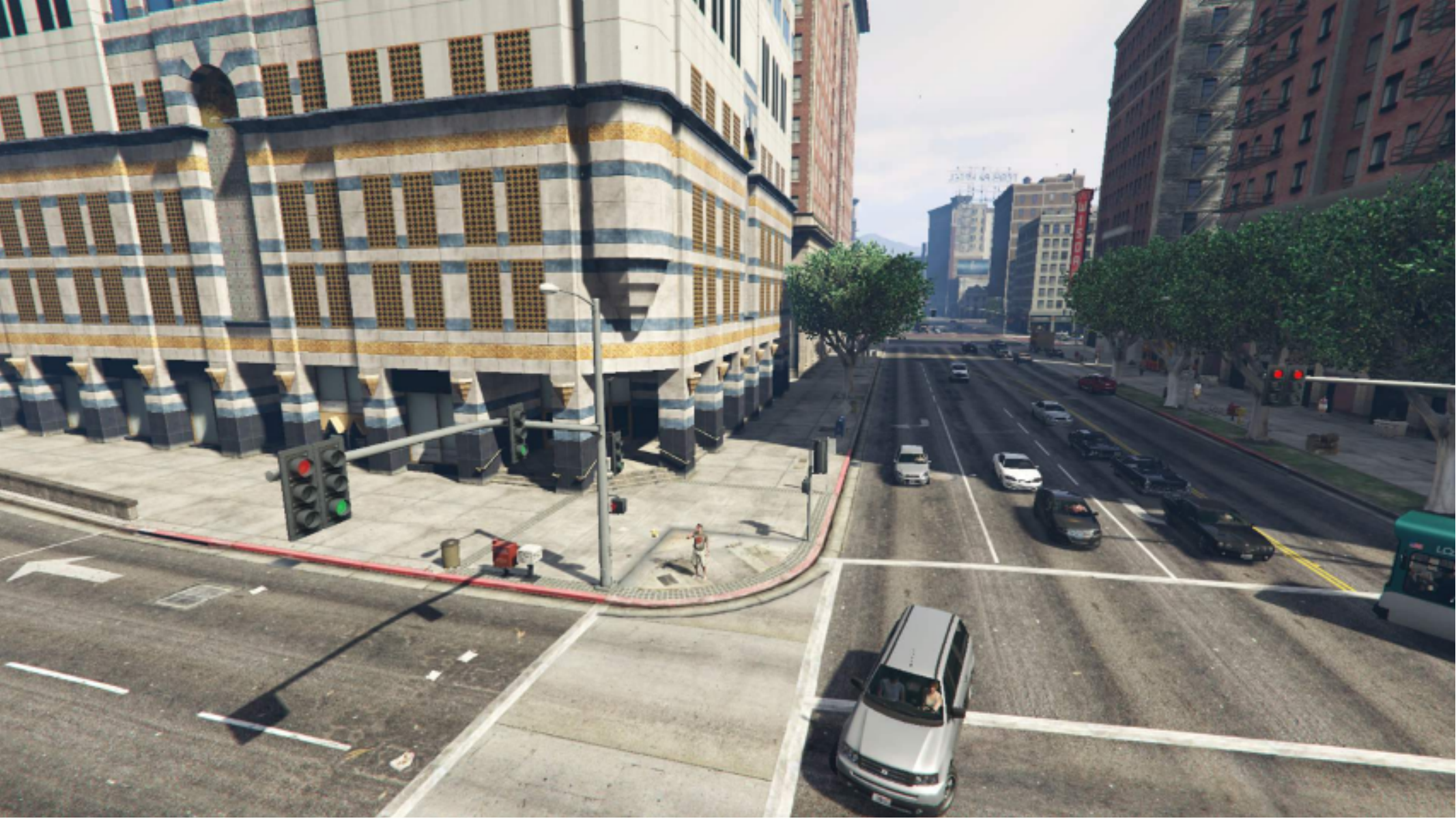} \\
\includegraphics[width=.22\textwidth,height=2.9cm]{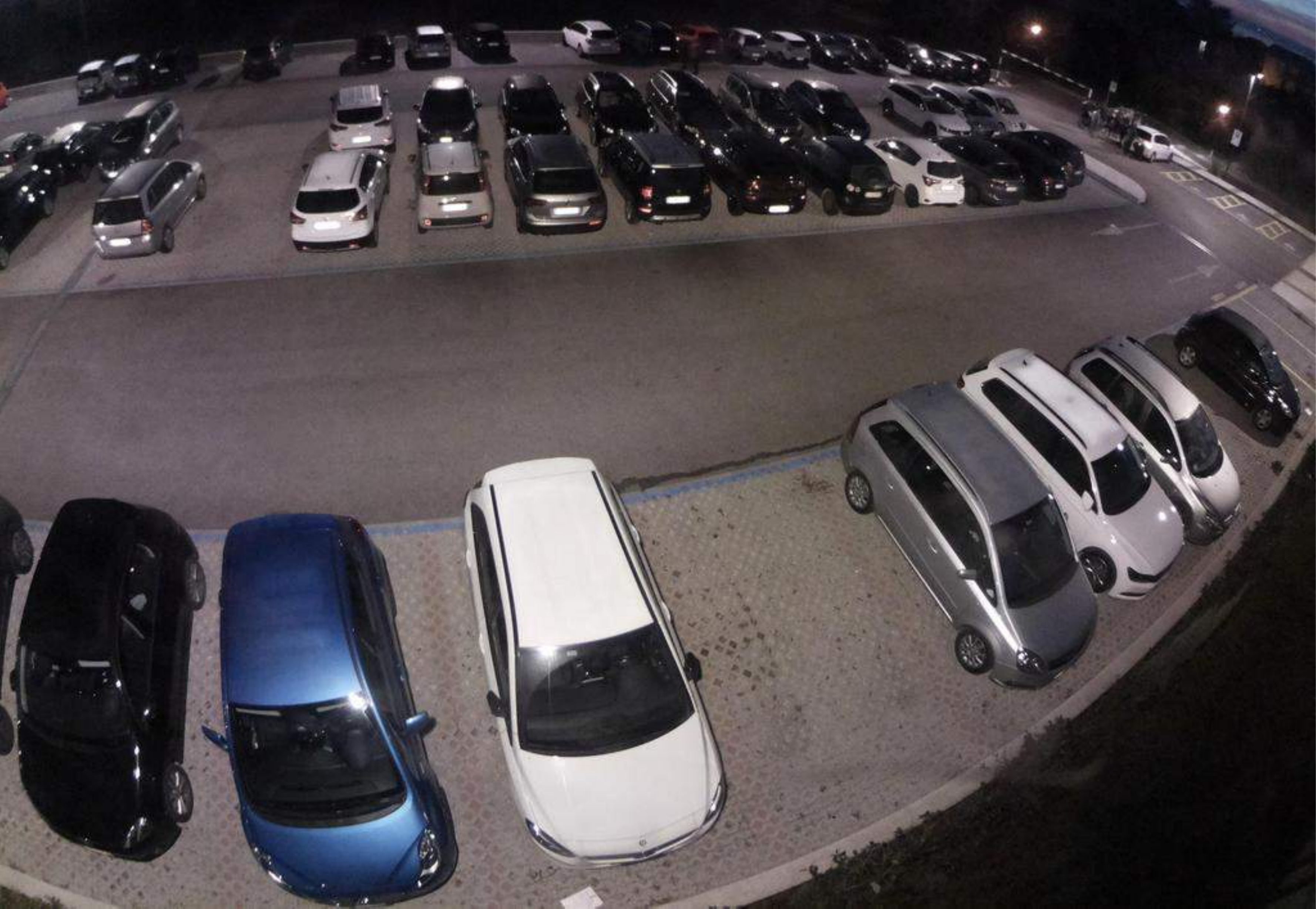} & 
\includegraphics[width=.22\textwidth,height=2.9cm]{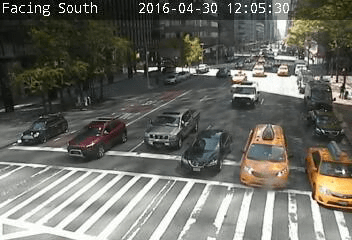} &
\includegraphics[width=.22\textwidth,height=2.9cm]{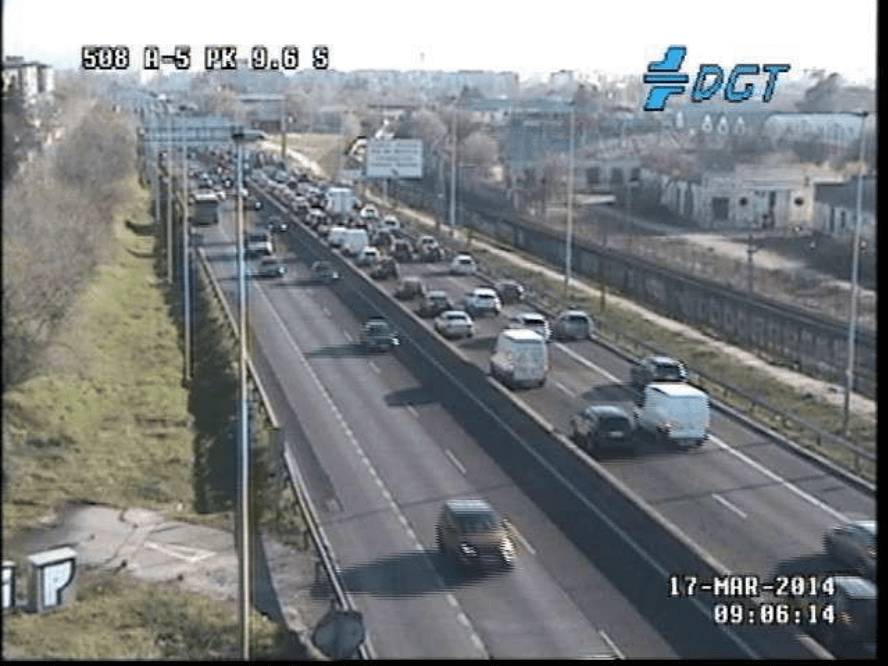} &
\includegraphics[width=.22\textwidth,height=2.9cm]{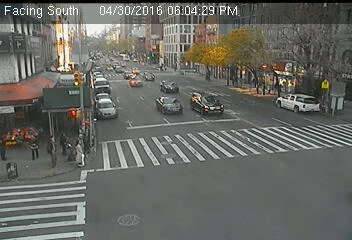} \\
(a) & (b) & (c) & (d)
\end{tabular}
\caption{\textbf{The Domain Shift scenarios that have been addressed in this work.} (a) \textit{Day2Night}; (b) and (c) \textit{Camera2Camera}; (d) \textit{Synthetic2Real}. The first row represents the labeled \textit{source} domain, while the second represents the unlabeled \textit{target} one used for our unsupervised domain adaptation.}
\label{examples_domain_gap}
\end{figure*}

We conducted experiments considering different domain shifts and validating our approach on various vehicle counting datasets. First, we employed two existing datasets for traffic density estimation, \textit{WebCamT} \cite{understandingCosteira} and \textit{TRANCOS} \cite{ExtremelyTrancos}. We used images acquired by a specific subset of cameras as the source domain. In contrast, to emphasize the domain shift problem, we represented the target domain with images captured by a different subset of cameras, seeing different perspectives and visual contexts. We called this type of domain shift \textit{Camera2Camera} domain shift. Comparisons with other techniques on these datasets showed the superiority of our approach. To test our solution with other types of domain shifts, we created and made publicly available two additional datasets, described in the following. The first one was the \textit{\acrfull{ndispark}} dataset, consisting of images captured by surveillance cameras in parking lots. Here, on the one hand, source data include annotated images collected by various cameras during the day. On the other hand, the unlabeled target domain contains images collected during the night in the same scenarios. We called this domain shift \textit{Day2Night}. The second one was instead the \textit{\acrfull{gta}} dataset, a vast collection of \textit{synthetic} images generated with the highly photo-realistic graphical engine of the \textit{Grand Theft Auto V} video game, developed by \textit{Rockstar North}. This dataset consists of urban traffic scenes, \textit{automatically} and precisely annotated with per-pixel annotations. To the best of our knowledge, this was the first \textit{instance} segmented synthetic dataset of traffic scenarios. We used this dataset to train the counting algorithm. Then, we performed \acrlong{uda} to be able to count in real-world images. In this case, the domain shift is represented by the \textit{Synthetic2Real} difference. Different from the previous chapter, in which we relied on a \textit{supervised} adaptation of the network using real-world data, here we dispensed the usage of the annotations of the target data. Using the proposed \acrshort{uda} strategy, we tackled the problem of data scarcity from two complementary sides: on the one hand, we exploited the great variability of the synthetic data, while, on the other hand, we mitigated the domain gap existing between the synthetic and the real-world data in an \textit{unsupervised} fashion. \ref{examples_domain_gap} summarizes the described domain shifts that we have addressed. In all the experiments, we showed that our \acrshort{uda} technique always outperformed the non-domain adapted models.

We summarize the main contributions presented in this chapter as follows:
\begin{itemize}
\item we introduce a \acrshort{uda} algorithm for traffic density estimation and counting, which can reduce the domain gap between a labeled source dataset and an unlabeled target one. To the best of our knowledge, this was the first time that \acrshort{uda} has been applied to counting; 
\item we present two new datasets, both having instance segmentation annotations. One was manually annotated, and it consists of images of parked cars collected during the day and night. The second is a synthetic collection of images in urban scenarios taken from a photo-realistic graphical engine, characterized by per-pixel annotations gathered automatically; 
\item we show through experimental evaluation over three different types of domain shifts and various vehicle counting datasets that our \acrshort{uda} schema significantly improved performances compared with the model without domain adaptation.
\end{itemize}

The chapter is organized as follows. In \ref{sec:uda-counting:related_works} we review some works related to the \acrshort{uda} and the counting by density estimation task. \ref{sec:uda-counting:datasets} describes the employed datasets, emphasizing the two ones introduced in this research. In \ref{sec:uda-counting:method} we introduce our \acrshort{uda} algorithm for traffic density estimation and counting. In \ref{sec:uda-counting:exp-settings} we describe the experimental settings, while in \ref{sec:uda-counting:exps-results} we show the experiments and the obtained results.

The research presented in this chapter was published in \cite{ciampi_ecai, ciampi_visapp}. All the associated resources are freely available at \href{https://ciampluca.github.io/unsupervised\_counting/}{https://ciampluca.github.io/unsupervised\_counting}.

\section{Related Works}
\label{sec:uda-counting:related_works}
This section reviews some previous work related to the \acrlong{uda} and the counting by density estimation task.

\paragraph{Unsupervised Domain Adaptation}
\acrfull{uda} is a particular case of \acrshort{da}, in which we rely only on supervision in the source domain without using the labels of the target data. For a more comprehensive survey about \acrshort{da} see \ref{sec:back:domain_adaptation}.
Traditional \acrshort{uda} approaches have been developed to address the problem of image classification, and they try to align features across the two domains (\cite{ganin2015unsupervised}, \cite{tzeng2017adversarial}). However, as pointed out in \cite{zhang2017curriculum}, they do not perform well in other tasks. More recent advances also involve the semantic segmentation task. In this case, adversarial training for \acrshort{uda} is the most employed approach. It includes two networks. The first predicts the segmentation maps for the input source image. The second acts as a discriminator, taking the feature maps from the segmentation network and trying to predict the input domain. The adversarial loss, computed from the discriminator output, tries to make the distributions of the two domains more similar. The first to apply such a technique is \cite{hoffman2016fcns}. More recently, the work proposed in \cite{hong2018conditional} employs a residual network and adversarial training to make the source feature maps closer to the target ones. The authors of \cite{chen2019learning} combine semantic segmentation and depth estimation to boost the adaptation performance, providing to the discriminator the segmentation and the depth prediction maps jointly. Another interesting work that inspired this paper is \cite{learning_to_adapt}, where the authors applied adversarial training to the output space taking advantage of the structural consistency across domains.

\paragraph{Counting by Density Estimation}
As already seen in \ref{sec:back:visual-counting}, counting by density estimation is a supervised learning approach that tries to establish a direct mapping from the image features to a corresponding density map (i.e., a continuous-valued function), skipping the challenging task of detecting instances of the objects. This approach is particularly effective in very crowded scenarios where the instances of the objects are sometimes not clearly visible due to occlusions. Recently, \acrlong{cnn}s have been heavily exploited for learning non-linear functions from crowd images to their corresponding density maps, and a multitude of methods have been proposed in the literature. One of the first works that employed a pure \acrshort{cnn} to estimate the density and count people in crowded contexts is presented by \cite{crowdnet}. A more efficient structure is proposed by \cite{multi_column} introducing a Multi-Column \acrshort{cnn}-based architecture (MCNN) for crowd counting. A similar idea is developed by \cite{towards_perspective} with a scale-aware, multi-column counting model named Hydra-CNN able to estimate traffic densities in congested scenes. More recently, the authors of \cite{csrnet} introduced CSRNet. This \acrshort{cnn}-based algorithm uses dilated kernels to deliver larger reception fields and replace pooling operations. We employ this network as the baseline in our work, and we briefly review its architecture in the next sections. The main limitations of these approaches are due to the scarcity of data. As a result, existing methods often suffer from overfitting, which leads to performance degradation while transferring them to other scenes. Besides, there is another inherent problem: the labels of these datasets are not very accurate. Most of the existing datasets are dot-annotated. Consequently, the ground truth density maps are just an approximation in which the sizes of the objects are estimated using some heuristics. In this chapter, we address both problems by proposing an \acrshort{uda} technique that exploits unlabeled data and by introducing two new datasets with per-pixel annotations that allow the creation of precise ground truth density maps.

\section{Datasets}
\label{sec:uda-counting:datasets}
\begin{figure*}
\centering
\begin{tabular}{ccc}
\includegraphics[width=.30\textwidth]{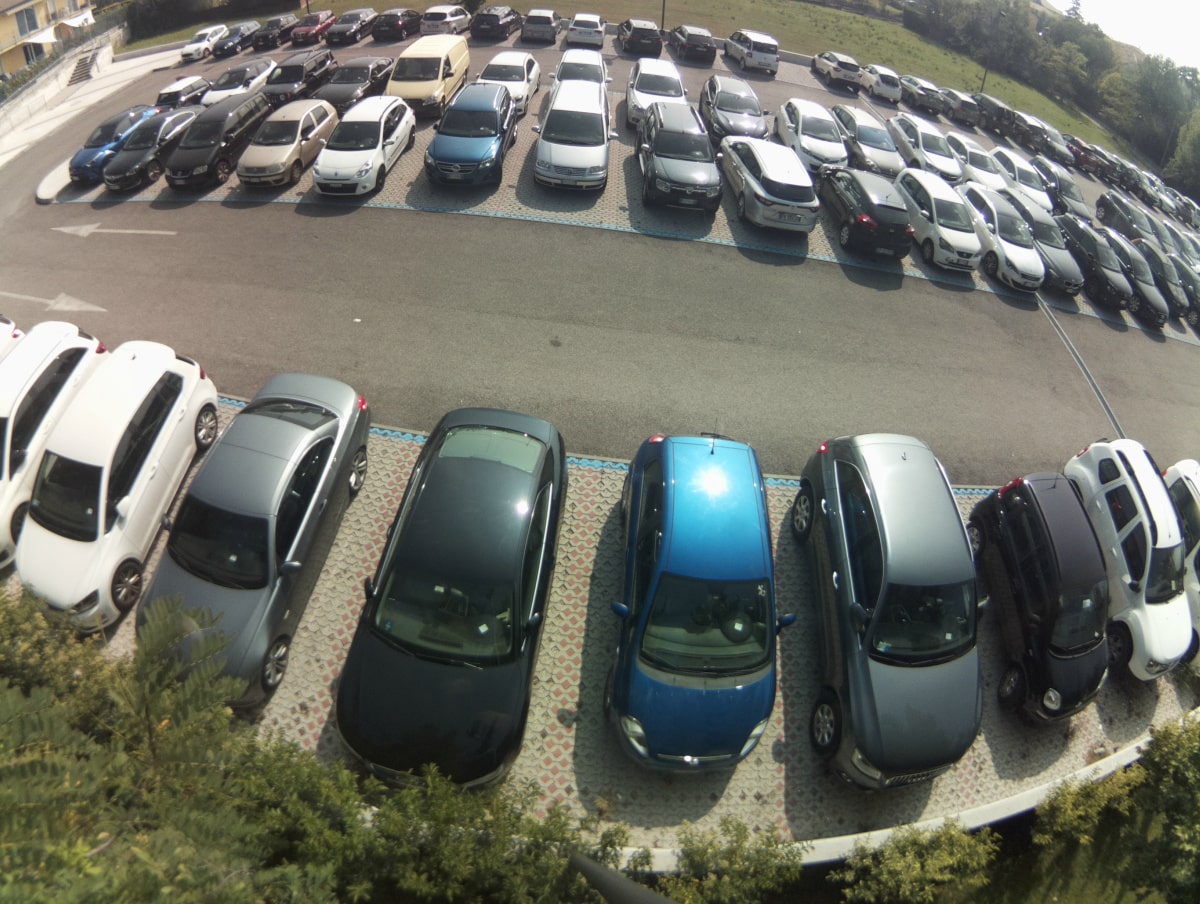} & 
\includegraphics[width=.30\textwidth]{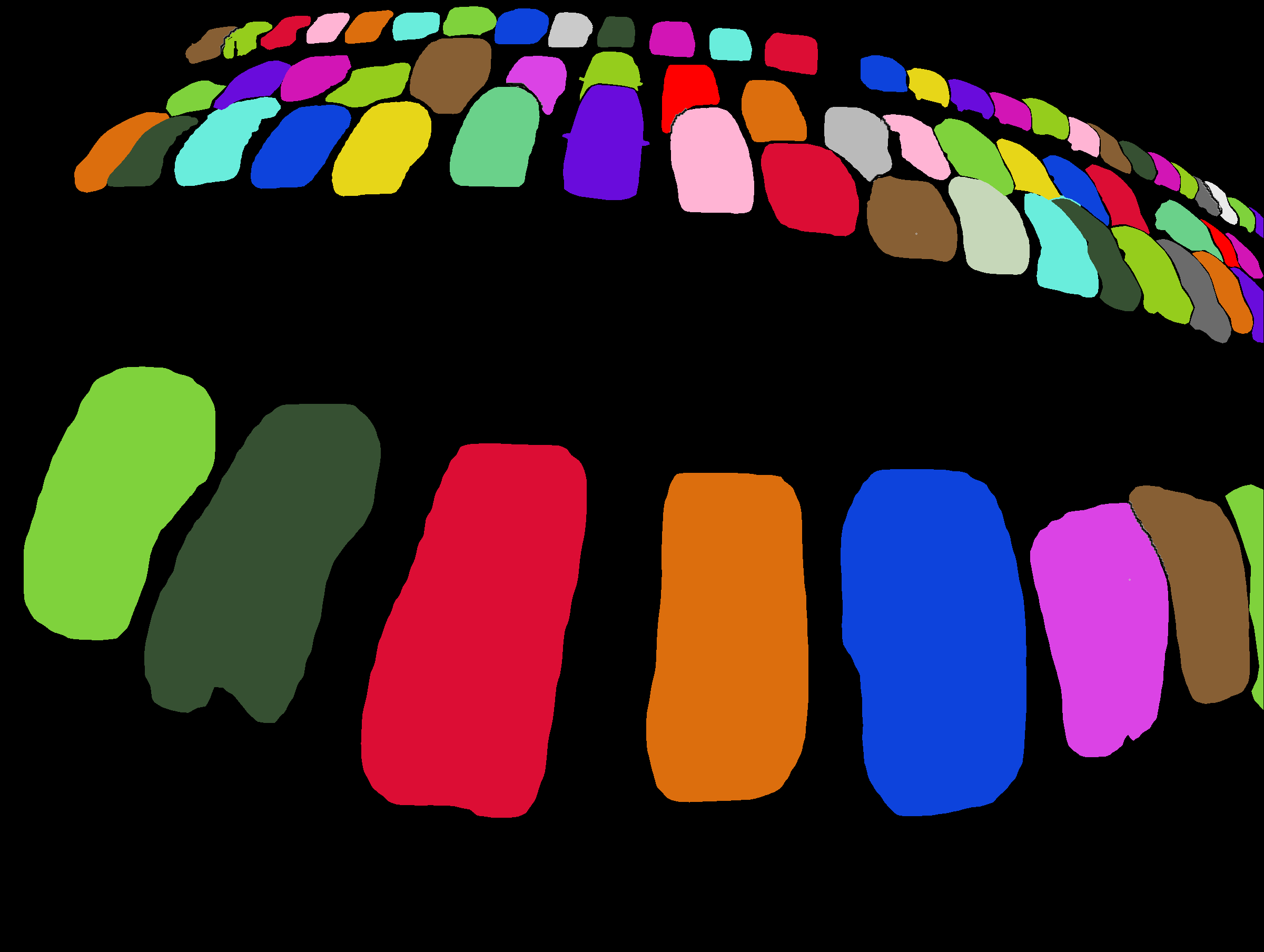} &
\includegraphics[width=.30\textwidth]{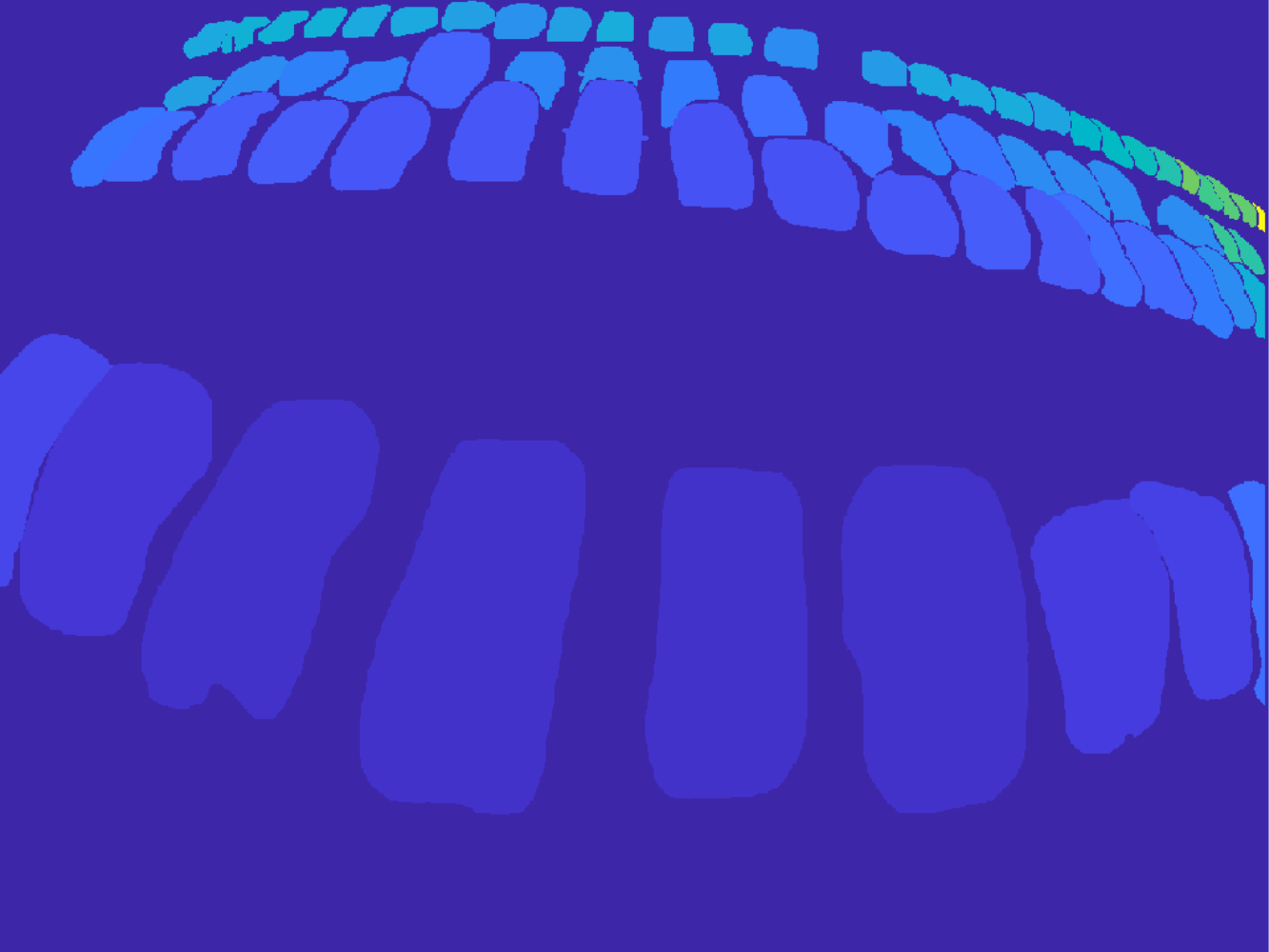}  \\
\includegraphics[width=.30\textwidth]{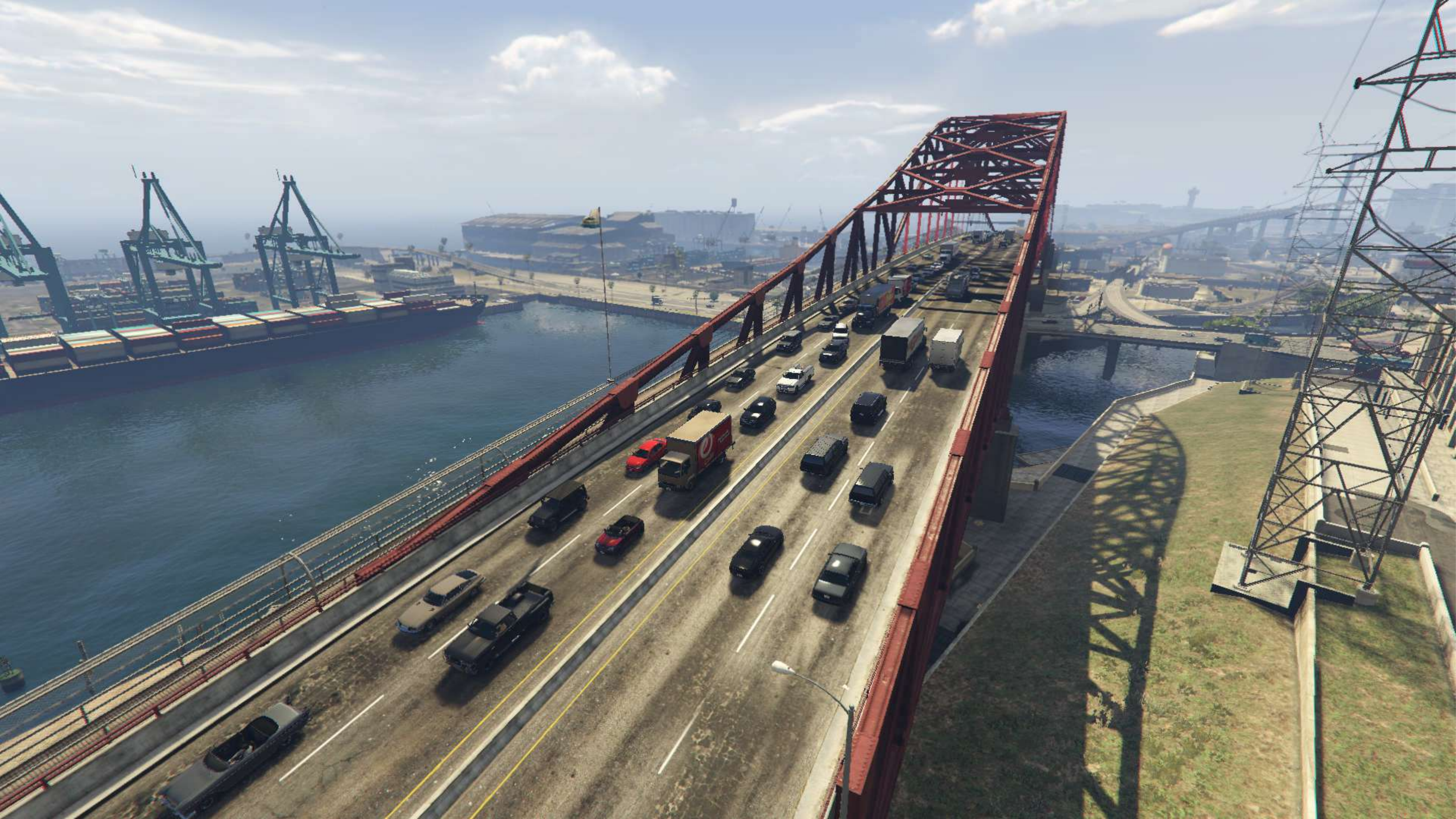} & 
\includegraphics[width=.30\textwidth]{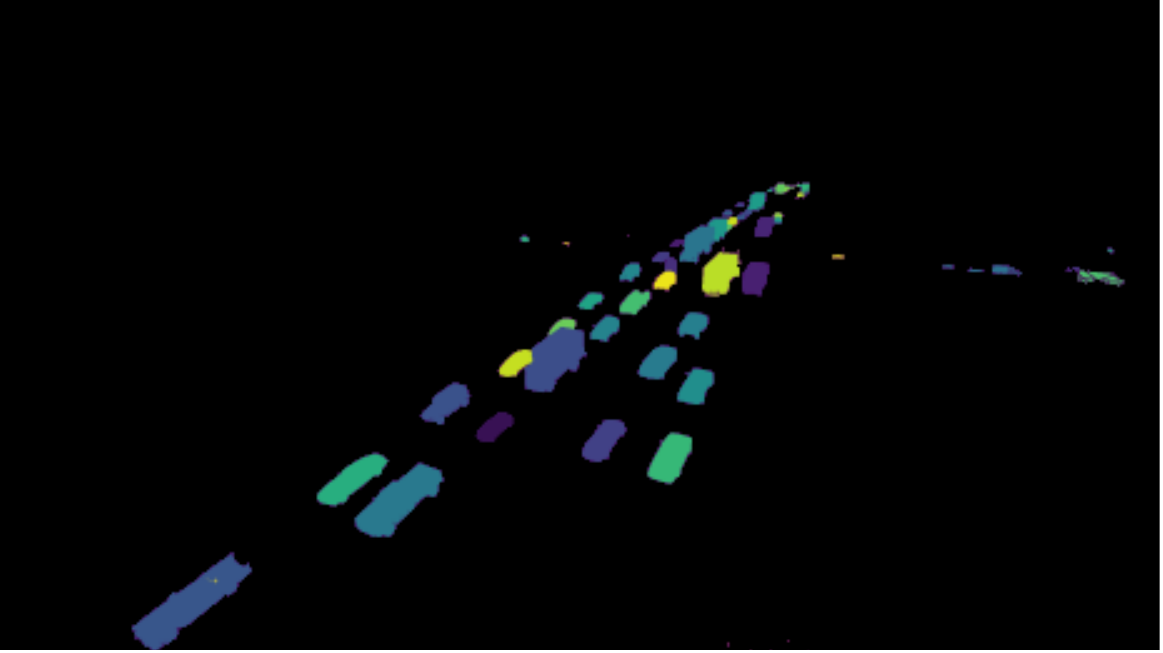} &
\includegraphics[width=.30\textwidth]{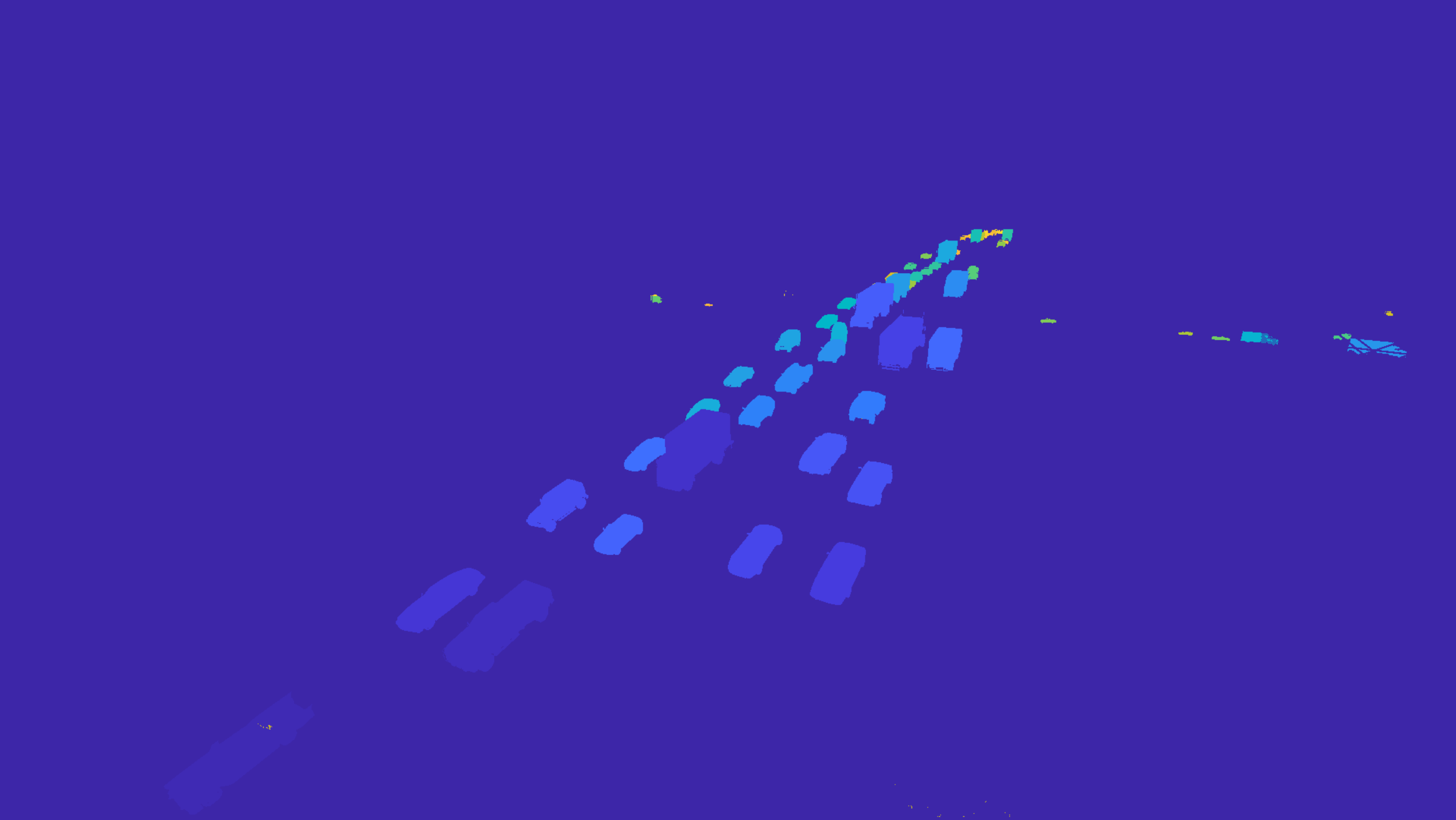} \\
\includegraphics[width=.30\textwidth]{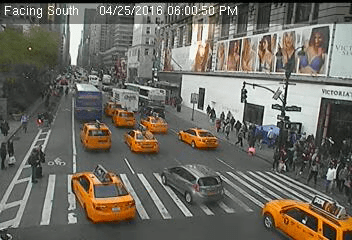} & 
\includegraphics[width=.30\textwidth]{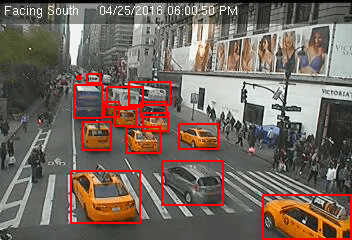} &
\includegraphics[width=.30\textwidth]{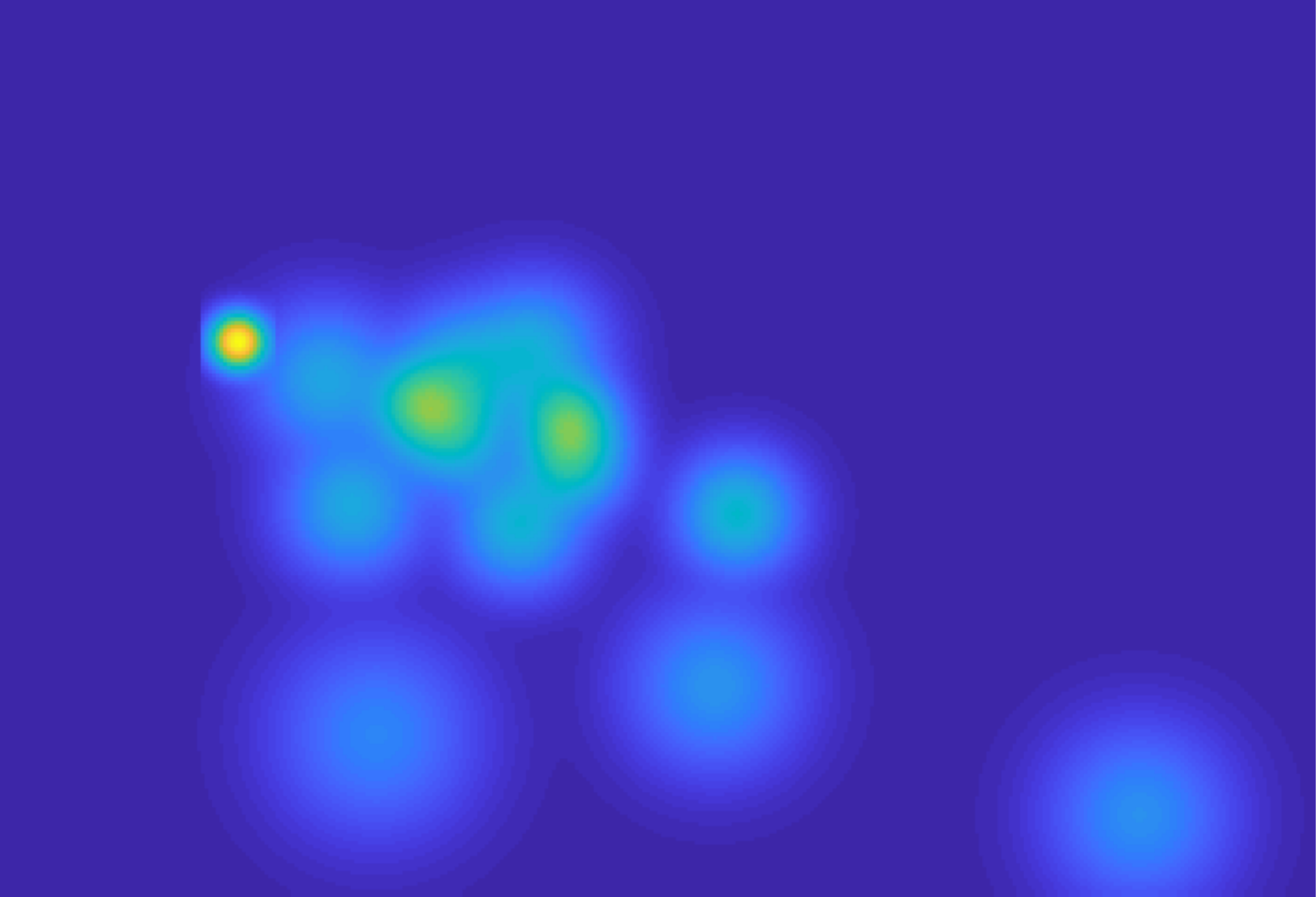}\\
\includegraphics[width=.30\textwidth]{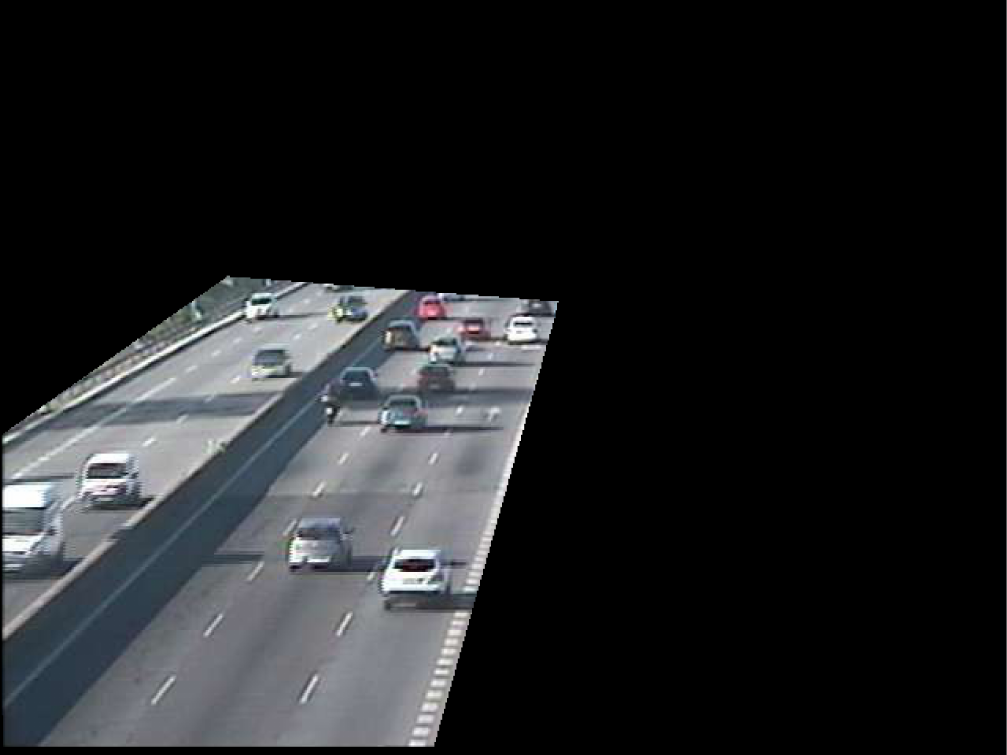} & 
\includegraphics[width=.30\textwidth]{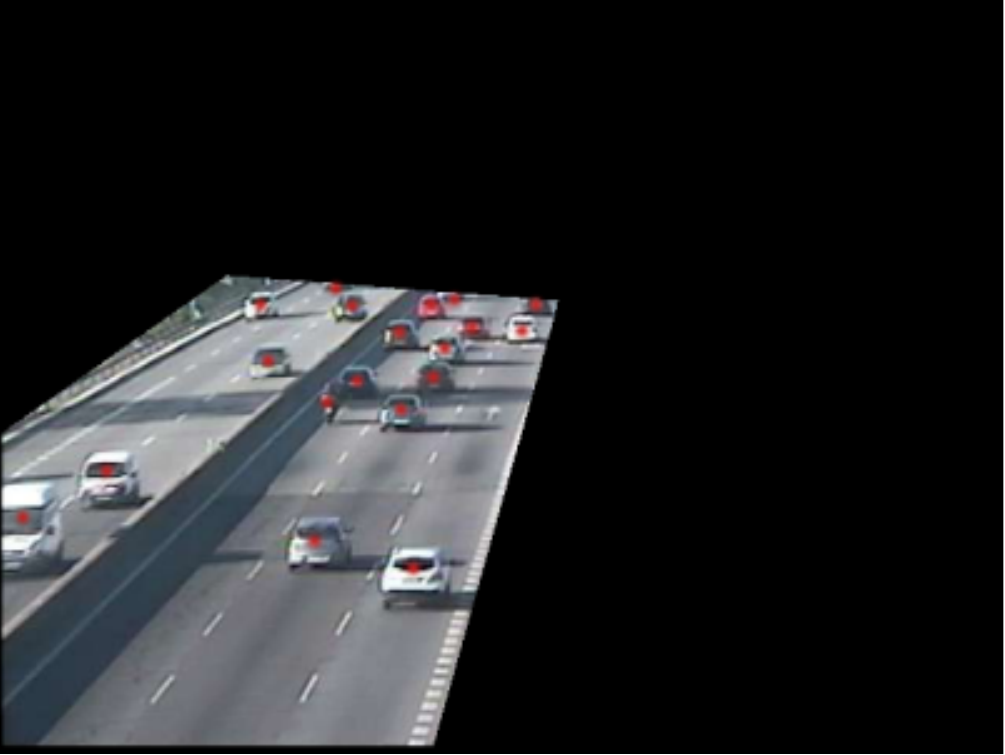} &
\includegraphics[width=.30\textwidth]{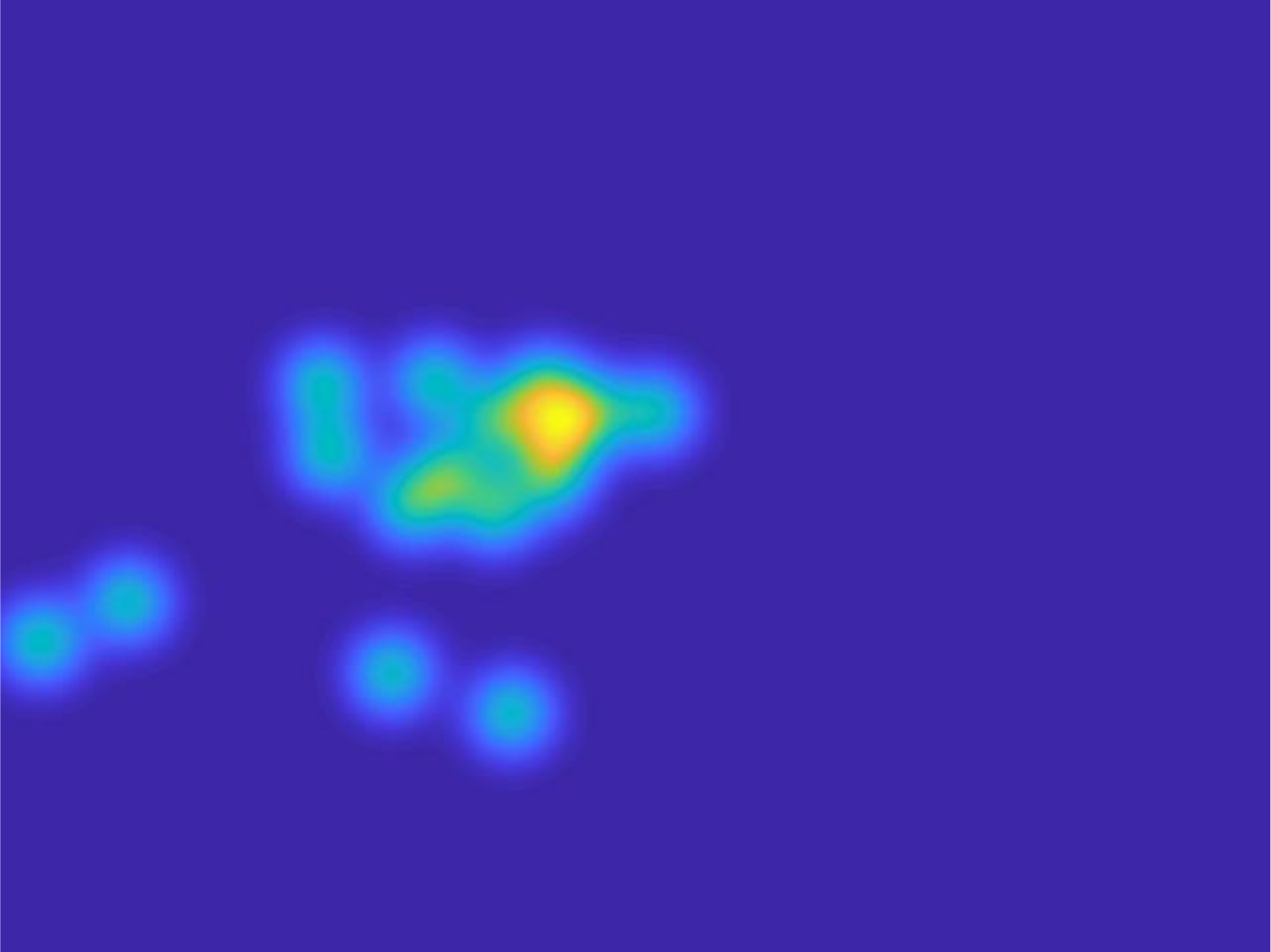}\\
   (a)  &  (b) & (c)
\end{tabular}
\caption{\textbf{Some examples taken by the four datasets used in this work.} (a) Images; (b) Labels; (c) Density Maps generated from the labels. Each row correspond to a specific dataset: from top to bottom, the \acrshort{ndispark} and the \acrshort{gta} datasets introduced in this work, the WebCamT dataset \cite{understandingCosteira} and the TRANCOS dataset \cite{ExtremelyTrancos}. Note that the densities maps generated in our datasets are accurate since we start from an instance segmentation annotations. Also notice that, in the case of the \acrshort{gta} dataset, annotations are \textit{automatically} generated without human effort.}
\label{fig:examples_images_instances_densities}
\end{figure*}

To prove the validity of our approach, we performed experiments on various vehicle counting datasets, offering different domain shift characteristics. Specifically, we exploited two existing datasets for traffic density estimation: \textit{WebCamT} \cite{understandingCosteira} and \textit{TRANCOS} \cite{ExtremelyTrancos}. Then, we used two additional datasets that we created on purpose and made publicly available: the \textit{\acrfull{ndispark}} dataset and the \textit{\acrfull{gta}} dataset. \ref{fig:examples_images_instances_densities} shows some images belonging to these datasets, together with the associated labels and the corresponding generated density maps used for the counting task. In the following, we describe more in detail each of them. 

\subsection{WebCamT Dataset}
The \textit{WebCamT} dataset is a collection of traffic scenes recorded using city-cameras introduced by \cite{understandingCosteira}. It is particularly challenging to analyze due to the low-resolution \((352\times240)\), high occlusion, and large perspective. We considered a total of about 40,000 images belonging to 10 different cameras and consequently having different views. We employed the existing bounding box annotations of the dataset to generate ground truth density maps. In particular, we considered one Gaussian Normal kernel for each vehicle present in the scene, having a value of \(\mu\) and \(\sigma\) equal to the center and proportional to the size of the bounding box surrounding the vehicle, respectively. We used this dataset to test performance with the \textit{Camera2Camera} domain shift.

\subsection{TRANCOS Dataset}
The \textit{TRANCOS} dataset is a public dataset containing 1244 dot-annotated images of different congested traffic scenes captured by surveillance cameras, introduced by \cite{ExtremelyTrancos}. The approximated ground truth density maps are generated by putting one Normal Gaussian kernel for each dot present in the scene, having a value of \(\sigma\) empirically decided by the authors. They also provided the regions of interest (ROIs) for each image. We used this dataset to test performance with the \textit{Camera2Camera} domain shift.

\subsection{The Night and Day Instance Segmented Dataset}
\begin{figure}
    \centering
  \subfloat{
       \includegraphics[width=0.48\linewidth]{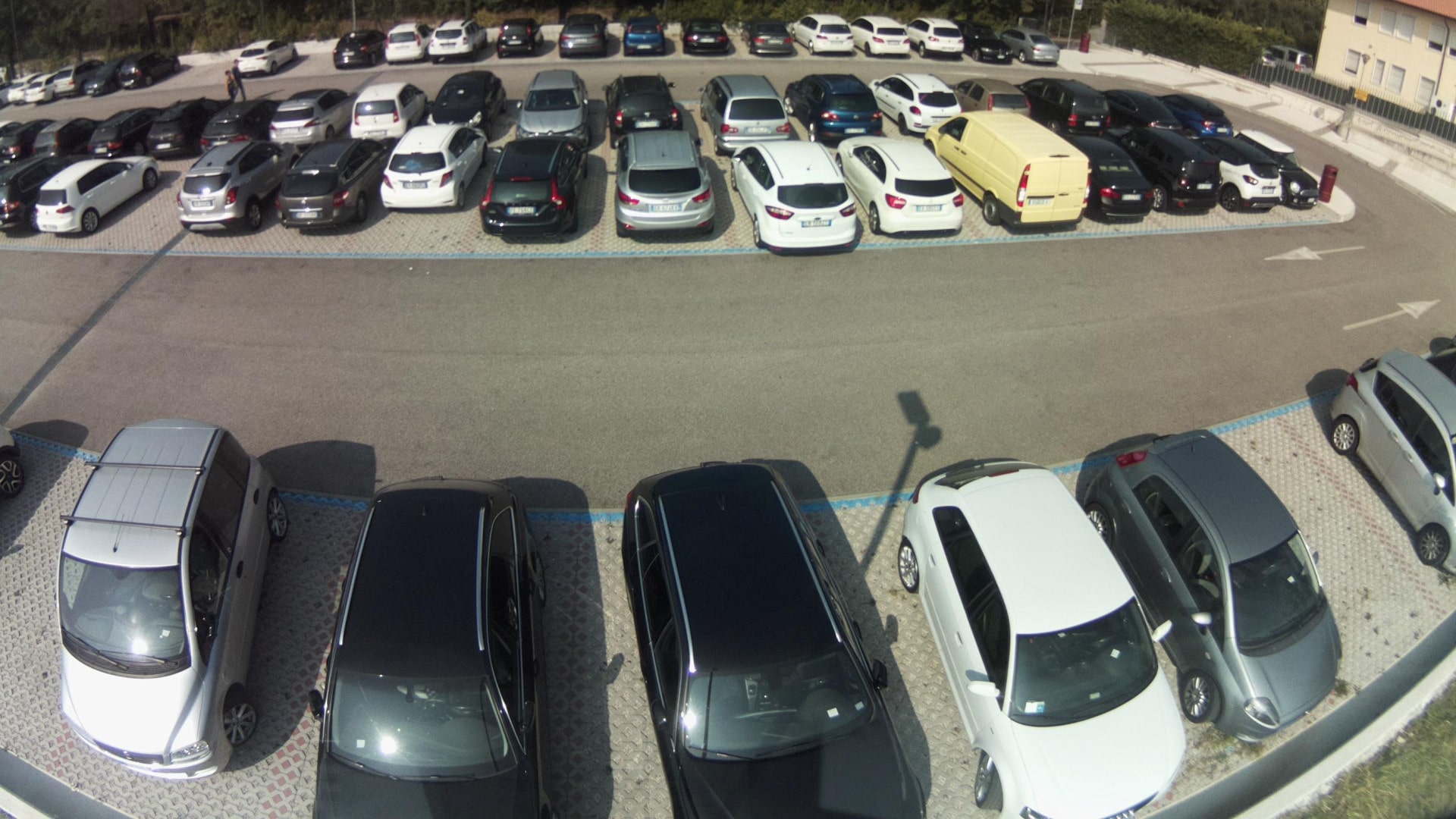}}
    \hfill
  \subfloat{
        \includegraphics[width=0.48\linewidth]{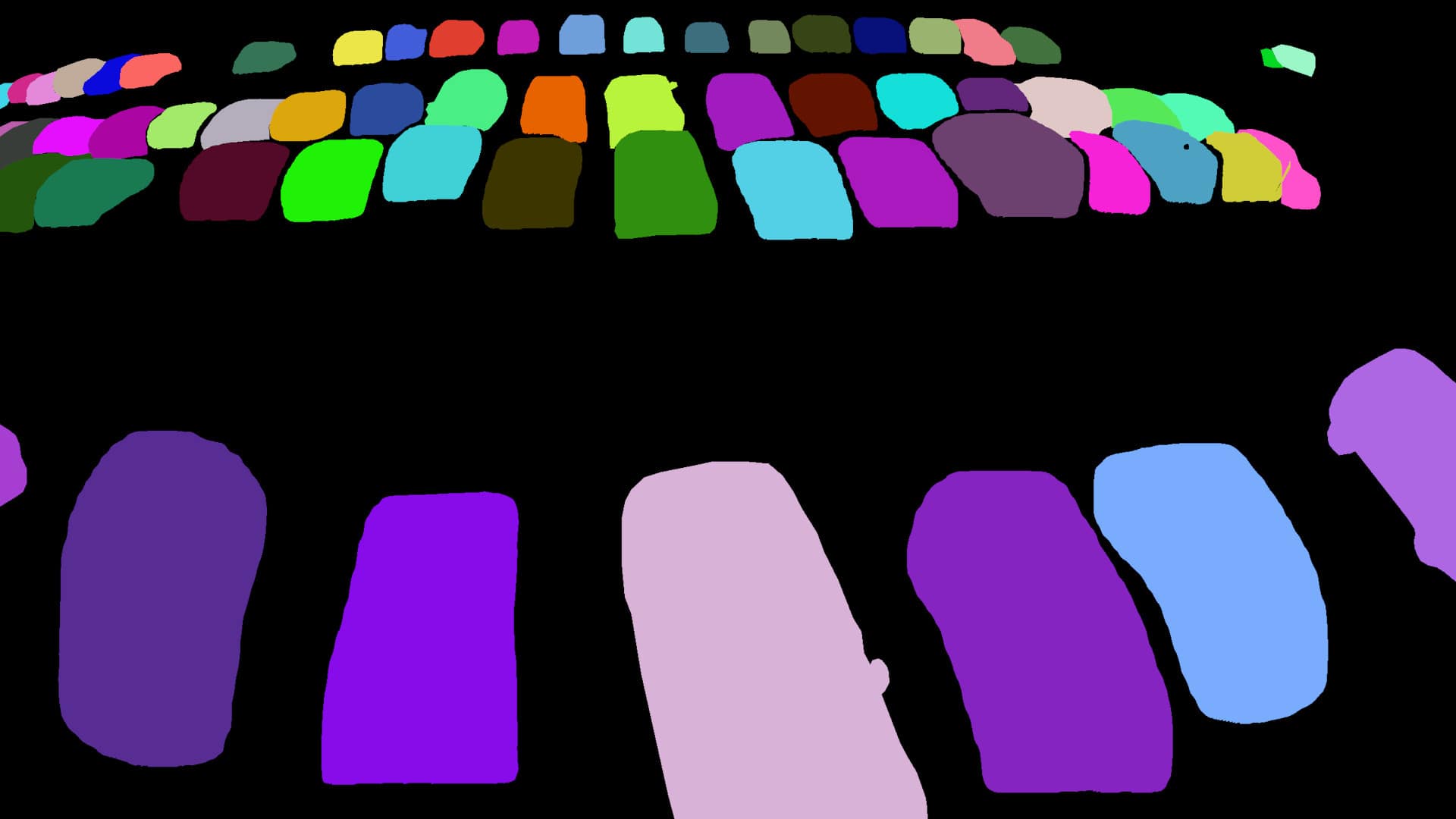}}
    \\ [2ex]
  \subfloat{
        \includegraphics[width=0.48\linewidth]{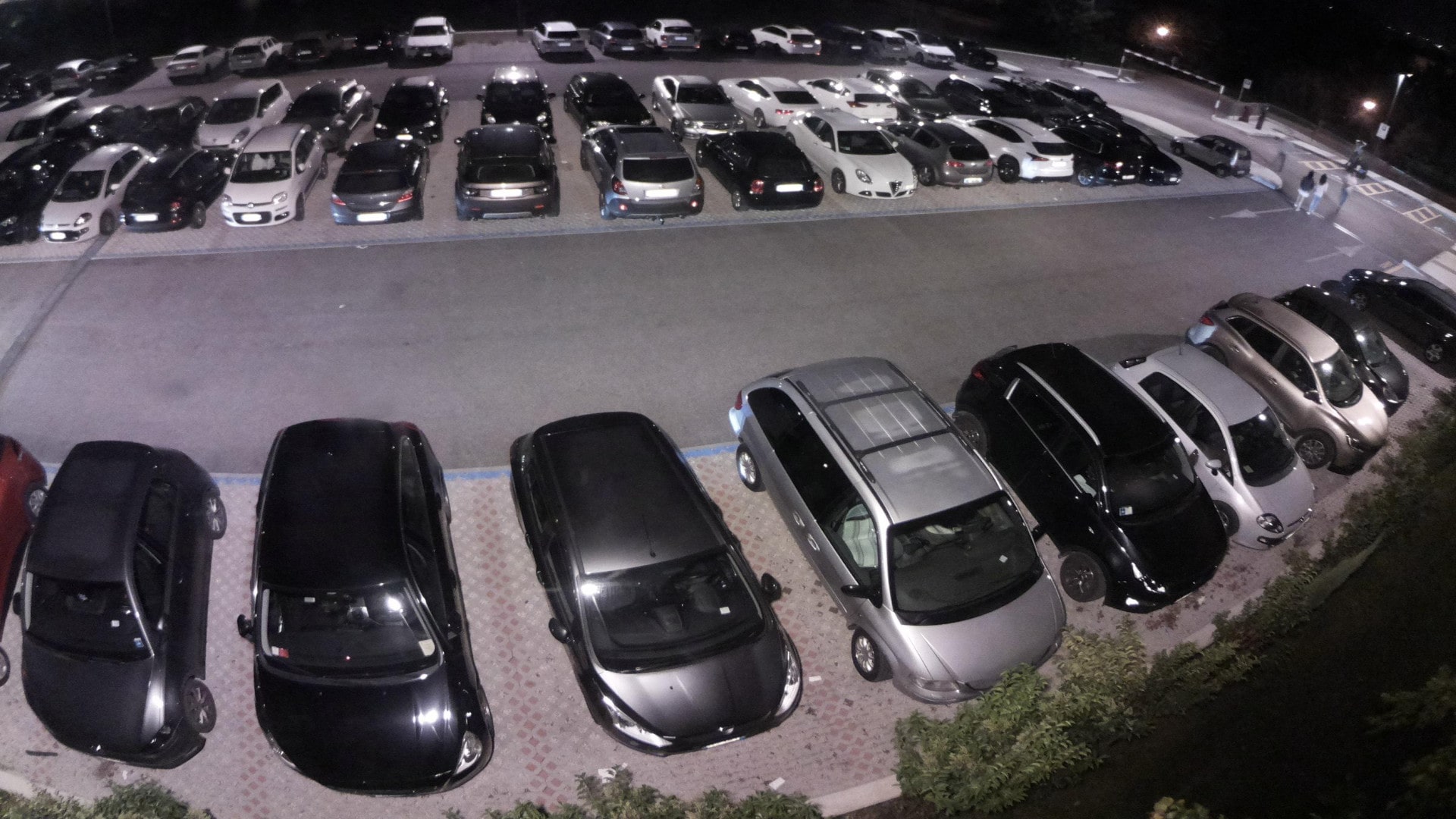}}
    \hfill
  \subfloat{
        \includegraphics[width=0.48\linewidth]{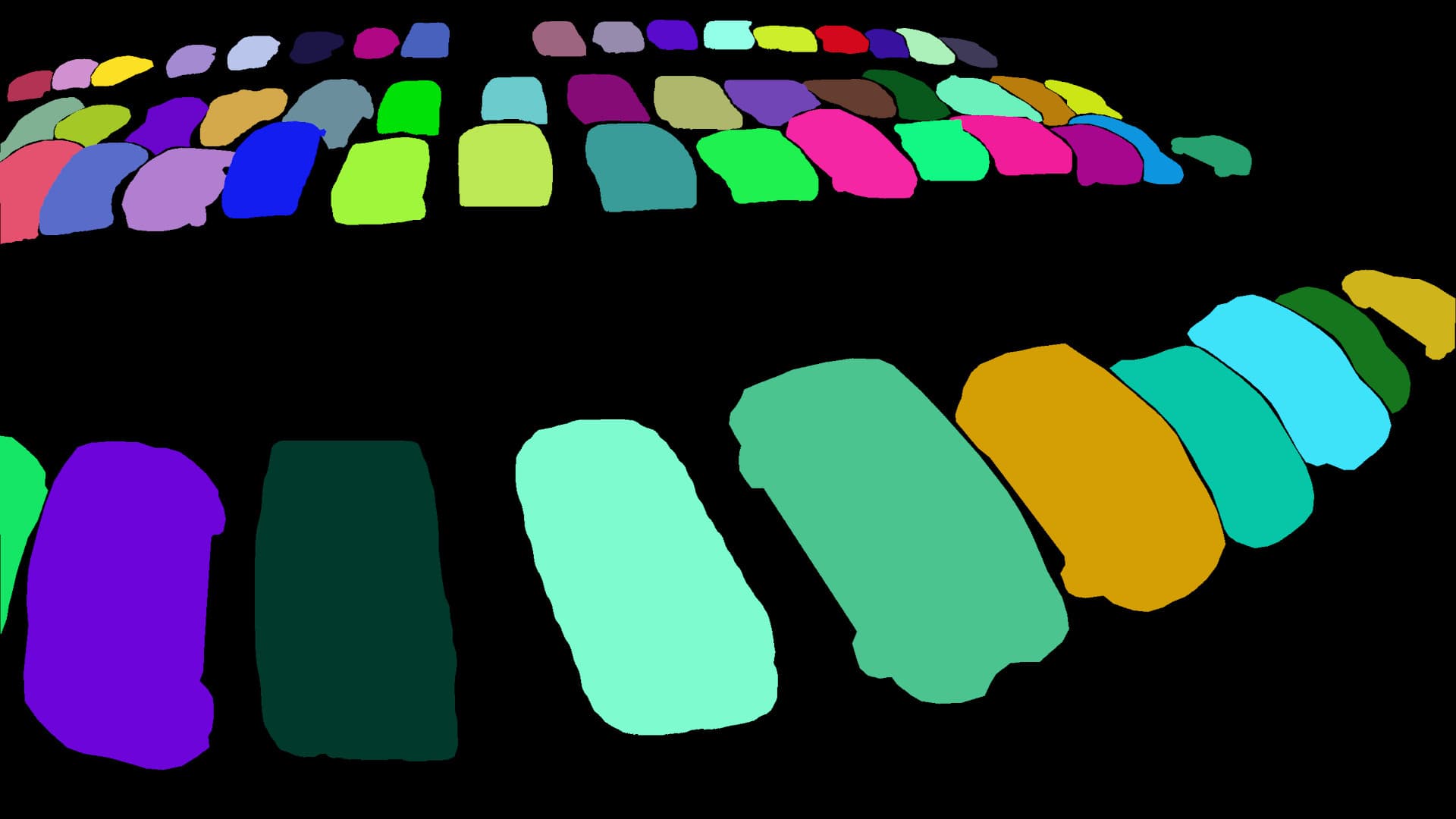}}
    \\ [2ex]
  \subfloat{
        \includegraphics[width=0.48\linewidth]{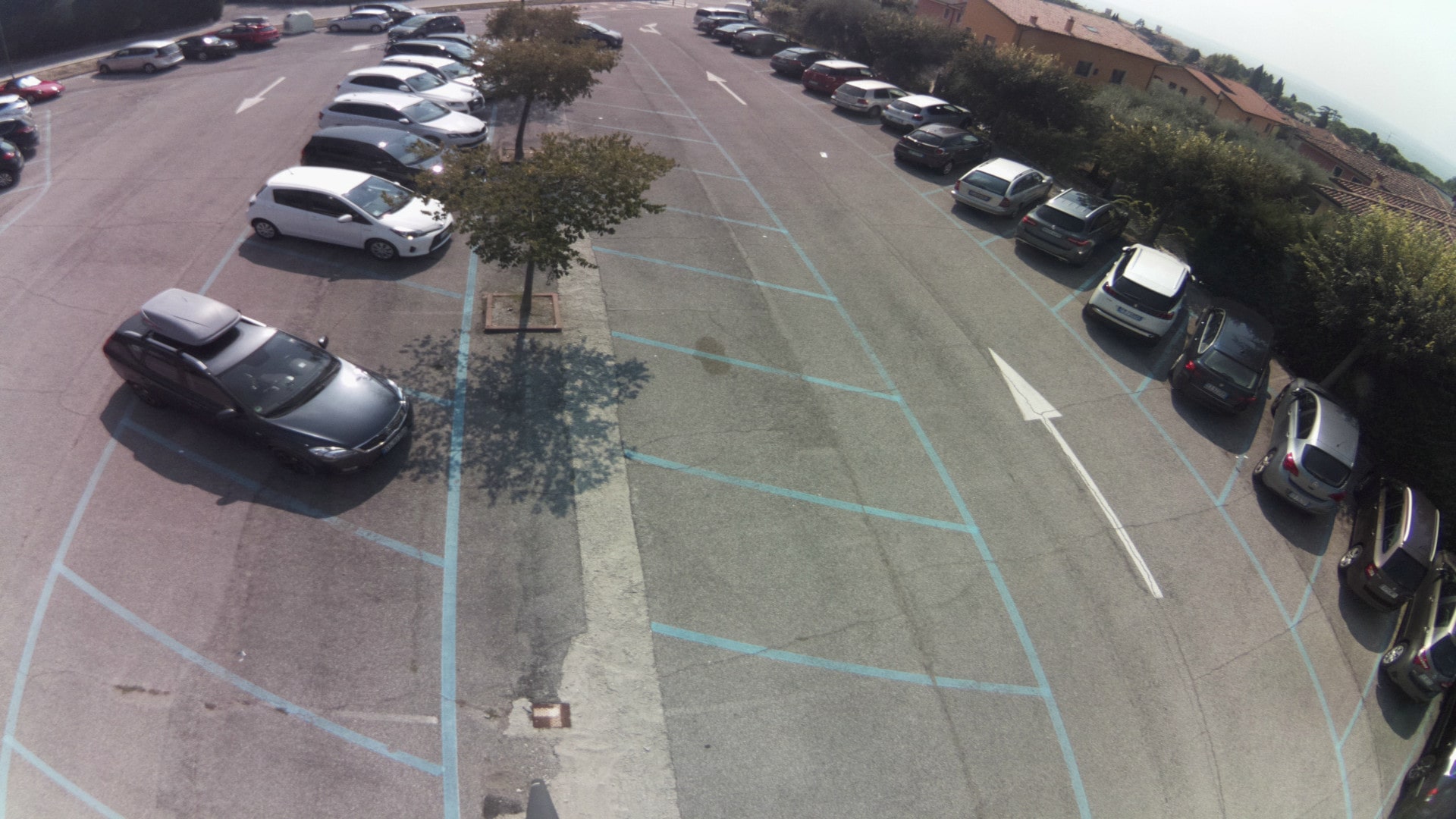}}
    \hfill
  \subfloat{
        \includegraphics[width=0.48\linewidth]{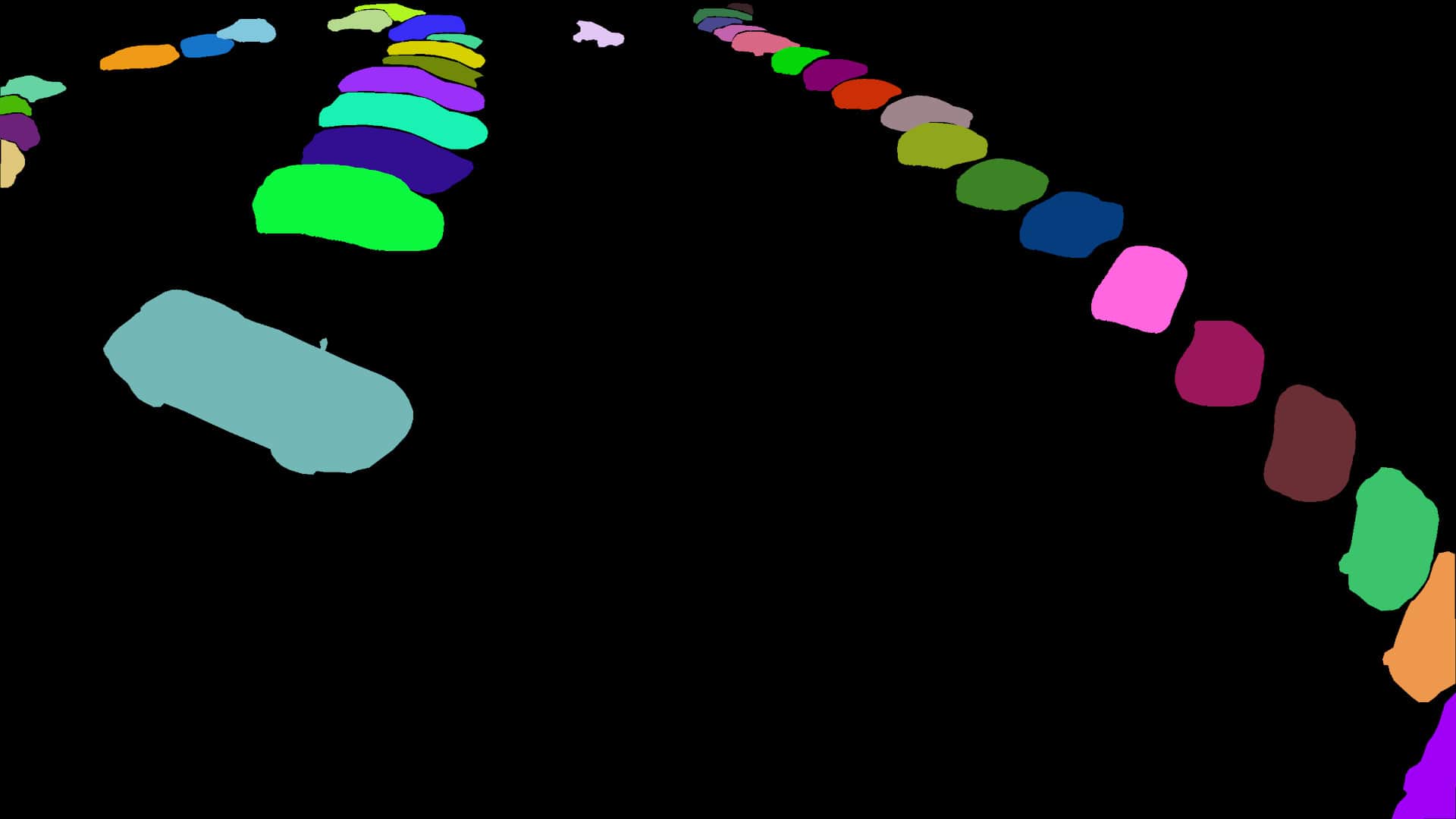}}
    \\ [2ex]
  \subfloat{
        \includegraphics[width=0.48\linewidth]{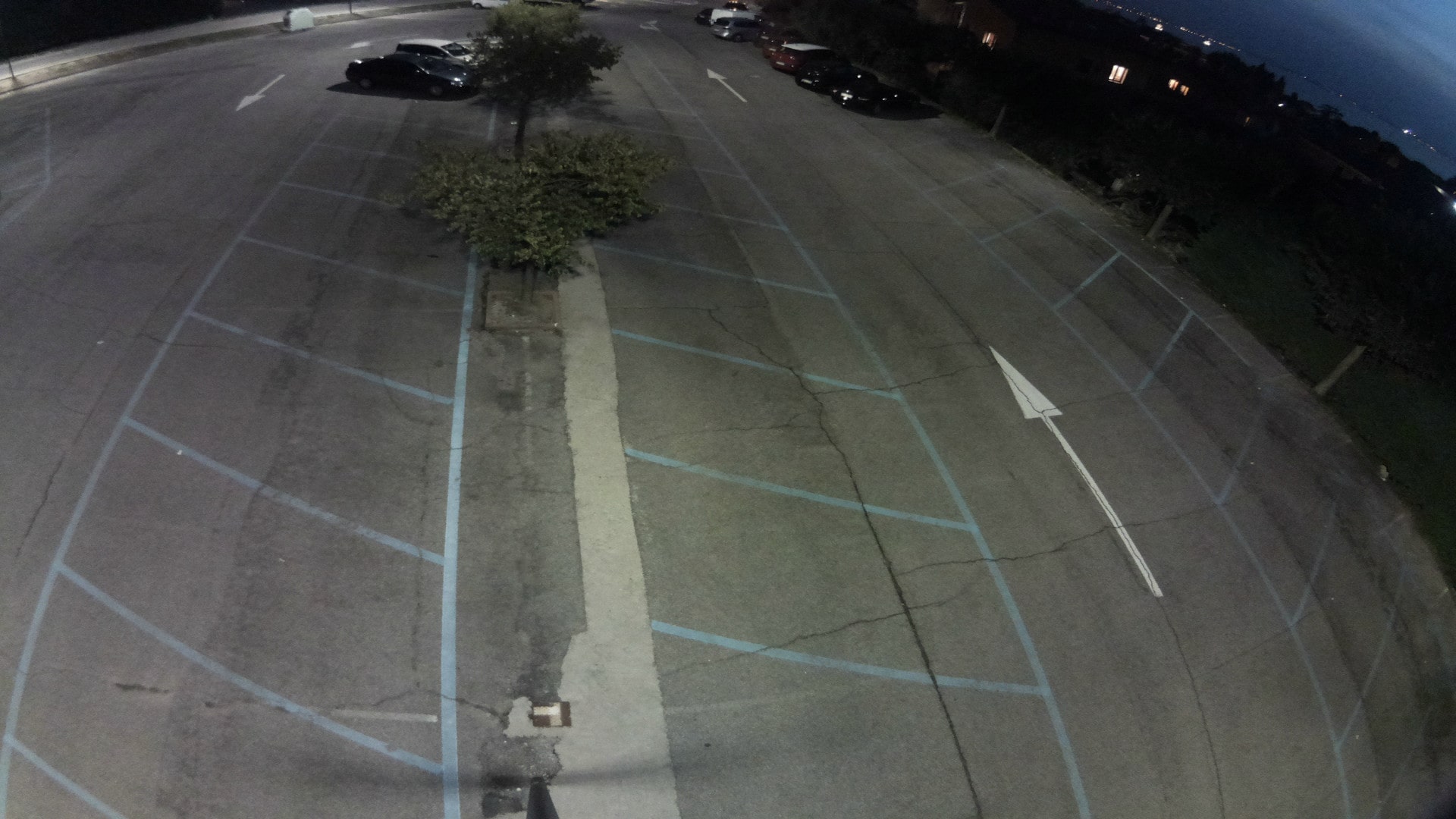}}
    \hfill
  \subfloat{
        \includegraphics[width=0.48\linewidth]{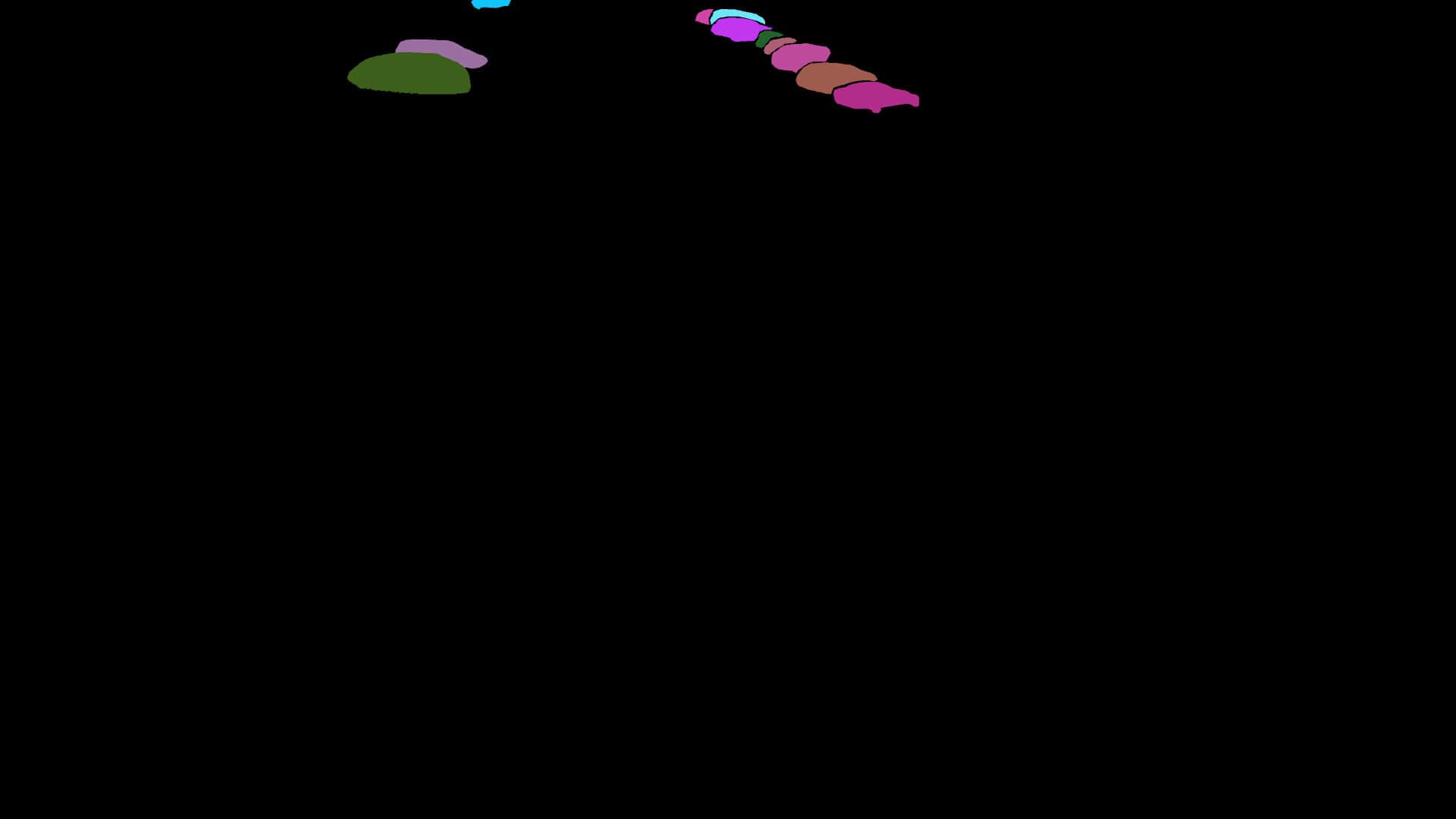}}

  \caption{\textbf{Samples of the \acrfull{ndispark} dataset.} We show images together with the associated per-pixel annotations.}
  \label{ndispark_examples} 
\end{figure}

The \textit{\acrfull{ndispark}} dataset was created by us on purpose and made publicly available. It is a small, manually annotated dataset for counting cars in parking lots, consisting of about 250 images. This dataset is challenging and describes most of the problematic situations we can find in a real scenario: seven different cameras capture the images under various weather conditions and viewing angles. Another challenging aspect is the presence of partial occlusion patterns in many scenes, such as obstacles (trees, lampposts, other cars) and shadowed cars. Furthermore, it is worth noting that images are taken during the day and the night, showing utterly different lighting conditions, and that, unlike most counting datasets, the \acrshort{ndispark} dataset is precisely annotated with \textit{instance} segmentation labels, which allowed us to generate accurate ground truth density maps for the counting task since the sizes of the vehicles were well-known. \ref{ndispark_examples} shows some samples of this dataset, together with the labels.

\subsection{The Grand Traffic Auto Dataset}
\begin{figure}
    \centering
  \subfloat{
       \includegraphics[width=0.48\linewidth]{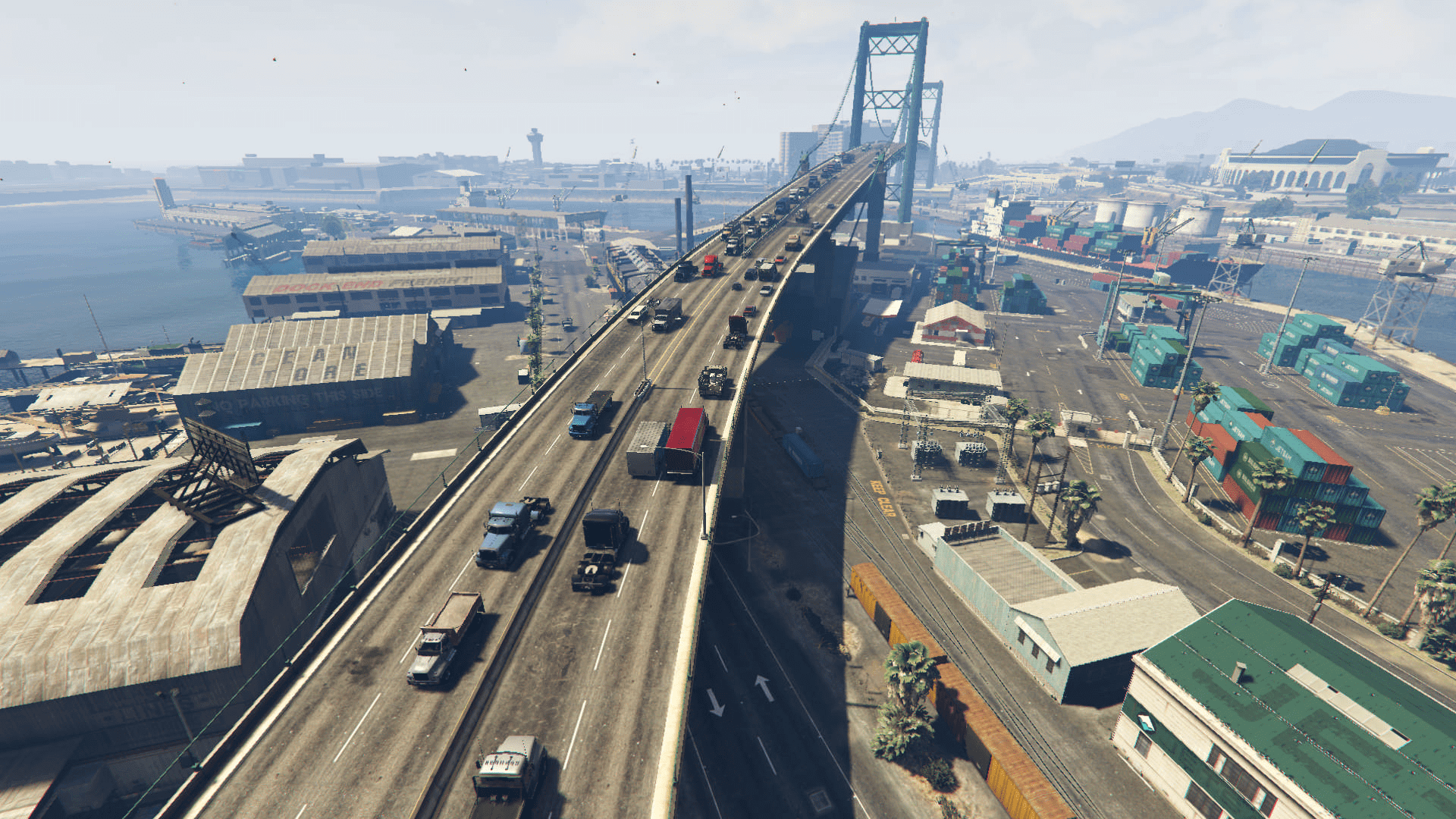}}
    \hfill
  \subfloat{
        \includegraphics[width=0.48\linewidth]{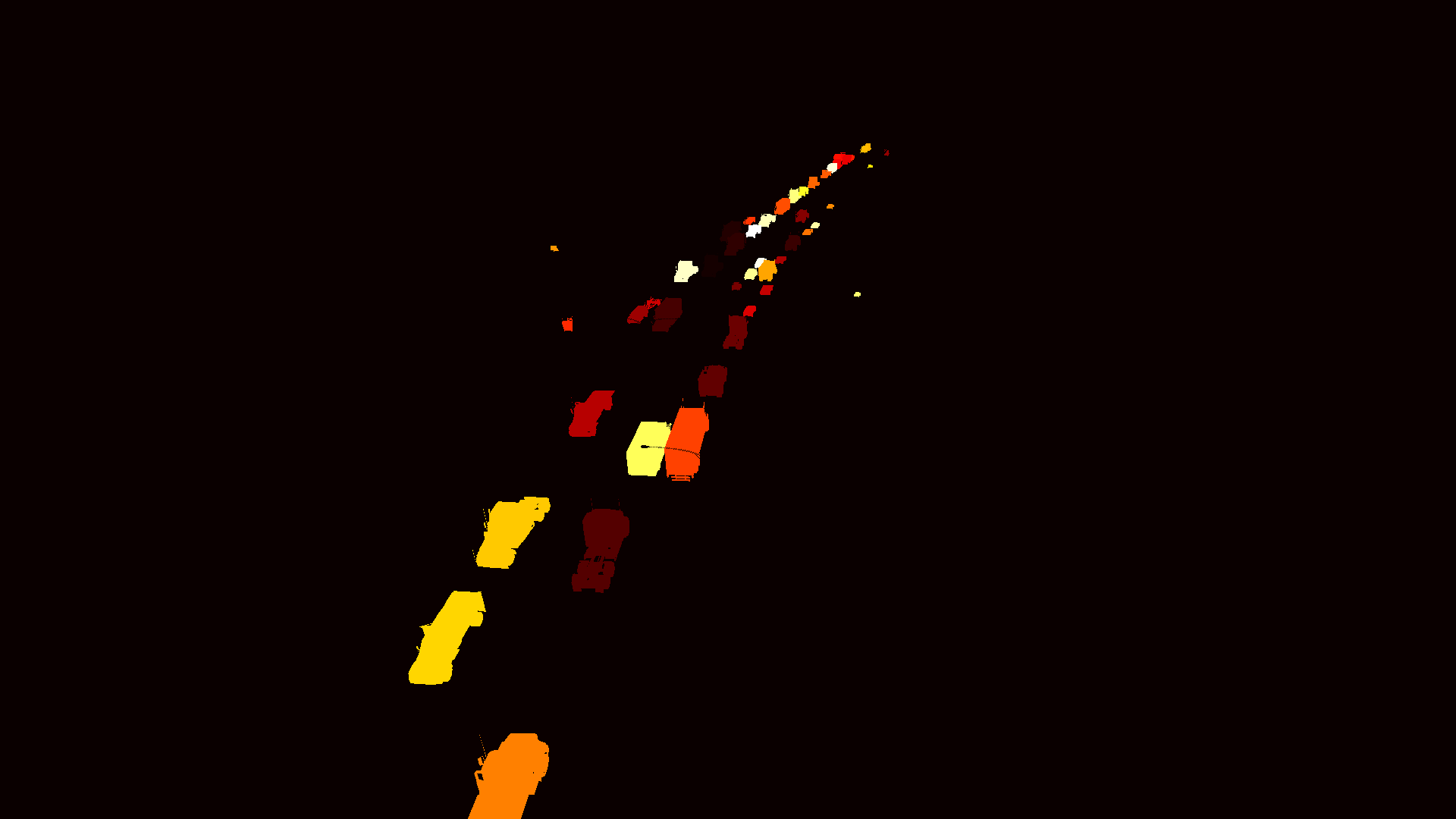}}
    \\ [2ex]
  \subfloat{
        \includegraphics[width=0.48\linewidth]{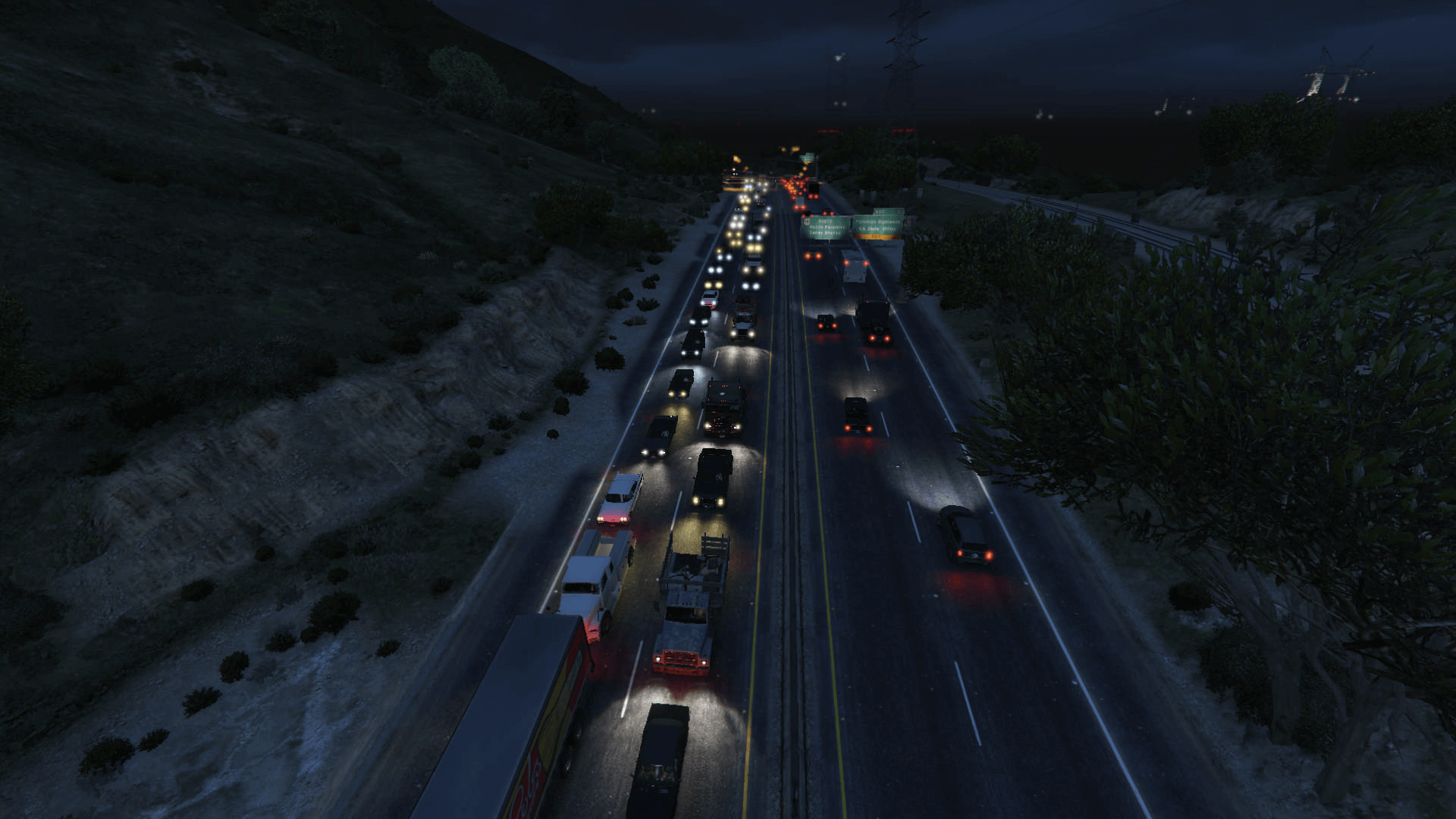}}
    \hfill
  \subfloat{
        \includegraphics[width=0.48\linewidth]{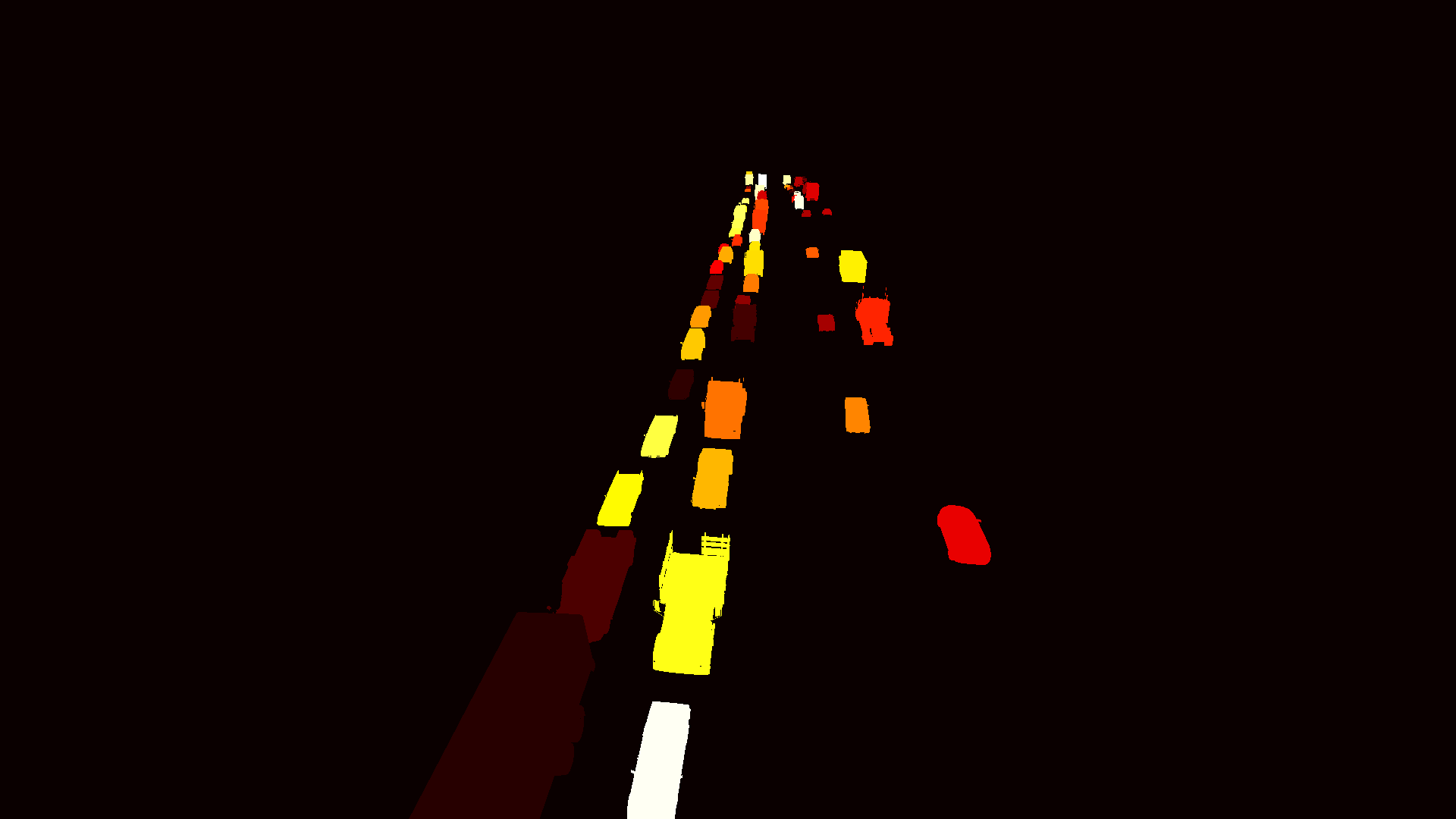}}
    \\ [2ex]
  \subfloat{
        \includegraphics[width=0.48\linewidth]{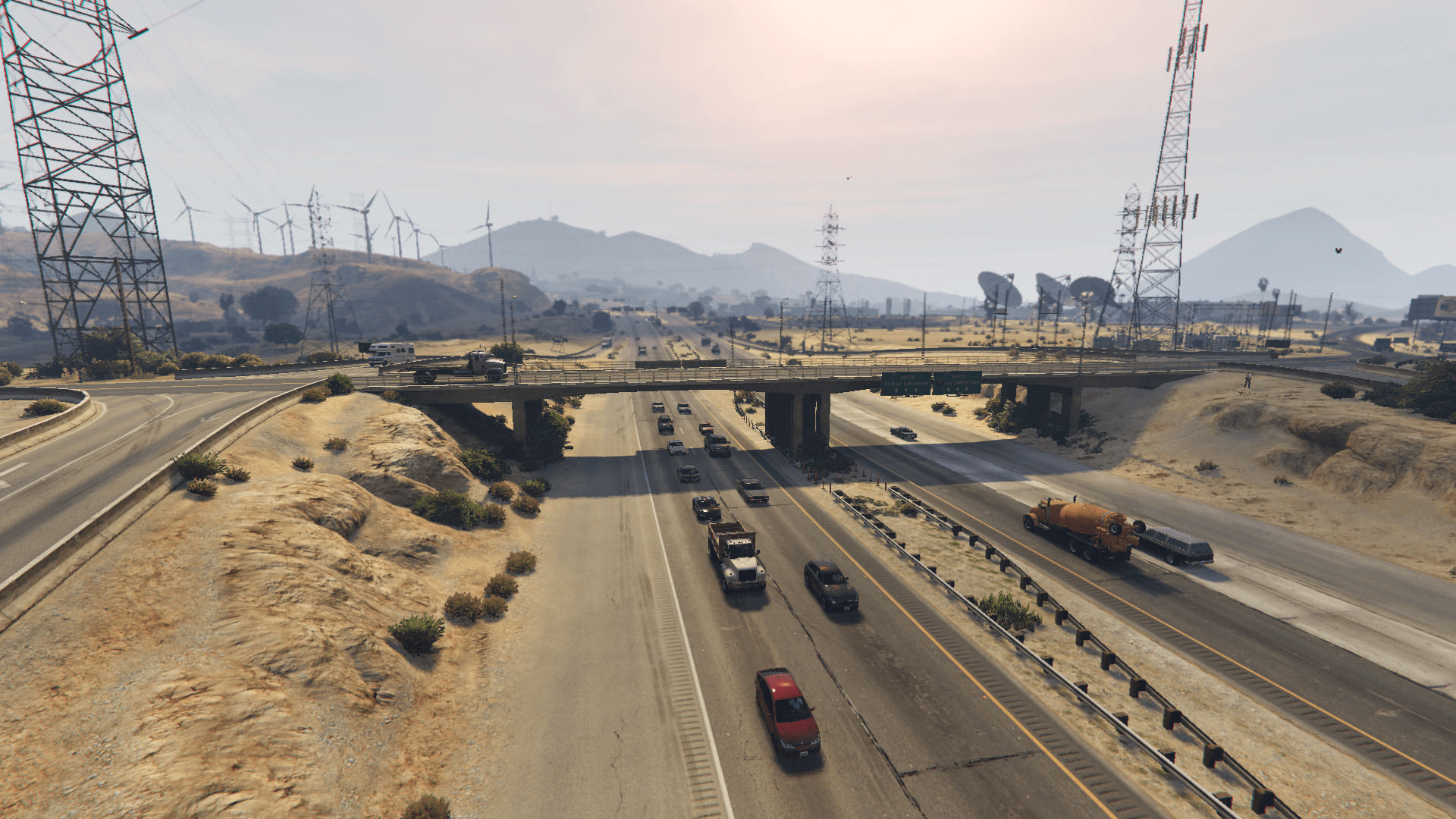}}
    \hfill
  \subfloat{
        \includegraphics[width=0.48\linewidth]{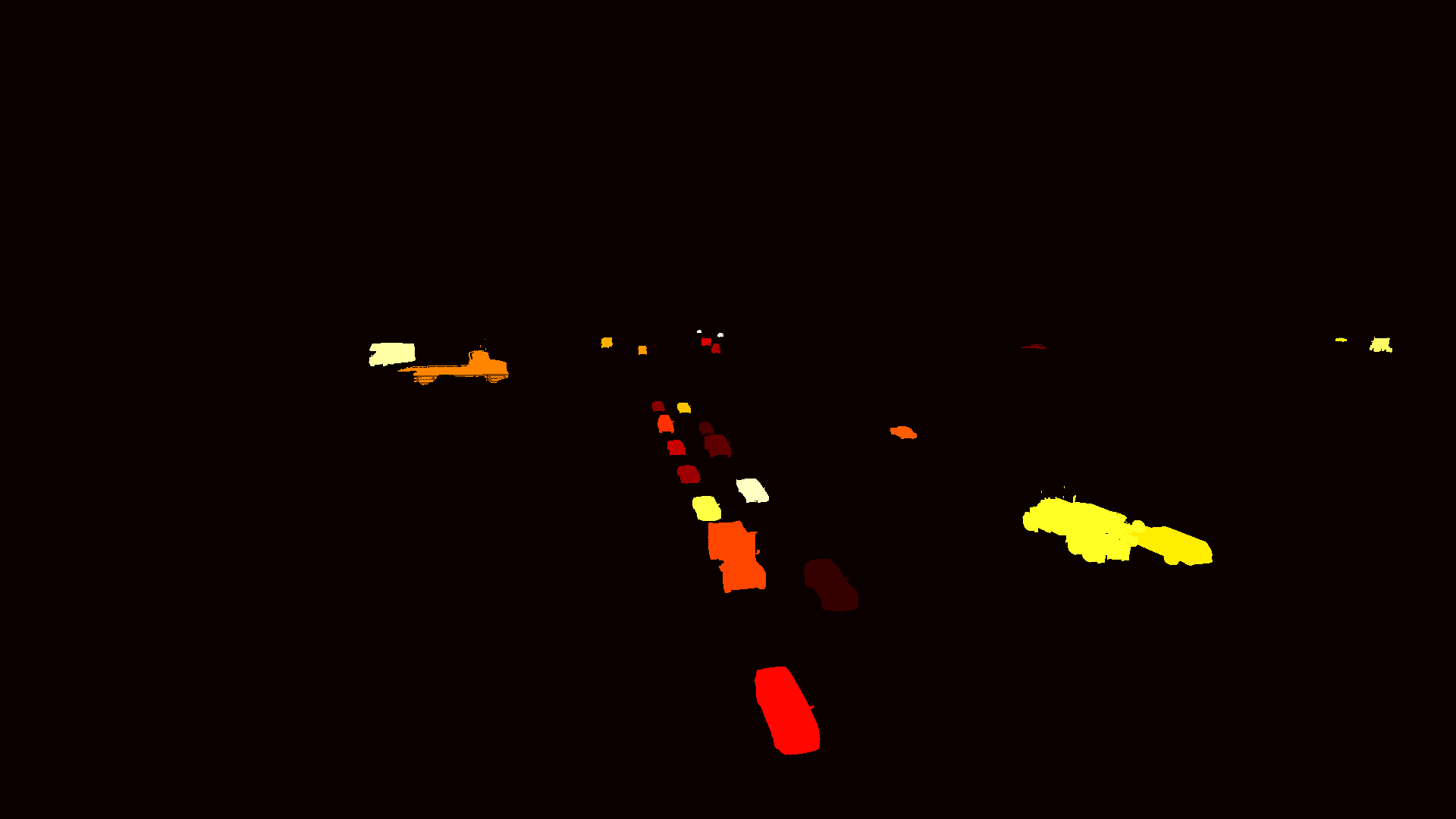}}
    \\ [2ex]
  \subfloat{
        \includegraphics[width=0.48\linewidth]{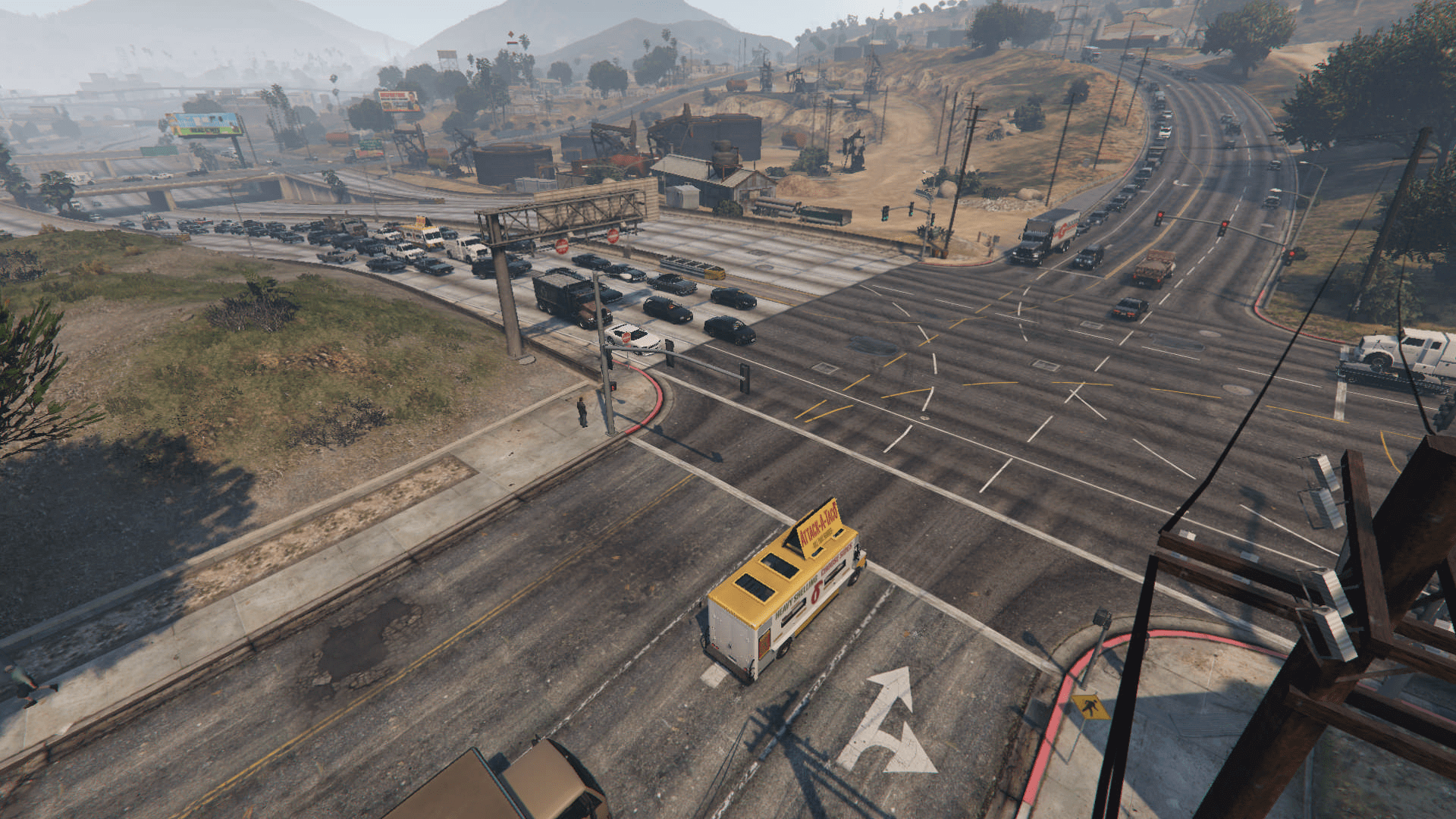}}
    \hfill
  \subfloat{
        \includegraphics[width=0.48\linewidth]{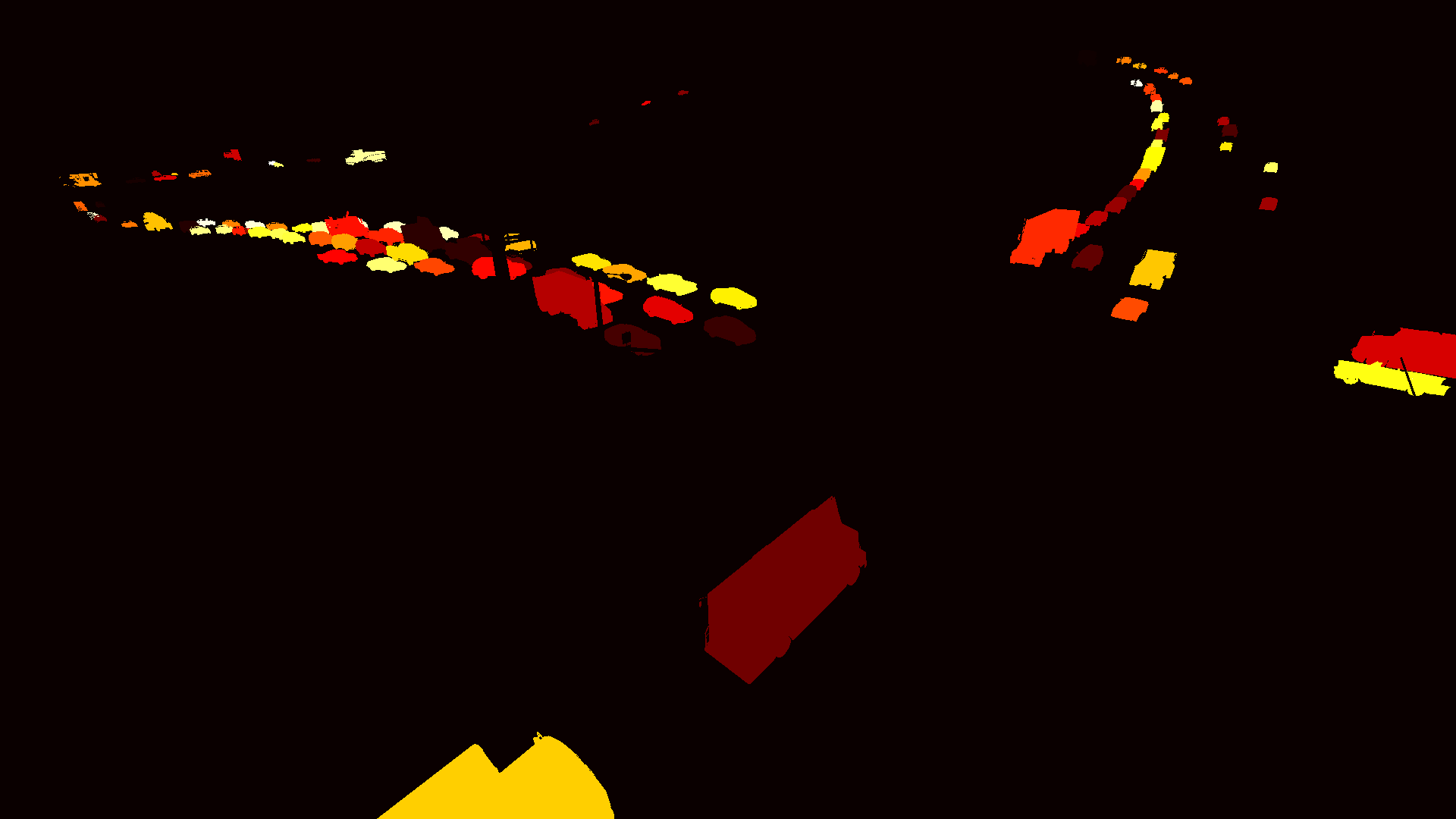}}
    \\ [2ex]
  \subfloat{
        \includegraphics[width=0.48\linewidth]{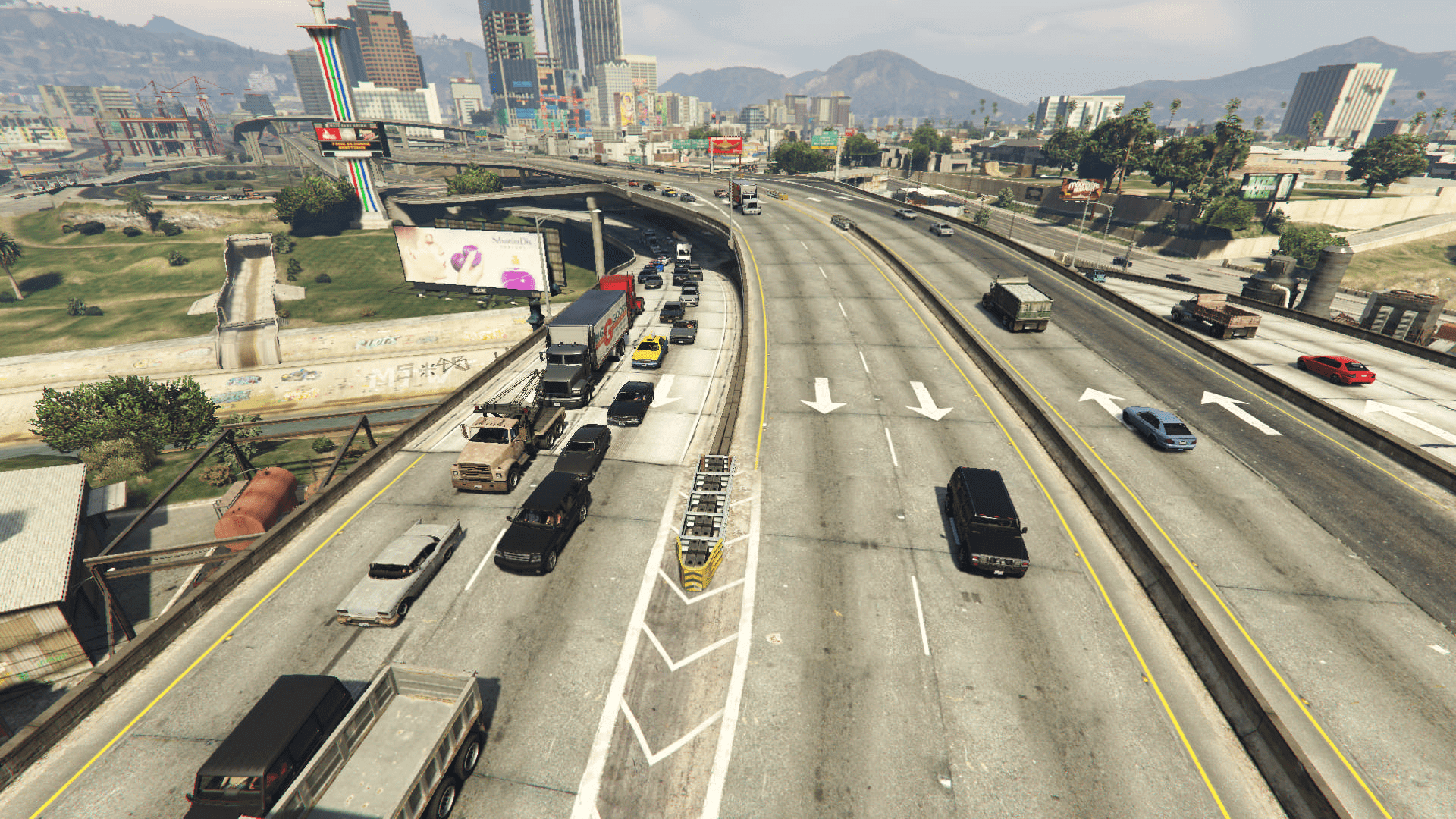}}
    \hfill
  \subfloat{
        \includegraphics[width=0.48\linewidth]{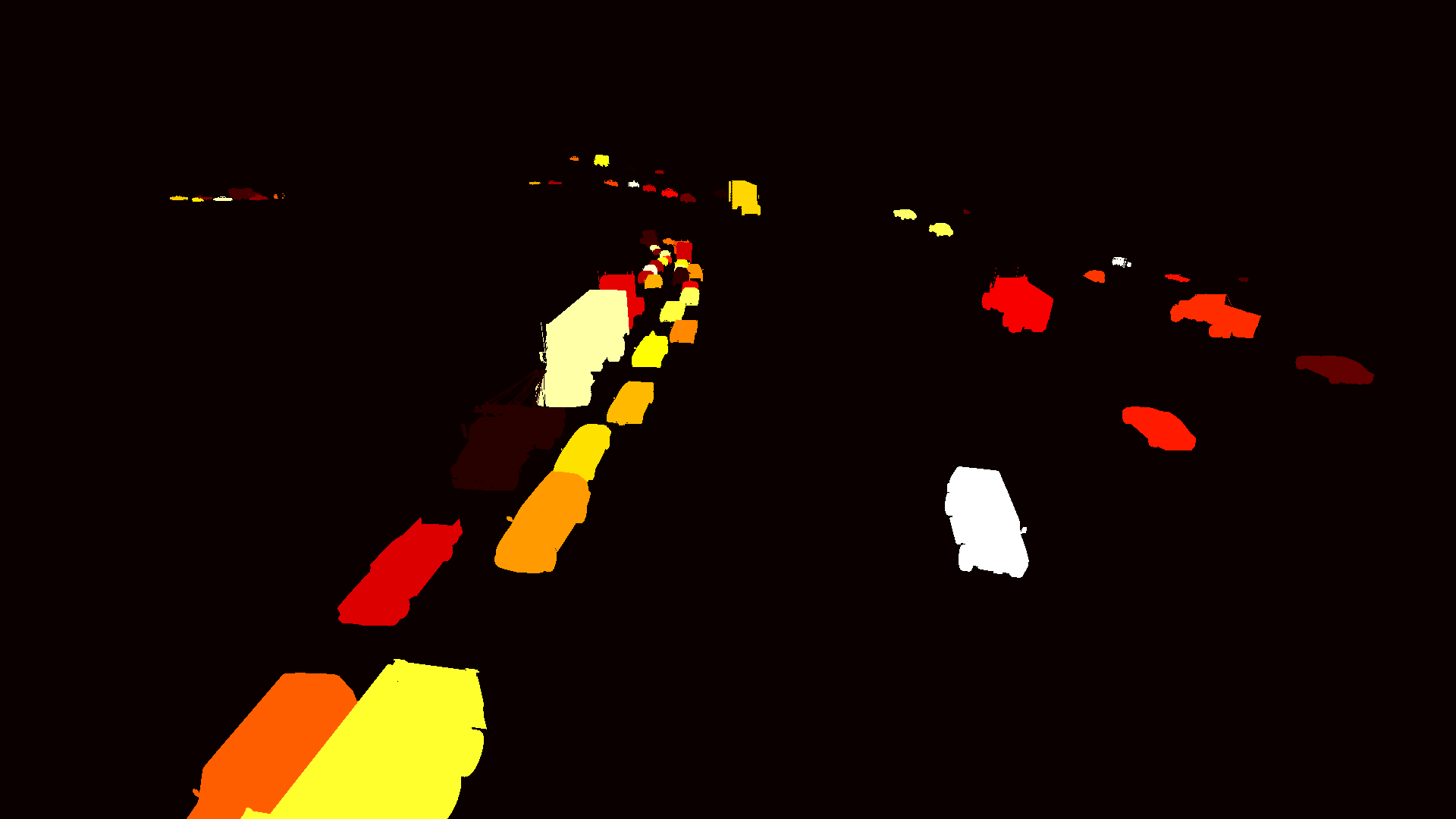}}

  \caption{\textbf{Samples of the \acrfull{gta} dataset.} We show images gathered from a video game together with the automatically-generated per-pixel annotations.}
  \label{gta_examples} 
\end{figure}

The \textit{\acrfull{gta}} dataset was also created by us on purpose and made publicly available. It is a vast collection of about 15,000 \textit{synthetic} images of urban traffic scenes collected using the highly photo-realistic graphical engine of the \textit{GTA V - Grand Theft Auto V} video game. About half of them concern urban city areas, while the remaining involve sub-urban areas and highways. To generate this dataset, we designed a framework that \textit{automatically} and precisely annotates the vehicles present in the scene with per-pixel annotations. To the best of our knowledge, this was the first \textit{instance} segmentation synthetic dataset of city traffic scenarios. As in the \acrshort{ndispark} dataset, the instance segmentation labels allowed us to produce accurate ground truth density maps for the counting task since the sizes of the vehicles were well-known.
In \ref{gta_examples}, we report some examples of images of the \acrshort{gta} dataset together with the annotations. 

\section{UDA for Traffic Density Estimation}
\label{sec:uda-counting:method}
Our method relies on a \acrshort{cnn} model trained end-to-end with adversarial learning in the output space (i.e., the density maps), which contains rich information such as scene layout and context. The peculiarity of our adversarial learning scheme is that it forces the predicted density maps in the target domain to have local similarities with the ones in the source domain. 

\begin{figure*}
\centerline{\includegraphics[width=.98\textwidth]{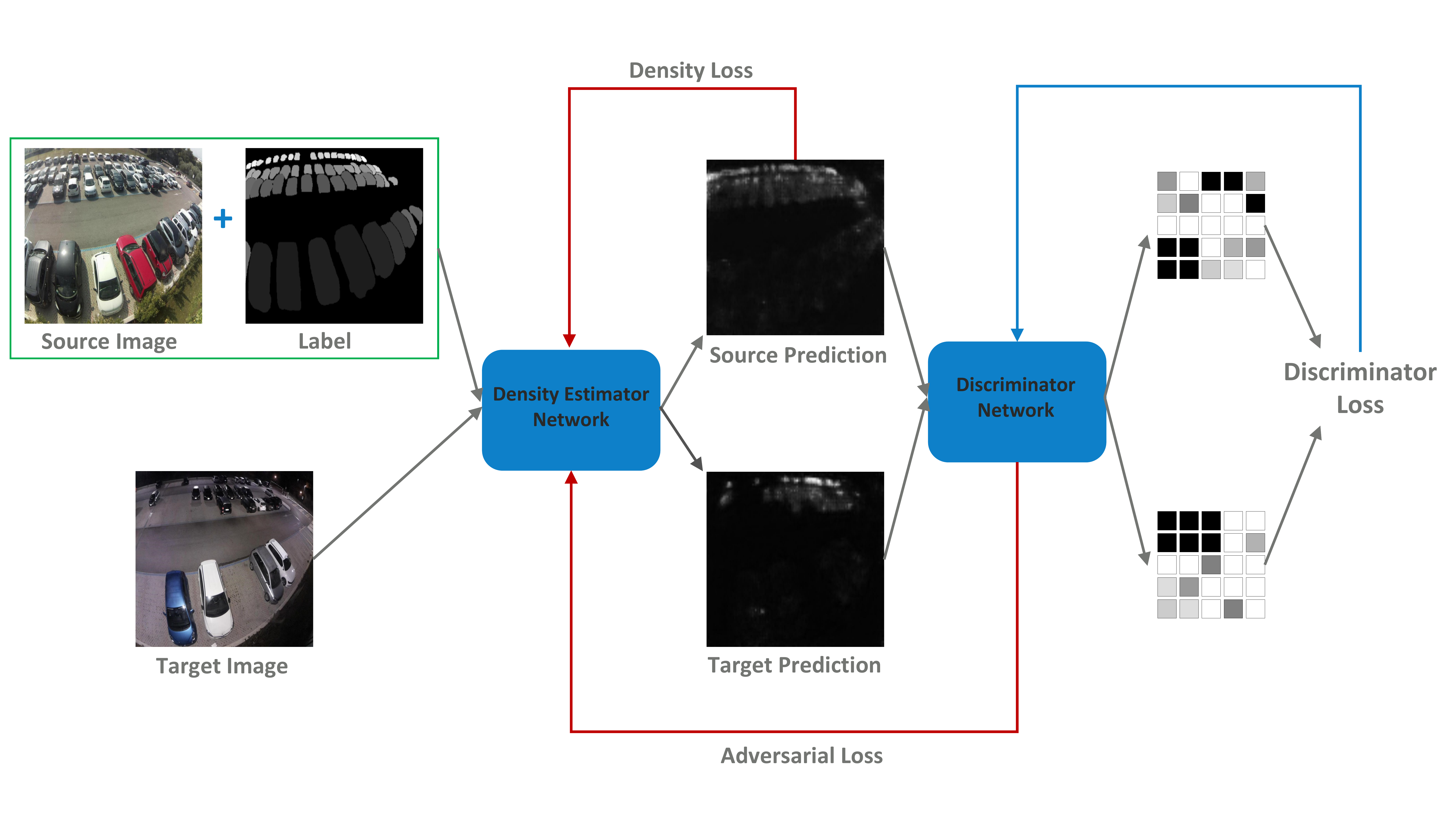}}
\caption{\textbf{High-level architecture of our proposed algorithm}. Given \(C \times H \times W\) images from source and target domains, we pass them through the density map estimation network to obtain output predictions. A density loss is computed for source predictions based on the ground truth. To improve target predictions, a discriminator is used to locally classify whether a density map belongs to the source or target domain. Then, an adversarial loss is computed on the target prediction and is back-propagated to the density map estimation and counting network.}
\label{overview_approach}
\end{figure*}

\ref{overview_approach} depicts the proposed framework consisting of two modules: i) a \acrshort{cnn} that predicts traffic density maps, from which we estimate the number of vehicles in the scene, and ii) a discriminator that identifies whether a density map (received by the density map estimator) was generated from an image of the source domain or the target domain. In the training phase, the density map predictor learns to map images to densities based on annotated data from the source domain. At the same time, it learns to predict realistic density maps for the target domain by trying to fool the discriminator with an adversarial loss. 
The output of the discriminator is a pixel-wise classification of a low-resolution map, where each pixel corresponds to a small region in the density map. Consequently, the output space is forced to be locally similar for both the source and target domains. In the inference phase, the discriminator is discarded, and only the density map predictor is used for the target images. We describe each module and how it is trained in the following subsections.

\subsection{Density Estimation Network}
We formulate the counting task as a density map estimation problem \cite{learning_to_count}. The density (intensity) of each pixel in the map depends on its proximity to a vehicle centroid and the size of the vehicle in the image so that each vehicle contributes with a total value of 1 to the map. Therefore, it provides statistical information about the location of the vehicles and allows the counting to be estimated by summing all density values (see also \ref{sec:back:visual-counting:traditional-approaches}.)

This task is performed by a \acrshort{cnn}-based model, whose goal is to automatically determine the vehicle density map associated with a given input image.
Formally, the density map estimator, $\Psi: \mathbb{R}^{C \times H \times W} \mapsto \mathbb{R}^{H \times W}$, transforms a $W \times H$ input image \(\mathcal{I}\) with $C$ channels, into a density map,  $D = \Psi(\mathcal{I}) \in \mathbb{R}^{H \times W}$. 

\subsection{Discriminator Network}
The discriminator network, denoted by \(\Theta\), also consists of a \acrshort{cnn} model. It takes as input the density map, \(D\), estimated by the network \(\Psi\). Its output is a lower resolution probability map where each pixel represents the probability that the corresponding region (from the input density map) comes either from the source or the target domain. The goal of the discriminator is to learn to distinguish between density maps belonging to source or target domains. Through an adversarial loss, this discriminator will, in turn, force the density estimator to provide density maps with similar distributions in both domains. In other words, the target domain density maps have to look realistic, even though the network \(\Psi\) was not trained with an annotated training set from that domain.

\subsection{Domain Adaptation Learning}

The proposed framework is trained based on an alternate optimization of the density estimation network, \(\Psi\), and the discriminator network, \(\Theta\). Regarding the former, the training process relies on two components: i) density estimation using pairs of images and ground truth density maps, which we assume are only available in the source domain; and ii) adversarial training, which aims to make the discriminator fail to distinguish between the source and target domains. As for the latter, images from both domains are used to train the discriminator on correctly classifying each pixel of the probability map as either source or target.

To implement the above training procedure, we use two loss functions: one is employed in the first step of the algorithm to train network \(\Psi\), and the other is used in the second step to train the discriminator \(\Theta\). These loss functions are detailed below.

\paragraph{Network \(\Psi\) Training.} We formulate the loss function for \(\Psi\) as the sum of two main components:

\begin{equation}
\mathcal{L}(\mathcal{I^\mathcal{S}}, \mathcal{I^\mathcal{T}}) = \mathcal{L}_{density}(\mathcal{I^\mathcal{S}}) + \lambda_{adv}\mathcal{L}_{adv}(\mathcal{I^\mathcal{T}}),
\label{global_loss_equation}
\end{equation}

\noindent where \(\mathcal{L}_{density}\) is the loss computed using ground truth annotations available in the source domain, while \(\mathcal{L}_{adv}\) is the adversarial loss that is responsible for making the distribution of the target and the source domain closer to each other. In particular, we define the density loss \(\mathcal{L}_{density}\) as the mean square error between the predicted and ground truth density maps, i.e. \(\mathcal{L}_{density} = MSE(D^{\mathcal{S}}, D^{\mathcal{S\_GT}})\).

To compute the adversarial loss \(\mathcal{L}_{adv}\), we first forward the images belonging to the target domain through network \(\Psi\), to generate the predicted density maps \(D^{\mathcal{T}}\). Then, we forward \(D^{\mathcal{T}}\) through network \(\Theta\), to generate the probability map \(P = \Theta(\Psi(\mathcal{I^\mathcal{T}})) \in [0,1]^{H' \times W'}\), where \(H'<H\) and \(W'<W\). The adversarial loss is given by
\begin{equation}
\mathcal{L}_{adv}(\mathcal{I^\mathcal{T}}) = - \sum_{h, w}\log (P_{h,w}),
\end{equation}
where the subscript \(h,w\) denotes a pixel in \(P\).
\noindent This loss makes the distribution of \(D^{\mathcal{T}}\) closer to \(D^{\mathcal{S}}\) by forcing \(\Psi\) to fool the discriminator, through the maximization of the probability of \(D^{\mathcal{T}}\) being locally classified as belonging to the source domain.

\paragraph{Network \(\Theta\) Training.} Given an image \(\mathcal{I}\) and the corresponding predicted density map \(D\), we feed \(D\) as input to the fully-convolutional discriminator \(\Theta\) to obtain the probability map \(P\). The discriminator is trained by comparing \(P\) with the ground truth label map \(Y \in \{0,1\}^{H' \times W'}\) using a pixel-wise binary cross-entropy loss
\begin{equation}
\begin{split}
\mathcal{L}_{disc}(\mathcal{I}) = - \sum_{h, w}(1-Y_{h,w})\log(1-P_{h,w}) + \\ 
+ Y_{h,w}log(P_{h,w}),
\end{split}
\end{equation}

\noindent where \(Y_{h,w} = 0 \;\; \forall \; h,w\) if \(\mathcal{I}\) is taken from the target domain and \(Y_{h,w} = 1\) otherwise.

\section{Experimental Evaluation}

\subsection{Implementation Details}
\label{sec:uda-counting:exp-settings}
This section provides some implementation details concerning the two modules making up our solution, i.e., the density map estimator and the discriminator.

\paragraph{Density Map Estimation and Counting Network.} We built our density map estimation network based on the Congested Scene Recognition Network (CSRNet) \cite{csrnet}. Here we briefly review some of the features characterizing this algorithm. For a more detailed description see also \ref{sec:back:visual-counting:cnn-based}. CSRNet provides a \acrshort{cnn}-based method that can understand highly congested scenes and perform accurate density estimation and counting. It is composed of two major components. The authors use the well-known VGG-16 network \cite{vgg_16} as the front-end for 2D feature extraction because of its strong transfer learning ability.
On the other hand, the back-end consists of dilated kernels. The basic concept of dilated convolutions is to deliver larger reception fields replacing the pooling operations. It is worth noting that the max pool operation is responsible for losing quality in the density generation procedure. Since the output size from VGG is reduced by a factor of 8 of the original input size, we up-sampled the final output to compare it with the ground truth density map.

\paragraph{Discriminator.} We used a \acrlong{fcn} similar to \cite{unsupervised_representation} and to \cite{learning_to_adapt}, composed of 5 convolution layers with kernel \(4\times4\) and stride of 2. The number of channels are \{64, 128, 256, 512, 1\}, respectively. Each convolution layer is followed by a leaky ReLU having a parameter equals to 0.2. \newline

We implemented the whole system using the PyTorch framework on a single Nvidia RTX 2080 GPU with 12 GB memory. To train the density estimator network and the discriminator, we used Adam optimizer \cite{adam_optimizer} with an initial learning rate set to \(10^{-5}\). During the training, it was crucial to balance the weight between density and adversarial losses. A small value of \(\lambda_{adv}\) may not help the training process significantly. In contrast, a larger value may propagate incorrect gradients to the density estimator. We empirically chose the value of \(\lambda_{adv}\) depending on the employed dataset.

\subsection{Experiments and Results}
\label{sec:uda-counting:exps-results}
\begin{figure*}
\centering
\begin{tabular}{cccc}
\includegraphics[width=.22\textwidth,height=2.4cm]{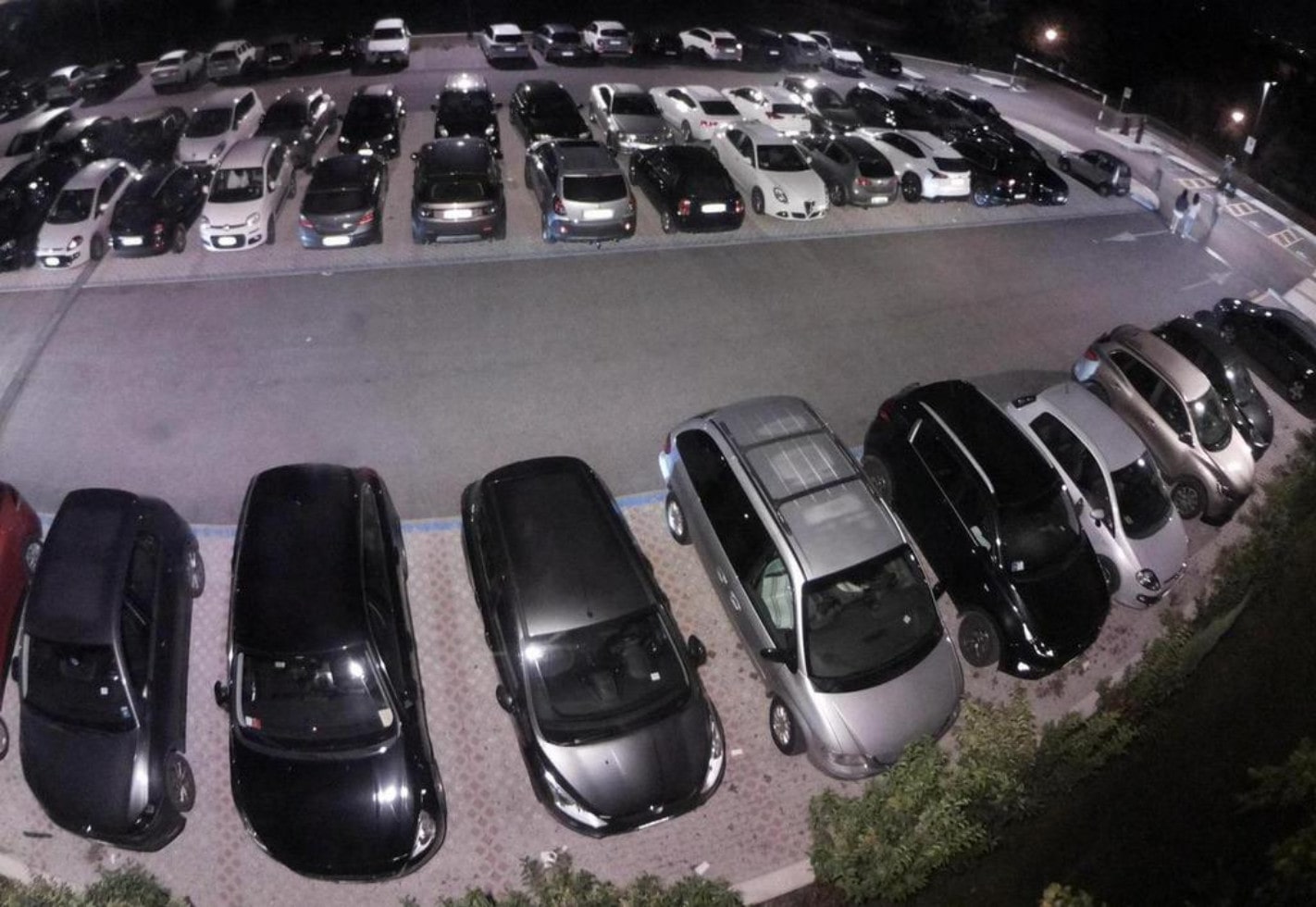} & 
\includegraphics[width=.22\textwidth,height=2.4cm]{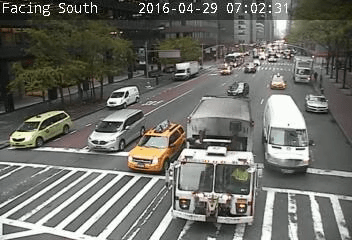} &
\includegraphics[width=.22\textwidth,height=2.4cm]{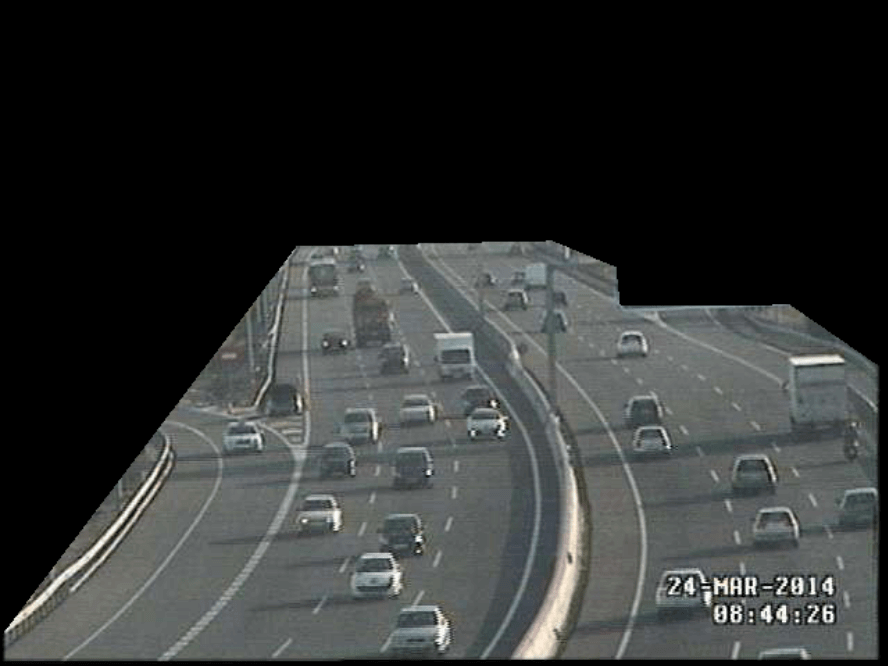} &
\includegraphics[width=.22\textwidth,height=2.4cm]{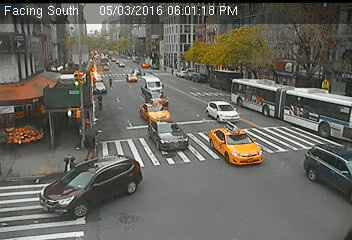}  \\
GT count: 56 & GT count: 13 & GT count: 35 & GT count: 12 \\
\includegraphics[width=.22\textwidth,height=2.4cm]{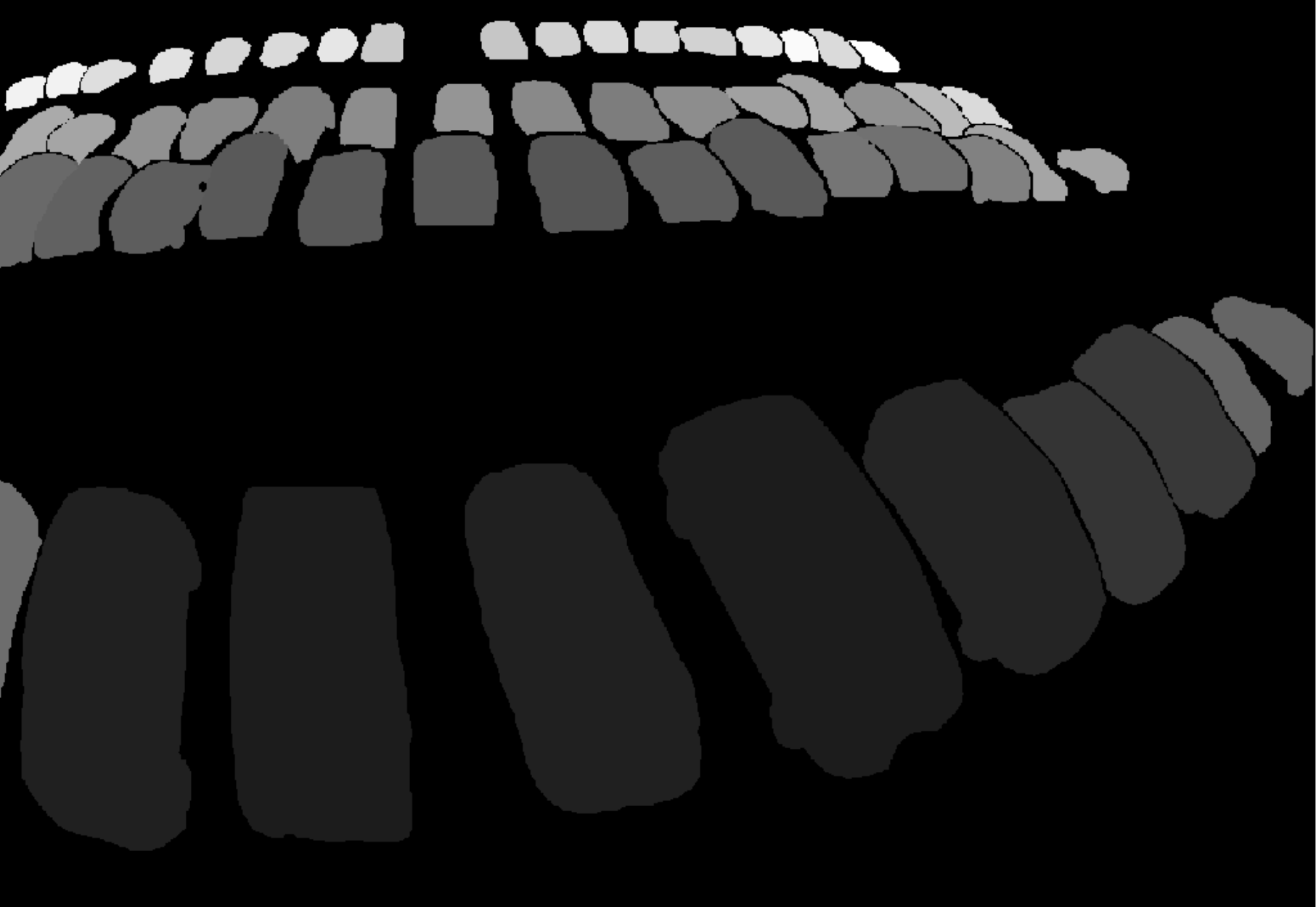} & 
\includegraphics[width=.22\textwidth,height=2.4cm]{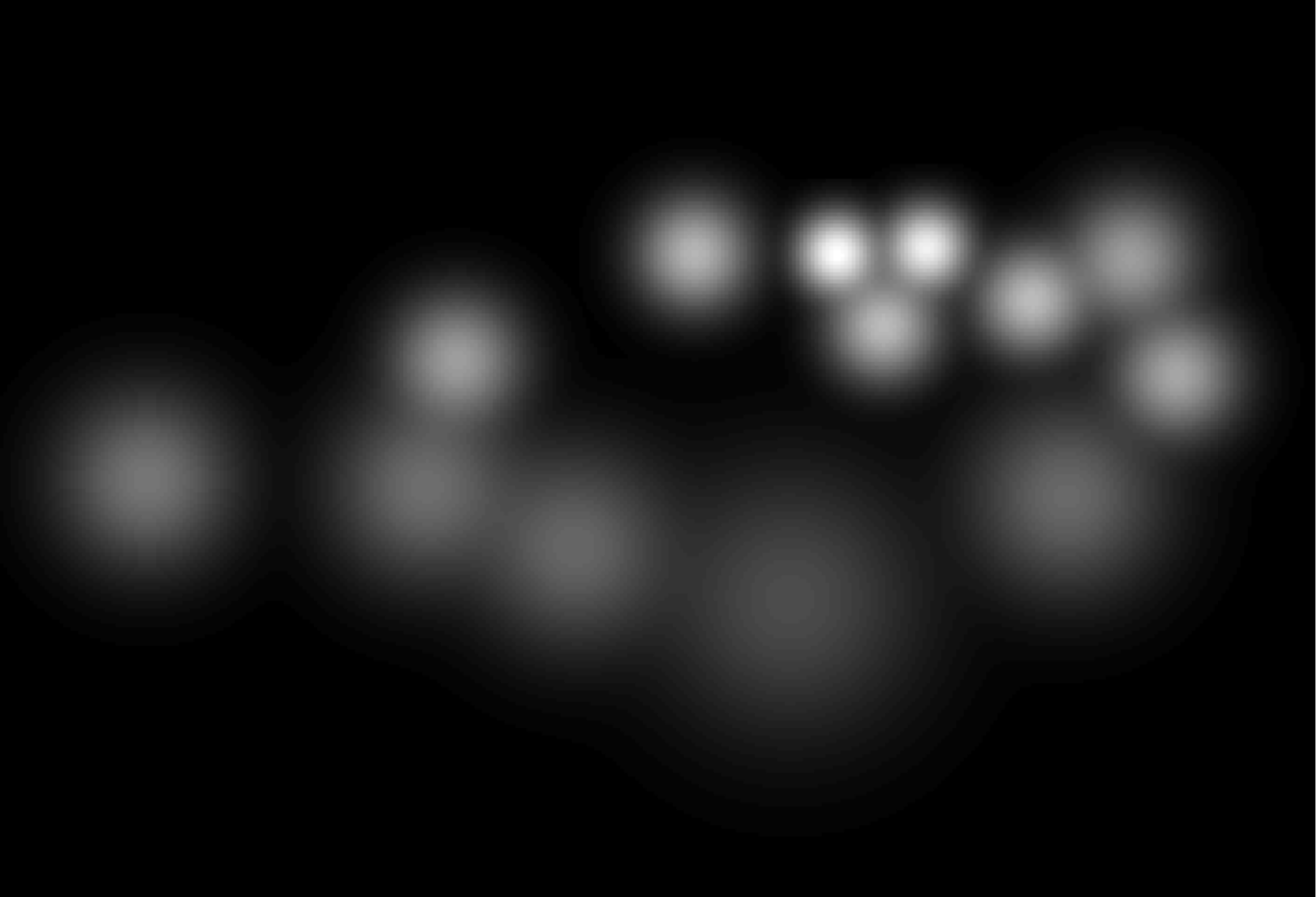} &
\includegraphics[width=.22\textwidth,height=2.4cm]{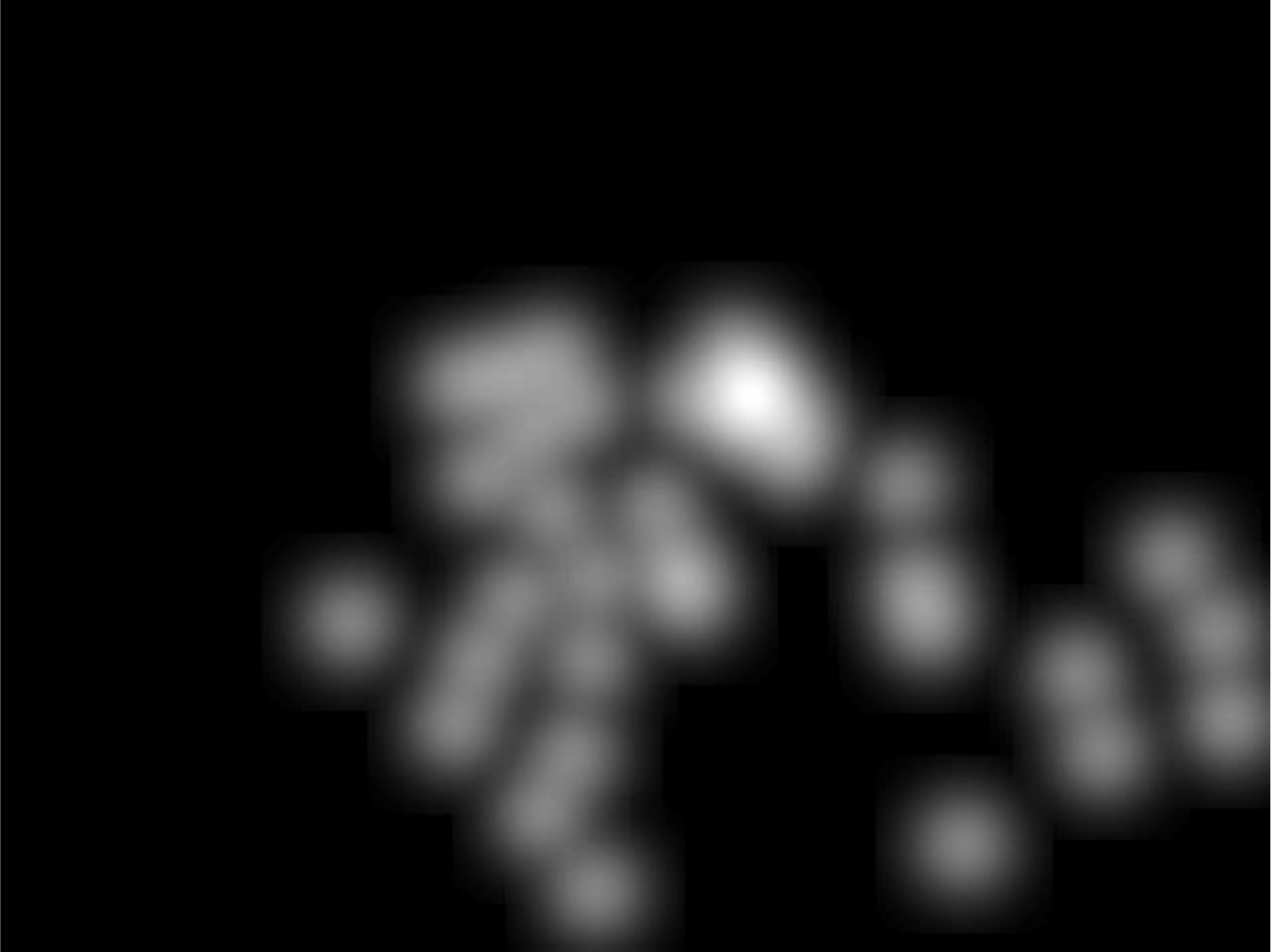} &
\includegraphics[width=.22\textwidth,height=2.4cm]{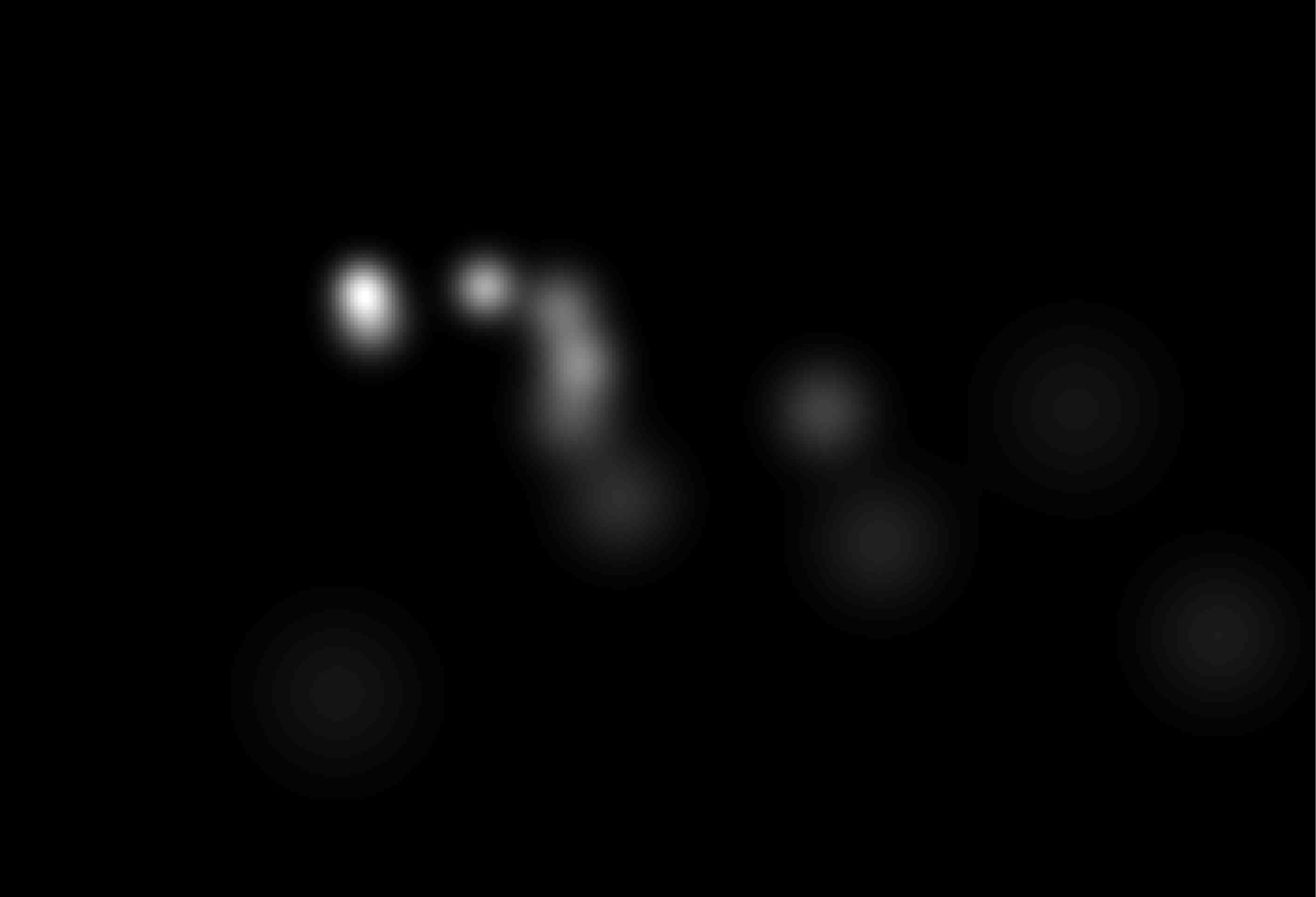} \\
Pred count: 53 & Pred count: 14 & Pred count: 38 & Pred count: 11 \\
\includegraphics[width=.22\textwidth,height=2.4cm]{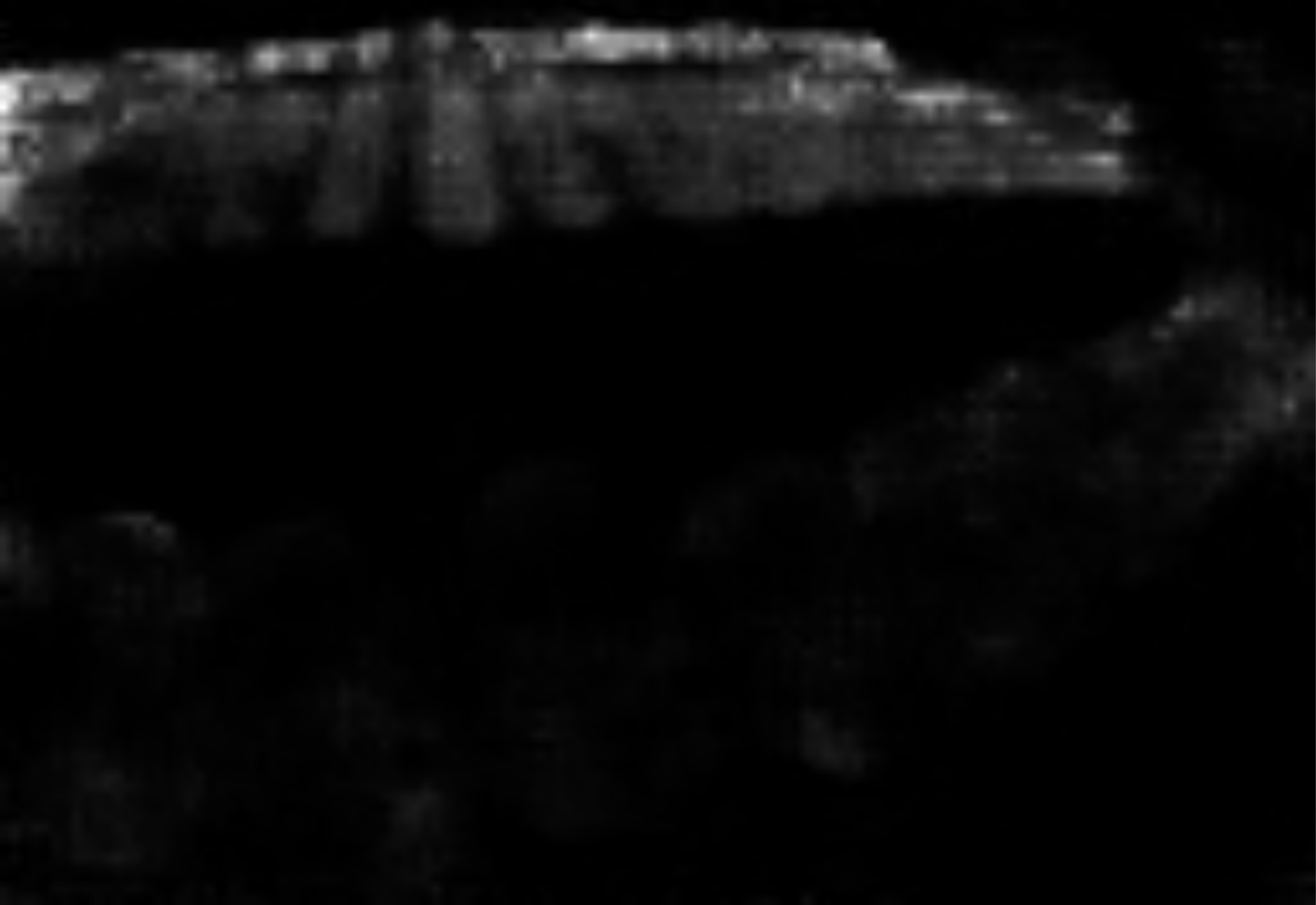} & 
\includegraphics[width=.22\textwidth,height=2.4cm]{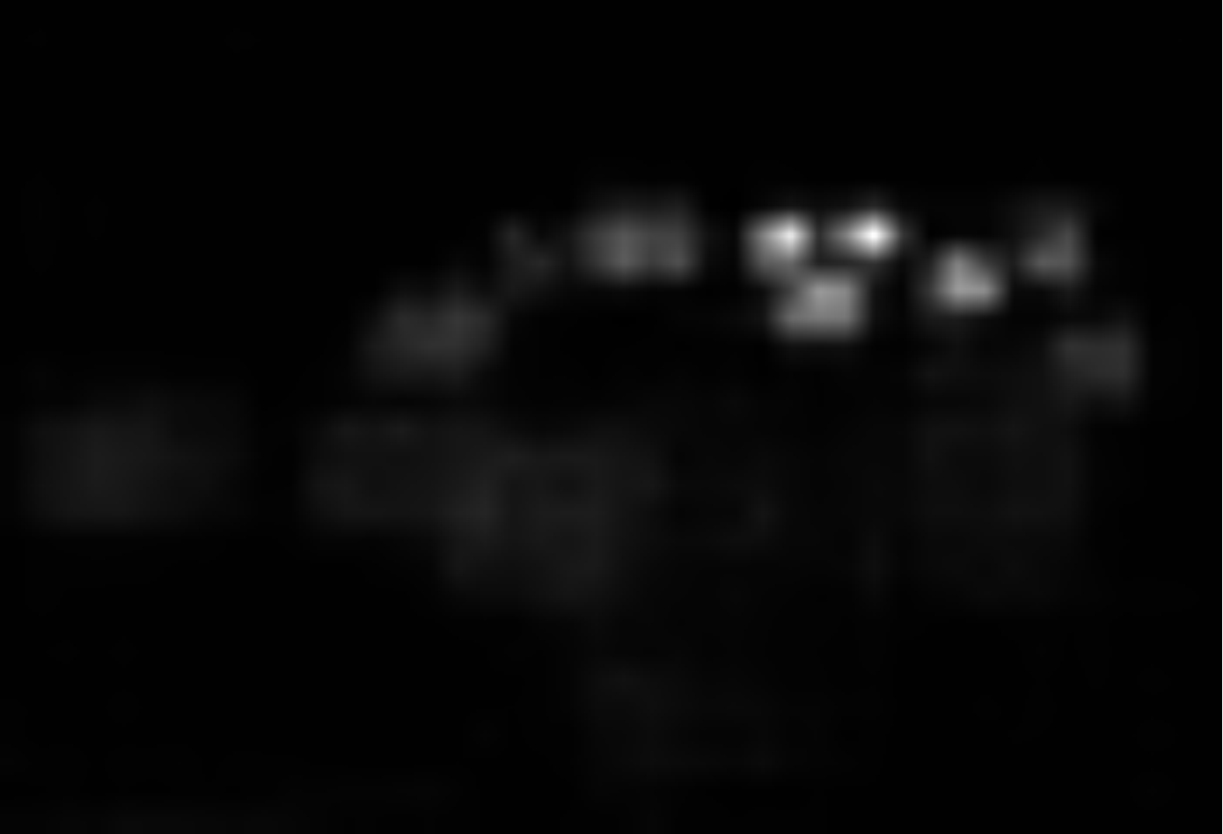} &
\includegraphics[width=.22\textwidth,height=2.4cm]{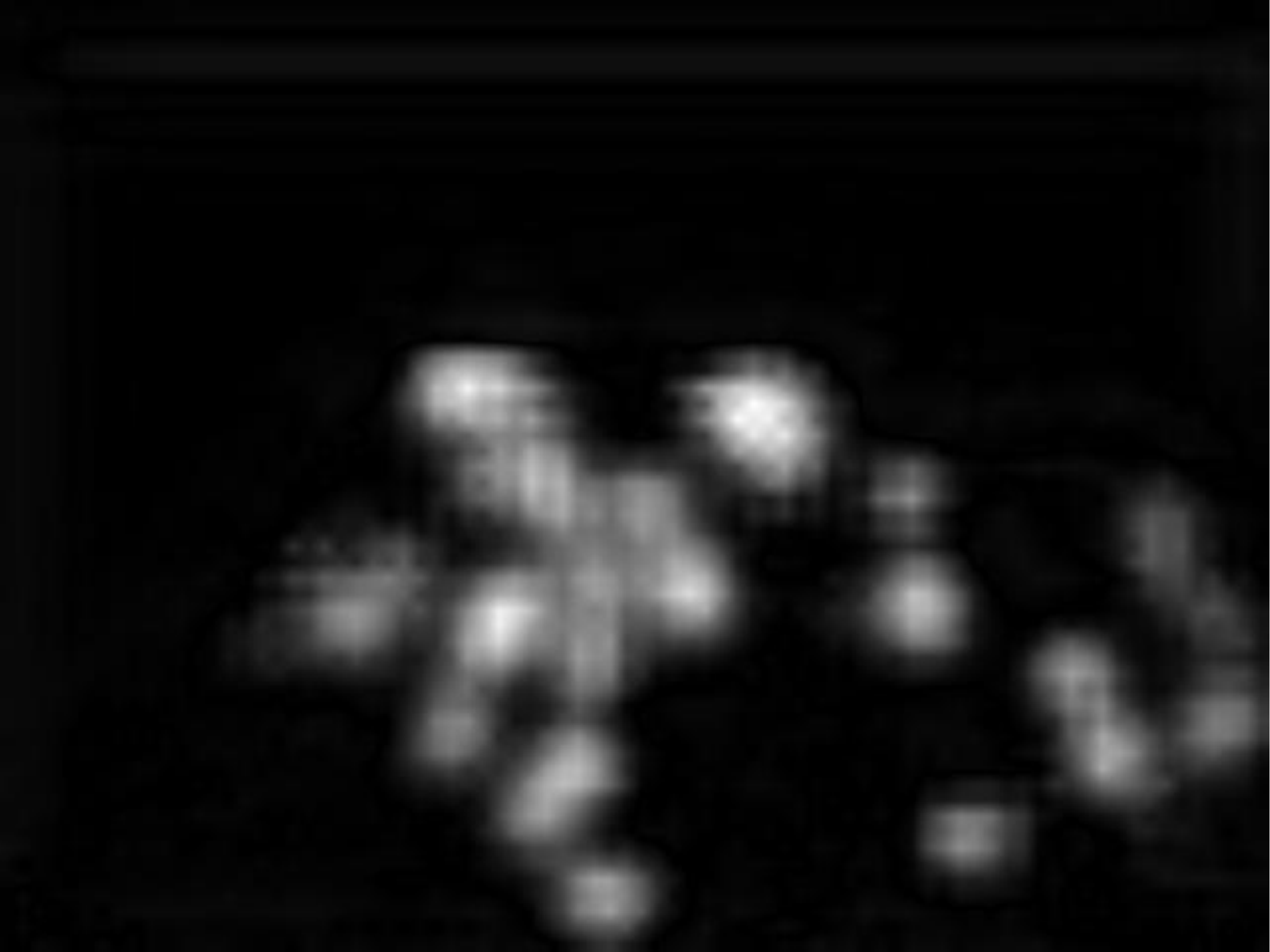} &
\includegraphics[width=.22\textwidth,height=2.4cm]{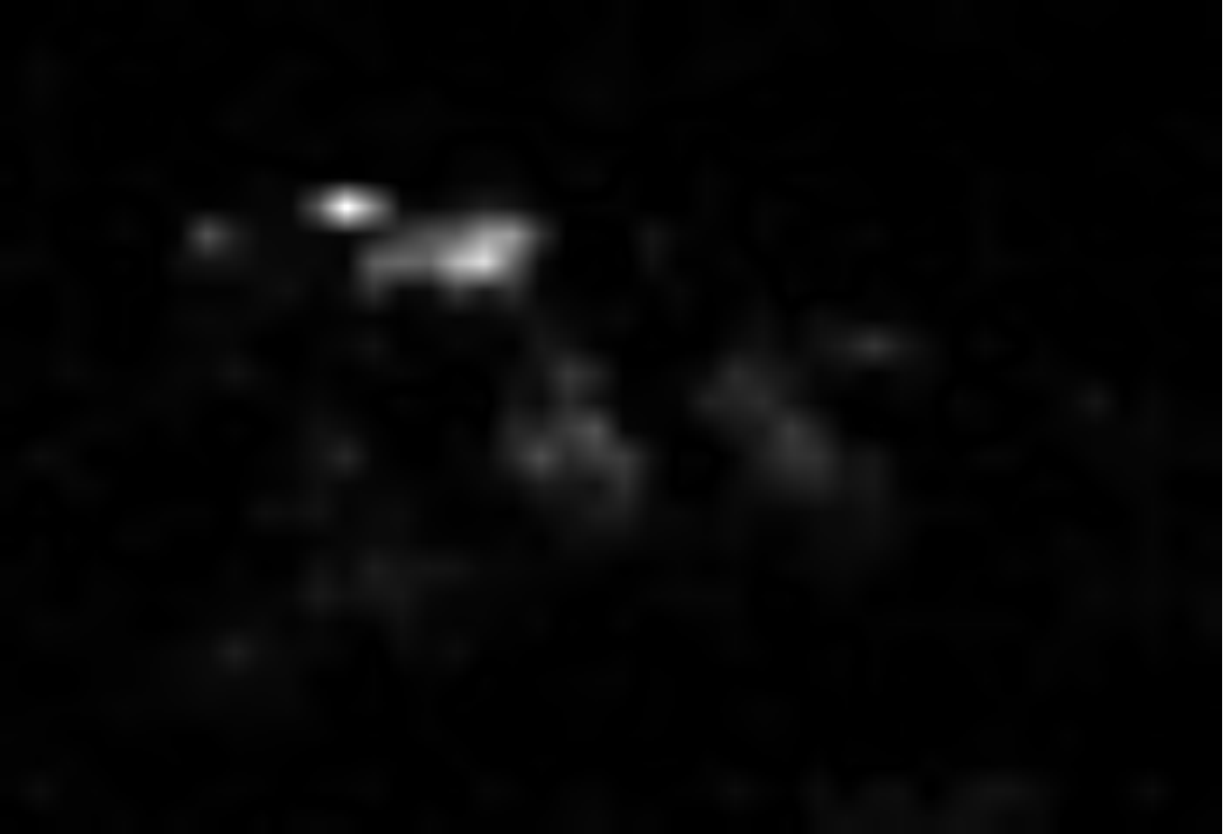} \\
   (a)  &  (b) & (c) & (d)
\end{tabular}
\caption{\textbf{Examples of the predicted density maps in the considered scenarios.} (a)  \textit{Day2Nigh} Domain Shift using the \textit{NDISPark} dataset; (b) and (c) \textit{Camera2Camera} Domain Shift employing the \textit{WebCamT} and \textit{TRANCOS} datasets, respectively; (d) \textit{Synthetic2Real} Domain Shift using the \textit{GTA} dataset for the training phase and the \textit{WebCamT} dataset for testing on real images. First row: input images; second row: ground truth; third row: predicted density maps.
}
\label{examples_pred_densities}
\end{figure*}

We validated the proposed \acrshort{uda} method for traffic density estimation and counting under different settings. First, we employed the \acrshort{ndispark} dataset, and we tested the \textit{Day2Night} domain shift; then, we utilized the WebCamT and the TRANCOS datasets to take into account the \textit{Camera2Camera} performance gap. Finally, we used the \acrshort{gta} dataset to consider the \textit{Synthetic2Real} domain difference. For all the experiments, we based the evaluation of the models on three metrics widely used for the counting task (see also \ref{sec:back:visual-counting:metrics} for more details): (i) \acrfull{mae} that measures the absolute count error of each image; (ii) \acrfull{mse} that instead quantifies the squared count error for each image; (iii) \acrfull{mare}, which measures the absolute count error divided by the true count. 
Results are summarized in \ref{results_table}, while in the following, we discuss the results obtained for every considered scenario. Finally, we show some examples of the outputs obtained using our models, showing their visual quality. In particular, \ref{examples_pred_densities} illustrates the ground truth and the predicted density maps for some random samples of the considered scenarios.

\begin{table*}
\caption{\textbf{Experimental results obtained for the four considered domain shift.} We employed three evaluation metrics: the \acrfull{mae}, the \acrfull{mse} and the \acrfull{mare}. We achieved performance improvements for all the scenarios, considering all the three metrics. }
\begin{center}
\begin{tabular}{ |p{5.5cm}||p{2.0cm}|p{2.0cm}|p{2.0cm}|  }
\hline
& \hfil MAE & \hfil MSE & \hfil MARE\\
 \hline
 \hline
 \multicolumn{4}{|c|}{\textit{Day2Night Domain Shift - \textit{NDISPark} Dataset}} \\
 \hline
Baseline - CSRNet \cite{csrnet}  & \hfil 3.95 & \hfil 27.45 & \hfil 0.43 \\
Our Approach & \hfil \textbf{3.49} & \hfil \textbf{20.90} & \hfil \textbf{0.39} \\
 \hline
 \hline
 \multicolumn{4}{|c|}{\textit{Camera2Camera Domain Shift - \textit{WebCamT} Dataset \cite{understandingCosteira}}} \\
 \hline
Baseline - CSRNet \cite{csrnet}  & \hfil 3.24 & \hfil 16.83 & \hfil 0.21 \\
Our Approach & \hfil \textbf{2.86} & \hfil \textbf{13.03} & \hfil \textbf{0.19} \\
\hline
 \hline
  \multicolumn{4}{|c|}{\textit{Camera2Camera Domain Shift - \textit{TRANCOS} Dataset \cite{ExtremelyTrancos}}} \\
 \hline
Hydra-CNN \cite{towards_perspective} & \hfil 10.99 & \hfil 68.70 & \hfil 0.71 \\
FCN-MT \cite{understandingCosteira} & \hfil 5.31 & \hfil - & \hfil 0.85 \\
LC-ResFCN \cite{counting_blobs} & \hfil 3.32 & \hfil - & \hfil - \\ 
Baseline - CSRNet \cite{csrnet}  & \hfil 3.56 & \hfil 30.64 & \hfil 0.10 \\
Our Approach & \hfil \textbf{3.30} & \hfil \textbf{23.60}  & \hfil \textbf{0.08} \\
 \hline
 \hline
  \multicolumn{4}{|c|}{\textit{Synthetic2Real Domain Shift - \textit{GTA} Dataset}} \\
 \hline
Baseline - CSRNet \cite{csrnet}  & \hfil 4.10 & \hfil 25.83 & \hfil 0.28 \\
Our Approach & \hfil \textbf{3.88} & \hfil \textbf{23.80} & \hfil \textbf{0.27} \\
 \hline
\end{tabular}
\end{center}
\label{results_table}
\end{table*}

\paragraph{Day2Night Domain Shift}
In this scenario, we split the \acrshort{ndispark} dataset into train, validation, and test subsets containing about 100, 50, and 100 images, respectively. The former had only pictures taken during the day (source domain), while the validation and the test subsets contained night images (target domain). To fairly evaluate our method, we first considered the baseline model without the domain adaptation module (i.e., putting the \(\lambda_{adv}\) value to zero). Then, we added the adversarial module to compare the results. In both cases, we trained the network for 300 epochs, validating at each iteration. We chose the best validation model in terms of \acrshort{mae}, and we tested it against the test set. As shown in \ref{results_table}, using our solution, we obtained performance improvements considering all three metrics.

\paragraph{Camera2Camera Domain Shift}
In this case, we performed two sets of experiments to test the domain shift that takes place when we consider a camera different from the ones used in the training phase. 

First, we considered the WebCamT dataset, and we split it into train, validation, and test subsets. In the former, we accounted for about 25,000 images belonging to 7 cameras (source domain). In the last two, we considered the remaining 15,000 pictures of 3 different cameras, having diverse contexts and slightly different angles of view (target domain). We compared the baseline and our solution when training for 20 epochs, validating it at each iteration and choosing the best model in terms of \acrshort{mae}.

Second, we took into account the TRANCOS dataset. We split it into train, validation, and test sets, following \cite{ExtremelyTrancos}. The train set represented the source domain, while the other two belonged to the target domain and were collected in different contexts. We trained our domain adaptation for 200 epochs, picking the best validation model in terms of \acrshort{mae}, and we evaluated it against the test set. We compared the obtained results with the ones claimed by \cite{csrnet} using only the state-of-the-art CSRNet algorithm (i.e., our baseline) and with other state-of-the-art techniques present in the literature.

As shown in \ref{results_table}, we obtained performance improvements in both cases, taking into account all three metrics. Considering the publicly available TRANCOS dataset, we achieved superior results not only concerning the baseline but also compared to the other considered approaches. 

\paragraph{Synthetic2Real Domain Shift}
In this scenario, we trained the algorithm using synthetic images. Then we tested it on real-world data. In particular, we considered a subset of the \acrshort{gta} dataset containing about 5,000 images of city traffic scenarios, and we used it as the training set (source domain). On the other hand, we accounted for the test and the validation subsets of the WebCamT dataset as the target domain. We compared the results obtained using the baseline model and our solution with the domain adaptation module. In both cases, we trained the algorithm for 20 epochs, validating at each iteration. We chose the best model in terms of \acrshort{mae}. 

Again, as shown in \ref{results_table}, we achieved better results compared to the basic model. We believe that this scenario is particularly interesting because we obtained comparable results with the previous one, but this time \textit{without} using manual annotations neither in the source domain nor in the target one.

\section{Summary}
\label{sec:uda-counting:summary}
In this chapter, we tackled the problem of determining the density and the number of objects present in large sets of images. The proposed methodology,  built on a \acrshort{cnn}-based density estimator, was able to generalize to new data sources for which there were no annotations available. We achieved this generalization by exploiting an \acrfull{uda} strategy, whereby a discriminator attached to the output forced similar density distribution in the target and source domains. Experiments showed a significant improvement relative to the performance of the model without domain adaptation. To the best of our knowledge, we were the first to introduce a \acrshort{uda} scheme for counting to reduce the gap between the source and the target domain without using additional labels. 

Another contribution is represented by the creation of two new per-pixel annotated datasets made available to the scientific community. One of the two novel datasets was a synthetic dataset created from a photo-realistic video game. Here the labels were automatically assigned while interacting with the API of the graphical engine. Using this synthetic dataset, we demonstrated that it is possible to train a model with data gathered from virtual worlds and to perform \acrshort{uda} toward a real-world scenario, obtaining outstanding performance \textit{without} using additional manual annotations. In other words, we addressed the problem of data scarcity from two complementary sides: on the one hand, we exploited the significant variability of the synthetic data, while, on the other hand, we mitigated the domain gap existing between the synthetic and the real-world images in an \textit{unsupervised} fashion, thus without using the labels of the test domain as in the previous chapter.

\graphicspath{{img/counting-with-uncertainty/}}

\chapter{Counting Biological Structures with Raters’ Uncertainty}
\label{ch:counting-with-uncertainty}

Detection and counting of biological structures are among the earliest fields revolutionized by artificial neural networks now dominating state of the art.
Several vision models (mostly convolutional networks) have been successfully adopted to localize, segment, and count cells or other structures from microscopy images and even provide counting-density estimation particularly effective in ``crowded'' scenarios. However, as already seen in the previous chapters concerning other tasks and applications, the success of these methods assumes the availability of a large-scale representative set of \textit{well-labeled} images.
In this chapter, compared to the previous ones, we consider the problem of data scarcity in a different setting.
Whereas, in most cases, objects to be counted can be unambiguously flagged by human raters, here we investigate cell detection and counting under the assumption of \textit{weak multi-rater labels}, that is, in the presence of non-negligible disagreement between multiple raters.
This often occurs when trying to detect and count cells with non-trivial patterns on a large scale, where several factors can produce weak labels;
raters can incur errors due to fatigue or inexperience (common when hiring less-experienced raters to reduce labeling time) or have different judgments that can span from conservative to liberal when assigning labels.

More reliable labels can be obtained by naively averaging the decisions taken by several raters on the same data, i.e., multi-rating can be leveraged to create stronger singular annotations.
However, such data are expensive to obtain and often available only in small quantities. On the other hand, given the scale of training sets needed for deep learning methodologies and the counting task, we consider here the case in which few expert raters, on a limited labeling budget, tend to label new data rather than label the same images more than once. This results in large, single-rater weakly labeled datasets and only small multi-labeled subsets~\cite{campagner2021ground}.

In this setting, we proposed a two-stage counting methodology for biological structures in microscopy images, where each stage is devised to fully exploit the annotations in each data subset.
The first stage adopts existing solutions on weakly-labeled data to detect and count cells. Specifically, we compared three common CNN-based methodologies already present in the literature --- a) \textit{segment and count}, b) \textit{detect and count}, and c) \textit{count by density estimation}. The goal was to investigate their counting ability when trained with data characterized by significant label noise from errors introduced by raters, and to derive \textit{uncalibrated} scores from the models' output that have not been designed to correlate with the quality of the predictions.
In the second stage, using a small set of multi-rater data, we defined a rescoring model that refines predictions of the first stage, increasing the correlation between the scores assigned by the model to the predictions and the raters' agreement on the sample labels. We referred to scores produced in this stage as \textit{calibrated} scores, in contrast with the uncalibrated ones previously assigned; these final scores can eventually be used to filter low-quality predictions by practitioners.
Advantages in operating in two-stage are twofold: i) the localization of objects are decoupled from their scoring, thus obtaining an overall improved counting model when the latter is fine-tuned even on a few multiple raters' judgments, and, ii) we can easily swap the first stage with any state-of-the-art localization and counting method, making the pipeline model-agnostic and ``future proof'' - any subsequent work can simply plug-in the best detector and still use the proposed pipeline when multi-rater data is available.

We evaluated the various stages of our pipeline on a novel weakly-labeled dot-annotated dataset that we publicly released.
It consists of a collection of fluorescence microscopy images of mice brain slices containing Perineuronal Nets (PNNs), extracellular matrix aggregates surrounding the cell body of a large number of neurons throughout the nervous system.
Multiple expert raters have labeled a small part of the dataset; nonetheless, the maximum agreement between raters is roughly 70\%, highlighting the need for an automated counting technique that accounts for uncertain patterns.
We showed through experimental evaluation that our proposed two-stage pipeline, independently from the specific implementation of each stage, improve the performance of several state-of-the-art counting methods on multiple ground-truth settings, from liberal to conservative ones.

The main contributions presented in this chapter can be summarized as follows:
\begin{itemize}
\item the proposal of a two-stage pipeline that improves biological structures counting in weak-labels settings,
\item the introduction of a novel dot-annotated dataset for cell counting in microscopy images (specifically, perineuronal nets) comprised of a large weakly-labeled single-rater subset and a smaller multi-rater subset, and
\item the public release of the first pre-trained models for automatic perineuronal nets counting in fluorescence images (at the best of our knowledge).
\end{itemize}

We organize the rest of the chapter as follows.
We review related work to our in \ref{sec:counting-with-uncertainty:related_works}.
In \ref{sec:counting-with-uncertainty:datasets}, we describe the datasets used in our experiments. 
\ref{sec:counting-with-uncertainty:method} formalizes the proposed methodologies, while \ref{sec:counting-with-uncertainty:experiments} outlines the performed experiments showing the obtained results.

The research presented in this chapter was published in \cite{DBLP:conf/visapp/CiampiCAG22, luca_ciampi_2021_5567032}.
Code and trained models are publicly available at \\ \href{https://github.com/ciampluca/counting\_perineuronal\_nets}{https://github.com/ciampluca/counting\_perineuronal\_nets}.

\section{Related Works}
\label{sec:counting-with-uncertainty:related_works}
This section overviews some works related to ours concerning counting approaches tailored to estimating the number of cells or structures in microscopy images. We also review some works focusing on learning with multi-rater data.

\paragraph{Microscope Cell Counting.}
Counting biological structures like cells in microscopy images is a crucial step to diagnose many diseases~\cite{venkatalakshmi2013automatic} and to understand cellular and molecular mechanisms~\cite{solnica2005conserved}.
Several automatic cell counting methods have been proposed over the years to facilitate this tedious and challenging task.
Compared to a typical counting task, microscopy images present different challenges, such as low image contrast, significant cell shape and count variance, and superposition of cells, leading to occlusions.
As such, both detection-based and regression-based methods have been proposed, each with the advantages and the drawbacks already highlighted in \ref{sec:back:visual-counting}. Basically, regression techniques have shown superior performance compared to the detection-based ones in crowded scenarios where the instances of the objects are sometimes not clearly visible due to occlusions and clumps; however, they cannot precisely localize the objects present in the scene, eventually providing only a coarse position of the area in which they are distributed. 
A relevant example belonging to the detection-based category is~\cite{Arteta_2016}, where the authors introduced a tree-structured discrete graphical model exploited to select and label a set of non-overlapping regions in the image by a global optimization of a classification score. More recently, the authors of~\cite{paulauskaite2019deep} exploited the popular Mask R-CNN~\cite{mask_rcnn} instance segmentation framework to detect overlapping cells, whereas~\cite{Dou_2017} used a \acrshort{cnn} to segment biological structures from 3D medical images. A comprehensive survey about deep learning algorithms used in medical image analysis, including cell detection and segmentation in microscopy images, is given by~\cite{med_survey}.

Recent efforts also focused on regression-based approaches that cope better with overlapped objects and crowded scenarios. For example,~\cite{saunet} proposed SAU-Net, an extension of the U-Net segmentation network~\cite{unet} with a Self-Attention module for counting by density regression. In~\cite{improving_with_heatmap}, another regression-based counting model is introduced, enhanced by regulating activation maps from the final convolution layer of the network with coarse ground-truth activation maps generated from simple dot annotations. More, in~\cite{count-ception}, the authors proposed a novel deep neural network architecture adapted from the Inception family~\cite{inception} of networks called Count-ception.
In~\cite{Huang_2020}, the so-called CSRNet~\cite{csrnet}, a regression-based CNN suitable for counting objects in several contexts (already employed in \ref{ch:uda-counting} for traffic density estimation), is employed to estimate cell densities in immunohistochemically stained sections of breast tissue. Jiang and Yu proposed two different regression-based cell counting approaches~\cite{two_path, foreground_mask}, again, based on the estimation of density maps.
Finally, authors in~\cite{he2021deeply} presented another regression model based on density estimation where auxiliary convolutional neural networks are employed to assist in the training of intermediate layers.
Other regression-based strategies have also been devised to deal with densely concentrated cells but still generating individual cell detections, such as~\cite{falk2019u},~\cite{tofighi2019prior},~\cite{Koyuncu_2020} and~\cite{xie2018efficient}.
These approaches first generate intermediate maps that indicate the likelihood of each pixel being the center of a cell in the image, and then convert them into detections by applying some form of Non-Maximum Suppression (NMS).

Concerning the automatic counting of PNNs, previous solutions are often based on brittle hand-crafted computer vision pipelines, such as in~\cite{Slaker_2016}.
To the best of our knowledge, we are the first to use deep-learning solutions to address the counting of perineuronal nets and its specific challenges, such as the extreme inter-image variance of the number and the non-trivial appearance of PNNs that cause difficulty to precisely count them, even for human experts.

\paragraph{Learning with multi-rater data}
When dealing with multi-rater data, most existing methodologies apply simple strategies like majority voting to obtain a unique set of ground-truth labels. However, approaches exploiting multi-rater data more effectively exist and are not new; in their seminal work, \cite{dawid1979maximum} proposed an Expectation-Maximization algorithm to estimate raters' error-rates in multinomial multi-rater data.
More recent works aim at modeling raters' reliability for aggregating or filtering labels, such as~\cite{rodrigues2013learning} and~\cite{zhang2012integration}. We refer the reader to \cite{zheng2017truth} for a review of approaches and challenges in inference with multi-rater data.
The recent trend is instead increasingly exploiting multi-rater data, when possible, to increase data efficiency;
in the biomedical context, Wei et al.~\cite{wei2021learn} proposed a curriculum learning approach on samples with increasing raters' agreement for histopathology image classification. In contrast, Mirikharaji et al.~\cite{mirikharaji2021d} tackles skin-lesion segmentation by building multiple models (one for each set of raters' labels) and then aggregating models predictions.
To the best of our knowledge, the only proposed counting approach dealing with multi-rater data is~\cite{counting_wild}, where authors train a supervised algorithm to count antarctic penguins in images dot-annotated by non-professional volunteers; multi-rater labels are mainly exploited to estimate the object scale, which varies wildly in their dataset (the diameter of a penguin varies between 15 and 700 pixels) and is instead fixed in our scenario.
When dealing with constant scale objects, as in our microscopy images scenario, their solution resembles~\cite{falk2019u}, a segmentation-based approach adopted and compared in this work.
Moreover, instead of requiring large multi-rater training sets, our approach is designed to train on a large single-rater set plus a small multi-rater set, lowering the total labeling cost.

\section{Datasets}
\label{sec:counting-with-uncertainty:datasets}
In this section, we describe the employed datasets, summarized in \ref{tab:datasets}. We considered four publicly available single-rater datasets widely used in the context of the microscope cell counting task that we exploit for comparing the adopted counting architectures against the state of the art. Those have served as baselines for our counting framework. Then, we illustrate our novel collection of fluorescence microscopy images containing perineuronal nets labeled by multiple professional raters, which we used for the experimental evaluation of our two-stage counting pipeline.

\begin{table}
\caption{\textbf{Summary of datasets.} We report some numerical characteristics on the top of the table. Below, we show a dataset image sample (for PNN, we show a 640x640 crop) and, in the last three rows, the associated targets exploited during training. Specifically, the targets are generated from dot annotations using different procedures: i) \textit{bounding boxes} are produced by generating squares with side \textit{s}, ii) \textit{density maps} are built by superimposing Gaussian kernels $G_\sigma$, and iii) \textit{segmentation maps} are generated drawing discs with radius \textit{r} separated by background ridges. Bounding boxes, Gaussian kernels and discs are centered in the dot-annotated locations; the \textit{s}, \textit{$\sigma$}, and \textit{r} parameters are fixed and dataset-specific, depending on the typical object size in the images. Targets in the multi-class BCData dataset are shown in false colors.}%
\label{tab:datasets}%
\renewcommand\tabularxcolumn[1]{m{#1}}%
\newcolumntype{C}{>{\centering\arraybackslash}m{.189\linewidth}}%
\newcolumntype{Y}{>{\centering\arraybackslash}m{.0945\linewidth}}%
\setlength{\tabcolsep}{0pt}%
\renewcommand*{\arraystretch}{0}%
\newcommand{\imgwidth}{.189\linewidth}%
\footnotesize%
\begin{tabularx}{\linewidth}{m{.0275\linewidth}CCCCYY}
\toprule
&   &   &   &   & \multicolumn{2}{c}{PNN}   \\[0.5ex]                                       \cmidrule{6-7}
& VGG~\cite{learning_to_count} & MBM~\cite{mbm_original} & ADI~\cite{count-ception} & BCData~\cite{Huang_2020} & 1 rater (PNN-SR)              & 7 raters (PNN-MR)       \\[0.5ex]
\midrule
subjects   & none (synthetic)   & 8     & N/A   & 394   & 1     & 1 \\[2ex]
images   & 200  & 44  & 200  & 1,338    & 25    & 12    \\[2ex]
size   & 256$\times$256     & 600$\times$600    & 150$\times$150    & 640$\times$640  & \scriptsize $\ge$8184$\times$6163 & \scriptsize 2000$\times$2000 \\[2ex]
objects    & 35,192       & 5,553       & 29,684    & 181,074   & 34,620    & 2,351 \\[2ex]
obj./img. & 176$\pm$61  & 126$\pm$33  & 148$\pm$32  & 135$\pm$68   & 1,385$\pm$590  & 196$\pm$43    \\
\bottomrule
\multicolumn{1}{c}{\hspace{-0ex}\rotatebox[origin=c]{90}{Sample}}%
&\multicolumn{6}{c}{%
\includegraphics[align=c,width=\imgwidth]{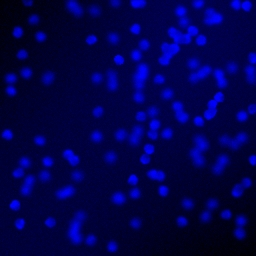}%
\includegraphics[align=c,width=\imgwidth]{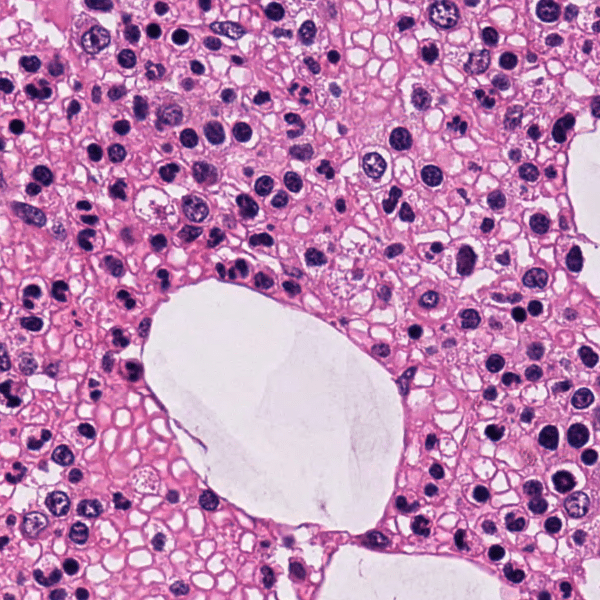}%
\includegraphics[align=c,width=\imgwidth]{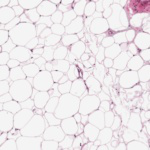}%
\includegraphics[align=c,width=\imgwidth]{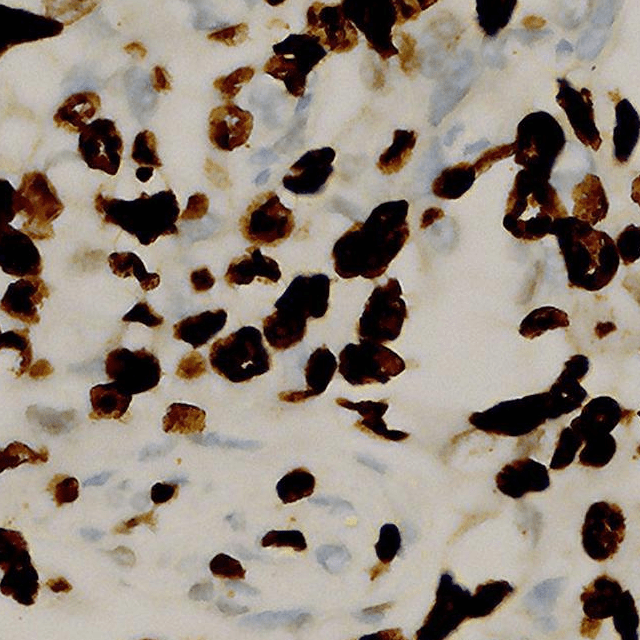}%
\includegraphics[align=c,width=\imgwidth]{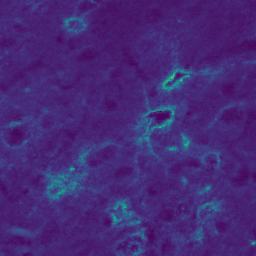}%
}\\%
\multicolumn{1}{c}{\hspace{-0ex}\rotatebox[origin=c]{90}{Bounding Boxes}}%
&\multicolumn{6}{c}{%
\includegraphics[align=c,width=\imgwidth]{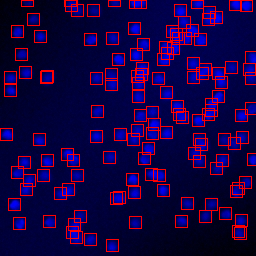}%
\includegraphics[align=c,width=\imgwidth]{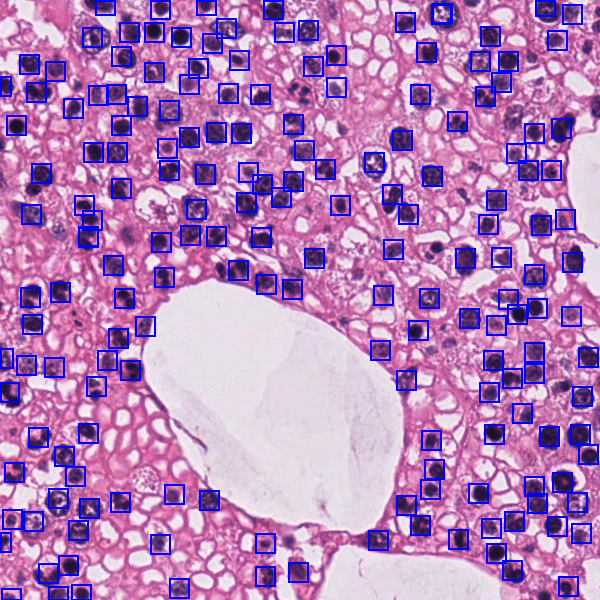}%
\includegraphics[align=c,width=\imgwidth]{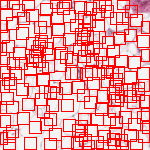}%
\includegraphics[align=c,width=\imgwidth]{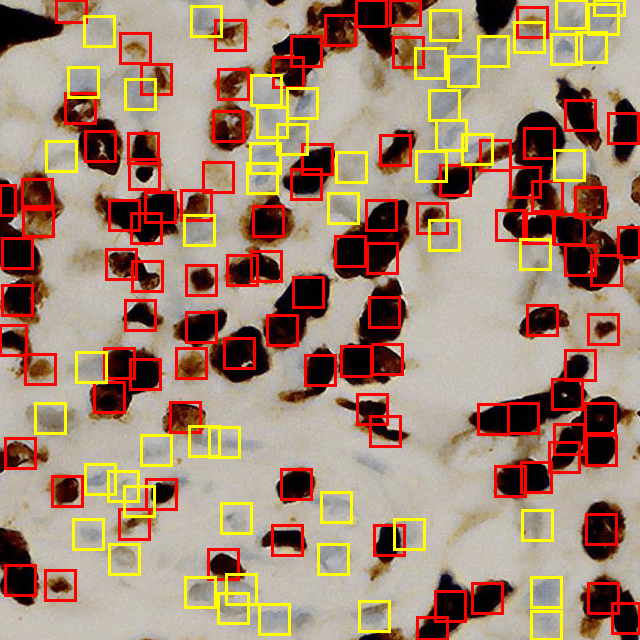}%
\includegraphics[align=c,width=\imgwidth]{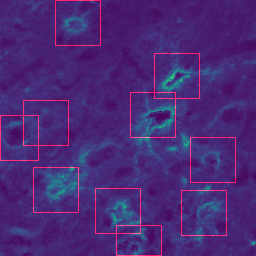}%
}\\%
\multicolumn{1}{c}{\hspace{-0ex}\rotatebox[origin=c]{90}{Density Map}}%
&\multicolumn{6}{c}{%
\includegraphics[align=c,width=\imgwidth]{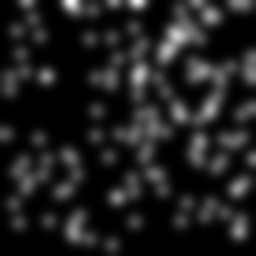}%
\includegraphics[align=c,width=\imgwidth]{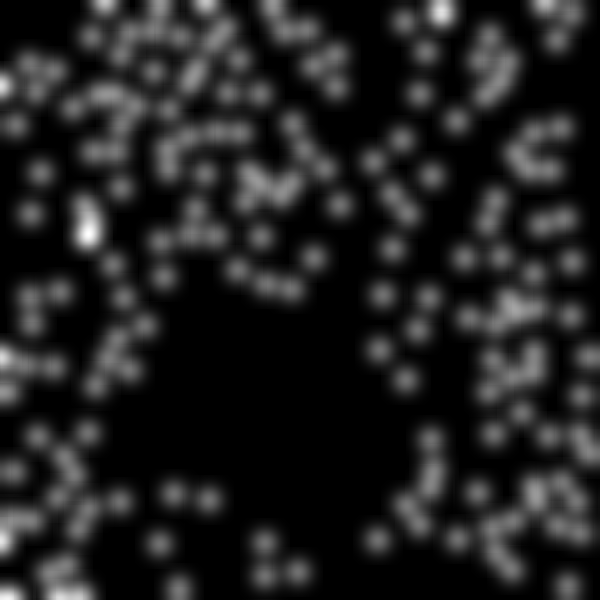}%
\includegraphics[align=c,width=\imgwidth]{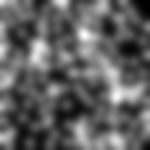}%
\includegraphics[align=c,width=\imgwidth]{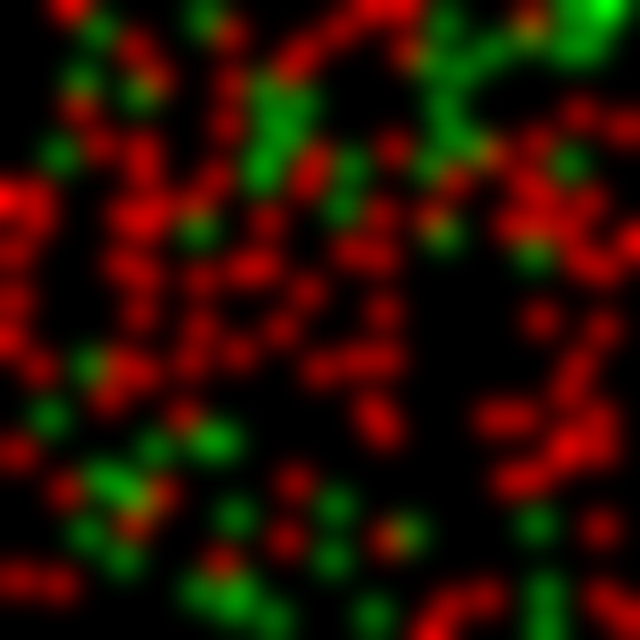}%
\includegraphics[align=c,width=\imgwidth]{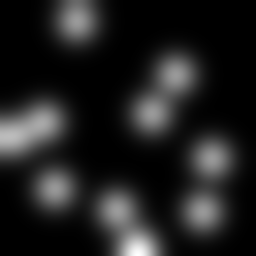}%
}\\%
\multicolumn{1}{c}{\hspace{-0ex}\rotatebox[origin=c]{90}{Segmentation Map}}%
&\multicolumn{6}{c}{%
\includegraphics[align=c,width=\imgwidth]{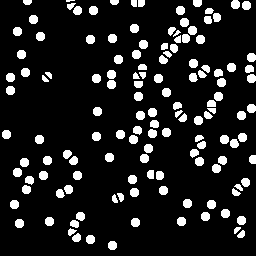}%
\includegraphics[align=c,width=\imgwidth]{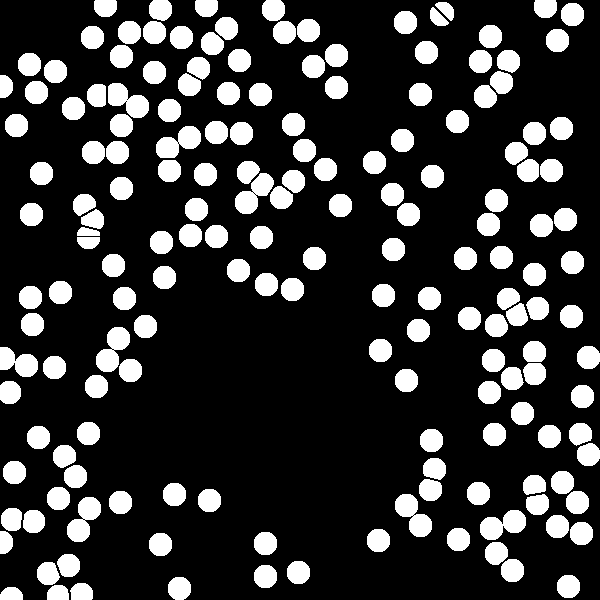}%
\includegraphics[align=c,width=\imgwidth]{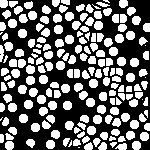}%
\includegraphics[align=c,width=\imgwidth]{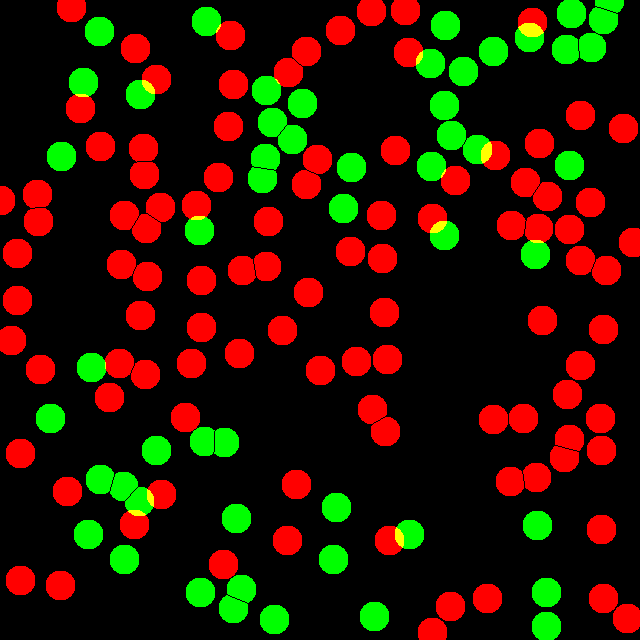}%
\includegraphics[align=c,width=\imgwidth]{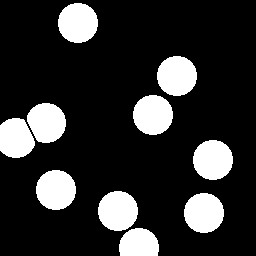}%
}\\%
\toprule
\end{tabularx}
\end{table}

\subsection{VGG Cells Dataset}
This public dot-annotated dataset was introduced by~\cite{learning_to_count}. It contains 200 RGB synthetic images simulating bacterial cells from fluorescence-light microscopy at various focal distances.
Images have a fixed size of 256×256×3 pixels, and the cells are designed to be clustered and occluded with each other.

\subsection{MBM Cells Dataset}
The \textit{\underline{M}odified \underline{B}one \underline{M}arrow (MBM)} dataset contains 44 RGB dot-annotated microscopy images of human bone marrow with various cell types stained blue. The original dataset was collected by~\cite{mbm_original}, acquiring 11 microscopy images from the human bone marrow tissues of 8 different patients. The original images are 1200×1200×3 pixels in size, but authors in~\cite{count-ception} split each of them into four images with the size of 600×600 pixels. 

{\subsection{ADI Cells Dataset}
The \textit{\underline{Adi}pocyte (ADI)} dataset is a human subcutaneous adipose tissue dot-annotated collection of microscopy images introduced by~\cite{count-ception}. It consists of 200 Regions Of Interest (ROI) of 150×150×3 pixels in size sampled from high-resolution histology slides representing adipocyte cells. The average cell count across all images is $165 \pm 44.2$, and the size of the biological structures can vary dramatically, representing a challenging test case for automated cell counting procedures.}

{\subsection{BCData}
The \textit{\underline{B}reast tumor \underline{C}ell \underline{Data}set (BCData)} \cite{Huang_2020} is a recent collection of 1,338 images based on Ki-67 staining with 181,074 dot-annotated cells divided into two classes (positive and negative tumor cells, i.e., malignant and not malignant, respectively). Unlike other datasets, BCData is not only large in scale concerning the labeled objects but also considering the number of different unique patient cases (that are 394). The size of each image is fixed to 640×640×3 pixels, and the authors divided the dataset into training, validation, and testing split at a ratio of approximately 6:1:3 (803, 133, and 402 images, respectively).}

\subsection{PNN Dataset}
\textit{\underline{P}eri\underline{n}euronal \underline{N}ets (PNNs)} are extracellular matrix aggregates surrounding the cell body of many neurons throughout the nervous system; their alterations are associated with several physiological processes and pathological conditions, e.g., psychiatric disorders such as schizophrenia~\cite{berretta2015losing,mauney2013developmental}. This contributed to the increasing interest in PNNs research spanning various conditions and animal models, including rodents~\cite{ueno2018expression, pnn_rodents_2, pnn_rodents_3}, primates~\cite{mueller2016distribution}, and even human brain samples~\cite{rogers2018normal}.
We collected and publicly released \cite{luca_ciampi_2021_5567032} a novel dataset of fluorescence microscopy images of mice brain slices containing annotations for perineuronal nets \footnote{The experiments were carried out in accordance with the directives of the European Community Council (2011/63/EU) and approved by the Italian Ministry of Health (Authorization \# 621/2020-PR).}.
Specifically, we obtained $50\mu m$ brain slices from C57BL6/J adult mice (transcardially perfused with $4\%$ paraformaldehyde). PNNs were stained with a green fluorescent marker by sequentially incubating them with biotinylated Wisteria floribunda Lectin (WFA) and streptavidin Alexa Fluor™ 488 conjugate. We acquired images with a fluorescence microscope (Zeiss Apotome.2). PNNs were manually annotated by neuroscientists and biologists from the laboratory of Prof. Pizzorusso, a leading expert in the field of the PNNs since 2002 \citep{Pizzorusso_2002}. 

\begin{figure}
\centering
\includegraphics[height=5.9cm]{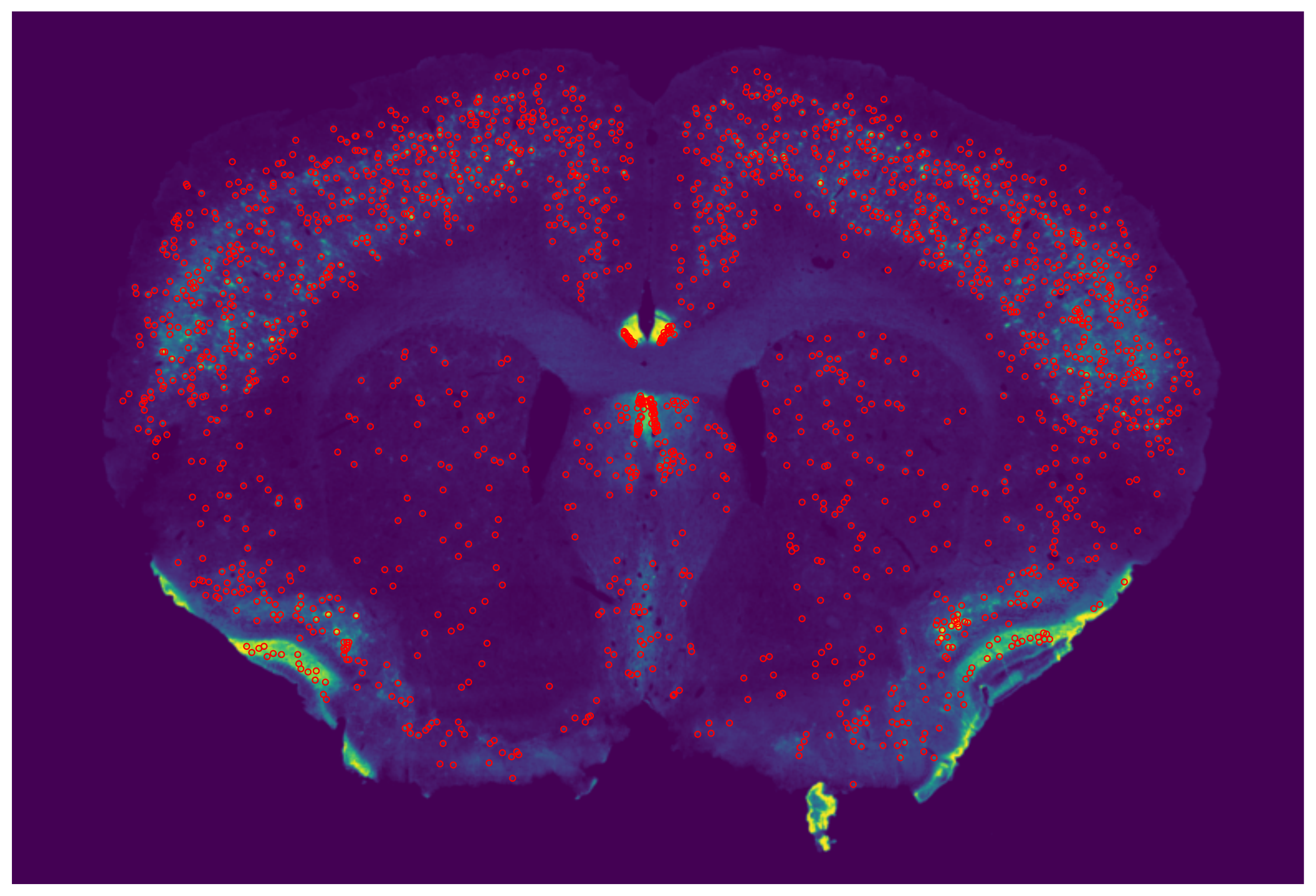}%
\hfill%
\includegraphics[height=5.9cm]{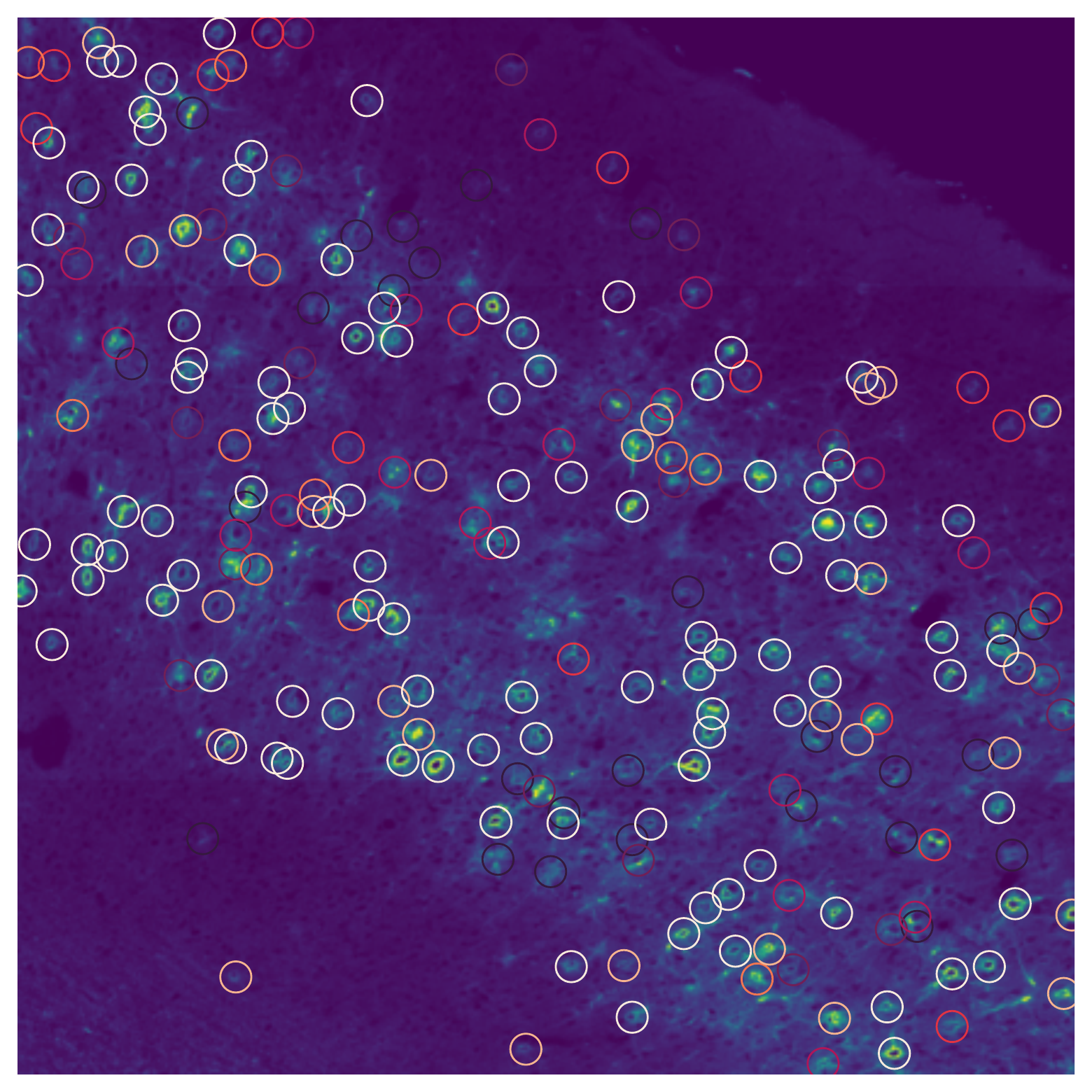}%
\\[.6cm]%
\includegraphics[height=5.5cm]{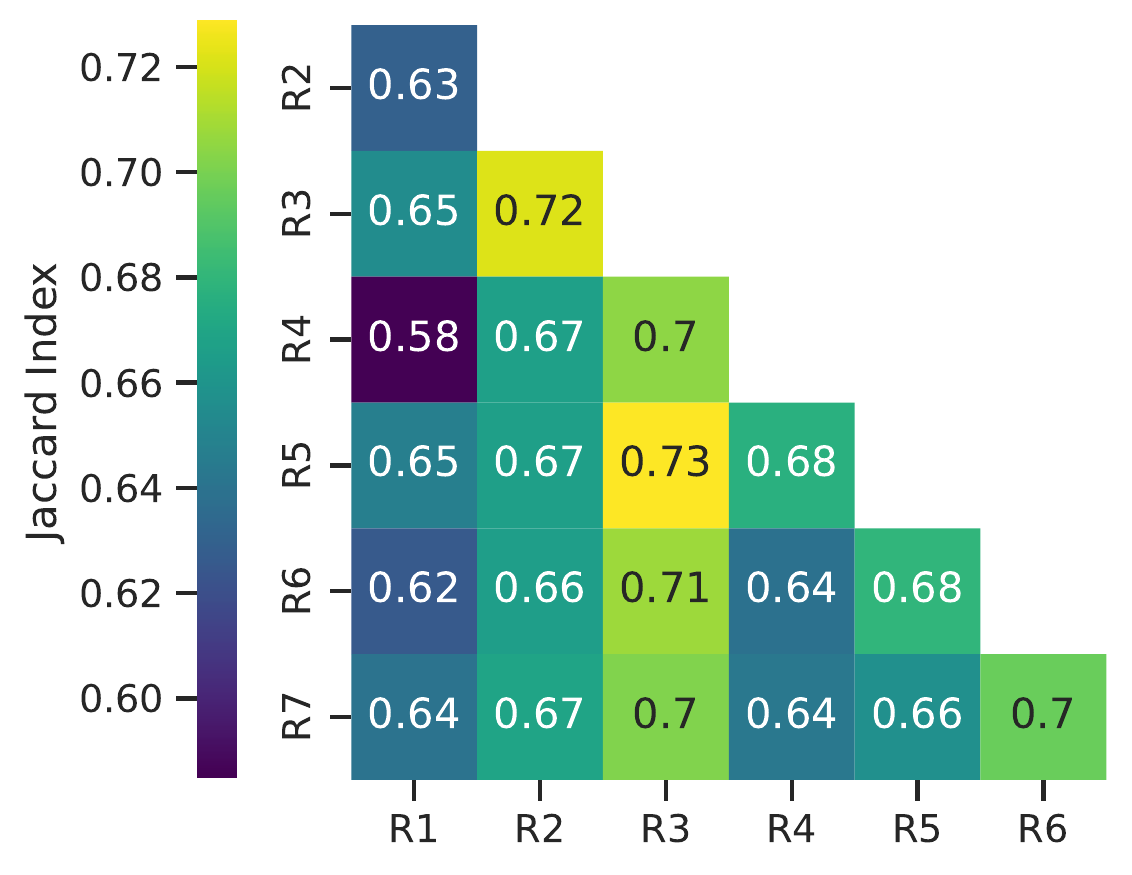}%
\hfill%
\includegraphics[height=5.5cm,trim={0 0 0.25cm 0},clip]{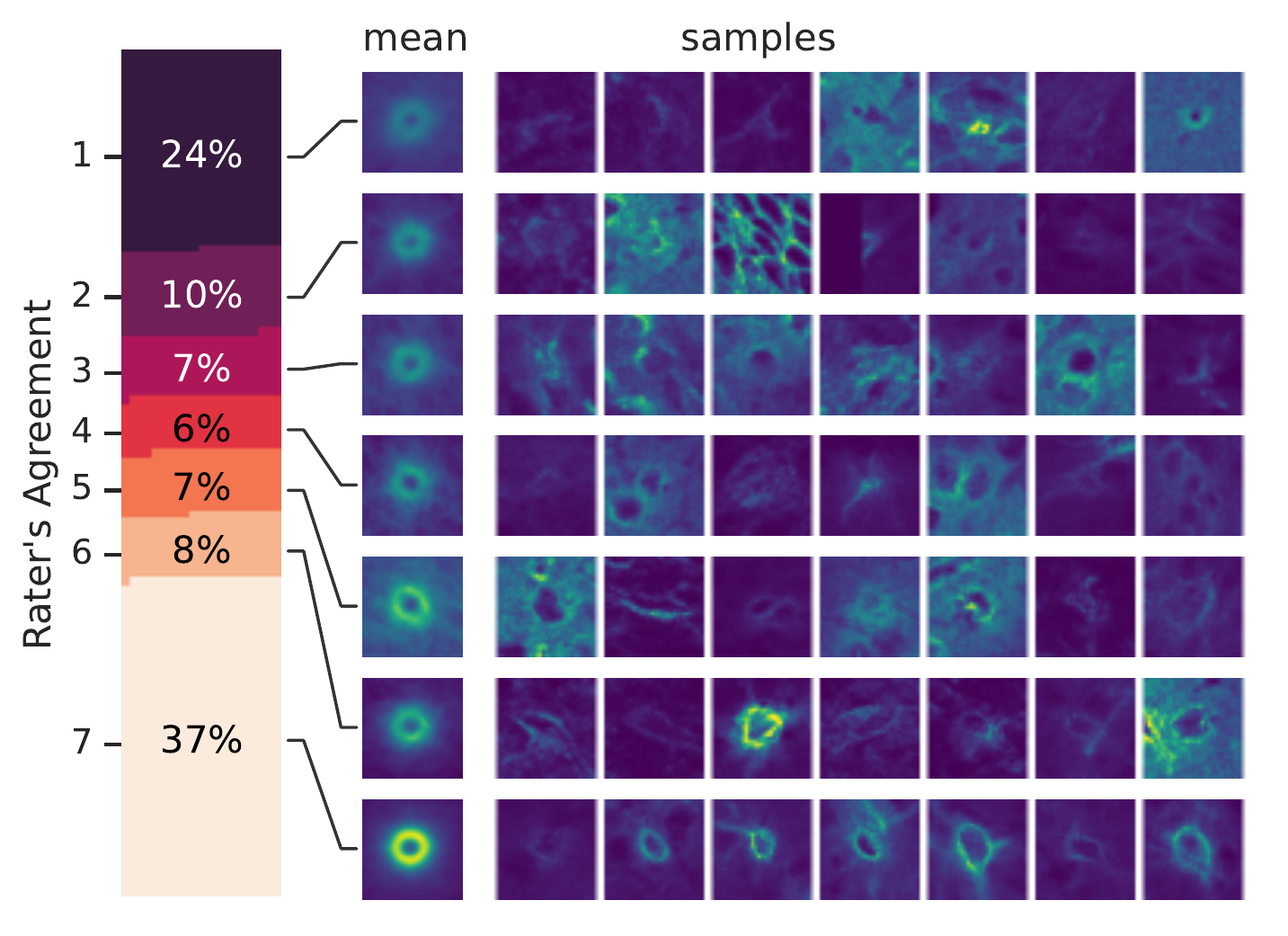}%
\caption{\textbf{PNN Dataset.}
\textit{Top Left}. A sample from the single-rater subset (PNN-SR) with dot annotations in red.
\textit{Top Right}. A sample from the multi-rater subset (PNN-MR, labeled by 7 raters);
the color of the circles encodes the number of raters' that identified that PNN following the legend on the below.
\textit{Bottom Left}. Jaccard Index between the PNN sets found by each rater.
\textit{Bottom Right}. Breakdown of the PNN-MR subset by raters' agreement level. We show some sample patches centered in the locations identifying PNNs, together with the mean patches; percentages represent the fraction of the total number of PNNs localized by $i$ raters, for $i = 1, \dots, 7$.}
\label{fig:pnn-dataset}
\end{figure}

The dataset is composed of two subsets --- a large \underline{s}ingle-\underline{r}ater subset (\textit{PNN-SR}) and a smaller \underline{m}ulti-\underline{r}ater subset (\textit{PNN-MR}) --- described below and depicted in \ref{fig:pnn-dataset}.

1) \textit{PNN-SR}: consists of 25 images having different sizes ranging from $8184 \times 6163$ to $15120 \times 9477$ pixels.
The extreme size of the images makes their use impracticable by AI-based Computer Vision tools unless dividing them into smaller regions.
Among all the images, there are roughly 34k annotated PNNs, varying from a few dozens to some thousand per image, depending on the considered portion of the brain.
An expert manually created annotations by putting a dot over the centroid of each identified PNN.
Since PNNs are often not easy to find and are subject to different judgments depending on the rater, the training labels are sure to contain errors. Thus, this subset can be considered \textit{weakly-annotated}.

2) \textit{PNN-MS}: comprises 12 microscopic images of $2000 \times 2000$ pixels representing different portions of a mouse brain, with a total of 2,532 dot-annotated PNNs.
The main peculiarity of this subset is that the annotation procedure has been performed by seven different raters, showing a remarkable discrepancy between the various judgments. 
As shown in \ref{fig:pnn-dataset} (bottom-right), more than $40\%$ of the PNN has not been annotated by the majority of raters (3 or less of the 7 raters), expressing the difficulty of achieving error-free assessments by a single rater. 

\section{Methodology}
\label{sec:counting-with-uncertainty:method}

Most counting approaches, both regression- or detection-based, can obtain good detections of the objects and a good prediction of the total count when using well-labeled training sets, as already demonstrated by cell counting literature.
However, under the presence of weak labels, these models tend to detect also low-confidence or spurious patterns with high confidence for multiple reasons;
for example, regression-based models such as density-map estimators do not model the confidence of a detected pattern, thus disabling any filtering step, and detection-based approaches often assign overestimated detection scores to maximize recall that does not correlate with the ``objectness'' of the pattern.
Although it is feasible to modify current models to better express this correlation, training them would necessitate large multi-rater datasets (where each pattern is labeled with a degree of objectness or quality) that are expensive to obtain.

\begin{figure}
\centering
\includegraphics[width=\linewidth]{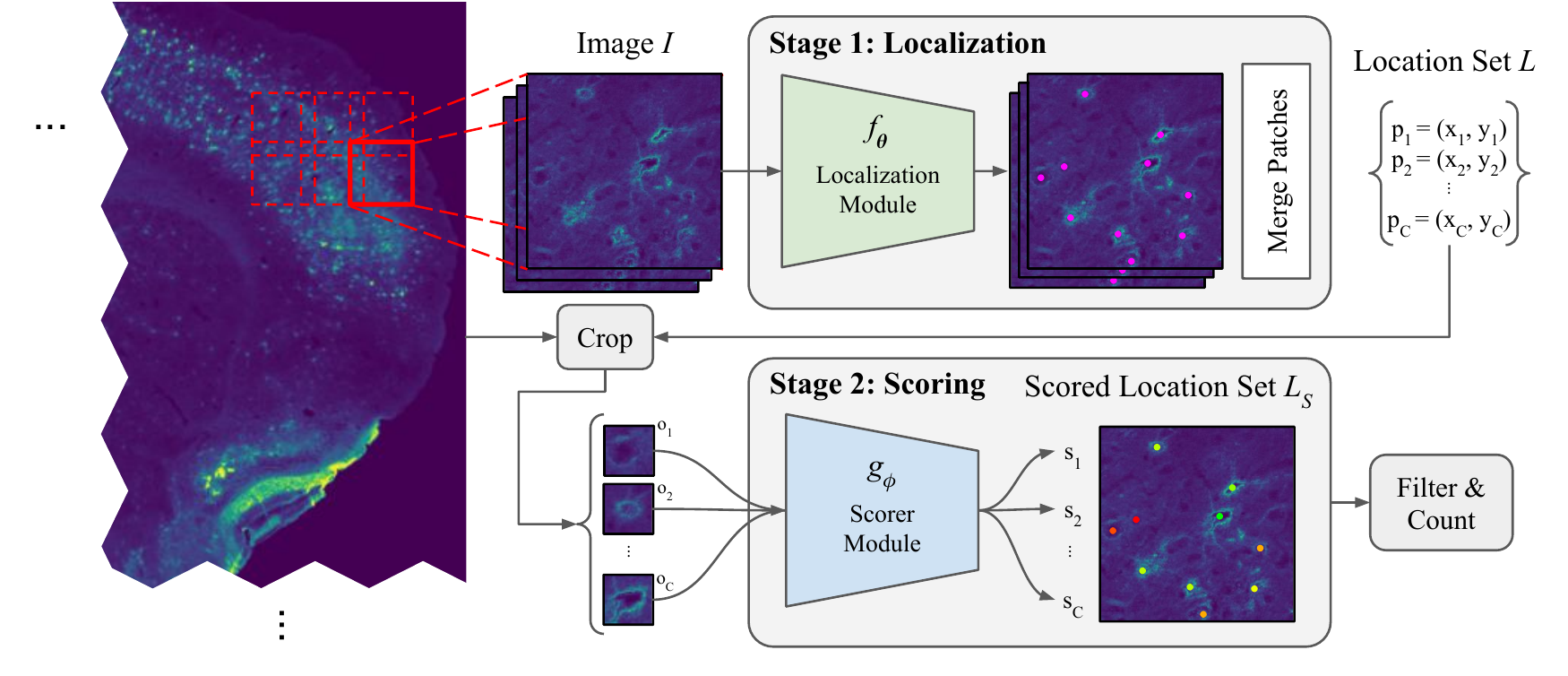}
\caption{\textbf{Proposed counting pipeline.} We model the task as a two-stage process. In the first one, we detect the objects exploiting a localization model $f_\theta$, previously trained on a large collection of dot-annotated images that may have weak labels. In the second stage, we employ a scorer model $g_\phi$ that assigns to the objects localized in the previous step an ``objectness'' score, which we correlate with the pattern uncertainty quantified by the agreement's level.}
\label{fig:overview}
\end{figure}

Here, we assume to have access to a large weakly-labeled single-rater dataset and only a small multi-rater subset.
With the former, we exploit the power of existing counting solutions, and with the latter, we devise an additional rescoring stage to cope with the problems discussed above.
Specifically, we model the counting task as a process (depicted in \ref{fig:overview}) comprised of two stages, each having its separate training phase.
The first stage follows standard approaches producing a set of coordinates localizing objects in the input image.
In the second stage, we consider the objects previously localized, and we assign them an ``objectness'' score that correlates with the raters' agreement on their detection, i.e., a higher score indicates a higher probability that most or all human raters detect that object.
To do so, we define a scorer module that inputs a small cropped patch containing the previously localized objects and outputs a scalar score.
We train it in a supervised fashion with a small set of multi-rater data, where the agreement between multiple raters reflects the pattern's certainty.
In practice, the output of the scorer model provides a new ``objectness'' score that practitioners can use to exclude or include samples from the total count.

We describe the two stages in more detail below.

\subsection{Localization Stage}
\label{sec:counting-with-uncertainty:method:localization-stage}
For this stage, we assume to have a collection of $N$ images with dot annotations $\mathcal{X} = \{(I_1, \hat{L_1}), \dots, (I_N, \hat{L_N})\}$, where $I_i$ is the $i$-th image and $\hat{L_i}$ 
is the set of coordinates of the structures to be counted in image $I_i$ labeled by a human rater.
We assume $\mathcal{X}$ is large and may have weak labels, e.g., it may contain spurious (false positives) and missing annotations (false negatives).

A localization model $f_\theta$ applied to the input image $I$ produces a set of coordinates
$L = \{p_1, \dots, p_C\ |\ p_i \in \mathbb{R}^2\}$ 
localizing the objects to be counted.
This model is trained using location data $\mathcal{X}$ and can be implemented following several different strategies;
here, we test three successful approaches from the literature, that are \textit{segmentation}, \textit{detection}, and \textit{density estimation}, described below.

\paragraph{Localization by Segmentation}
For this approach, we follow~\cite{falk2019u}, i.e., we first produce a segmentation map $S = f_\theta(I) \in [0,1]^{H \times W}$ for the input image $I$ having height $H$ and width $W$.
$S$ is then thresholded and further processed to extract connected components.
The centroids of those components form the output localizations $L$.
This solution can accommodate variable-shaped objects, but segmentation annotations are usually very expensive to produce.
Here, we generate the target segmentation maps $\hat{S} \in [0,1]^{H \times W}$ by imposing a disc centered in the dot-annotated position.
The radius of the disc is fixed and depends on the typical object size in the dataset.
A narrow ridge separates overlapping discs.
In case of multiple object classes, the network outputs one segmentation map per class, and target generation is performed independently for each class.
An example of a target segmentation map is reported in \ref{tab:datasets}.
The model is trained to minimize the weighted binary cross-entropy between pixels of the output and target maps;
more weight is assigned to more important pixels of the map, such as background ones near foreground objects.
More details of the generation procedure of segmentation targets are available in~\cite{falk2019u}.
As in~\cite{falk2019u}, we implement $f_\theta$ as a standard U-Net~\cite{unet}. In the following, we will refer to this method as \textit{S-UNet}.

\paragraph{Localization by Detection}
For this approach, we employ the Faster-RCNN model for visual object detection~\cite{faster_rcnn} (see \ref{sec:back:cnn-based-detectors:two-stages-detectors} for further details);
$f_\theta$ produces a list of bounding boxes following the standard two-stage detection paradigm.
In the first step, a Region Proposal Network (RPN) generates
the region proposals that might contain objects, slicing pre-defined region boxes (called anchors);
in the second step, these priors are refined, performing a regression to the coordinates of bounding boxes precisely localizing the objects inside these Regions of Interest (RoIs).
The centers of the boxes comprise the final localization of the entities.
Targets are produced by generating squared bounding boxes centered in the dot-annotated data with a fixed side, again, depending on the typical object size in the dataset.
We implement $f_\theta$ as a Faster-RCNN network with a Feature Pyramid Network module and a ResNet-50 backbone. From now on, we will refer to this method as \textit{FRCNN}.

\paragraph{Localization by Density Estimation}
We also tested density-estimation approaches known for delivering excellent counting performances, especially in ``crowded'' scenarios.
Using this approach, we learn a regression model producing a density map $D = f_\theta(I) \in \mathbb{R}^{H \times W}$ from an input image of height $H$ and width $W$.
Each pixel of $D$ corresponds to the \textit{quantity} of the objects present at that precise point.
Thus, the notion of density map loosely corresponds to the physical/mathematical notion of density;
the number of objects $n$ in an image sub-region $P \subseteq I$ is estimated by integrating $D$ over $P$, i.e., summing up pixel values in the considered region, $n = \sum_{p \in P} D_p$. We refer the reader to \ref{sec:back:visual-counting:traditional-approaches} for further details.
Although these approaches are not intended to localize objects, a coarse localization can be obtained by analyzing the estimated density map, in particular by finding the top-$n$ maximum local peaks of it, as already done in \cite{counting_cells_zisserman}.
During training, the target density maps are produced by superimposing Gaussian kernels $G_\sigma$ centered in the dot-annotated locations;
the spread parameter $\sigma$ is fixed and depends on the typical object size in the dataset.
In case of multiple object classes, the network outputs one density map per class, and target generation is performed independently for each class.
An example of a target density map is reported in \ref{tab:datasets}.
The model is trained by minimizing the mean squared error loss between target and output density maps.
We implement $f_\theta$ exploiting the Congested Scene Recognition Network (CSRNet)~\cite{csrnet}, a CNN for density estimation comprised of a modified VGG-16 network~\cite{vgg_16} for feature extraction and a series of dilated convolutional layers~\cite{dilated_1} to extract deeper information of saliency and, at the same time, maintaining the output resolution (see \ref{sec:back:visual-counting:cnn-based} for further details). We will refer to this method as \textit{D-CSRNet}.

Once objects are localized, the ``objectness'' of each prediction needs to be quantified to permit filtering of false positives or negatives that inevitably leak from labeling errors.
We cope with this task in the subsequent separate stage, but for comparison, we also consider deriving objectness scores from the three accounted localization models alone as a baseline.
Among the considered models, \textit{FRCNN} is the only one that natively outputs a score $\in [0 ,1]$ stating the probability of containing an object inside the bounding box that we can use as objectness score.
On the other hand, \textit{S-UNet} and \textit{D-CSRNet} do not provide directly a score that can be used for filtering predictions.
For \textit{S-UNet}, we derive a score $\in [0 ,1]$ associated with each localized object by considering the maximum values of the connected components found in the predicted segmentation map.
Regarding the \textit{D-CSRNet} model, we instead infer a score by taking the local maximum peak values of the predicted density map.
However, none of these scores are defined to correlate with the pattern uncertainty that instead needs to be explicitly modeled.
We do that in the second stage of our pipeline.

\subsection{Scoring Stage}
\label{sec:counting-with-uncertainty:method:score-calibration-stage}
The goal of this stage is to define a model that scores the certainty of a pattern;
higher scores should represent objects localized by most human raters, while lower scores should indicate dubious patterns.

Given the coordinates $p$ of an object in image $I$ localized with one of the approaches in the previous stage, we define a scorer model $g_\phi$ that assigns to the object a scalar objectness score $s = g_\phi(o)$, with $o$ the squared sub-patch of the image $I$ centered in $p$ containing the object.
To train $g_\phi$, we assume to have a small set of images where objects have been labeled by $K$ different raters; this produces a training set $\mathcal{X}' = \{(o_1, a_1), \dots, (o_M, a_M)\}$, where $o_i \in \mathbb{R}^{l \times l}$ is the image sub-region containing the $i$-th localized object, and $a_i \in \{0,\dots,K\}$ is the raters' agreement, i.e., the number of raters who localized that object.
Regions containing no localized objects ($a = 0$) are used as negative samples during training.

Although this rescoring stage is novel in counting pipelines, we can formulate it as well-known problems and implement it following existing solutions.
Below we propose several methodologies that can be adopted for training the $g_\phi$ model. It is worth noting that the $s$ score takes on different values depending on the adopted method.

\paragraph{Agreement Regression (AR)}
A simple baseline is directly regressing scores from the input patches.
In this formulation, $g_\phi$ produces a scalar output and is trained to directly regress the normalized raters' agreement $\sfrac{a}{K}$ from the object patch.
Specifically, we minimize
\begin{equation}
    \mathcal{L}(\mathcal{X}'; \phi) = \frac{1}{2}\sum_{(o, a) \in \mathcal{X}'} \left( \frac{a}{K} - g_\phi(o) \right)^2\,,
\end{equation}
where $o$ is a squared image patch containing a localized object and $\sfrac{a}{K}$ is the fraction of raters localizing that object.

\paragraph{Agreement Classification (AC)}
Another simple baseline comprises classifying the input patches in agreement levels.
In this formulation, we consider the $K+1$ agreement values $a \in \{0,\dots,K\}$ (including the $0$ value as background samples) as separate classes into which objects can be classified.
The model $g_\phi$ produces a $(K+1)$-way softmax output that is trained with standard cross-entropy loss
\begin{equation}
    \mathcal{L}(\mathcal{X}'; \phi) = - \sum_{(o, a) \in \mathcal{X}'} \log(g^a_\phi(o))\,,
\end{equation}
where $g^i_\phi(o)$ indicates the $i$-th output of the model.
The final scalar score $s$ is obtained 
as the (normalized) expected value of the class over the output categorical distribution
\begin{equation}
    s(o) = \frac{1}{K} \sum_{i=0}^K i \cdot g^i_\phi(o)\,.
\end{equation}

\paragraph{Agreement Ordinal Regression (OR)}
We formulate the scoring problem as an ordinal regression problem with $K+1$ ordered categories from the lowest to the highest agreement.
Similarly to agreement regression, $g_\phi$ produces a scalar output but is trained following~\cite{pedregosa2017consistency}.
Along with model parameters, a set of $K$ ordered thresholds $\Theta = \{\theta_i\}_{\scriptscriptstyle i=0}^{\scriptscriptstyle K-1}, \theta_0 < \theta_1 < \dots < \theta_{K-1}$ are defined as learnable parameters.
Given the model scalar output $s = g_\phi(o)$, we model
\begin{equation}
P(a \leq k | o) = \sigma(\theta_k - s)\quad  k = 0, \dots, K-1\,,
\end{equation}
where $\sigma$ is the sigmoid function, and thus
\begin{equation}
\begin{aligned}
y_k(o) &= P(a = k | o) \\
    &= P(a \leq k | o) - P(a \leq k-1 | o) \\
    &= 
    \begin{cases}
    \sigma(\theta_0 - s) & \text{if } k=0\,, \\
    \sigma(\theta_k - s) - \sigma(\theta_{k-1} - s) & \text{if } k=1 \dots K-1\,\text{, and} \\
    1 - \sigma(\theta_{K-1} - s)  & \text{if } k=K\,.
    \end{cases}
\end{aligned}
\end{equation}
The models parameters $\phi$ and thresholds $\Theta$ are optimized by minimizing the negative log likelihood of observed samples
\begin{equation}
    \mathcal{L}(\mathcal{X}'; \phi, \Theta) = - \sum_{(o, a) \in \mathcal{X}'} \log(y_a(o)) \,.
\end{equation}
The values of $\theta_i$ are optionally clipped after each update to kept them ordered.
Once trained, we discard $\Theta$ and adopt only $g_\phi$ to output the score $s$ for an object.

\paragraph{Agreement Rank Learning (RL)}
Here, we model agreement by learning to rank a tuple of samples with increasing agreement values.
Our formulation instantiates a standard pairwise learning to rank approach~\cite{burges2005learning} with a custom sample loss definition.
Specifically, we still define a model with a scalar output $s = g_\phi(o)$, but we employ a different training scheme;
given a $(K+1)$-tuple of ordered samples $O = (o_0, \dots, o_K)$ containing one sample per agreement class (i.e., $o_i$ has an agreement value $a_i = i$), we ask our model to produce scores  
$s_i = g_\phi(o_i)$ that are sorted $s_0 < s_1 < \dots < s_K$.
Translating this constraint in a loss function for the single tuple, we obtain a class-balanced pairwise margin loss
\begin{equation}
    \mathcal{L}(o_0, \dots, o_K; \phi) = \frac{1}{K} \sum_{i=1}^{K} \max( m - g_\phi(o_i) + g_\phi(o_{i-1}), 0 )\,,
\end{equation}
where $m$ is a margin hyper-parameter empirically set to $0.1$.
A dataset of tuples is obtained by repeatedly drawing $K+1$ random samples, one for each agreement class, from the training set $\mathcal{X}'$. This has the advantage to produce large training datasets even when dealing with a small initial multi-rater dataset.
The batch loss is obtained as the mean loss over a batch of tuples.

\section{Experiments and Results}
\label{sec:counting-with-uncertainty:experiments}

In this section, we describe the experiments performed to validate our approach and discuss the obtained results. We divided them into three parts. First, we evaluated the considered counting architectures, i.e., the segmentation-based \textit{S-UNet}, the detection-based \textit{FRCNN}, and the density-based \textit{D-CSRNet} approaches, against standard single-rater cell counting benchmarks. The aim was to demonstrate that they work plausibly fine, i.e., they produced comparable results against the state-of-the-art, warding off that results provided by our counting pipeline are not due to a weak poorly-trained baseline.
Then, we evaluated the first stage of our pipeline, i.e., the localization stage. Specifically, we performed experiments on our novel PNN dataset, training the three adopted counting architectures with single-rater data having significant label noise from errors introduced by raters. The goal was to detect and count perineuronal nets under this weakly labeled setting, deriving \textit{uncalibrated} scores from the models' output that have not been designed to correlate with the quality of the predictions. 
Finally, we performed experiments with our multi-rater PNN-MR subset to validate our proposed second stage, i.e., the score calibration stage. Here, we refined predictions of the previous stage, producing \textit{calibrated} scores that increase the correlation with the raters' agreement.
We compared it to several baselines, showing that it improves counting performances when dealing with uncertain patterns.

\subsection{Evaluation of the adopted counting architectures}
\label{sec:counting-with-uncertainty:experiments:state-of-the-art}
We evaluated the three adopted counting approaches against the state of the art using VGG Cells, MBM Cells, ADI Cells, and BCData counting benchmarks described in \ref{sec:counting-with-uncertainty:datasets}. All these collections of images are single-rater, i.e., the final available labels belong to a single rater. 
Even when multiple raters have been employed during the annotation procedure, the final annotations are squashed into a single label per object. In other words, multi-rater annotations are not leveraged if not for the creation of stronger annotations at the expense of the dataset scale.

We followed the evaluation protocol introduced by \cite{learning_to_count} and adopted by most subsequent works; we considered a testing subset fixed for all the experiments (100 images for VGG Cells and ADI Cells, and 10 images for MBM Cells) and training and validation subsets of varying size ($N$ images for each subset) to simulate lower or higher numbers of labeled examples.
Following previous work, we set $N$ to 16, 32, and 50 for VGG Cells, to 10, 25, and 50 for ADI Cells, and to 5, 10, 15 for MBM Cells. Concerning BCData, we instead used the training, validation and testing splits provided by \cite{Huang_2020}.
As performance metrics, we adopted the widely used \acrfull{mae} already defined in \ref{sec:back:visual-counting:metrics}.
For VGG Cells, MBM Cells, and ADI Cells, we repeated the experiment 10 times, randomly sampling different splits for each configuration, and we reported the mean and standard deviation of the evaluation metric. To check the consistency of the results on these random splits, we also re-implemented the original FCRN-A method presented in \cite{counting_cells_zisserman}, thus performing an exact head-to-head comparison with the same samples being used for training and testing. 
Concerning BCData, we reported the mean and the standard deviation of the MAE calculated between 10 runs over the 402 images comprising the test split, changing the random initialization seed each time.

\ref{tab:sota} reports results on the four datasets.
The density-based solution \textit{D-CSRNet} performed best among tested solutions on the VGG dataset and comparably to state of the art. 
The other two adopted methods, i.e., the segmentation-based \textit{S-UNet} and the detection-based \textit{FRCNN}, exhibited slightly larger errors, consistently with their inherent limitations when applied to ``crowded'' scenarios with occluded objects like VGG Cells. On the same grounds, \textit{D-CSRNet} achieved best performances on the BCData dataset, and the detection-based \textit{FRCNN} approach is the one that faced more difficulties.
On the other hand, in the MBM and the ADI datasets, where the challenges are more related to the object shape variations, all the approaches showed competitive results, outperforming state-of-the-art solutions in some cases (e.g., \textit{S-UNet} in MBM and \textit{FRCNN} in ADI). Overall, the tested approaches performed in line with state-of-the-art, and thus we proceeded to adopt them in the first localization stage of our pipeline.

\begin{table}
\caption{\textbf{Comparison of the adopted architectures on standard single-rater counting benchmarks.}
For VGG, MBM and ADI we varied the training and validation subsets ($N$ images for each subset), repeating the experiments 10 times. For BCData we used the splits provided by \cite{Huang_2020}, performing 10 runs changing the seed each time. Mean$\pm$st.dev. of MAE is reported.}
\label{tab:sota}
\footnotesize
\newcolumntype{C}{>{\centering\arraybackslash}X}
\setlength{\tabcolsep}{2pt}
\begin{subtable}{\linewidth}
\caption{\textbf{VGG Cells} (200 images in total - 100 test images)}
\label{tab:sota:vgg}
    \begin{tabularx}{\linewidth}{l@{}CCC}
        \toprule
        Method & N = 16 & N = 32 & N = 50 \\
        \midrule
        \citet{Arteta_2016} & N/A & 5.06 $\pm$ 0.2 & N/A \\
        GMN \cite{class_agnostic} & N/A & 3.6 $\pm$ 0.3 & N/A \\
        \citet{learning_to_count} & 3.8 $\pm$ 0.2 & 3.5 $\pm$ 0.2 & N/A \\
        VGG-GAP-HR \cite{improving_with_heatmap} $^*$ & N/A & 2.95$^{**}$ & 2.67 \\
        SAU-Net \cite{saunet} $^\dagger$ & N/A & N/A & 2.6 $\pm$ 0.4 \\
        FCRN-A \cite{counting_cells_zisserman} & 3.4 $\pm$ 0.2 & 2.9 $\pm$ 0.2 & 2.9 $\pm$ 0.2$^\ddagger$ \\
        FCRN-A \cite{counting_cells_zisserman} $^\mathsection$ & 4.7 $\pm$ 0.7 & 3.3 $\pm$ 0.2 & 3.1 $\pm$ 0.1 \\    
        Count-Ception \cite{count-ception} & 2.9 $\pm$ 0.5 & 2.4 $\pm$ 0.4 & 2.3 $\pm$ 0.4\\
        CCF \cite{jiang2020cell} & 2.8 $\pm$ 0.1 & 2.6 $\pm$ 0.1 & 2.6 $\pm$ 0.1 \\
        C-FCRN+Aux \cite{he2021deeply} $^\$$ & \multicolumn{3}{c}{2.3 $\pm$ 2.2} \\
        \midrule
        S-UNet \cite{falk2019u} \textit{(our)} &  7.7 $\pm$ 2.0 &  6.8 $\pm$ 1.0 &  4.5 $\pm$ 1.0 \\ 
        D-CSRNet \cite{csrnet} \textit{(our)} & 3.7 $\pm$ 0.3 & 2.9 $\pm$ 0.3 & 2.5 $\pm$ 0.1 \\
        FRCNN \cite{faster_rcnn} \textit{(our)} & 9.3 $\pm$ 0.7 & 7.5 $\pm$ 0.6 & 7.0 $\pm$ 0.4 \\ 
        \midrule
    \end{tabularx}
    *  They did not report standard deviation. 
    ** They used a validation subet of $100 - N$ images. 
    $\dagger$ They did not use a test subset, but only a $100 - N$ images validation subset. 
    $\ddagger$ Reported in their work as N = 64. 
    $\$$ They used a 5-fold cross validation-based evaluation protocol considering the whole dataset. 
    $\mathsection$ Re-implemented in this work
\end{subtable}\\[6ex]
\begin{subtable}{\linewidth}
\caption{\textbf{MBM Cells.} (44 images in total - 10 test images)}
\label{tab:sota:mbm}
    \begin{tabularx}{\linewidth}{l@{\hspace{3pt}}CCC}
        \toprule
        Method & N = 5 & N = 10 & N = 15 \\
        \midrule
        \citet{xie2018efficient} $^\ddagger$ & \multicolumn{3}{c}{36.3 $\pm$ 19.4}\\
        FCRN-A \cite{counting_cells_zisserman} $^\$$ & 28.9 $\pm$ 22.6 & 22.2 $\pm$ 11.6 & 21.3 $\pm$ 9.4 \\
        FCRN-A \cite{counting_cells_zisserman} $^\mathsection$ & 15.6 $\pm$ 4.3 & 12.4 $\pm$ 4.0 & 12.2 $\pm$ 2.9 \\    
        \citet{DBLP:conf/cvpr/MarsdenMLKO18} $^*$ & 23.6 $\pm$ 4.6  & 21.5 $\pm$ 4.2 & 20.5 $\pm$ 3.5 \\
        Count-Ception \cite{count-ception} & 12.6 $\pm$ 3.0 & 10.7 $\pm$ 2.5 & 8.8 $\pm$ 2.3 \\
        CCF \cite{jiang2020cell} $^*$ & 9.3 $\pm$ 1.4 & 8.9 $\pm$ 0.9 & 8.6 $\pm$ 0.3 \\
        C-FCRN+Aux \cite{he2021deeply} $^{**}$ & \multicolumn{3}{c}{6.5 $\pm$ 5.2} \\
        SAU-Net \cite{saunet} $^\dagger$ & N/A & N/A & 5.7 $\pm$ 1.2 \\
        Jiang and Yu \cite{two_path} & 8.2 $\pm$ 1.1 & 6.9 $\pm$ 0.9 & 6.0 $\pm$ 0.6 \\
        Jiang and Yu \cite{foreground_mask} & - & - & 6.0 $\pm$ 0.2 \\
        \midrule
        S-UNet \cite{falk2019u} \textit{(our)} & 5.5 $\pm$ 1.9 & 5.9 $\pm$ 4.2 & 5.7 $\pm$ 0.9 \\ 
        D-CSRNet \cite{csrnet} \textit{(our)} & 9.4 $\pm$ 3.4 & 7.2 $\pm$ 2.0 & 6.4 $\pm$ 1.4 \\ 
        FRCNN \cite{faster_rcnn} \textit{(our)} & 9.3 $\pm$ 1.4 & 9.0 $\pm$ 1.4 & 8.6 $\pm$ 0.8 \\ 
        \midrule
    \end{tabularx}
    * They used 14 test images. 
    ** They used a 5-fold cross validation-based evaluation protocol considering the whole dataset. 
    $\dagger$ They did not use a test subset, but only a $44 - N$ images validation subset.
    $\ddagger$ They used a train/test split of 8/3 using full-size images. 
    $^\$$ Implemented by \citep{count-ception}.
    $\mathsection$ Re-implemented in this work.
\end{subtable}\\[6ex]
\hfill
\end{table}

\begin{table}\ContinuedFloat
\footnotesize
\newcolumntype{C}{>{\centering\arraybackslash}X}
\setlength{\tabcolsep}{2pt}
\begin{subtable}{\linewidth}
\caption{\textbf{ADI Cells.} (200 images in total - 100 test images)}
\label{tab:sota:adi}
    \begin{tabularx}{\linewidth}{lCCC}
        \toprule
        Method & N = 10 & N = 25 & N = 50 \\
        \midrule
        FCRN-A \cite{counting_cells_zisserman} $^\mathsection$ & 21.1 $\pm$ 4.7 &  13.1 $\pm$ 0.7 &  11.3 $\pm$ 1.1 \\    
        Count-Ception \cite{count-ception} & 25.1 $\pm$ 2.9 & 21.9 $\pm$ 2.8 & 19.4 $\pm$ 2.2 \\
        CCF \cite{jiang2020cell} & 16.9 $\pm$ 1.9 & 14.5 $\pm$ 0.4 & 14.5 $\pm$ 0.4 \\
        SAU-Net \cite{saunet} $^\dagger$ & N/A & N/A & 14.2 $\pm$ 1.6 \\
        Jiang and Yu \cite{two_path} & 13.8 $\pm$ 0.7 & 11.6 $\pm$ 0.4 & 10.6 $\pm$ 0.3 \\
        Jiang and Yu \cite{foreground_mask} & - & - & 10.1 $\pm$ 0.1 \\
        \midrule
        S-UNet \cite{falk2019u} \textit{(our)} & 16.6 $\pm$ 5.5 & 13.6 $\pm$ 1.8 & 13.7 $\pm$ 4.9 \\
        D-CSRNet \cite{csrnet} \textit{(our)} & 12.6 $\pm$ 1.3 & 10.8 $\pm$ 1.5 &   8.8 $\pm$ 1.0 \\
        FRCNN \cite{faster_rcnn} \textit{(our)} & 10.0 $\pm$ 0.9 & 9.1 $\pm$ 0.7 & 8.7 $\pm$ 0.8 \\
        \midrule
    \end{tabularx}
    $\dagger$ They did not use a test subset, but only a $200 - N$ images validation subset. $\mathsection$ Re-implemented in this work.
\end{subtable}\\[6ex]
\begin{subtable}{\linewidth}
\caption{\textbf{BCData} (1,338 images in total - 803 train, 133 val, 402 test); positive and negative cells are malignant and not malignant tumor cells, respectively.}
\label{tab:sota:bcd}
    \begin{tabularx}{\linewidth}{lCCC}
        \toprule
        Method & Positive & Negative & All  \\
        \midrule
        \citet{Sirinukunwattana_2016} $^*$ $^\dagger$ & 9.1 & 20.6 & 14.8 \\
        CSRNet \cite{csrnet} (integration) $^*$ $^\$$ & 9.2 & 24.8 & 14.8 \\
        U-CSRNet \cite{Huang_2020} (integration) $^*$ & 10.0 & 18.0 & 14.0 \\
        CSRNet \cite{csrnet} (detection) $^*$ $^\$$ & 7.7 & 14.1 & 10.9 \\
        U-CSRNet \cite{Huang_2020} (detection) $^*$ & 6.8 & 14.1 & 10.5 \\
        \midrule
        S-UNet \cite{falk2019u} \textit{(our)} & 8.3 $\pm$ 0.5 & 19.7 $\pm$ 0.9 & 14.4 $\pm$ 0.7 \\ 
        D-CSRNet \cite{csrnet} \textit{(our)} & 8.3 $\pm$ 0.9 & 16.6 $\pm$ 1.4 & 12.5 $\pm$ 0.9 \\ 
        FRCNN \cite{faster_rcnn} \textit{(our)} & 10.3 $\pm$ 0.4 & 30.9 $\pm$ 2.3 & 20.6 $\pm$ 1.3 \\ 
        \midrule
    \end{tabularx}
    \footnotesize  
    * They did not report standard deviation. 
    $^\$$ Implemented by \citep{Huang_2020}, they used ResNet-50 \citep{resnet} instead of VGG-16 \citep{vgg_16} for feature extraction. 
    $\dagger$ Implemented by \citep{Huang_2020}.
\end{subtable}
\end{table}

\subsection{Localization Stage Evaluation}
\label{sec:counting-with-uncertainty:experiments:exp-pnn}
For these experiments, we applied the three previously evaluated solutions to our novel PNN dataset, and we investigated their counting ability in the presence of weakly-labeled data, i.e., under significant label noise introduced by errors of raters.

We considered the large single-rater subset PNN-SR to train the models, whereas we evaluated them on the multi-rater subset PNN-MR.
Since the training set contains weak labels, for each solution, we derived a scalar score $s$ from the models' output that can be used to filter low-quality predictions.
We refer to the scores obtained in this stage as \textit{uncalibrated} scores, since they have not been designed to correlate with the quality of predictions (we did this in the subsequent scoring stage).
For \textit{S-UNet} we set $s$ as the maximum value of the connected component found in the segmentation map. For \textit{FRCNN}, we set $s$ as the classification score that the network already outputs together with the regressed bounding box coordinates localizing the object.
Finally, for \textit{D-CSRNet}, we considered as $s$ the value of the higher local peak in the density maps localizing the object.

During the training phase, we split the data into training and validation parts.
We did not adopt the common per-image split strategy, as the number of PNNs vastly varies depending on the particular considered brain slice (i.e., image), and thus this strategy would have produced unbalanced splits.
Instead, we split each image vertically in half, including one half in the training set and the other in the validation set in an alternate fashion.

Due to the extreme size of the images, we processed them in patches.
During the training phase, we cropped squared randomly localized patches from training images.
We experimented with different patch sizes of 256, 320, 480, 640, and 800 pixels.
At validation time, we divided and processed the image in regularly-spaced overlapped patches of the same size used during training (see also \ref{fig:overview}), we reconstructed the global output by combining patch predictions, and we computed metrics at the entire image level. For segmentation-based and density-based solutions, image-level maps were obtained by stitching back together the patch-level maps and taking the mean pixel values in the overlap areas. For the detection-based solution, we performed non-maximum suppression of all the bounding boxes predicted in the overlap areas.

\begin{figure}
\centering
\includegraphics[width=\linewidth]{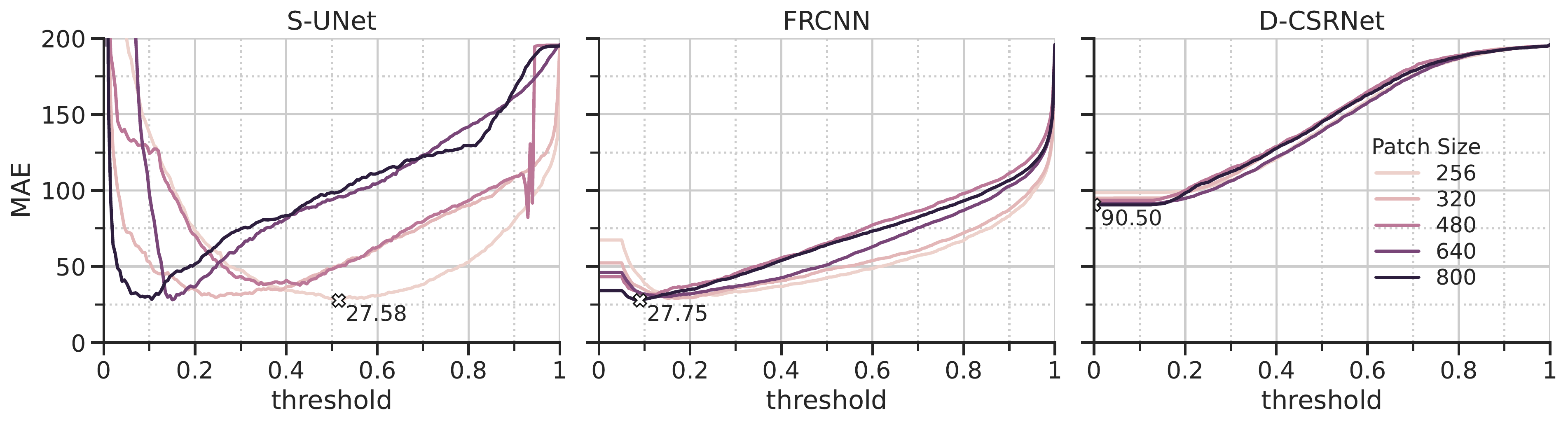}\\
\includegraphics[width=\linewidth]{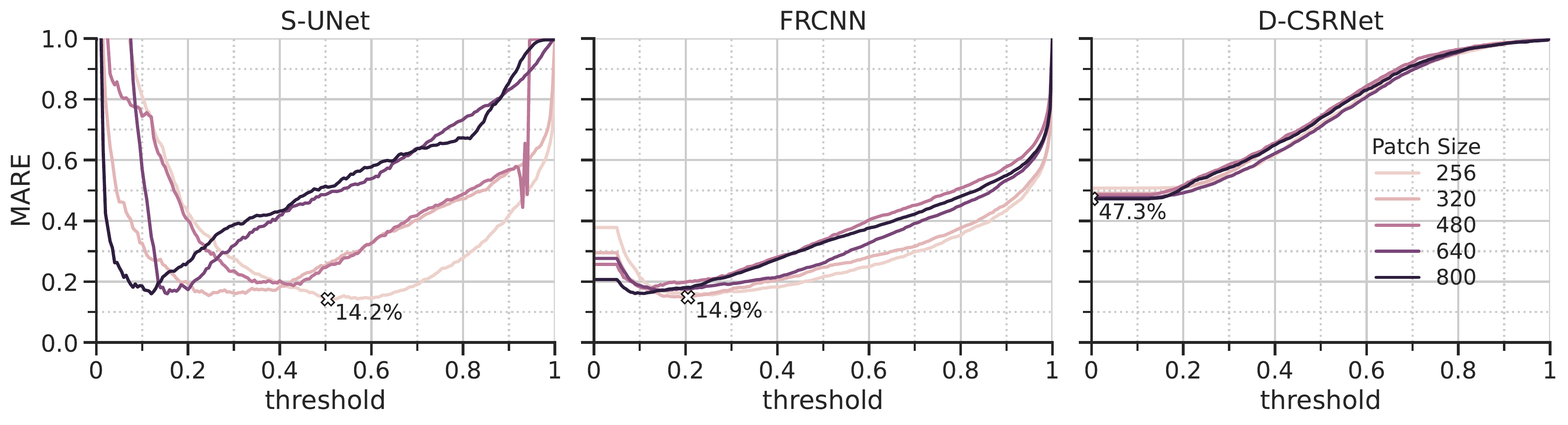}\\
\includegraphics[width=\linewidth]{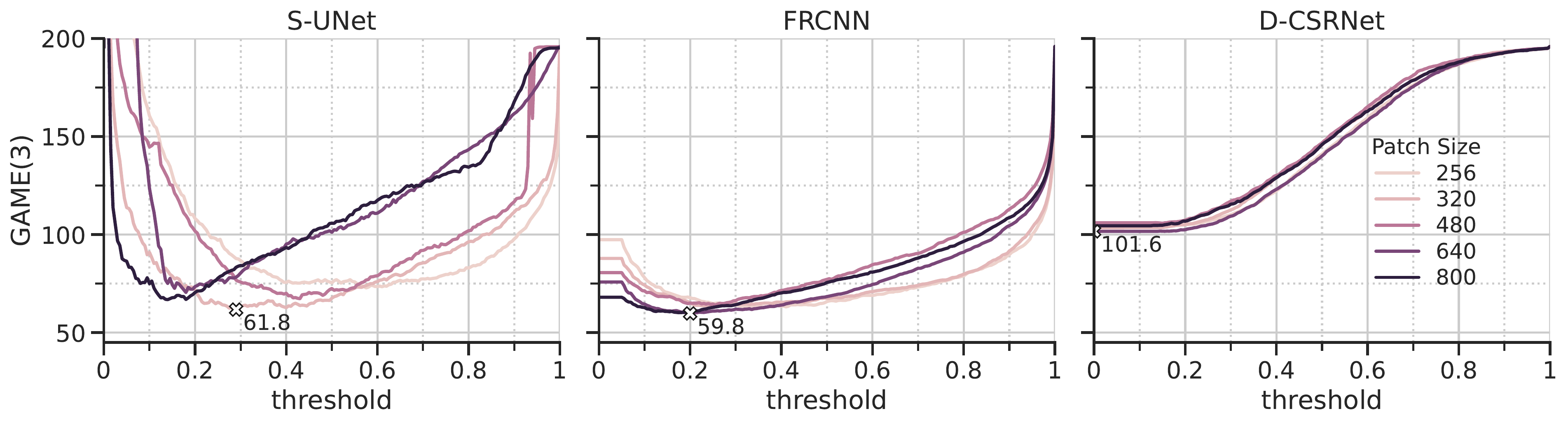}\\
\includegraphics[width=\linewidth]{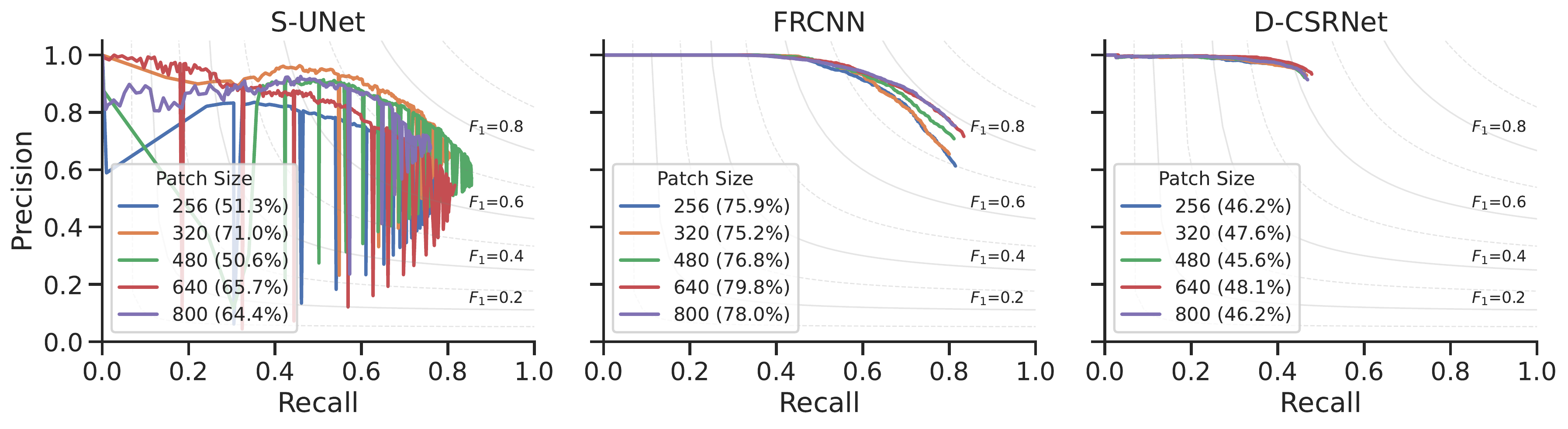}\\
\caption{\textbf{PNN-MR: Impact of patch size and threshold to counting and localization performance}
on standard segmentation-based (S-UNet), detection-based (FRCNN), and density-based (D-CSRNet) approaches (without score calibration).}
\label{fig:pnn-patch-size}
\end{figure}

In \ref{fig:pnn-patch-size}, we show the results obtained by the three solutions (one per column) on the whole multi-rater PNN-MR subset when varying the patch size and the threshold on the scalar score $s$. To compare the counting errors we used, in addition to the \acrshort{mae}, the \acrfull{mare} that provides a percentage relating the absolute counting error to the number of objects to be counted, and the \acrfull{game} \cite{ExtremelyTrancos} that accounts for localization errors, as it is computed by sub-dividing the image in $4^L$ non-overlapping regions and computing the \acrshort{mae} in each of these sub-regions (see also \ref{sec:back:visual-counting:metrics} for further details).
As depicted in the first three rows of \ref{fig:pnn-patch-size}, patch size did not significantly influence the performance of \textit{FRCNN} and \textit{D-CSRNet}. Thus, for these models, we suggest opting for bigger patch sizes that reduce processing overhead. 
On the other hand, the \textit{S-UNet} solution was more sensitive to this aspect;
due to artifacts in the overlap regions of the segmentation map, different patch sizes induced different score distributions that responded differently to score thresholding.
For \textit{S-UNet}, the best performance is obtained with smaller patch sizes together with more conservative threshold values.
Note that the density-based solution \textit{D-CSRNet} was the most strained by weakly-labeled training data, achieving the worst overall performance primarily due to a low recall.
Moreover, as expected, score thresholding was not effective since the density peak value employed as the score was not expected to locate the PNN precisely and correlate with its ``visual quality''.
Even when counting was performed via density map integration, instead of peak localization and counting, the best counting performance that \textit{D-CSRNet} achieves is an MAE of $90.99$.
Finally, in the last row of \ref{fig:pnn-patch-size}, we show Precision-Recall (PR) curves, with the goal of better highlighting the influence of the considered threshold.
Precision was computed as the percentage of model predictions that corresponded to a ground-truth object, whereas Recall was the percentage of ground-truth objects identified by the model.
Predictions and ground-truth objects were matched using the Hungarian algorithm based on the 2D Euclidean distance between pixel coordinates; we assigned an infinite cost to pairs separated by a distance greater than 1.25 times the typical radius of objects.
As depicted in the figure, the PR curves concerning the \textit{FRCNN} and the \textit{D-CSRNet} are monotonic, while the \textit{S-UNet} does not have this property, again due to artifacts in the segmentation maps and on merging/separating components.

So far, we have experimented on the entire PNN-MR subset, thus including every PNN found by at least one rater in the ground-truth set.
Next, we illustrate how the trained models behave when asked to localize only PNNs on which at least $a$ raters agree on their presence.
Specifically, we defined four sets of ground truth labels for PNN-MR comprising PNNs labeled by at least 1, 4, 5, and 7 out of 7 raters, respectively, simulating different counting policies, from more liberal to more conservative ones;
the choice of 4 and 5 raters reflects the rater's agreement above $50\%$ and $70\%$, respectively, which are two thresholds widely adopted to legitimize labels.
In \ref{tab:pnn-agree-mae}, we report the results in terms of MAE, MARE, and GAME(3), obtained on these four sets by the three tested models in their most effective combination of patch size and threshold values.
We observe that models tended to correctly identify and count the PNNs found by more raters, as these are also the clearer and easier-to-spot samples.
Although all the models delivered a similar performance at the higher agreement levels, at the lowest agreement level ($a \geq 1$), the \textit{D-CSRNet} tended to have a higher error due to its inability to achieve a high recall on low-agreement samples.
We observe that \textit{FRCNN} tended to achieve lower errors when increasing the rater's agreement $a$ with respect to the other tested models.
We also note that all models showed high variability in MAE values computed on different test set images.
We deem this was due to particular brain regions where PNNs appeared dimmer and thus more difficult for both models and human raters to cope with (see \ref{fig:pnn-prediction-examples}).

\begin{table}
\caption{\textbf{PNN-MR: Performance on different agreement levels.}
We report the mean $\pm$ st.dev. of MAE, MARE, and GAME(3) considering four sets of ground-truth labels for the whole PNN-MR dataset composed by objects labeled by any number of raters (Any), at least 50\%, at least 70\%, or all raters, respectively, to simulate different counting policies, from liberal to conservative ones. Models are trained once on the weakly-labeled PNN-SR dataset.
}
\small
\centering
\newcolumntype{G}{D{;}{\,\pm\,}{4.3}}
\newcolumntype{C}{>{\centering\arraybackslash}X}
\begin{tabularx}{\linewidth}{ll*4G}
\toprule
&& \multicolumn{4}{c}{Raters' Agreement} \\ 
   \cmidrule(lr){3-6}
&& 
\multicolumn{1}{C}{\ Any\ \ {\footnotesize $(a \ge 1)$}} & 
\multicolumn{1}{C}{$\ge 50\%$ {\footnotesize $(a \ge 4)$}} & 
\multicolumn{1}{C}{$\ge 70\%$ {\footnotesize $(a \ge 5)$}} & 
\multicolumn{1}{C}{$100\%$ {\footnotesize $(a = 7)$}} \\    
&& 
\multicolumn{1}{C}{{\footnotesize 2,351 obj.}} & 
\multicolumn{1}{C}{{\footnotesize 1,384 obj.}} & 
\multicolumn{1}{C}{{\footnotesize 1,234 obj.}} & 
\multicolumn{1}{C}{{\footnotesize 880 obj.}} \\    
\midrule
\multirow{3}{*}{MAE} 
& S-UNet &  27.6;24.8 &  15.1;14.1 &  15.8;11.6 &  13.8;12.8 \\
& FRCNN &  27.8;21.6 &  15.5;13.4 &  13.3;12.6 &    7.8;9.9 \\
& D-CSRNet &  91.5;43.6 &  21.1;23.0 &  15.3;20.1 &   10.9;8.9 \\
\midrule
\multirow{3}{*}{MARE (\%)} 
& S-UNet &  14.2;11.3 &  12.6;12.9 &   14.5;9.6 &  19.0;13.3 \\
& FRCNN &  14.9;11.0 &  13.6;10.9 &  12.7;12.5 &  11.8;13.6 \\
& D-CSRNet &  47.8;20.3 &  18.9;18.0 &  14.3;14.8 &  15.7;11.7 \\
\midrule
\multirow{3}{*}{GAME(3)}
& S-UNet &  61.8;18.7 &  36.8;15.2 &  34.9;15.4 &  29.0;13.3 \\
& FRCNN &  60.2;12.9 &  31.7;13.5 &  28.5;12.6 &   23.2;7.8 \\
& D-CSRNet &  99.3;38.5 &  41.2;18.4 &  35.0;14.8 &  28.1;11.8 \\
\bottomrule
\end{tabularx}
\label{tab:pnn-agree-mae}
\end{table}

\begin{figure*}
\centering
\newcolumntype{C}{>{\centering\arraybackslash}X}
\setlength{\tabcolsep}{1pt}
\begin{tabularx}{\linewidth}{l@{\hspace{5pt}}CCCC}
& \multicolumn{2}{c}{Easy Sample}
& \multicolumn{2}{c}{Hard Sample} \\
\cmidrule(lr){2-3} \cmidrule(lr){4-5}
& $a\ge1$
& $a = 7$
& $a\ge1$
& $a = 7$ \\
\rotatebox[origin=c]{90}{\small Groundtruth}&%
\includegraphics[align=c,width=\linewidth]{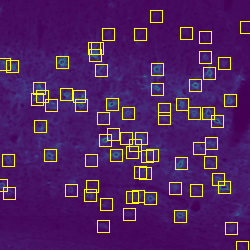}&%
\includegraphics[align=c,width=\linewidth]{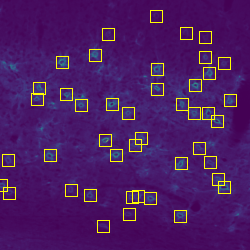}&%
\includegraphics[align=c,width=\linewidth]{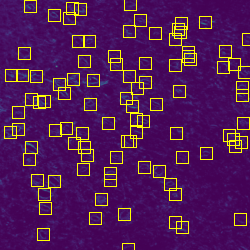}&%
\includegraphics[align=c,width=\linewidth]{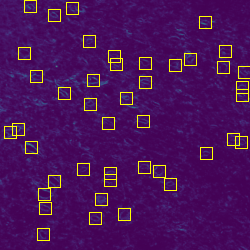}\vspace{0.5ex}\\%
\rotatebox[origin=c]{90}{\small S-UNet}&%
\includegraphics[align=c,width=\linewidth]{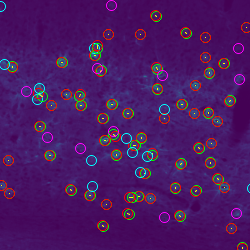}&%
\includegraphics[align=c,width=\linewidth]{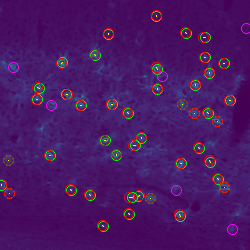}&%
\includegraphics[align=c,width=\linewidth]{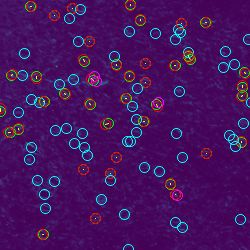}&%
\includegraphics[align=c,width=\linewidth]{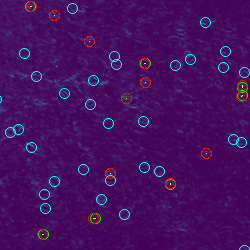}\vspace{0.5ex}\\%
\rotatebox[origin=c]{90}{\small FRCNN}&%
\includegraphics[align=c,width=\linewidth]{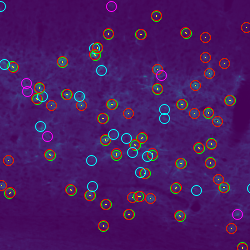}&%
\includegraphics[align=c,width=\linewidth]{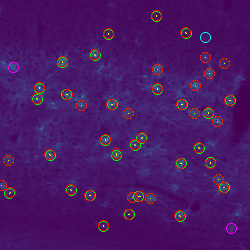}&%
\includegraphics[align=c,width=\linewidth]{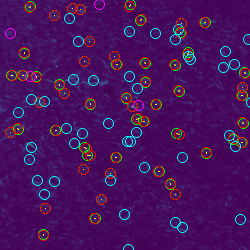}&%
\includegraphics[align=c,width=\linewidth]{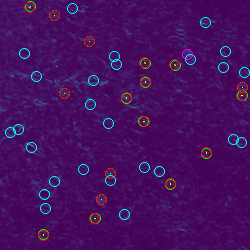}\vspace{0.5ex}\\%
\rotatebox[origin=c]{90}{\small D-CSRNet}&%
\includegraphics[align=c,width=\linewidth]{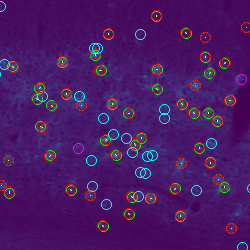}&%
\includegraphics[align=c,width=\linewidth]{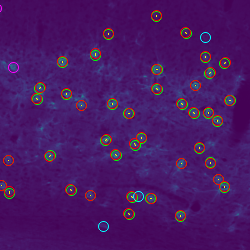}&%
\includegraphics[align=c,width=\linewidth]{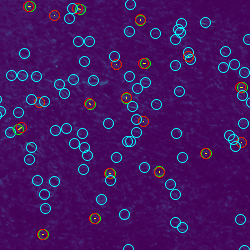}&%
\includegraphics[align=c,width=\linewidth]{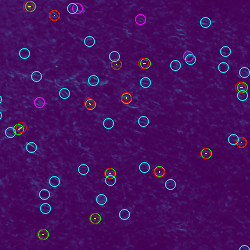}
\end{tabularx}
\definecolor{mypurple}{rgb}{1,0,1}
\definecolor{mycyan}{rgb}{0,1,1}
\caption{\textbf{PNN-MR: Examples of localization predictions of tested models.}
We show portions of two samples; an easy one (first two columns) and a hard one (last two columns), with different ground truths defined by including only PNNs with a minimum raters' agreement $a$.
In the first row, we highlight the \colorbox{yellow}{ground truth} in yellow squares.
In the rest of the rows, we indicate \colorbox{mypurple}{false positives} in purple, \colorbox{mycyan}{false negatives} in cyan, and \colorbox{green}{true positives} in green, with the corresponding \colorbox{red}{ground-truth position} drawn in red and connected via a thin yellow line.
Best viewed in electronic format.
}
\label{fig:pnn-prediction-examples}
\end{figure*}

\subsection{Scoring Stage Evaluation}
\label{sec:counting-with-uncertainty:experiments:exp-scoring}
Here, we describe the experiments to evaluate the proposed additional scoring stage.
The goal was to produce new ``objectness'' scores that correlate with the raters' agreement;
we refer to scores produced in this stage as \textit{calibrated} scores, in contrast with the uncalibrated ones derived in the previous stage.
This stage required a small multi-rater dataset to be used as a training set, and thus we adopted the PNN-MR dataset for both the training and testing phases.
To this end, we randomly split the images comprising PNN-MR into train, validation, and test sets following the widely employed 70/15/15 proportion. \ref{tab:pnn-mr-split} reports the average number of objects per agreement level in the PNN-MR train/validation/test splits.

\begin{table}[t]
\caption{\textbf{PNN-MR Splits Statistics}. Mean and standard deviation of number of objects in the five PNN-MR 70/15/15 random splits.}
\label{tab:pnn-mr-split}
\centering
\setlength{\tabcolsep}{3pt}
\newcolumntype{G}{D{;}{\,\pm\,}{4.3}}
\newcolumntype{C}{>{\centering\arraybackslash}X}
\begin{tabularx}{\linewidth}{l*4G}

\toprule
& \multicolumn{4}{c}{Raters' Agreement} \\ 
  \cmidrule(lr){2-5}
&
\multicolumn{1}{C}{       Any} & 
\multicolumn{1}{C}{$\ge 50\%$} & 
\multicolumn{1}{C}{$\ge 70\%$} & 
\multicolumn{1}{C}{$   100\%$} \\
&
\multicolumn{1}{C}{\footnotesize $(a \ge 1)$} &
\multicolumn{1}{C}{\footnotesize $(a \ge 4)$} &
\multicolumn{1}{C}{\footnotesize $(a \ge 5)$} &
\multicolumn{1}{C}{\footnotesize $(a  =  7)$} \\
\midrule
Total        &   2351 &  1384 &  1234 &   880 \\
\midrule
Train      &  1167;70 &  678;56 &  606;50 &  428;33 \\
Validation &   569;60 &  351;38 &  314;33 &  232;28 \\
Test       &   615;58 &  356;27 &  315;24 &  220;25 \\
\bottomrule
\end{tabularx}
\end{table}

\begin{figure}
\centering
\includegraphics[width=\linewidth]{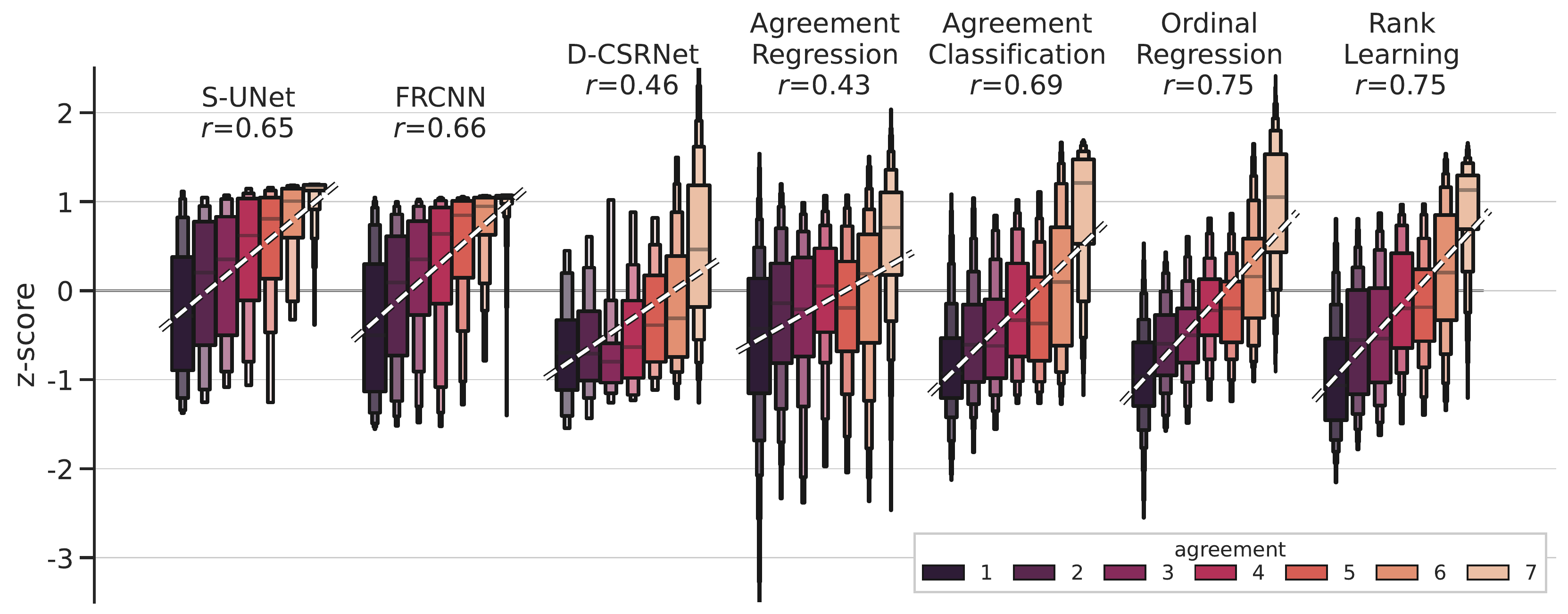}
\caption{\textbf{PNN-MR: Correlation between scores and raters' agreement.} We show the distribution of (z-normalized) scores per agreement level, the regression lines, and the
Pearson correlation coefficient $r$ between scores and raters' agreement for each tested methods.}
\label{fig:score-vs-agreement}
\end{figure}

First, we assessed this scoring stage in a stand-alone way, considering it independently from our overall counting pipeline. 
We implemented the scorer model $g_\phi$ as a small convolutional network with 8 Conv-GroupNorm-ReLU blocks followed by average pooling and a linear projection producing the desired number of outputs (8 for Agreement Classification, 1 for the rest).
We trained each calibration methodology presented in \ref{sec:counting-with-uncertainty:method:score-calibration-stage} by providing, as inputs, small patches around PNNs centered in the locations provided by the ground-truth labels.
Due to the limited size of the dataset, we performed five runs with randomly generated splits.
In \ref{fig:score-vs-agreement}, we report the distribution of (z-normalized) scores obtained by the tested methods on the PNN patches of the test splits.
In addition, we also report the distributions of the uncalibrated scores obtained from the localization models as described in \ref{sec:counting-with-uncertainty:experiments:exp-pnn}.
We notice that scores obtained by Agreement Ordinal Regression (OR) and Agreement Rank Learning (RL) strategies behave best in terms of correlation with the raters' agreement $a$ achieving the highest Pearson's correlation index $r$ of 75\%.
Those are also the most data-efficient methods, as they operate on pairs or tuples of samples, while Agreement Regression (AR) and Classification (AC) seem to suffer from the limited number of samples.
Thus, opting for OR or RL is suggested, as multi-rater data is often limited.
Moreover, the mean scores per agreement level of OR and RL tend to follow a steep regression line. In contrast, in other methods like \textit{D-CSRNet} and AR, the score distributions of nearby agreement levels are more overlapped and often with non-monotonic means.

\begin{figure}
\centering
\includegraphics[width=\linewidth]{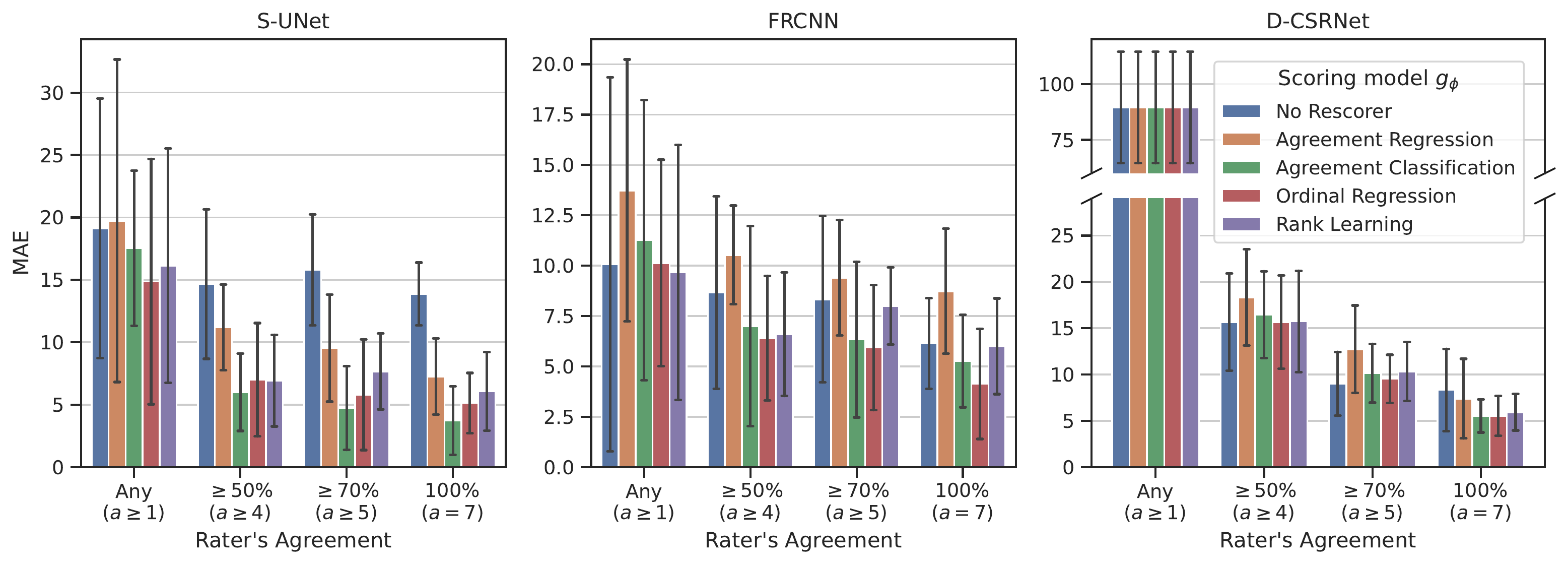}
\caption{
\textbf{PNN-MR (Test Subset): Impact of the rescoring stage $g_\phi$ on counting performance in terms of MAE.}
We show mean values $\pm$ standard deviation over five runs with randomized train/val/test splits of the PNN-MR subset.
Numerical values in tabular format can be found in \ref{tab:calibrated-counting}.
}
\label{fig:calibrated-counting}
\end{figure}

Next, we evaluated the effect of the scoring stage on the whole counting pipeline.
We first localized PNNs in test images with the three localization models trained in \ref{sec:counting-with-uncertainty:experiments:exp-pnn};
for this operation, we set a high-recall threshold for each model, such that filtering can be postponed after rescoring by $g_\phi$.
We then scored the locations found using the trained scoring models and evaluate counting performance for each combination of localization and scorer models.
\ref{fig:calibrated-counting} and \ref{tab:calibrated-counting} compare the achieved performance in terms of MAE when choosing the best threshold value for the rescored predictions. As baseline, we report also the counting performance of using uncalibrated scores, i.e., without using the proposed rescoring stage.
Again, we report results for different ground-truth settings defined by the minimum desired raters' agreement.
\textit{S-UNet} combined with Agreement Classification (AC) achieves the best counting performance among most ground-truth configurations, whereas Agreement Ordinal Regression (OR) is the second-best rescoring solution.
Despite providing significant boosts compared to other tested rescoring approaches, AC can suffer when multi-rater samples are unbalanced (or missing) among agreement levels, which is fairly common in this application.
In those cases, rank-based methods (OR or RL) are known to behave better under these scenarios. However, we leave to future work the evaluation of sample efficiency and of robustness to class unbalance.
Note that for \textit{S-UNet} and \textit{FRCNN}, score calibration generally improves the counting performance, specifically for the former where we achieve MAE reductions up to 11.07.
On the other hand, when adopting the \textit{D-CSRNet} localization method, we achieve an improvement only on the highest agreement test set;
this is mainly due to the limited recall of \textit{D-CSRNet} on the PNN dataset.

\begin{table}
\caption{\textbf{PNN-MR (Test Subset): Impact of the rescoring stage $g_\phi$ on counting performance in terms of MAE.}
We report mean and standard deviation over five runs with randomized train/val/test splits of the PNN-MR subset.
`w/o' indicates results without the rescoring stage, i.e., filtering is performed on method-specific scores extracted in the first localization stage. Results are also showed using plots in \ref{fig:calibrated-counting}.
}
\small
\centering
\newcolumntype{G}{>{\centering\arraybackslash}X}
\newcolumntype{C}{>{\centering\arraybackslash}X}
\begin{tabularx}{\linewidth}{ll*4G}
\toprule
&& \multicolumn{4}{c}{Raters' Agreement} \\ 
   \cmidrule(lr){3-6}
&& 
\multicolumn{1}{C}{\ Any\ \ {\footnotesize $(a \ge 1)$}} & 
\multicolumn{1}{C}{$\ge 50\%$ {\footnotesize $(a \ge 4)$}} & 
\multicolumn{1}{C}{$\ge 70\%$ {\footnotesize $(a \ge 5)$}} & 
\multicolumn{1}{C}{$100\%$ {\footnotesize $(a = 7)$}} \\    
&
$g_\phi$ \\
\midrule
\multirow{5}{*}{\rotatebox[origin=c]{90}{S-UNet}} 
& w/o &  19.13 $\pm$ 11.63 &  14.67 $\pm$ 6.70 &  15.80 $\pm$ 4.98 &  13.87 $\pm$ 2.81 \\
& AR &  19.73 $\pm$ 14.44 &  11.20 $\pm$ 3.83  &  9.53 $\pm$ 4.80 &  7.27 $\pm$ 3.41 \\
& AC &  17.53 $\pm$ 6.93  & 6.00 $\pm$ 3.46 & 4.73 $\pm$ 3.75 & 3.73 $\pm$ 3.07 \\
& OR &  14.87 $\pm$ 10.98  & 7.00 $\pm$ 5.07 & 5.80 $\pm$ 4.96 &  5.13 $\pm$ 2.69 \\
& RL &  16.13 $\pm$ 10.48 & 6.93 $\pm$ 4.10 &  7.67 $\pm$ 3.39 &   6.07 $\pm$ 3.51 \\
\midrule
\multirow{5}{*}{\rotatebox[origin=c]{90}{FRCNN}} 
& w/o &  10.07 $\pm$ 10.38 &  8.67 $\pm$ 5.33 &  8.33 $\pm$ 4.61 &  6.13 $\pm$ 2.51 \\
& AR &  13.73 $\pm$ 7.27 &  10.53 $\pm$ 2.73 &  9.40 $\pm$ 3.21 &  8.73 $\pm$ 3.47 \\
& AC &  11.27 $\pm$ 7.77 & 7.00 $\pm$ 5.55 & 6.33 $\pm$ 4.31 & 5.27$\pm$ 2.56 \\
& OR &  10.13 $\pm$ 5.73 &  6.40 $\pm$ 3.46 &  5.93 $\pm$ 3.47 &  4.13 $\pm$ 3.06 \\
& RL & 9.67$\pm$ 7.08 & 6.60 $\pm$ 3.43 &  8.00 $\pm$ 2.13 &   6.00 $\pm$ 2.66 \\
\midrule
\multirow{5}{*}{\rotatebox[origin=c]{90}{D-CSRNet}} 
& w/o &  89.53 $\pm$ 27.92 &  15.67 $\pm$ 5.85 &  9.00 $\pm$ 3.85 &  8.33 $\pm$ 4.96 \\
& AR &  89.53 $\pm$ 27.92 &  18.33 $\pm$ 5.80  &  12.73 $\pm$ 5.28 &  7.40 $\pm$ 4.80  \\
& AC &  89.53 $\pm$ 27.92  & 16.47 $\pm$ 5.24  & 10.13 $\pm$ 3.55 & 5.53 $\pm$ 1.99 \\
& OR &  89.53 $\pm$ 27.92 &  15.67 $\pm$ 5.62 &  9.53 $\pm$ 2.90 &  5.53 $\pm$ 1.99 \\
& RL & 89.53 $\pm$ 27.92 & 15.73 $\pm$ 6.12 &  10.33 $\pm$ 3.55 &   5.93 $\pm$ 2.20 \\
\bottomrule
\end{tabularx}\\[1ex]
{\footnotesize
AR = Agreement Regression.
AC = Agreement Classification.
OR = Agreement Ordinal Regression.
RL = Agreement Rank Learning.}
\label{tab:calibrated-counting}
\end{table}

\section{Summary}
\label{sec:counting-with-uncertainty:conclusion}
In this chapter, we tackled the task of counting biological structures from microscopy images under the assumption that training datasets were characterized by weak multi-rater labels, i.e., in the presence of non-negligible disagreement between multiple raters. This is a hypothesis that often occurs in medical images where intrinsically non-trivial patterns can produce weak annotations due to raters' judgment differences, even among experts.
More robust annotations can be obtained by aggregating and averaging the decisions provided by multiple raters regarding the same data.
However, the scale of the counting task and the limited resources dedicated to the labeling process put a damper on this solution since raters prefer to label new data rather than annotate the same data more times. Consequently, the challenge here has been to deal with weakly-labeled large sets of data, i.e., characterized by errors due to differences in raters' judgment, and very small, more reliable multi-rater data, from which it becomes crucial to make the best fully exploiting the redundancy of the information represented by the different raters' opinions. Thus, as in the previous chapters, also in this case, we faced the counting task in the presence of data scarcity, but in a different setting that characterized medical images and, in general, all the contexts where non-trivial patterns prevent the possibility of easily identifying the objects we want to count.

In this setting, we proposed a two-stage counting strategy, where each stage is devised to make the best of the annotations available in each data subset.
The first stage exploited large single-rater data to bootstrap state-of-the-art counting methodologies; we evaluated three CNN-based methods i.e., segmentation-based \textit{S-UNet}, detection-based \textit{FRCNN} and density-based approaches \textit{D-CSRNet}.
We showed that this step alone leads to sub-optimal results due to the underlying noisy nature of the employed single-rater data.
Thus, we introduced a second rescoring stage that harnesses a small multi-rater subset and refines the previously computed predictions.

We performed an extensive experimental evaluation of our pipeline on a novel weakly-labeled dot-annotated dataset introduced on purpose, consisting of a collection of fluorescence microscopy images of mice brains containing biological structures.
Results showed that rescoring strategies improved the correlation between the scores and the raters' agreement. Using the proposed pipeline, we enhanced counting performance, in some cases significantly reducing the MAE. Whereas even simple rescoring methods such as Agreement Classification is beneficial, we deem the rank-based ones, like Agreement Ordinal Regression and Agreement Rank Learning, to be also data-efficient and robust to data unbalance, operating on pairs or tuples of samples. However, we leave a rigorous evaluation of those aspects to future work.

\graphicspath{{img/counting-for-covid/}}

\chapter{An Embedded Toolset for Human Activity Monitoring in Critical Environments}
\label{ch:counting-for-covid}

As occurs during a severe health emergency event, there exist scenarios in which ensuring compliance to a set of guidelines becomes crucial to secure a safe living environment in which human activities can be conducted. In fact, as evidenced during the recent COVID-19 pandemic, wearing medical masks, avoiding the creation of large gatherings in confined places, and keeping a certain physical distance among people were the most common rules every government applied in their jurisdiction territories. However, human supervision could not always guarantee this task, especially in crowded scenes where checking personal protection equipment or enforcing strict social behaviors must be continuously assessed to preserve global health. 

As we have already seen in the previous chapters, Computer Vision applications have shown outstanding results in several daily life tasks. Automatic image analysis aimed at classifying, locating, and counting objects, as well as estimating the distance between different instances of objects, are typical examples of applications of Computer Vision technology, which can be a valuable tool to automatically monitor human activities in critical environments through images captured by networked cameras.

In this chapter, we present an embedded modular Computer Vision-based and AI-assisted system that puts into practice some of the techniques described in the previous chapters, carrying out several tasks to help monitor individual and collective human safety rules. We strive for a real-time but low-cost system, thus complying with the compute- and storage-limited resources availability typical of off-the-shelves embedded devices, where images are captured and processed directly onboard. Our solution consists of multiple modules, each responsible for a specific functionality that the user can easily enable and configure. One of them aims at estimating the number of people present in a region of interest, a piece of crucial information to monitor the occupancy area. By measuring, and eventually limiting, the number of people who can be in a specific site at any one time, it is possible to reduce the likelihood of setting up gatherings. Other helpful capabilities that our system makes available are the skill to estimate the so-called social distance (i.e., the physical distance among pedestrians), as well as the possibility to localize and classify \acrfull{ppe} worn by people (such as helmets, high-visibility clothing, and face masks) that the World Health Organizations has recommended as one of the primary tools able to curb the spread of the disease, like, for example, the recent COVID-19 pandemic.

To validate our solution, we test all the functionalities that our framework, deployed on an embedded device, makes available, exploiting two novel datasets that we collected and annotated on purpose and that represent another scientific contribution. Specifically, we gathered a first dataset comprising images captured by a smart camera located in a public square in the city of Pisa, Italy, that represents a typical scenario for which it is crucial to check compliance with the safety rules. Moreover, we collected and annotated a second dataset comprising synthetic and real-world images containing pedestrians with and without \acrshort{ppe}, such as helmets, high-visibility vests, and face masks. 
Experiments show that our system can effectively carry out all the functionalities that the user can set up, providing to be a valuable asset to automatically monitor compliance with safety rules.

To summarize, the main contributions introduced in this chapter are the following:
\begin{itemize}
\item We introduce an expandable and flexible Computer Vision-based and \acrshort{ai}-assisted embedded system, deployed in a real use-case scenario, capable of automatically monitoring human activities in critical environments, where individual and collective safety must be constantly checked. We base our solution on modules responsible for specific tasks that the user can easily configure and add to the whole system, making available many functionalities such as estimating the number of pedestrians present in the monitored scene.
\item We collect and annotate two novel datasets that we exploit to validate our framework. One, named \textit{CrowdVisorPisa}, is gathered from a camera in a public square of the city of Pisa, Italy, and represents a typical scenario for which it can be essential to monitor compliance with the safety rules, such as the respect of a ruled maximum number of people allowed to stay in the site.
\item We conduct experiments evaluating all the modules and the functionalities, which our framework makes available \emph{in an embedded and deployed off-the-shelf device}, showing that our solution may be a valid aid to monitor and handle critical environments drastically reducing human supervision. 
\end{itemize}

We organize the rest of the chapter as follows. We review similar works in \ref{sec:counting-for-covid:related_work}, and we introduce our modular framework and its plug-ins in \ref{sec:counting-for-covid:framework}. In \ref{sec:counting-for-covid:datasets}, we describe the exploited datasets along with the adopted training procedures. In \ref{sec:counting-for-covid:experiments}, we show our experiments, also discussing and analyzing the obtained results. 

The research presented in this chapter was published in \cite{crowdvisor_eswa}.

\section{Related Work}
\label{sec:counting-for-covid:related_work}

Due to the COVID-19 pandemic, many Computer Vision-based works have been recently published to help monitor human activities analyzing images, especially on the specific task of evaluating the social distance between people. For example, the Inter-Homines system, presented in \cite{cucchiara_1}, evaluates in real-time the contagion risk in a monitored area by analyzing video streams. The system includes occlusion correction, homography transformation, and people anonymization. People are located in the space exploiting the CenterNet \cite{centernet} object detector, and interpersonal distances are then calculated. Results are evaluated on the \acrshort{jta} dataset \cite{fabbri_jta} (i.e., in a virtual world). In \cite{saponara2021implementing}, the YOLO9000 detector \cite{yolo9000} has been exploited to detect people, and centroids of the found bounding boxes are computed to evaluate the distance between them. Similarly, in \cite{AHMED2021102571}, a platform for social distance tracking in top perspective video frames based on YOLOv3 \cite{yolo_v3} was presented. Here too, centroids of the bounding boxes are used to estimate distances. A subset of the authors of \cite{AHMED2021102571} also presented in \cite{AHMED2021102777} a social distance framework based on the Faster-RCNN detector \cite{faster_rcnn}. Instead, YOLOv3 \cite{yolo_v3} to detect humans and Deepsort \cite{deepsort} to track people are exploited in \cite{punn2020monitoring}. Experiments are conducted on the Oxford town center surveillance footages \cite{Benfold2011StableMT}. 
Monitoring of workers to detect social distancing violation that uses Mobilenet-V2 \cite{mobilnet_v2} to detect people is introduced in \cite{khandelwal2020using}.

Another task recently tackled in literature, again related to the COVID-19 pandemic, is face mask detection. For example, authors \cite{mask_det} present an edge computing-based mask identification framework (ECMask). It consists of three main stages: video restoration, face detection (inspired by FaceBoxes \cite{faceboxes}), and mask identification (based on Mobilenet-V2 \cite{mobilnet_v2}). \acrlong{dl} models were trained and evaluated on the Bus Drive Monitoring Dataset, which unfortunately is not publicly available. Authors in \cite{DBLP:journals/corr/abs-2103-08773} developed a deep learning-based computer vision system able to perform face mask detection but also face-hand interaction detection. A more comprehensive literature review of applications of \acrlong{ai} in battling against COVID-19 is given in \cite{TAYARANIN2021110338} including social distancing and face mask detection.

Differently from most other works, in this chapter, we present a \textit{modular} and \textit{expandable} Computer Vision-based embedded system that can fulfill \textit{multiple} tasks to help monitor compliance of individual and collective human safety rules in critical scenarios, like the one caused by the COVID-19 pandemic. The main peculiarities are that it runs directly on a low-cost computing device and that the user can easily enable and configure the available functionalities, ranging from estimating the number of people present in the scene, computing social distances, or detecting \acrshort{ppe}s, combining more modules and building more complex tasks.

\section{Modular Framework}
\label{sec:counting-for-covid:framework}

The general purpose of our monitoring system is to be embeddable on low-cost devices and, above all, to be expandable to different features in demanding situations. To this end, we designed a framework able to orchestrate a set of internal and user-defined \emph{plugins}, each dedicated to a single task. Specifying inputs and outputs makes it possible to create a dependency graph. Each sub-module represents a node, and each pair of matching input-output represents an edge. In this way, given the desired output, a \emph{topological sort} is executed to minimize and linearize the sequential execution of computations. Although an easier solution may exist in the context of the plugins, our methodology is relatively simple to implement and allows low-cost systems to execute any complex compute graphs in a sequential and semantically-correct way.

\begin{figure}[t]
\centering
\includegraphics[page=4,width=\linewidth]{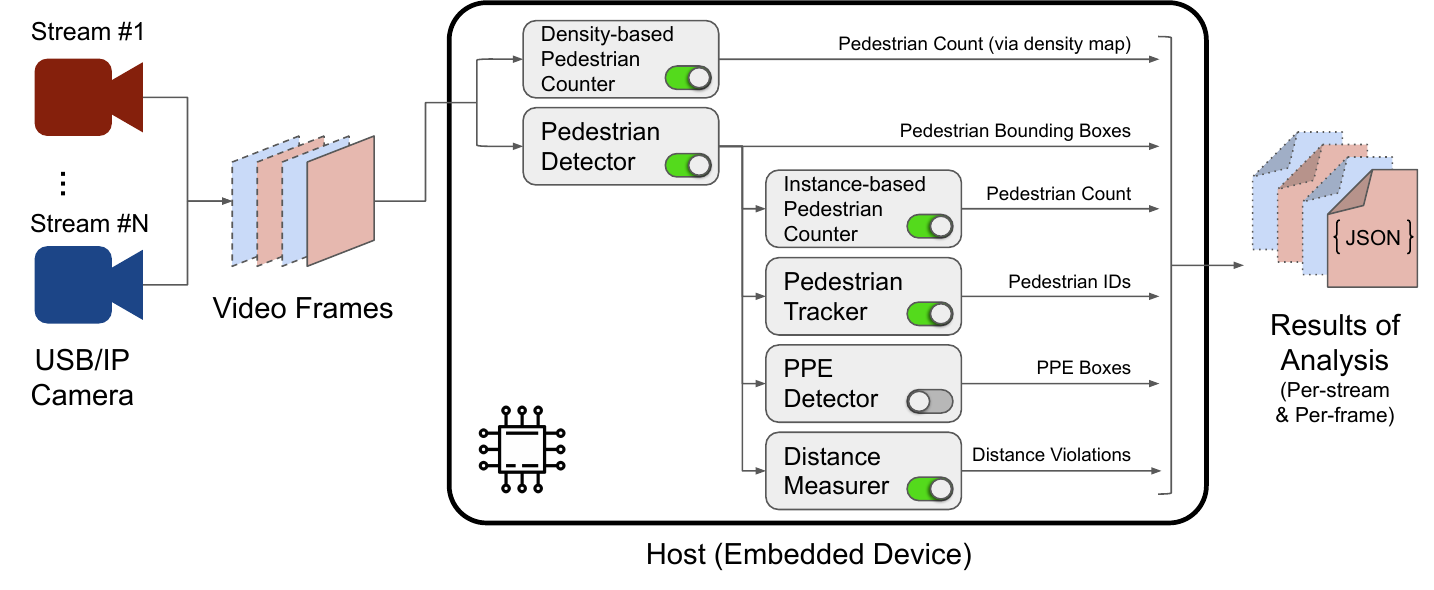}
\caption{\textbf{Overview of our modular framework.}
Multiple video streams can be multiplexed and handled by a single instance. Results are generated in JSON format and can be routed to downstream services.
Our system is \textit{flexible} and \textit{expandable}, as modules can be activated or deactivated depending on the user's needs, and novel functionalities can be introduced with additional custom modules.}
 \label{fig:overview_covid}
\end{figure}

\begin{figure}[t]
\centering
\includegraphics[page=5,width=\linewidth]{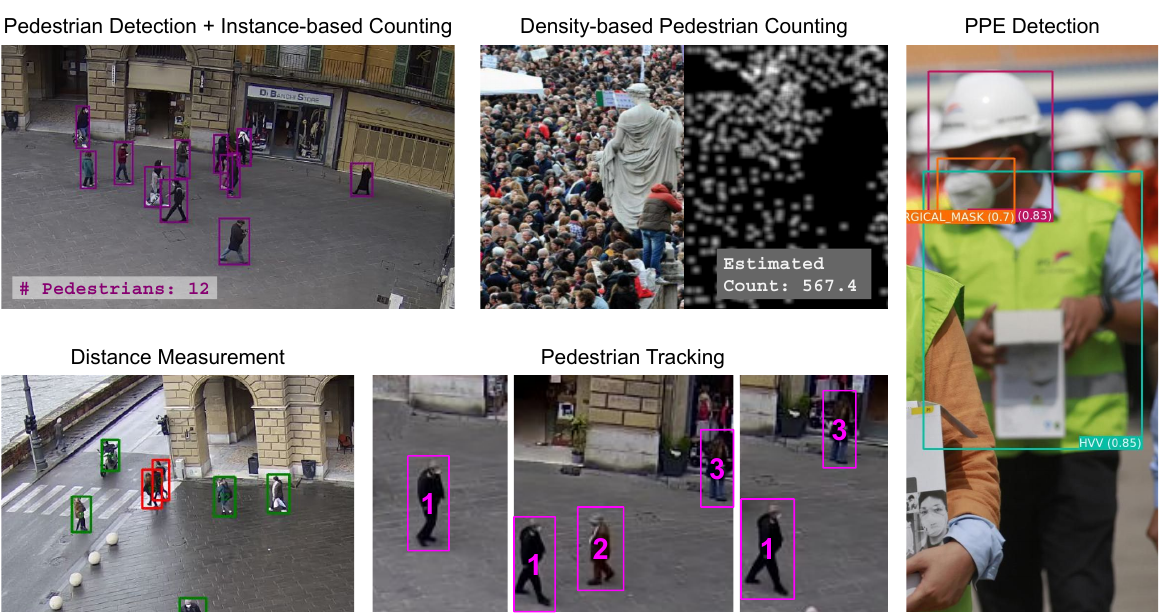}
\caption{\textbf{Visualization of output examples of the modules currently available in our system.} The outputs of each module are the following.
\textbf{Pedestrian Detection \& Instance-based Counting}: list of pedestrian bounding boxes and respective count.
\textbf{Pedestrian Tracking}: numeric ID assigned to detected pedestrians persisting through frames.
\textbf{Density-Based Pedestrian Counting}: estimated number of pedestrians (and, optionally, the density map).
\textbf{Distance Measurement}: IDs of groups of pedestrians violating a predefined distance.
\textbf{PPE Detection}: list of PPE bounding boxes detected per pedestrian.}
\label{fig:sample-predictions}
\end{figure}

An overview of our modular framework is depicted in Figure~\ref{fig:overview_covid}.
Video frames are taken at regular intervals from one or more cameras and processed locally.
Multiple video streams can be multiplexed and handled by a single system instance.
Current modules include a) Pedestrian Detector, b) Density-based Pedestrian Counter, c) Instance-based Pedestrian Counter, d) Pedestrian Tracker, e) PPE Detector, and f) Interpersonal Distance Measurer; Figure~\ref{fig:sample-predictions} exemplifies the results of the analyses performed by each module, whereas their detailed description is reported in the following sections.
All the modules are toggleable; the Instance-based Pedestrian Counter, Pedestrian Tracker, Interpersonal Distance Measurer, and PPE Detector modules depend on the output of the Pedestrian Detector module and require it to be active.
Results of the active modules are combined and provided in JSON format to be consumed by downstream services. Note that video frames are analyzed onboard and never stored; this enables privacy-aware solutions where captured images never leave the edge devices.

\paragraph{Pedestrian Detector}
\label{sec:counting-for-covid:framework:object_detector}
The pedestrian detector is the system's main component on which almost all other plug-ins rely. Its primary purpose is to localize and classify pedestrian instances from input images. These detections constitute the main data that will be exploited, in different ways, by the other nodes of the system.

We base our pedestrian detector on \textit{Faster R-CNN} \cite{faster_rcnn}, a popular state-of-the-art CNN-based object detection system belonging to the two-stage paradigm, already described in \ref{sec:back:cnn-based-detectors:two-stages-detectors}, and that we also exploited in \ref{ch:virtual-to-real} and \ref{ch:counting-with-uncertainty}. We preferred a two-stage detector to single-shot detectors, as we could easily extract features from the region proposal stage and make them directly available for subsequent processing. We used the extracted features when tracking pedestrians, and they could be used in other future modules, such as cross-camera pedestrian re-identification. Moreover, Faster R-CNN is widely adopted and usually guarantees a state-of-the-art detection performance. However, the modularity of our system does not limit us to Faster R-CNN, and another object detector can be easily adopted in the future, just replacing the related module.

We specialize this detection system to localize pedestrian instances through a supervised domain adaption technique that exploits several pedestrian datasets, in a similar way to what we have done in \ref{ch:virtual-to-real}. 

\paragraph{Pedestrian Tracker}
\label{sec:counting-for-covid:framework:Tracker}
Object tracking can be an essential tool to increase the robustness to spurious detections and achieve temporal consistency in video analysis. To this end, we implement and apply an object tracker over pedestrian detection to reidentify people among consecutive video frames. This step is beneficial for assessing temporal rules, such as raising alarms after the same pedestrian occupies a forbidden area for more than a predefined amount of time. 

The implementation of the tracker follows the formulation of DeepSort~\cite{deepsort}; it is based on SORT~\cite{sort}, a simple causal tracking algorithm for 2D objects in which targets are represented by the position and area of the bounding box and their speed of variation. The state of each target (also known as \textit{tracklet}) is updated with available detections using a Kalman filter framework.
DeepSort builds upon SORT by adding a matching scheme between predicted and actual targets based on feature vectors that describe the appearance of tracked objects; tracklets can be confirmed if the cosine score between feature vectors of the predicted and actual target is above a programmable threshold. Feature vectors in DeepSort must be provided by extracting representations from detected regions with an additional pretrained network. In our implementation, we avoid this step by reusing the feature vectors of detected regions that the object detection network has already extracted; in particular, we perform a Region of Interest (RoI) average pooling of the features extracted by the CNN backbone using only the regions provided by the pedestrian detection module.

\paragraph{Instance-based and Density-based Pedestrian Counter}
\label{sec:counting-for-covid:framework:crowd_counter}
In some scenarios where individual and collective safety has to be constantly monitored, like people aggregations during the recent COVID-19 pandemic, estimating the number of people present in a region of interest is crucial to monitor the occupancy area. By measuring, and limiting, the number of people who can visit a location at any one time, it is possible to reduce the likelihood of setting up people gatherings and, consequently, minimize human virus transmission. Our solution relies on a dedicated plug-in that can work in two different modalities that the user can conveniently pick out, depending on the considered scenario. The first one, named \textit{Instance-based Pedestrian Counter}, is better suited for not particularly crowded environments and relies on the pedestrian detector component previously described. The second, named \textit{Density-based Pedestrian Counter}, is instead a more holistic approach more appropriate for highly crowded scenarios; it aims to compute a mapping between the features of the captured image and its pedestrian density maps, skipping the detection of the single instances. The estimated number of people present in the controlled area can then be obtained by integrating this density map. More information about this technique are provided in Section \ref{sec:back:visual-counting:traditional-approaches}. We build our density map estimator upon the Congested Scene Recognition Network (CSRNet) \cite{csrnet}, a CNN-based algorithm that can understand highly congested scenes and perform accurate density estimation, already described in \ref{sec:back:visual-counting:cnn-based} and that we exploited also in \ref{ch:uda-counting} and \ref{ch:counting-with-uncertainty}, for estimating urban traffic density and counting cells in microscopy images, respectively. 

\paragraph{Interpersonal Distance Measurer}
\label{sec:counting-for-covid:framework:distance_measurer}
A critical condition that must be kept under control in dynamic environments where an infection is ongoing is represented by the physical distance among individuals. In the case of air-borne diseases, it is thus very common to issue rules to avoid people gatherings in confined places and keep a specific reciprocal separation to contrast the spread of pathogen agents. Although crowd counting is effective in monitoring aggregations, measuring distances among people becomes critical during pandemic events. Assuming that individuals mostly hang out on the same planar floor, we decided to measure their actual distance by applying a simple pre-calibrating step to the fixed monitoring camera, using a proper geometrical transformation that places detected items on a common system of reference. Our solution is to pre-compute a mapping between real points in the scene whose relative position is known and their projection on the acquired frame image. This process is well known in Computer Vision and consists of finding a \emph{homography}, i.e., a perspective transformation that projects on two different points of view a set of 3D points lying on the same plane.

\paragraph{Personal Protective Equipment Detector}
\label{sec:counting-for-covid:framework:ppe_detector}
A simple intervention for protecting health and well-being is wearing \acrfull{ppe}; this is particularly true in dangerous working environments, such as wearing harnesses and helmets on construction sites. Still, it also became evident in light of the recent COVID-19 pandemic, where wearing face masks can prevent infections.
Therefore, we implement a module dedicated to detecting worn \acrshort{ppe}, essential for ensuring compliance with regulations that imply personal protection. 

Our solution for \acrshort{ppe} detection follows the same methodology already adopted for pedestrian detection: specifically, we assume the same detector architecture based on Faster-RCNN. Differently from the pedestrian detector, the \acrshort{ppe} detector network takes as input a rather small image depicting a pedestrian, and it is trained to distinguish and detect several classes of worn \acrshort{ppe}, i.e., surgical/face masks, helmets, and high-visibility vests.
The module's input is provided by the pedestrian detection module: once detected, the patch of the video frame depicting a pedestrian is given as input to the \acrshort{ppe} detector that provides bounding boxes of \acrshort{ppe} if the pedestrian wears them. 

Conceptually, the detection of \acrshort{ppe} could be tackled by the object detector module by adding the \acrshort{ppe} classes to the base detection model.
However, we empirically noticed that merging the \acrshort{ppe} detection with the pedestrian detection module leads to performance degradation in both tasks.
Using separate modules provides a more flexible solution in which the input image resolution of both detectors can be adjusted separately and better adapted to the monitored scenario.
For example, when wide areas are monitored, \acrshort{ppe} detection can be performed several times at each processed video frame depending on how many pedestrians have been detected, while the \acrshort{ppe} detection network can be configured to operate on smaller input sizes, maintaining an affordable computational cost.

\section{Datasets and Architecture Adaptions}
\label{sec:counting-for-covid:datasets}
A key point in producing a verification system that can generalize on a broad spectrum of working conditions is to generate a training set based on an adequately large amount of environmental conditions. In our case, this means being able to access a massive amount of images involving people under different environmental scenarios. As we have already seen in the previous chapters, this point represents one of the main challenges related to the \acrshort{cnn}s, and in general to all supervised learning approaches, since manually annotating new collections of images is expensive and requires a notable human effort. Thus, to train the modules of our framework, we build vast training datasets, covering a multitude of different scenarios and contexts. We consider images from public datasets when available, and we collect others when needed. As we have already seen in \ref{ch:virtual-to-real} and \ref{ch:uda-counting}, a promising approach is to gather data from virtual world environments that mimics as much as possible all the characteristics of the real-world scenarios, and where the annotations can be acquired with an automated process. In this way, it is possible to collect vast quantities of annotated data covering a massive number of different scenarios. Thus, we also consider synthetic datasets like \acrshort{viped} introduced in \ref{ch:virtual-to-real}, providing a high variability of the training data. Hereafter, we briefly describe the adopted datasets, dividing them according to the module for which they are employed. Furthermore, we describe the exploited training procedures, highlighting the changes we made to the architectures to adapt them to our specific scenarios.

\begin{figure}
\centering
\includegraphics[width=.95\linewidth]{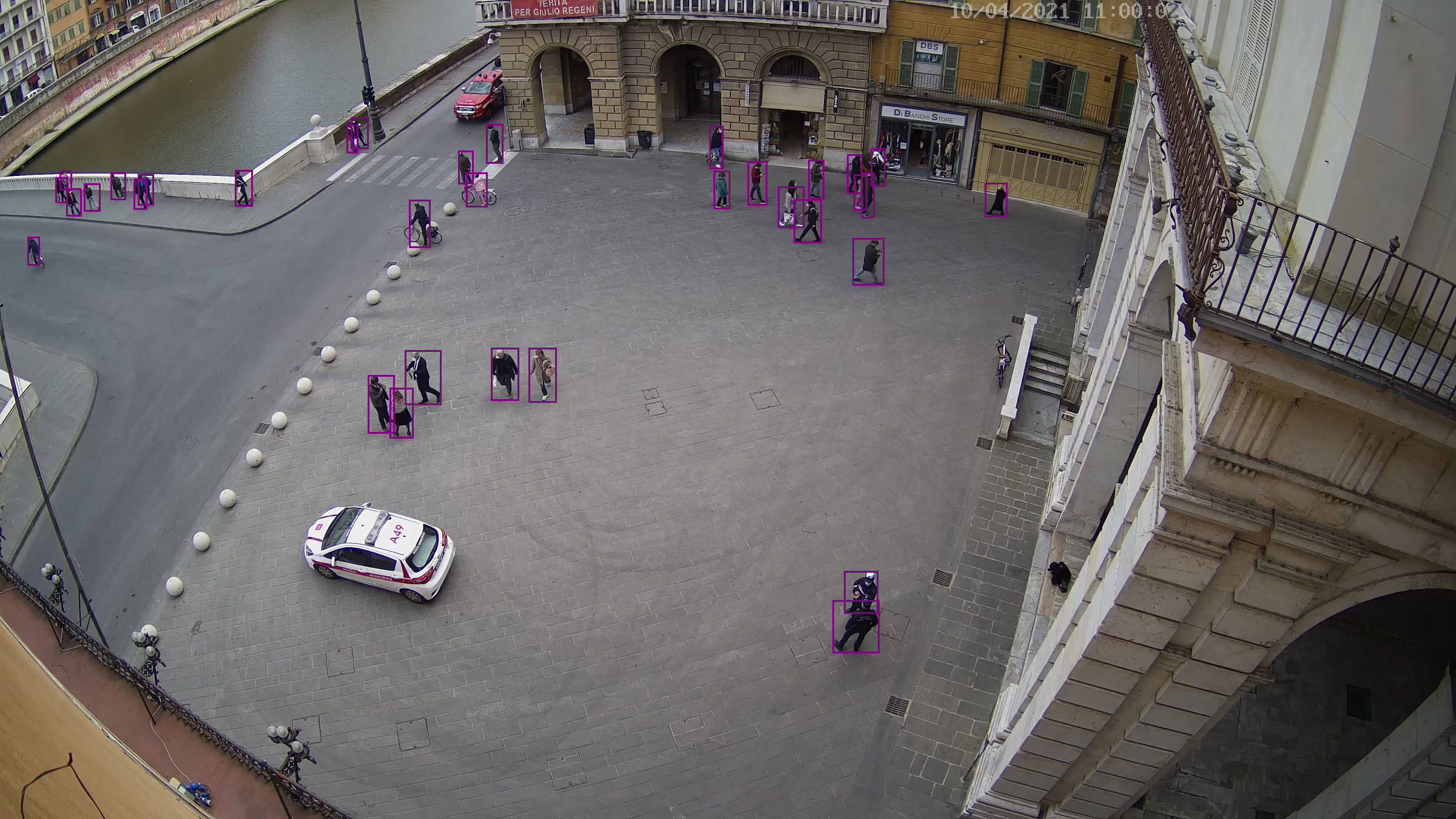}
\caption{\textbf{A sample from our novel CrowdVisorPisa dataset.} We show an image together with bounding box annotations localizing pedestrians.}
\label{fig:crowdvisorpisa_sample}
\end{figure}

\paragraph{Pedestrian Detection}
\label{sec:counting-for-covid:datasets_object_detection}
To train the pedestrian detector module, we use many popular publicly available pedestrian detection datasets: MOT17Det \cite{mot17_dataset}, MOT20Det \cite{mot20_dataset}, CityPersons \cite{citypersons_dataset}, CrowdHuman \cite{crowdhuman_dataset}, PRW \cite{prw_dataset} and CUHK-SYSU \cite{cuhk-sysu_dataset}. These datasets are described in \ref{sec:back:datasets}. We exploit also \acrshort{viped}, a synthetic dataset introduced in \ref{ch:virtual-to-real}. Furthermore, we introduce a novel dataset that we collected and annotated on purpose, named \textit{CrowdVisorPisa}, also employed for evaluating our solution. In particular, it comprises 15 different sequences gathered from a webcam located in a public square of the city of Pisa, Italy, each of which consists of ten images captured with a time interval of 1 second. We manually labeled all frames, localizing the pedestrian instances with bounding boxes. More, we also annotated each sequence taking track of the different pedestrian entities entering or exiting the scene. We divided the dataset into train and test splits, considering 10 and 5 different sequences, respectively. The former split is exploited to train the object detector module, while the latter is used to evaluate the performance of some modules of our framework. It is worth noting that, due to camera positioning not modifiable for local restrictions, this dataset represents a particularly challenging scenario as people instances are small and sometimes difficult to localize. A sample of the dataset is shown in \ref{fig:crowdvisorpisa_sample}.

To make the pedestrian detector able to run efficiently directly on computational- and resource-limited devices, we adopt a strategy similar to the one described in \ref{ch:counting-on-the-edge}, employing, as the backbone of Faster R-CNN, the \textit{ResNet50} architecture, a lighter version of the popular \textit{ResNet101} \cite{resnet}, and reducing the final fully-connected layers responsible for classifying the detected objects since, in our case, we have to localize and identify objects belonging to just one class (i.e., pedestrian). We call \textit{Light} this modified version of the pedestrian detector module to distinguish it from the \textit{Full} original one, having instead the \textit{ResNet101} backbone and a larger number of fully connected layers. To train this module, we adopt the supervised \emph{domain adaptation} strategy described in \ref{ch:virtual-to-real}, where, during the training phase, we mix the synthetic data taken from ViPeD, and the real-world images gathered from the remaining datasets. In this way, we take advantage of the great variability and size of ViPeD, and at the same time, we mitigate the existing domain shift between these synthetic data and the real-world ones. In particular, during the training phase, we exploit batches composed of $2/3$ of synthetic images and $1/3$ of real-world data, thus considering statistics from both domains throughout the entire procedure and where the real-world data acts as a regularization term over the synthetic data training loss.

\paragraph{PPE Detection}
To train and evaluate our \acrshort{ppe} detection module, we collect and annotate a novel dataset (named \textit{CrowdVisorPPE}). It comprises 54,017 images representing pedestrians with and without wearable PPE. Roughly half of the dataset comprises synthetic images procedurally generated using the GTA V video game engine as in~\cite{ppe_detection}, whereas the other half comprises photographic real-world images taken from the Web and manually annotated. The \acrshort{ppe} classes of interest, i.e., helmets, high-visibility vests (HVVs), and face masks, are annotated with bounding boxes. The real-world subset is the only source of face mask instances since they are not available for rendering in GTA V. We hold out a subset of real images as the test split, whereas synthetic images and the remaining real ones form the training split. 

As in the pedestrian detector, we adopt the Faster R-CNN model with the \textit{ResNet50} backbone as the \acrshort{ppe} detector architecture, and we use a similar domain adaptation strategy exploiting synthetic and real data. The only difference concerning the pedestrian detector is that we perform \acrshort{ppe} detection only on pre-segmented patches containing a single pedestrian instead of searching for \acrshort{ppe} in the full-frame. This simplifies the task for the model and enables us to save computational budget by processing smaller images.

\paragraph{Crowd Counting by Density Estimation}
To train the module responsible for crowd counting by density estimation, we exploit many popular publicly available real-world datasets: ShanghaiTech \cite{multi_column}, UCF-QNRF \cite{composition_loss} and NWPU-Crowd \cite{nwpu_dataset}. Furthermore, also in this case, we exploit also a synthetic dataset, named \acrfull{gcc} \cite{gcc_dataset}. These datasets are described in details in \ref{sec:back:datasets}. To train this module, we adopt the supervised domain adaptation strategy illustrated in \ref{ch:virtual-to-real}, consisting of training the network with the synthetic data and then fine-tuning it exploiting the real-world images. In particular, we set the initial weights of the network layers with values coming from a Gaussian distribution with 0.01 standard deviation. Then, we train the network exploiting the GCC dataset, and, finally, we fine-tune it using the real-world data.

\section{Experiments and Results}
\label{sec:counting-for-covid:experiments}
We evaluate all the modules making up our framework, considering different scenarios and exploiting appropriate metrics depending on the considered task. For all the experiments, we consider the \textit{Light} version of our object detector module since it has shown similar performance compared with the \textit{Full} version, and it is more appropriate to be used in combination with low-cost and computational-limited hardware.

Being our target a deployable monitoring system, we selected the NVIDIA Jetson TX2 embedded device as the hardware host. It is composed of two 64 bit CPUs with two and four cores each, an NVIDIA Pascal GPU with 256 CUDA cores, 8 GB of RAM shared between the system and the graphics accelerator, and a 32 GB solid-state storage volume.
The operating system is the NVIDIA's Linux4Tegra (L4T) distribution based on Ubuntu. We installed Python 3.8 with OpenCV 4.5 and the deep learning framework PyTorch 1.8. As detailed in \ref{tab:memory_usage}, memory usage is kept within 5 GB of both system and GPU RAM. An external USB camera completes the whole installation.

\begin{table}
\centering
\caption{\textbf{System and GPU Memory Usage in GB.} \textbf{PD} = pedestrian detector model type; \textbf{DC} = whether the density-based counter module is active; \textbf{PPE} = whether the PPE detector module is active.
The modular framework is assumed to always use the object detector in its \emph{Light} or \emph{Full} models, along with the enabled distance measure plug-in that consumes a fixed and negligible (less than 1 MB) amount of memory.
Video stream size is $1173 \times 880$ RGB pixels. System memory is calculated with \texttt{/usr/bin/time -f "\%M"}, GPU memory with \texttt{torch.cuda.max\_memory\_allocated()}.}

\newcommand{\cmark}{\ding{51}}
\newcommand{\xmark}{\ding{55}}
\newcommand{\tbh}[1]{\small \textbf{#1}}
\newcolumntype{C}{>{\centering\arraybackslash}X}

\begin{tabularx}{.5\linewidth}{cccCC}
    \toprule
	  \tbh{PD}
	& \tbh{DC}
	& \tbh{PPE}
	& \tbh{SysRAM}
	& \tbh{GpuRAM} \\
    \midrule
	\multirow{4}{*}{Light}  & \xmark & \xmark & 2.36 &  0.55 \\
	                        & \cmark & \xmark & 2.44 &  0.86 \\
	                        & \xmark & \cmark & 2.35 &  2.10 \\
	                        & \cmark & \cmark & 2.44 &  2.20 \\
    \midrule
	\multirow{4}{*}{Full}   & \xmark & \xmark & 2.51 &  0.62 \\
	                        & \cmark & \xmark & 2.52 &  0.94 \\
	                        & \xmark & \cmark & 2.51 &  2.20 \\
	                        & \cmark & \cmark & 2.51 &  2.30 \\
	\bottomrule
\end{tabularx}
\label{tab:memory_usage}
\end{table}

\subsection{Counting by Instances}
In this setting, we test and evaluate the counting by instance functionality. We consider our \textit{CrowdVisorPisa} dataset and, in particular, the five sequences belonging to the test subset, performing two different sets of experiments over it; the first one involves only the pedestrian detector module, and the second instead takes also into account the tracker module. More in detail, in the first case, we evaluate the effectiveness of our framework to estimate the number of people present in the single frames. On the other hand, in the second scenario, we also consider the temporal relation existing between consecutive images, tracking the found pedestrian instances over time. 

\begin{figure}
\begin{subfigure}{0.49\linewidth}
\centering
Number of detected people\\[2ex]
\includegraphics[width=\linewidth]{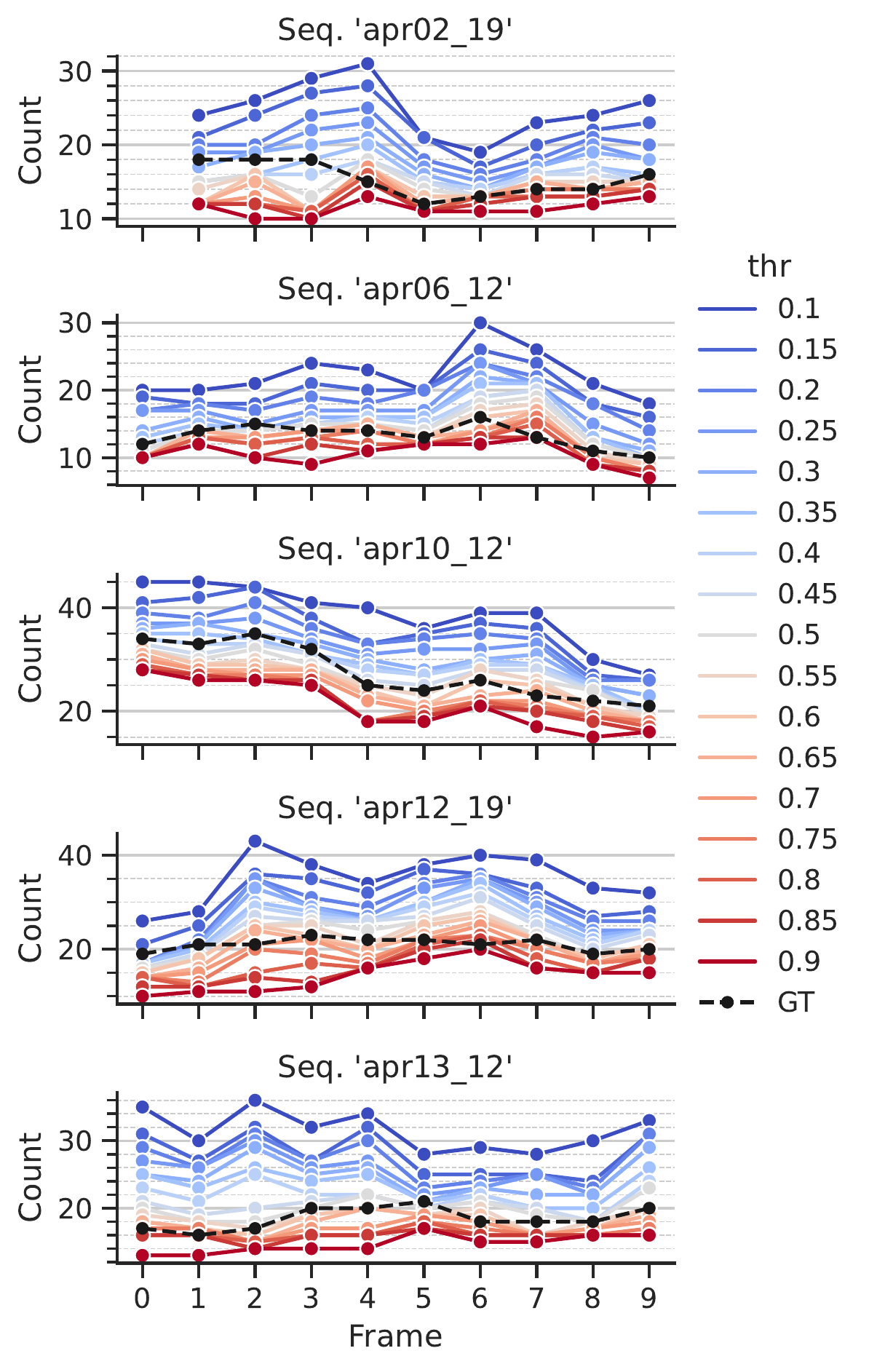}
\end{subfigure}
\hfill
\begin{subfigure}{0.49\linewidth}
\centering
Counting Errors\\[2ex]
\includegraphics[width=\linewidth]{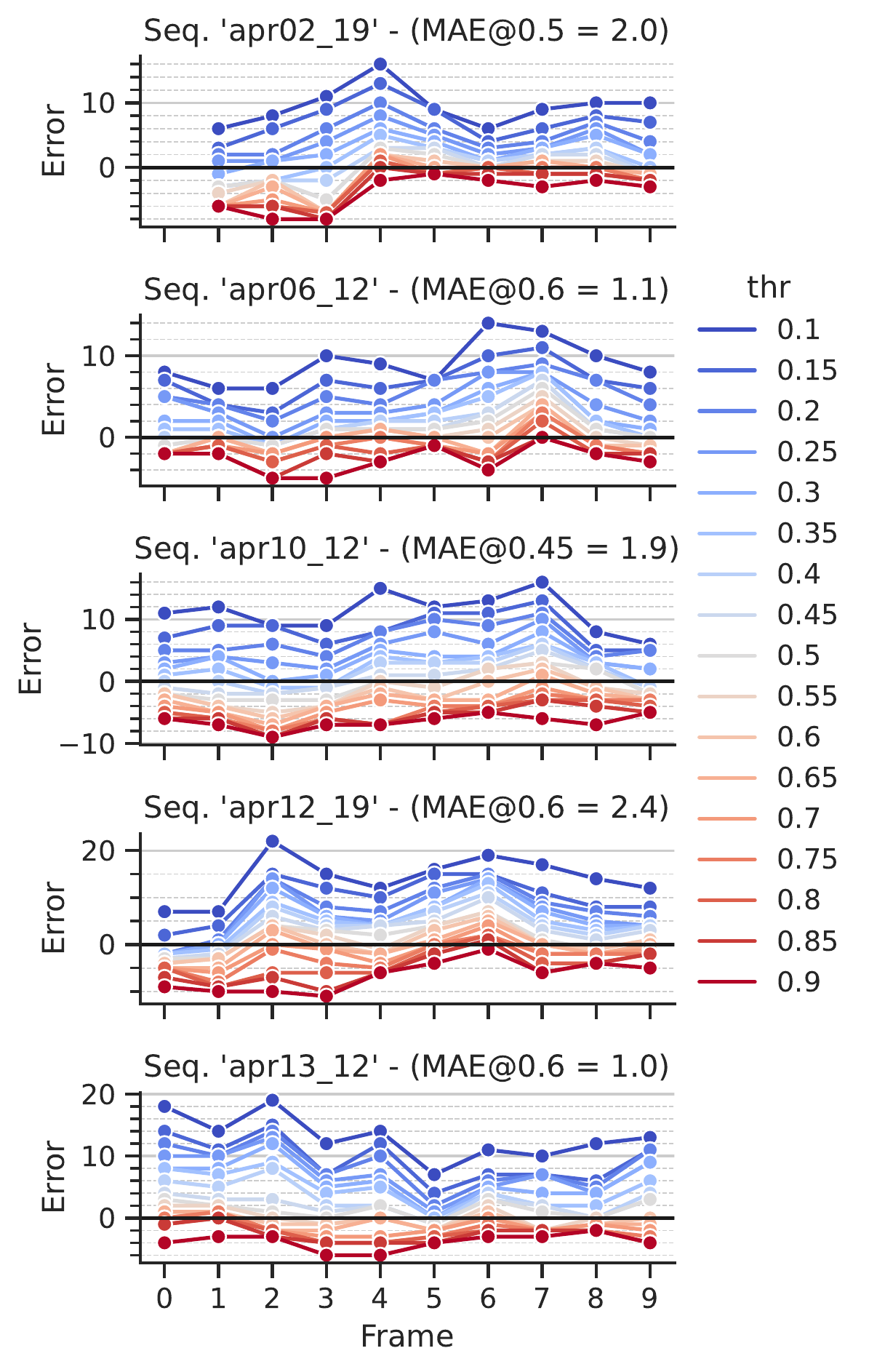}
\end{subfigure}
\caption{\textbf{Evaluation of \emph{counting by instances} functionality of our framework, considering the single still frames of the five test sequences of our \textit{CrowdVisorPisa} dataset.}
In the first column, we report the number of people located by our detector, varying the detection thresholds. The black line (GT) indicates the actual number of pedestrians in the frame.
The second column shows the counting errors and the best MAE obtained with a specific detection threshold.}
\label{fig:sequences-count}
\end{figure}

We report in Figure \ref{fig:sequences-count} the obtained results concerning the first scenario. Each row of the figure represents a different sequence. The first column shows the number of people that our detector module can localize for each frame comprising a sequence, varying the detection threshold. On the other hand, in the second column, we illustrate the errors in terms of counting. We also report, for each sequence, the best Mean Absolute Error (MAE), i.e., the mean of the sum of the absolute errors, obtained with a specific threshold. As can be seen, we get an MAE close to 1 or 2, depending on the considered scenario, demonstrating that the module provides a reliable estimation of the number of pedestrians present in the monitored scene. The optimal threshold may vary depending on the scenario and the desired behavior, e.g., the user may prefer under- or over-estimation in case of errors. Due to the empirical nature of its choice, in our system, we provide the dynamical configuration of several parameters, including detection thresholds.

\begin{figure}
\centering
\includegraphics[width=\linewidth]{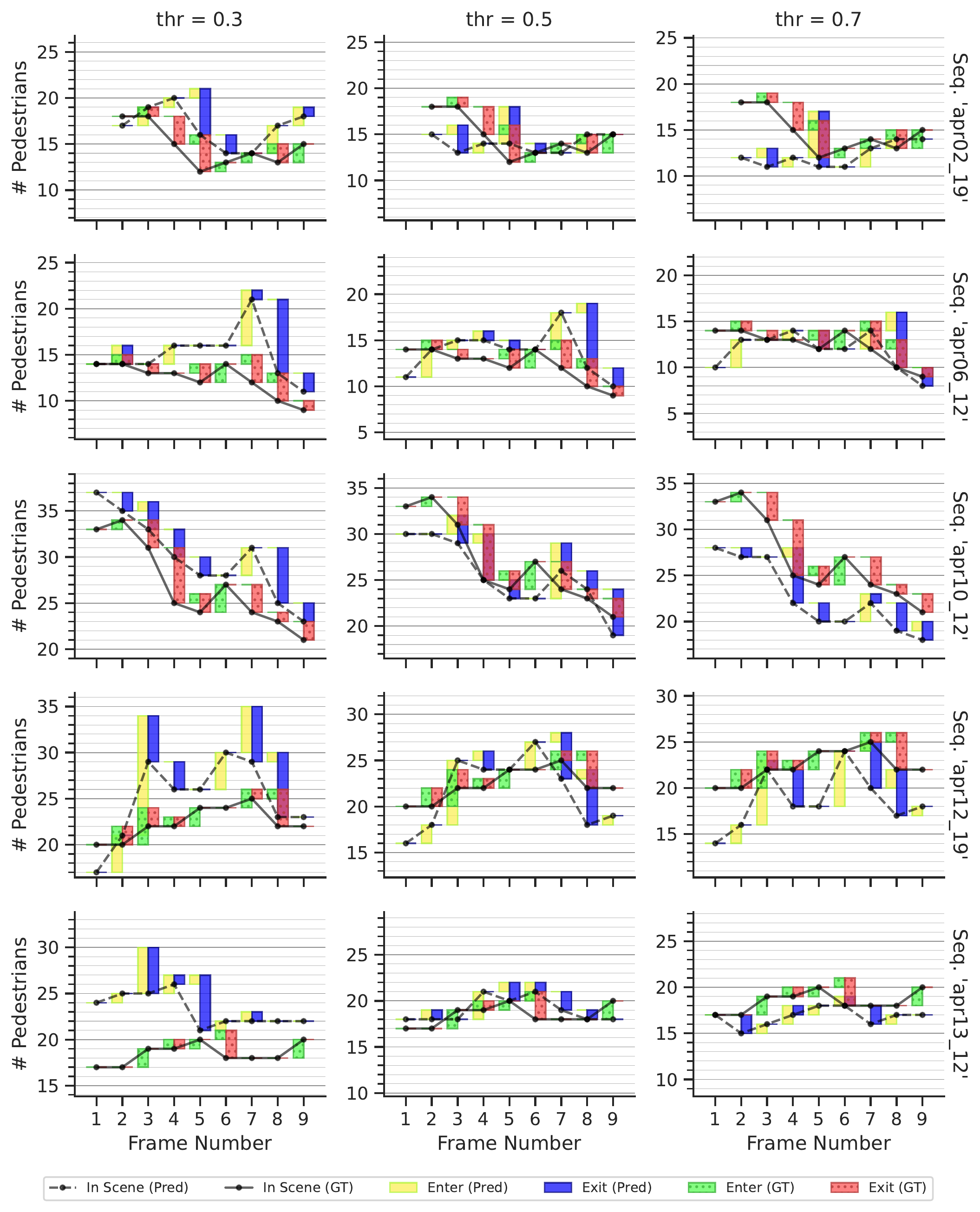}
\caption{\textbf{Evaluation of the counting by instances functionality of our framework, taking into account also the tracker module.} We considered the five test sequences of our \textit{CrowdVisorPisa} dataset reported one for each row. Columns represent the results obtained for three different detection thresholds. For each plot, we show the pedestrians that enter and exit from the scene, relying on the tracklets describing the recognized identities of the people instances over time.}
\label{fig:track-error}
\end{figure}

On the other hand, in Figure \ref{fig:track-error}, we show the results concerning the second scenario. Each row of the figure corresponds to a different sequence. We report the results about the single frames making up a sequence for three different detection thresholds, one for each column. In particular, we indicate the pedestrians that enter and exit from the scene at each frame, exploiting the tracklets provided by the tracker module that represents the recognized identities of the people instances over time.
Solid lines represent the actual number of people present in a frame over time, and green/red candles represent the number of people entering/exiting the scene.
Similarly, dashed lines represent the number of people predicted by our system, and yellow/blue candles represent the estimated number of people entering/exiting the scene as predicted by the tracker module.
We notice that with a low (resp. high) threshold value, our system tends to overestimate (resp. underestimate) the total number of people present in a sequence, thus finding its optimal threshold values in the 0.5 - 0.6 range.
We also note that false-positive detections tend to create spurious peaks in the people count. However, they often recovered in the immediately following frame.

\subsection{Counting by Density Estimation}
We validate the counting by density estimation, performed by the Density Counter module, by exploiting several test subsets belonging to publicly available datasets (described in \ref{sec:counting-for-covid:datasets}). This choice has been driven by the fact that our \textit{CrowdVisorPisa} dataset does not provide the needed features, i.e., the scenarios are not highly crowded. Given that the annotation procedure for labeling datasets having these characteristics is highly costly in terms of manual human effort, we exploited some already existing collections of annotated images available in the literature.

\begin{figure}
\begin{subfigure}{0.33\linewidth}
\centering Image
\includegraphics[width=\linewidth,height=2.9cm]{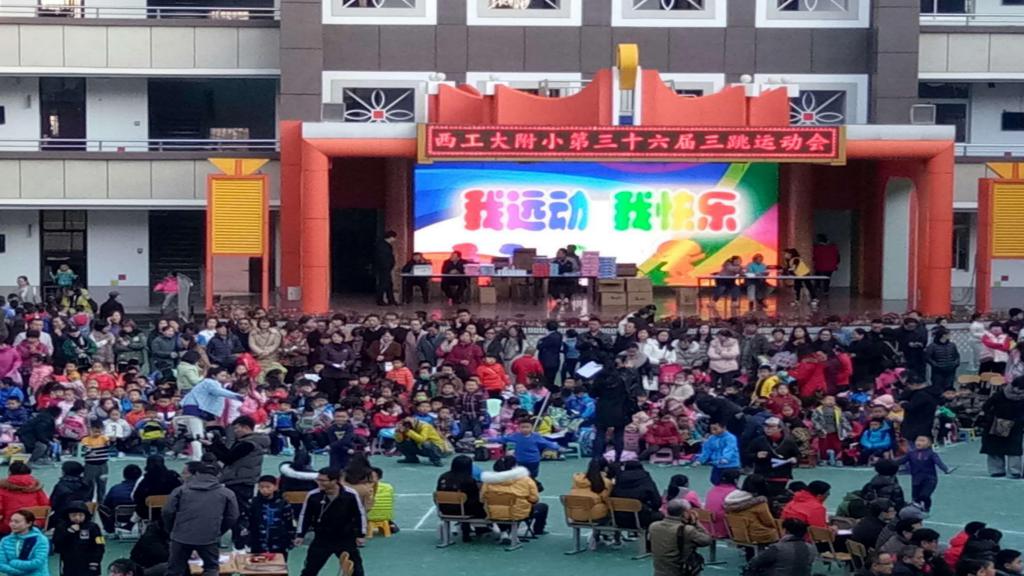}
\includegraphics[width=\linewidth,height=2.9cm]{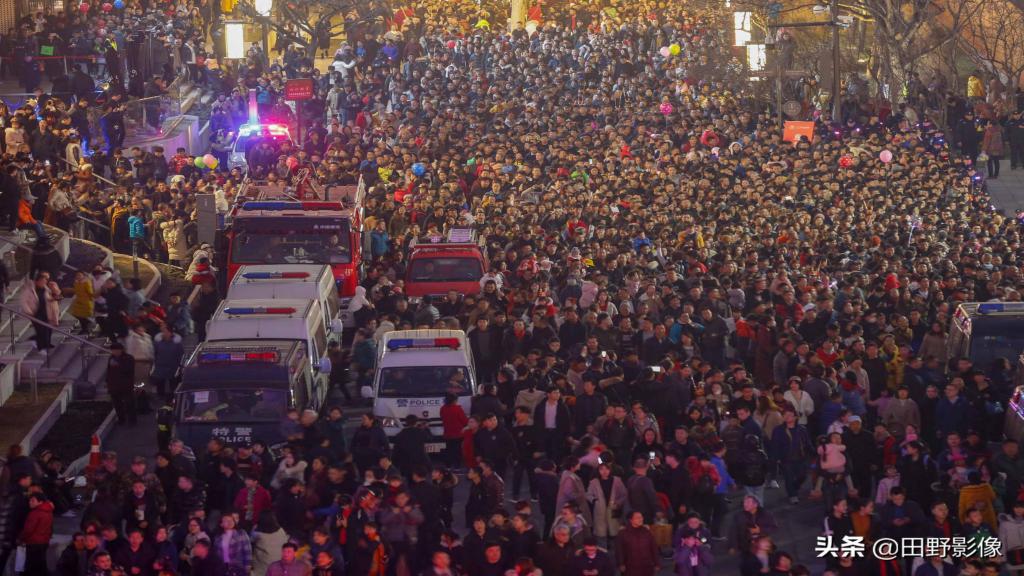}
\includegraphics[width=\linewidth,height=2.9cm]{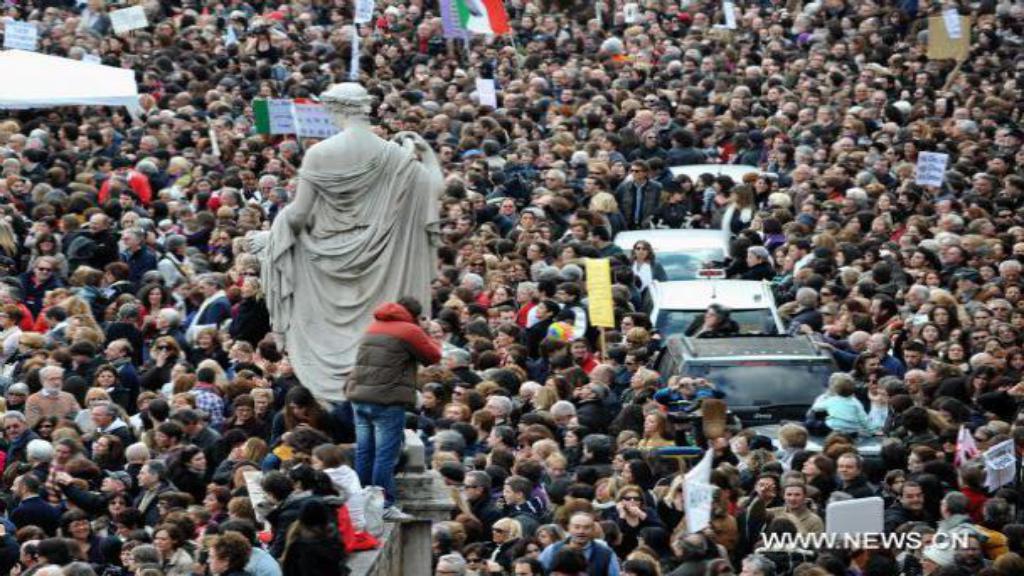}
\end{subfigure}%
\hfill%
\begin{subfigure}{0.33\linewidth}
\centering Groundtruth Density Map
\includegraphics[width=\linewidth,height=2.9cm]{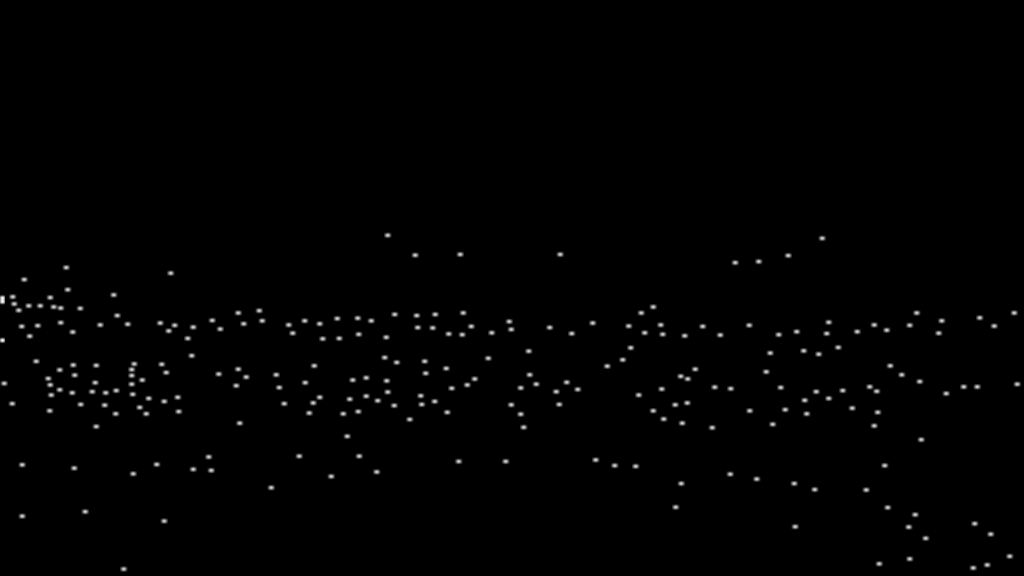}
\includegraphics[width=\linewidth,height=2.9cm]{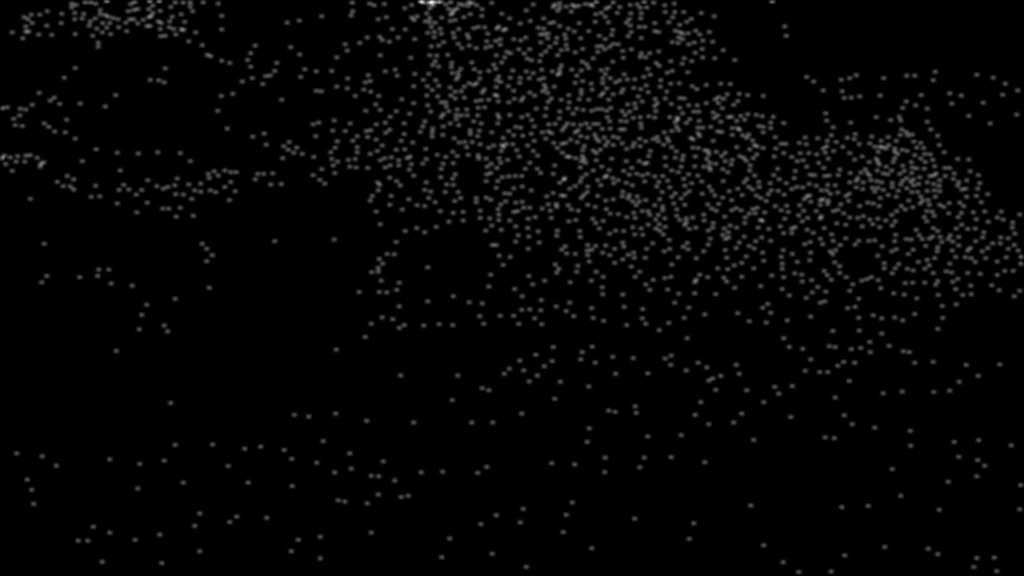}
\includegraphics[width=\linewidth,height=2.9cm]{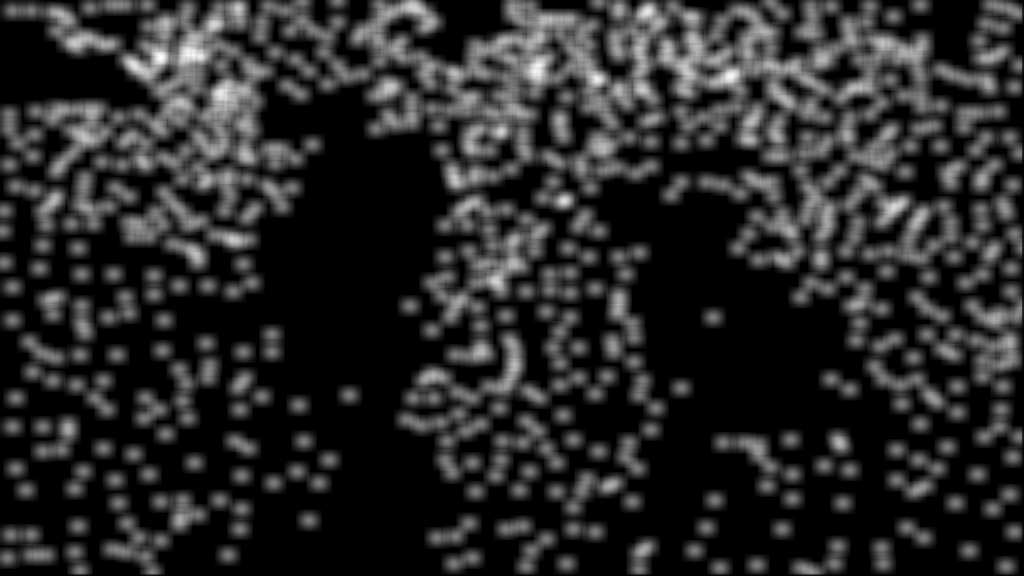}
\end{subfigure}%
\hfill%
\begin{subfigure}{0.33\linewidth}
\centering Predicted Density Map
\includegraphics[width=\linewidth,height=2.9cm]{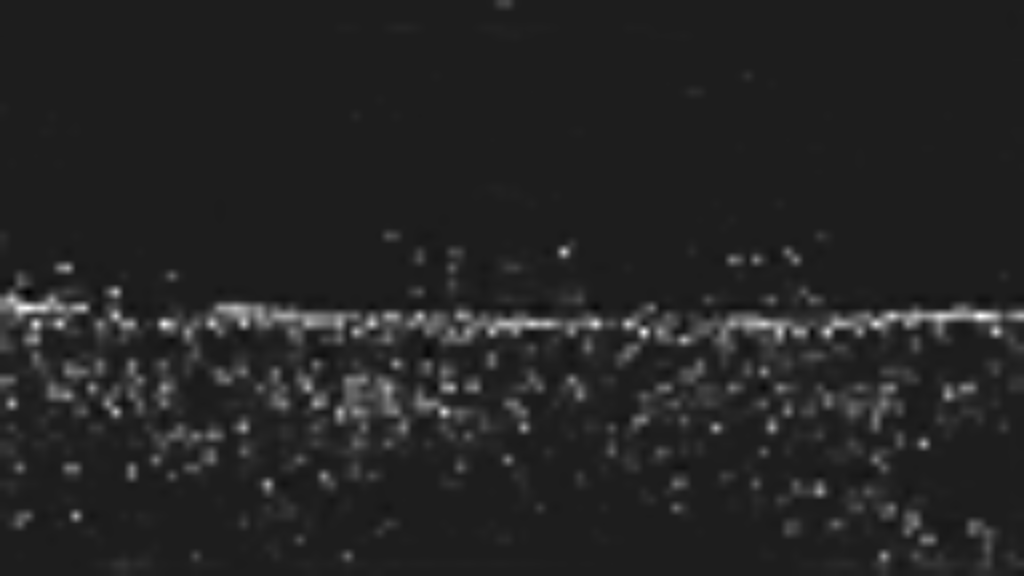}
\includegraphics[width=\linewidth,height=2.9cm]{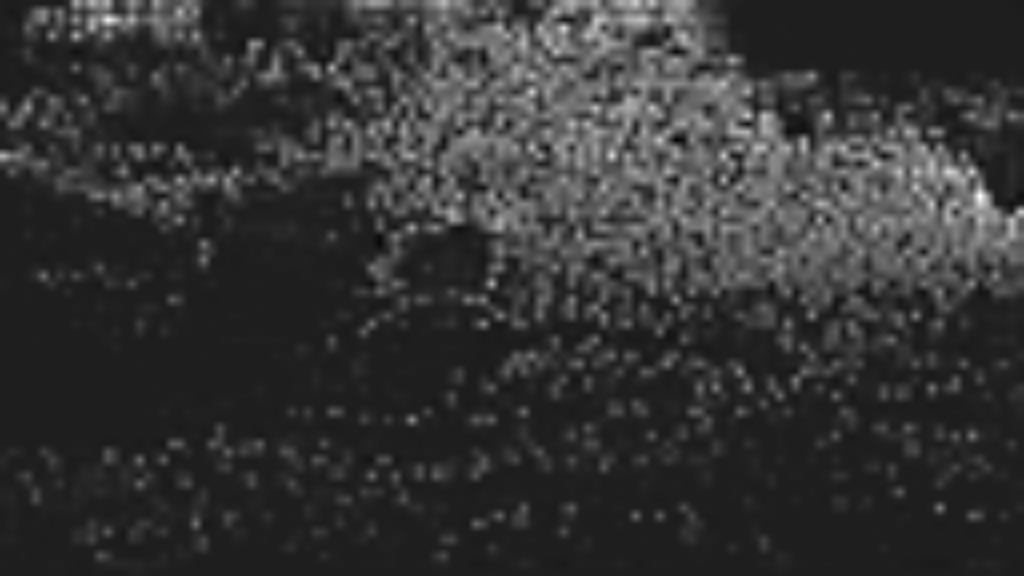}
\includegraphics[width=\linewidth,height=2.9cm]{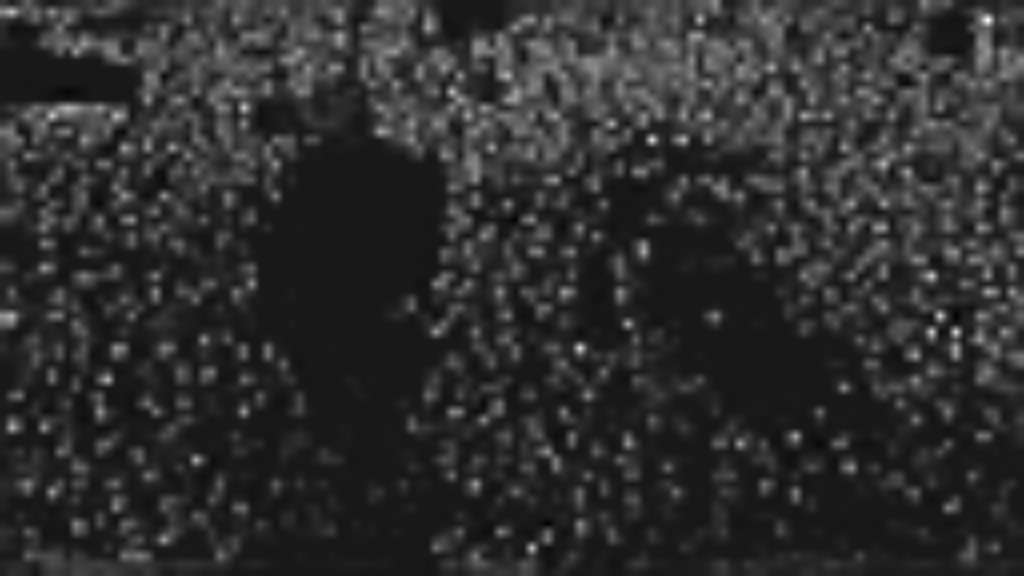}
\end{subfigure}
\caption{\textbf{Some samples of the test subsets exploited for the evaluation of our \emph{counting by density} estimation functionality.} We show some images together with the ground truth and the predicted density maps. Integrating these density maps, i.e., summing up the pixel values, we can obtain an estimation of the people present in the image.}
\label{fig:crowd_counter_density_output_examples} 
\end{figure}

We report in \ref{tab:modules-metrics} the obtained results in terms of \acrshort{mae} and \acrfull{rmse}. It is worth noting that, as a result of the squaring operation, \acrshort{rmse} effectively penalizes large errors more heavily than small ones, thus more suitable when outliers are particularly undesirable. Furthermore, we also compute the \acrfull{ssim} \cite{ssim} to measure the density map quality, which measures images' similarities under three aspects: brightness, contrast, and structure. The value of \acrshort{ssim} is in the $[0, 1]$ range: the larger it is, the less distortion of the image is measured (see also \ref{sec:back:visual-counting:metrics} for further details). Finally, in \ref{fig:crowd_counter_density_output_examples} we show some examples of the considered images, together with the ground truth and the predicted density maps.

\begin{table}
\caption{\textbf{Evaluation of the Pedestrian Detection (\textbf{PD}), Density-based Counter (\textbf{DC}), and \textbf{PPE} Detector modules, measured on the corresponding test sets.} The mean Average Precision (mAP) measures the average precision of the detection when varying the score threshold in detection-based modules (\textbf{PD}, \textbf{PPE}). For \textbf{DC}, MAE and RMSE measure the counting error, while SSIM measures the quality of the predicted density map.}
\centering
\begin{tabular}{ccccc}
\toprule
\textbf{PD} & \textbf{PPE} & \multicolumn{3}{c}{\textbf{DC}} \\
\cmidrule(lr){1-1} \cmidrule(lr){2-2} \cmidrule(lr){3-5}
mAP $\uparrow$ & mAP $\uparrow$ & MAE $\downarrow$ & RMSE $\downarrow$ & SSIM $\uparrow$ \\
\midrule
0.836   & 0.606 &  92.28 &  365.4 &  0.79 \\
\bottomrule
\end{tabular}
\label{tab:modules-metrics}
\end{table}

\subsection{Detecting Pedestrians and Personal Protection Equipment}
Here, we validate the detection of pedestrian and worn PPE, performed respectively by the Pedestrian Detector and the Personal Protection Equipment Detector modules. For the former, we focus on the five test sequences of our CrowdVisorPisa dataset, whereas for the latter, we consider the CrowdVisorPPE test subset.

\begin{figure}
\includegraphics[height=5.35cm]{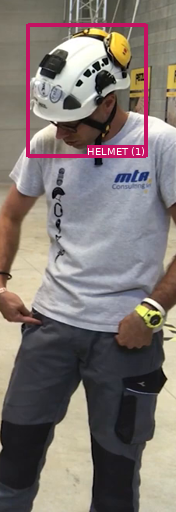}\hfill
\includegraphics[height=5.35cm]{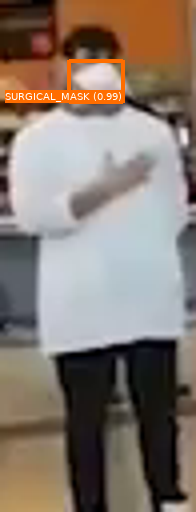}\hfill
\includegraphics[height=5.35cm]{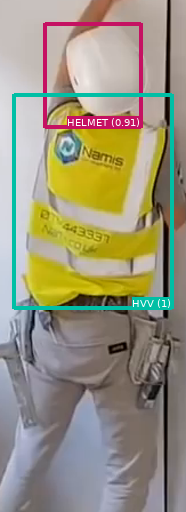}\hfill
\includegraphics[height=5.35cm]{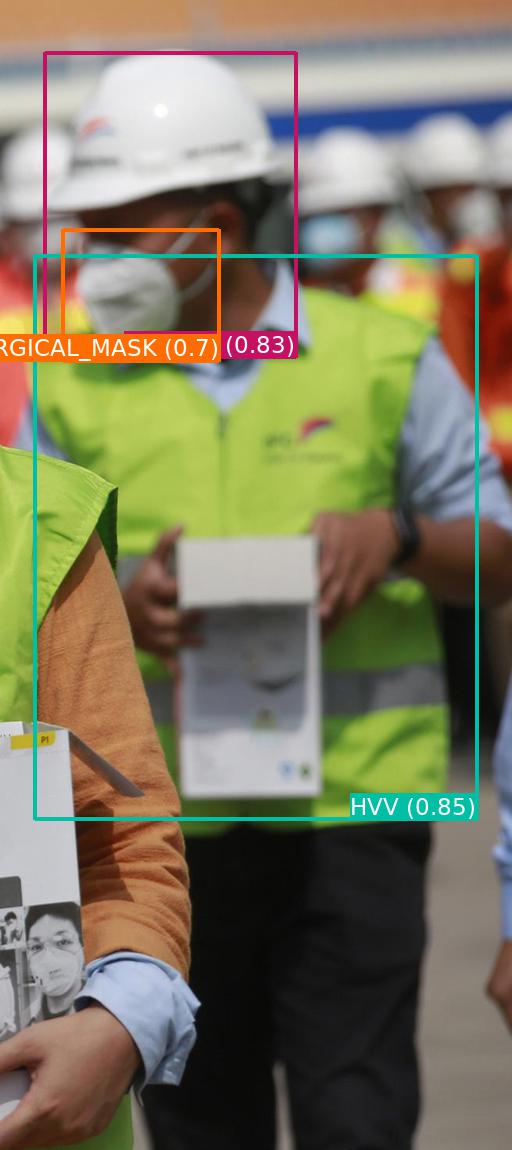}\hfill
\includegraphics[height=5.35cm]{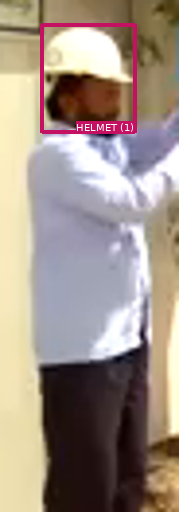}\hfill
\includegraphics[height=5.35cm]{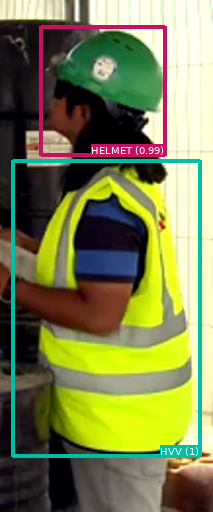}\hfill
\includegraphics[height=5.35cm]{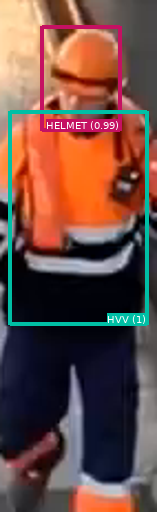}
\caption{\textbf{Examples of predictions of the PPE Detection module on our CrowdVisorPPE test set.} PPE classes are color coded (\colorbox{helmet}{\textsf{helmet}}, \colorbox{hvv}{\textsf{high-visibility vest}}, \colorbox{mask}{\textsf{face mask}}), and the detection score is reported in parenthesis.}
\label{fig:ppe-predictions}
\end{figure}

We evaluate the two modules using the \acrfull{map}, a popular metric in measuring the accuracy of object detectors that computes the average precision value for recall values spanning 0 to 1 (see also \ref{sec:back:cnn-based-detectors:metrics} for further details). We prefer the mAP over other threshold-dependent metrics, such as True Positive Rate, False Positive Rate, or F1-score. Indeed, threshold-dependent metrics are scenario- and application-specific; end-users may accept different levels of false positive or negative rates and may decide to tune thresholds differently depending on their needs.
On the other hand, mAP provides a unique metric summarizing the performance at multiple operational points. We report the obtained results in \ref{tab:modules-metrics}, showing that our modules can reach a \acrshort{map} of 0.836 and 0.606 for the pedestrian detection and the \acrshort{ppe} detection tasks, respectively. \ref{fig:ppe-predictions} shows some predictions of \acrshort{ppe} detections on the CrowdVisorPPE test set.

\subsection{Measuring Interpersonal Distances}
To establish a correspondence between the acquired image and a planar \emph{metric} surface onto which objects (i.e., pedestrians) positions can be evaluated, we used a known-sized manhole in the monitored scene to calculate the homography. The computed matrix is exploited to unwarp pixel position into real-world relative locations. The homography reprojection is a closed-form mathematical process, so no previous training is needed. 

\begin{figure}
\centering
\includegraphics[height=4cm, width=6.5cm]{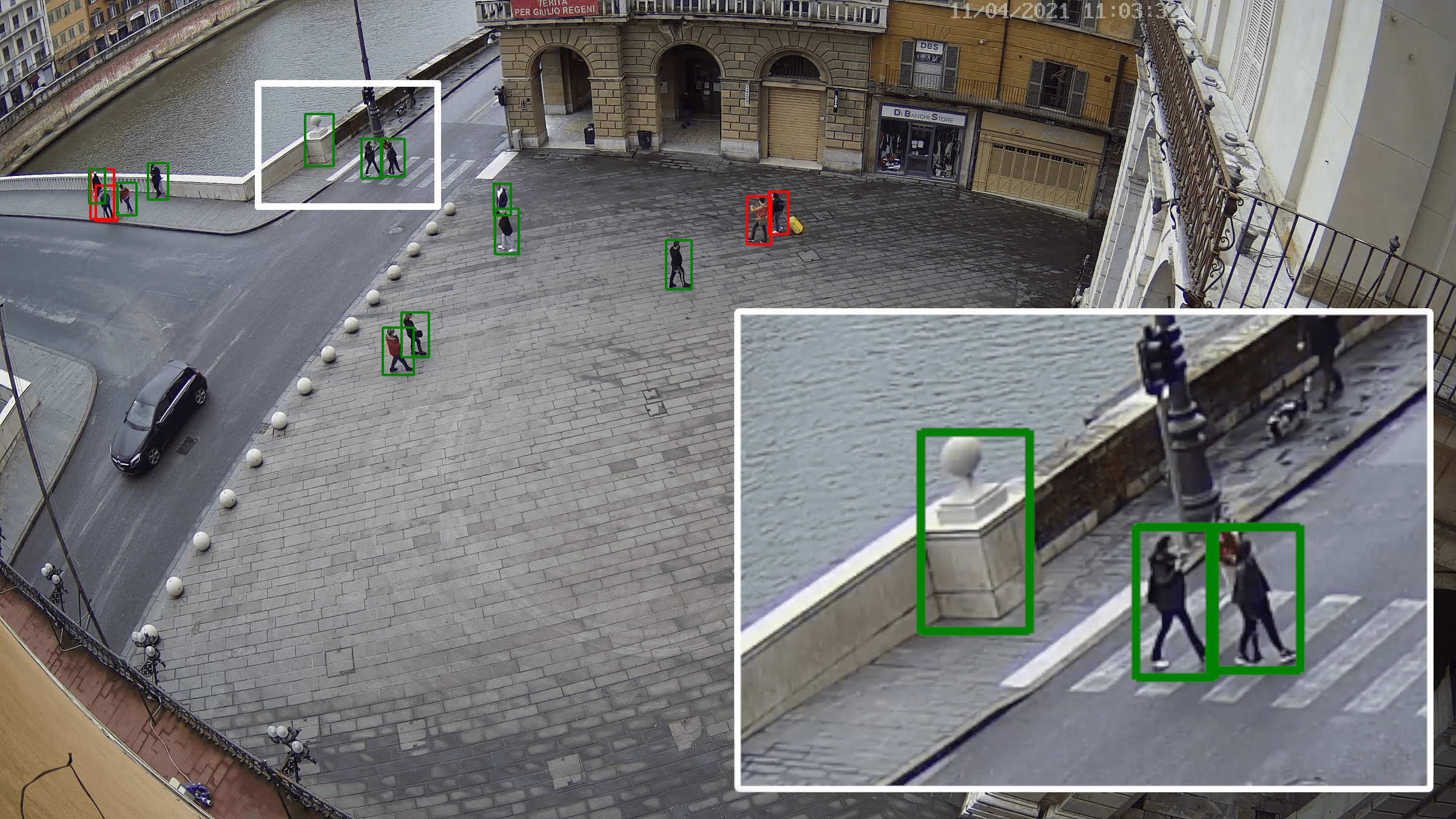}
\includegraphics[height=4cm]{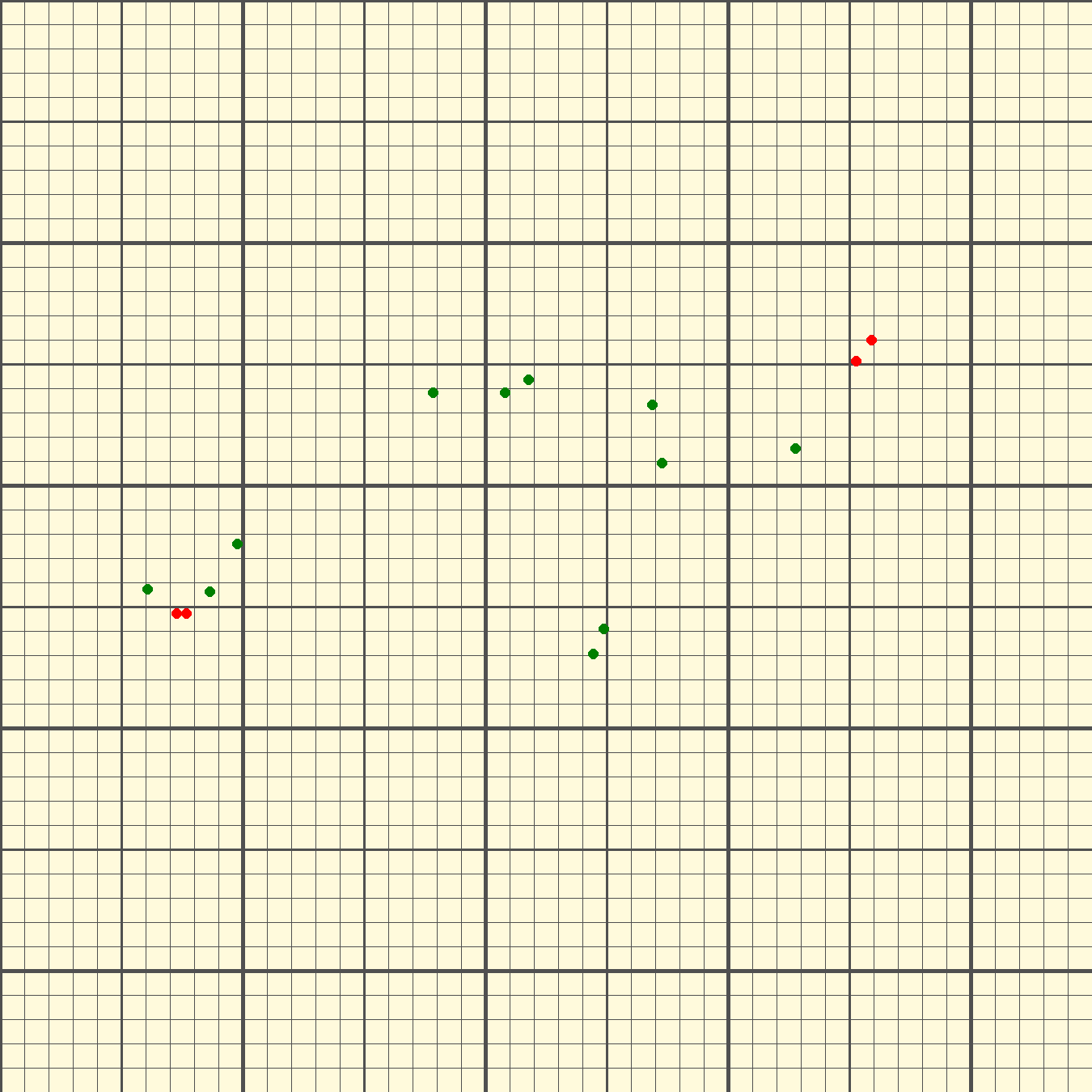}\\[1ex]
\includegraphics[height=4cm, width=6.5cm]{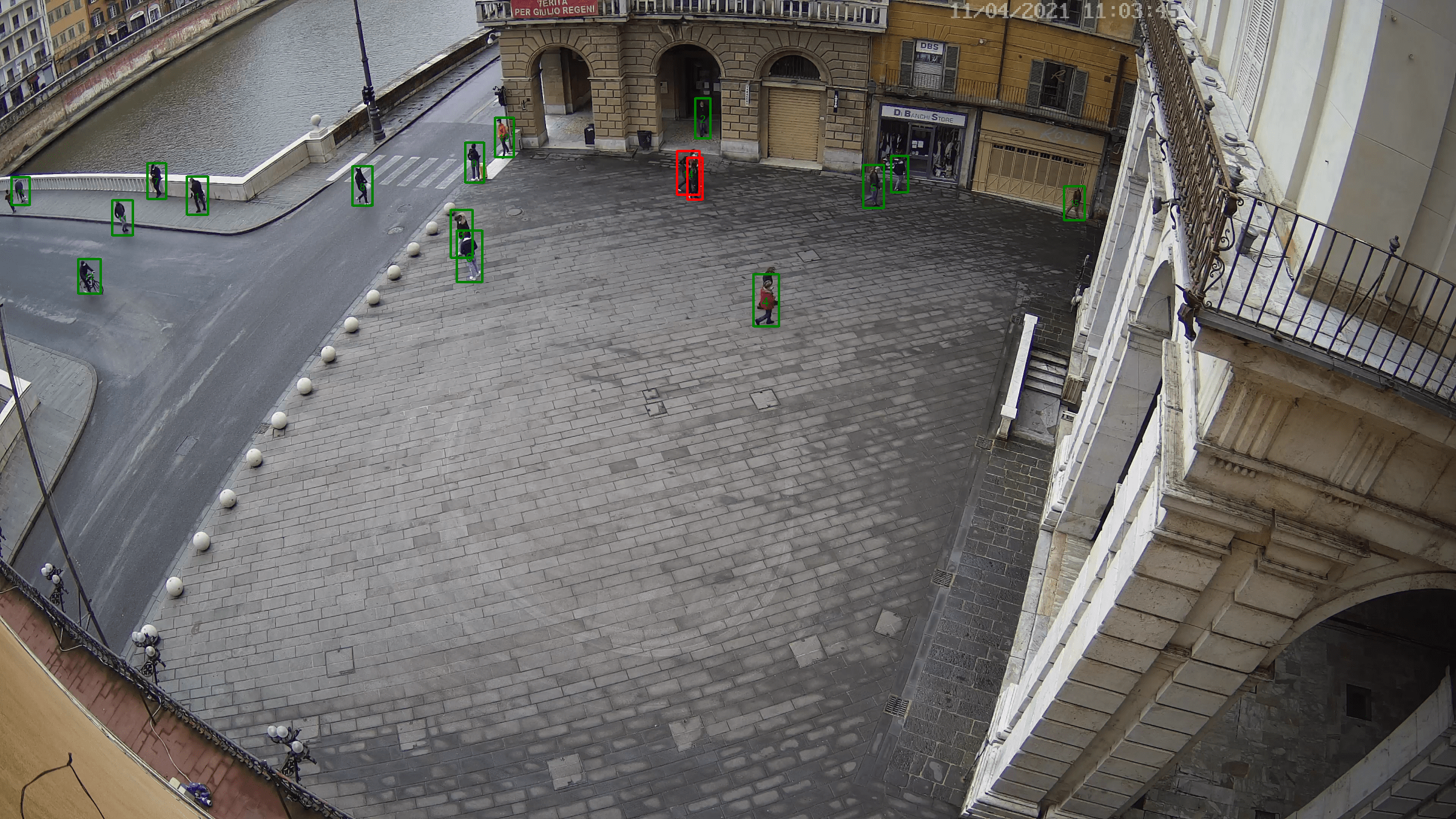}
\includegraphics[height=4cm]{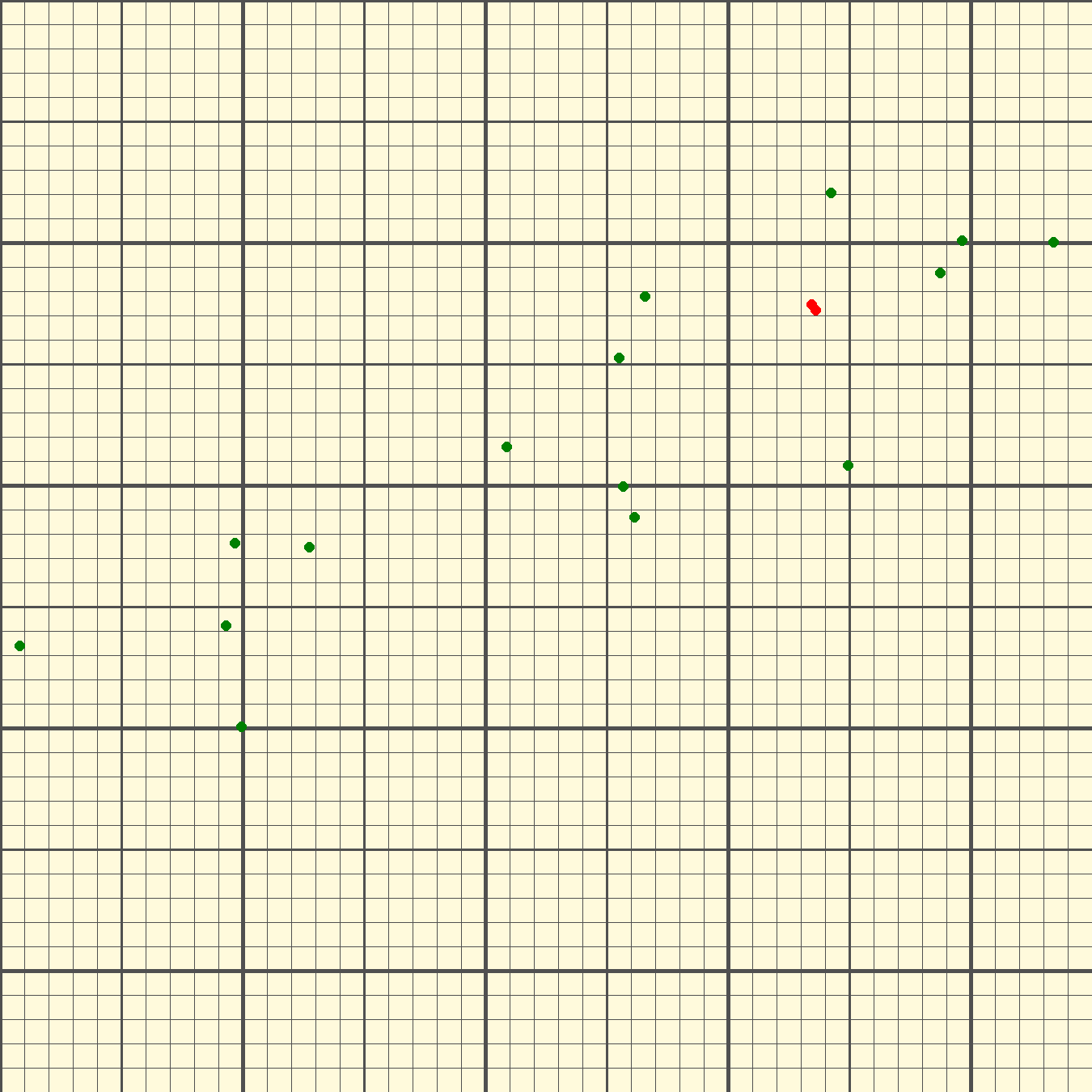}\\[1ex]
\includegraphics[height=4cm, width=6.5cm]{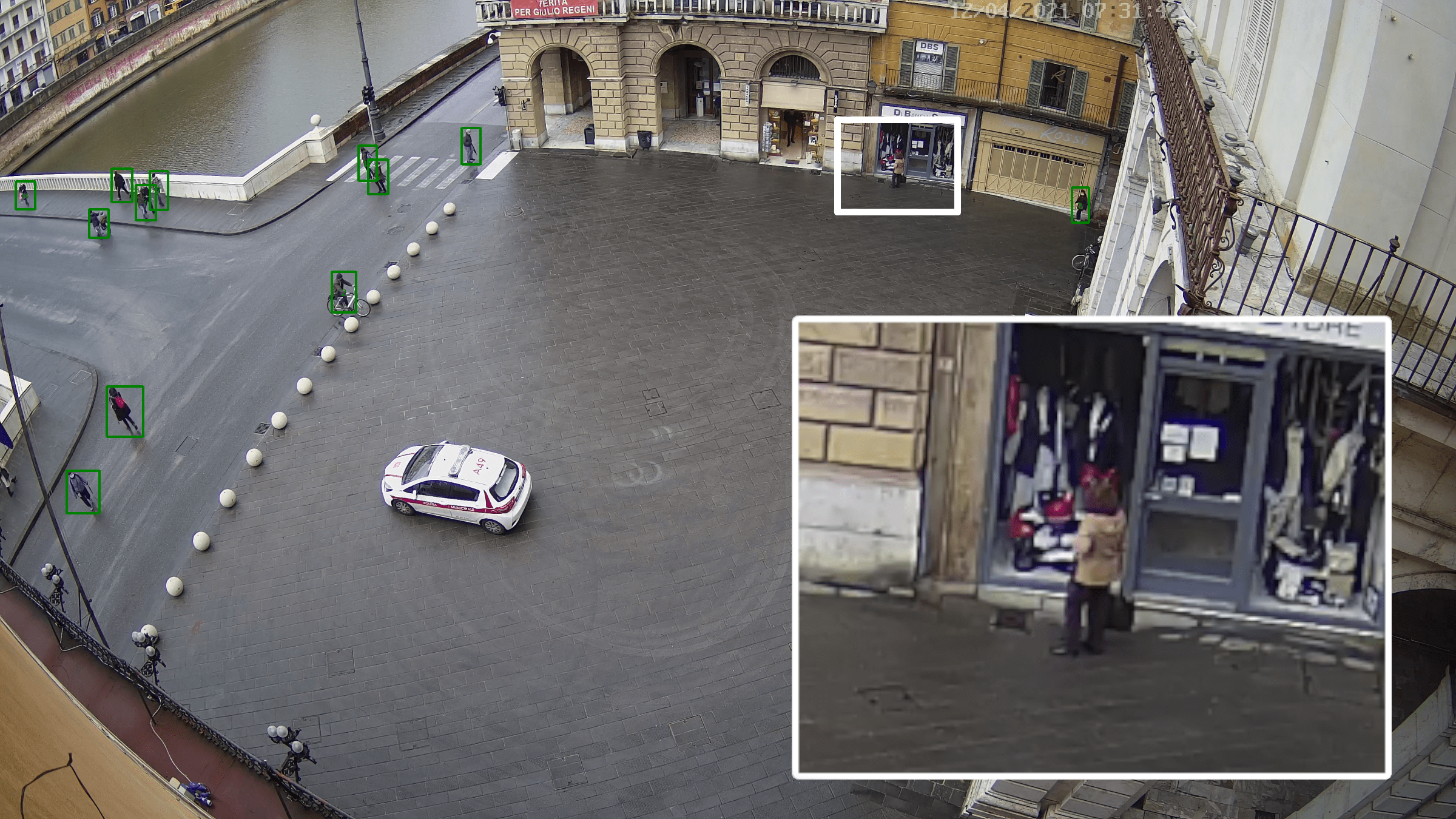}
\includegraphics[height=4cm]{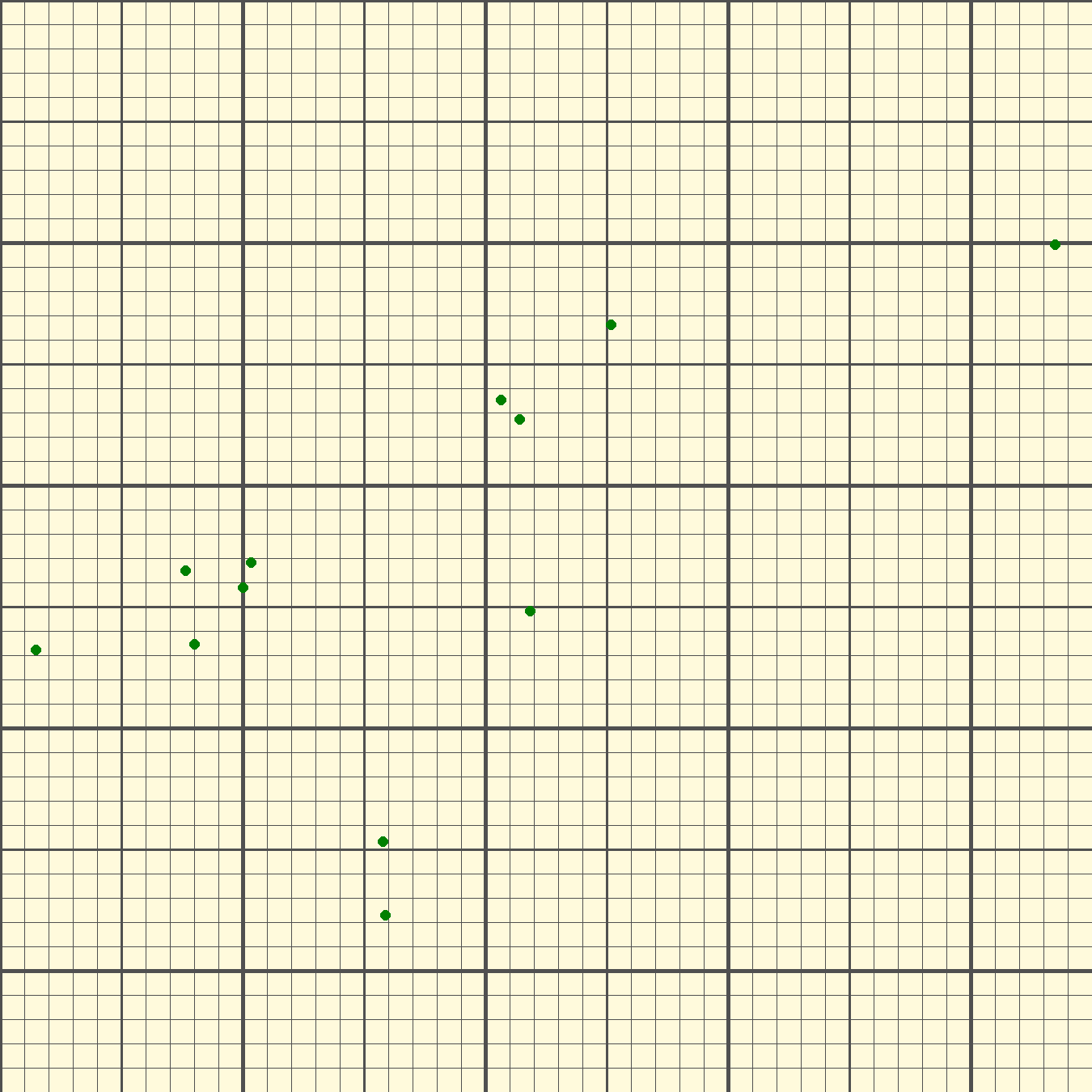}\\[1ex]
\includegraphics[height=4cm, width=6.5cm]{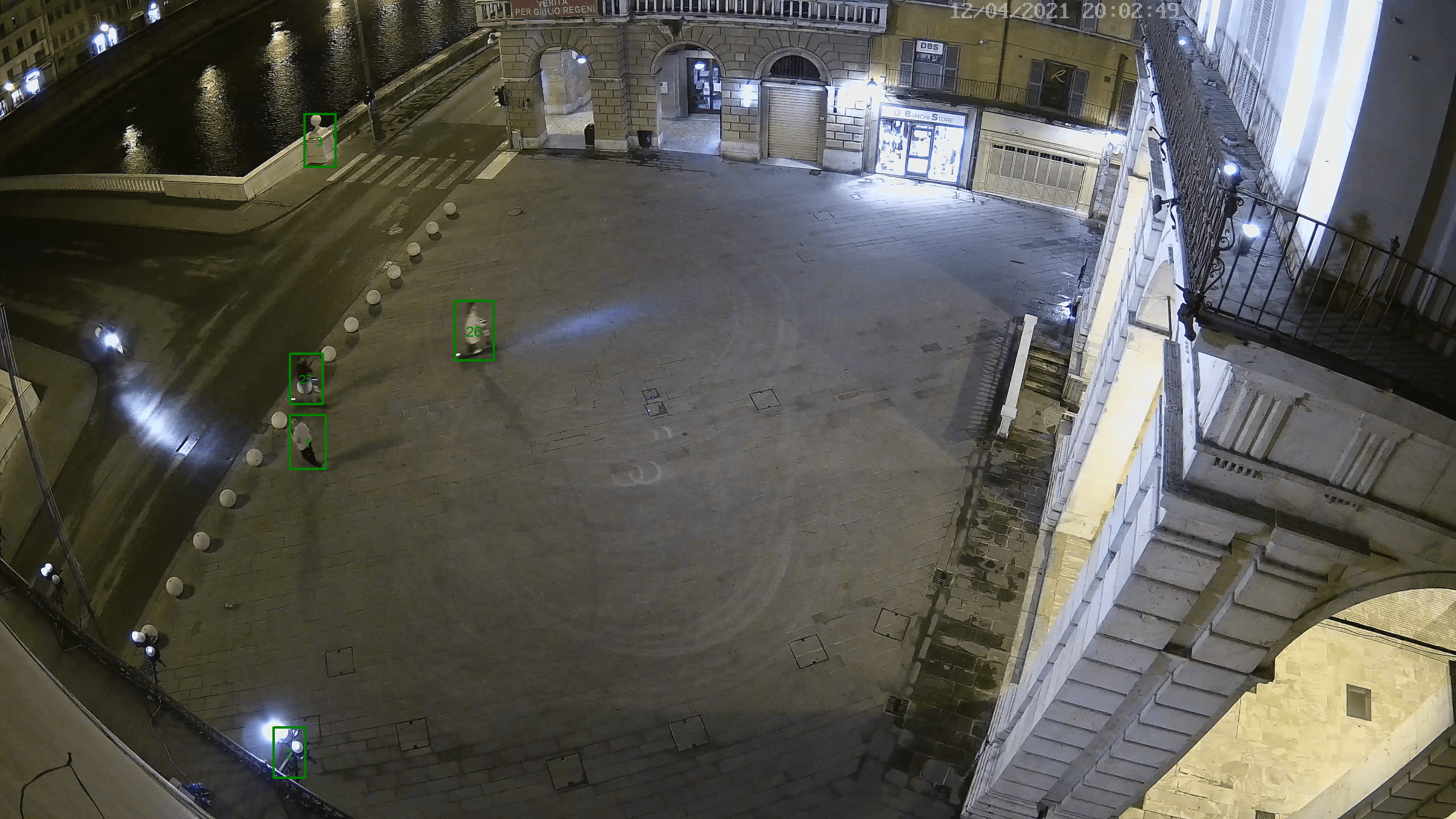}
\includegraphics[height=4cm]{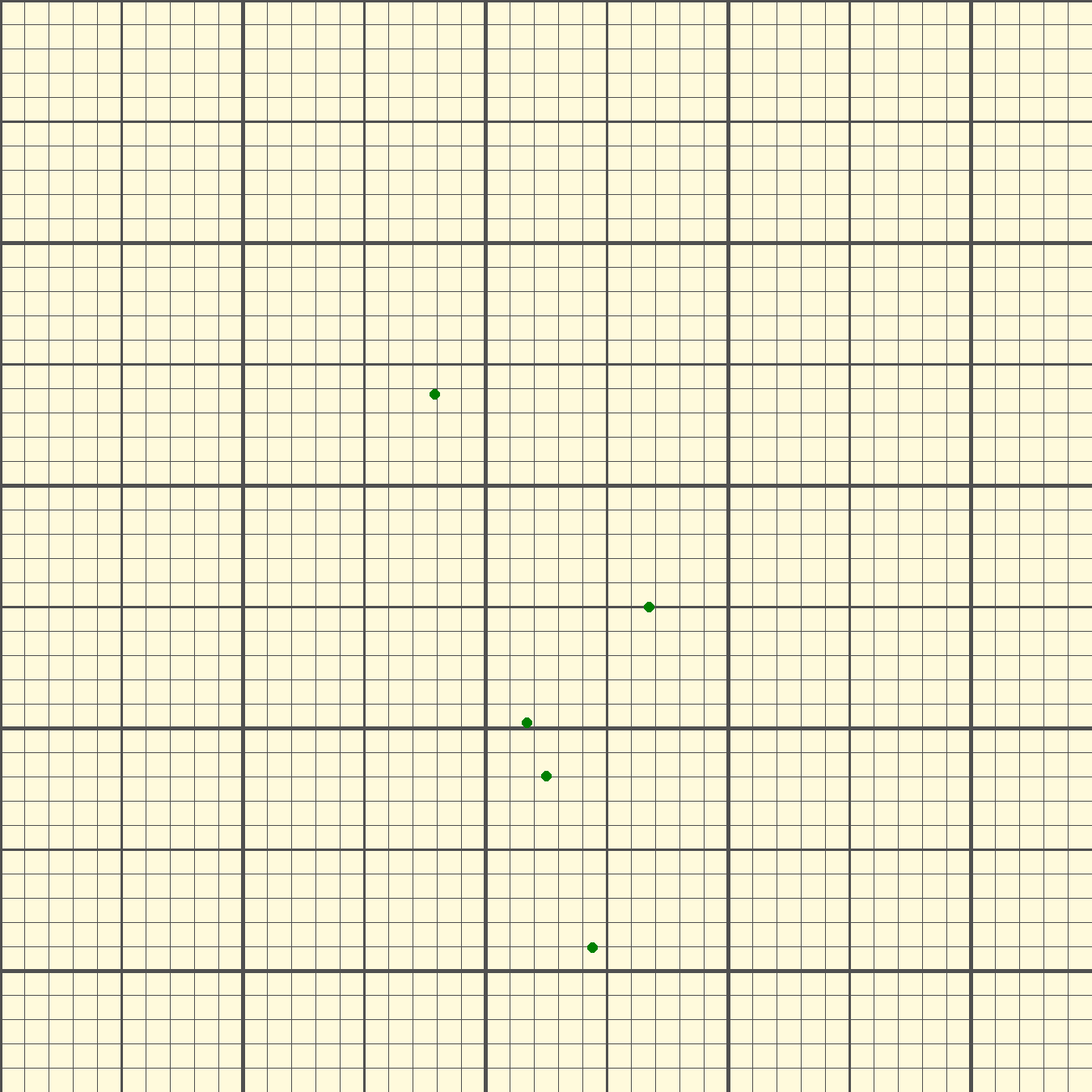}
\caption{\textbf{Examples of detections and distance warnings under different lighting conditions.} Each of the eight images represents, on its left side, pedestrians detected and tracked in the example scenario, while showing on its right side their 2D projection on a virtual planar surface obtained through homography, with a reference 1-meter-spacing overlay grid. The green color means a safe placement, and the red indicates violations of the 1-meter physical distance rule. Some failure examples are outlined in white and zoomed.}
\label{fig:det_proj}
\end{figure}

Our measuring results can be seen in Figure \ref{fig:det_proj}: in the examples shown, the precision of measurements is relative to both the initial calibration (i.e., manhole real size mapped to its projection on the input image) and the accuracy of the pedestrian bounding boxes (i.e., rectangles) predicted by the object detector. We measured the manhole with an upper bound precision of 1 cm and a pixel area error of about 3 cm, thus confining the overall measurements below the 10 cm error. For pedestrian positions, we used the midpoint of the lower edge of its predicted bounding box. As can be seen from the gridded unprojection of the examples, results are consistent within the above error gap.

\section{Summary}
\label{sec:counting-for-covid:conclusion}
In this chapter, we presented, from a more practical perspective, a modular framework based on Computer Vision and \acrshort{ai} technologies, deployed in a real use-case scenario on a low-cost off-the-shelf embedded platform and aimed at monitoring human activities in critical conditions. Our goal was to provide a system having the peculiarity to be \textit{expandable} in the future, simply adding new modules in charge of performing new functionalities that the user can easily enable (or disable) according to their needs. In this way, our framework turns out to be \textit{flexible} since the tasks to be accomplished can change in time, and it is always possible to insert a new module responsible for performing it.
As an effective setup, we implemented a set of visual-based modules for pedestrian detection, tracking, aggregation counting (based on instances or density maps), social distancing calculations, and personal protection environment detection. Specifically, we exploited some of the techniques illustrated in the previous chapters, training artificial neural models with publicly available and, for the purpose of the physical device installation, custom datasets; at the same time, we applied a transfer learning approach to expand detection capabilities by using computer-generated training imagery. To test the effectiveness of our solution, we monitored a known place in Italy during the restrictions imposed during the COVID-19 pandemic, proving satisfactory accuracy in terms of detection, counting, and physical distance measurements. 


\chapter{Conclusions}
\label{ch:conclusion}

In this thesis, we investigated and enhanced the adoption of \acrlong{dl}-based solutions for the visual counting task that automatically estimates the number of objects in still images or video frames. In particular:
\begin{itemize}
\item we tackled and proposed mitigations to the problem related to the lack of data needed for training current \acrshort{dl}-based solutions. Since the labeling budget is limited, data scarcity represents a severe issue that puts a damper on the scalability of supervised neural networks. This is particularly evident in the counting task, where the annotation procedure for the generation of the training data is particularly costly in terms of human effort, considering that images often contain hundreds or even thousands of objects to be labeled;
\item we proposed engineered solutions for the adoption of \acrshort{cnn}-based techniques in environments with limited power resources, moving all the computations directly onboard embedded vision systems.
\end{itemize}

\ref{ch:introduction} and \ref{ch:background} provided an introduction and background knowledge about the main visual counting approaches, both traditional and \acrshort{cnn}-based, and the \acrlong{da} topic. 

In \ref{ch:counting-on-the-edge} we explored  \acrshort{dl}-based solutions for counting vehicles onboard embedded vision systems, i.e., devices equipped with limited computational capabilities that can capture images and elaborate them. Specifically, we proposed a \acrshort{cnn}-based technique aiming at detecting and estimating the number of vehicles present in images taken by a \acrlong{uav}. We showed through experimental evaluation that our solution outperforms state-of-the-art in various real-world datasets, running at a speed of 4 \acrshort{fps} on an NVIDIA Jetson TX2 board. More, we introduced a novel multi-camera system that, combining a \acrshort{cnn}-based detector and a decentralized geometry-based approach, can precisely and automatically estimate the number of cars present in an entire parking lot, analyzing images gathered from multiple smart cameras. We made available a novel dataset that we exploited for the experiments, and we demonstrated that our solution benefits from redundant information from different cameras while improving overall performance.

To tackle the problem of data scarcity, in \ref{ch:virtual-to-real} we explored the adoption of synthetic data gathered from virtual environments resembling the real world. Here, the main advantage is that by interacting with the graphical renderer, it is possible to automatically collect the annotations associated with the images, which can then be used to train \acrshort{dl}-based algorithms. Specifically, in this chapter, we introduced \acrfull{viped}, a novel synthetic collection of images with precise bounding boxes localizing people in urban scenarios. We exploited this great amount of labeled data to train \acrshort{cnn}-based detector for the pedestrian detection task, the main building block for a myriad of applications, including people counting. We showed through experiments on some real-world datasets present in the literature that the detector achieves comparable or better results using synthetic data during the training phase rather than relying solely on real-world images. However, the potential of synthetic data cannot be fully exploited due to the Synthetic-to-Real Domain Shift, i.e., the image appearance difference between the synthetic training data and the real-world ones. Thus, to mitigate this domain gap, we also proposed two different supervised \acrfull{da} strategies, suitable for the pedestrian detection task but possibly applicable to general object detection. An extensive experimental evaluation demonstrated that the two proposed techniques boost the performance over specific real-world scenarios, reducing the Synthetic2Real Domain Shift by bringing the two domains closer together and thus achieving better results.

In \ref{ch:uda-counting}, we overcame the limitations that turned out from the adoption of the two supervised \acrshort{da} strategies presented in the previous chapter. Indeed, employing these two techniques to mitigate the domain gap between the synthetic and real-world data distributions has a price since it still relies on labeled data from the real-world domain. On the other hand, in this chapter, we proposed an end-to-end \acrshort{cnn}-based \acrshort{uda} algorithm for traffic density estimation and counting. During the training phase, we relied on the supervision provided by the labels automatically collected from the \acrfull{gta} dataset, a novel collection of synthetic images we gathered on purpose from a virtual environment; at the same time, we inferred some knowledge from real-world unlabeled images. In other words, we tackled the problem of data scarcity from two complementary sides: on the one hand, we exploited the significant dimension and variability of the synthetic data, while, on the other hand, we mitigated the \textit{Synthetic2Real} domain gap in an unsupervised fashion. To the best of our knowledge, our solution was the first attempt to introduce a \acrshort{uda} scheme for the counting task. More, we conducted experiments not only considering the \textit{Synthetic2Real} domain gap, but we also took into account other domain shifts. Specifically, we accounted for i) the \textit{Camera2Camera} domain shift, where training images are gathered from a set of cameras, while test data comes from different sources having different perspectives, and ii) the \textit{Day2Night} domain shift, where training images are collected during the night while test data is gathered during the night. An extensive performance evaluation over many datasets showed the superiority of our approach compared to state-of-the-art techniques.

Whereas in most cases, the objects to be counted can be unambiguously flagged by human raters during the annotation procedure, there are some circumstances in which this is hard to achieve. For instance, this often occurs in medical images where non-trivial intrinsic patterns mislead annotators. In \ref{ch:counting-with-uncertainty}, we tackled the task of counting biological structures from microscopy images under the assumption of having training datasets characterized by weak labels, that is, in the presence of non-negligible disagreement between multiple raters. Essentially, also in this chapter, we addressed the problem of data scarcity, but under a different setting, i.e., in the presence of weak annotations due to raters’ judgment differences. To this end, we proposed a two-stage counting strategy. In the first stage, we trained state-of-the-art \acrshort{dl}-based methodologies to detect and count biological structures exploiting a large set of single-rater labeled data sure to contain errors; in the second stage, using a small set of multi-rater data, we refined the predictions, increasing the correlation between the scores assigned to the samples and the agreement of the raters on the annotations. We assessed our pipeline on commonly available cell counting datasets, and we also created on purpose a novel dataset comprising fluorescence microscopy images of mice brains containing extracellular matrix aggregates named perineuronal nets. Through an extensive experimental evaluation, we demonstrated that we significantly enhanced counting performance, improving confidence calibration by taking advantage of the redundant information characterizing the multi-rater data.
 
Finally, from a more practical view, in \ref{ch:counting-for-covid} we presented a modular Computer Vision-based framework, deployed in a real use-case scenario on a low-cost off-the-shelf embedded platform, capable of automatically monitoring human activities in critical environments, where individual and collective safety must be constantly checked. Indeed, as evidenced during the recent COVID-19 pandemic, there exist scenarios in which ensuring compliance to a set of guidelines becomes crucial to secure a safe living environment in which human activities can be conducted. Our solution put in practice some of the techniques described in the previous chapters and consists of multiple modules, each responsible for specific functionality, such as counting people present in a region of interest, a piece of crucial information to monitor the area occupancy and consequently to drastically reduce the likelihood of setting up people gatherings. 
To validate our solution, we tested all the functionalities that our framework makes available, exploiting two novel datasets that we collected and annotated on purpose. One of them comprises images captured by a smart camera located in a public square in the city of Pisa, Italy, and it represents a typical scenario for which it is crucial to monitor the compliance of a ruled maximum number of people allowed to stay in the site. 
Experimental evaluation showed that our system can effectively carry out all the functionalities that the user can set up, providing a valuable asset to automatically monitor compliance with safety rules.

\section{Future Work}
There are multiple aspects discussed in this thesis that are worth further investigating. We report in the following the most promising research directions.

\paragraph{Unsupervised domain adaptation for density estimation and counting.}
In \ref{ch:uda-counting}, we addressed the problem of determining the density and the number of vehicles present in large sets of images captured by city cameras, tackling the domain shift existing between the training and the test data distributions without using additional labels. We achieved this generalization by exploiting an \acrshort{uda} strategy, attaching a discriminator to the output, and forcing similar density distribution in the target and source domains. Given the generality of the proposed approach, one line of future research could be to apply our strategy to more general counting tasks, not limited to estimating the number of vehicles in urban scenarios. More, given the conventional structure of the estimator, the improvement obtained by just monitoring the output suggests us the application of similar principles to the inner layers of the network, further enhancing the adapted model by constructing a multi-level adversarial network that performs domain adaptation at different feature levels, involving more robustly lower-level features that are far away from the high-level output labels. Finally, very recently, Kang et al. \cite{DBLP:conf/nips/KangW0ZH20} proposed a pixel-level \acrshort{da} technique for the semantic segmentation task, where a cycle association between source and target pixel pairs contrastively strength their connections to diminish the domain gap making the features more discriminative; this technique should be further explored to be applied in the counting by density estimation task.

\paragraph{Virtual To Real Adaptation of Pedestrian Detectors.} \acrshort{da} applied to the object detection task remains a relatively unexplored field. Most of the existing techniques rely on two-stage detectors because they exploit \acrshort{da} at the classification level, considering the object proposals produced in the first step of the detection pipeline. On the other hand, one-stage detectors are not involved in \acrshort{da} strategies due to their architecture that is not flexible enough to adapt it to this task. A possible research line is to account for anchor-less detectors and apply \acrshort{uda} strategies to the heatmaps they produce to localize the objects, thus considering the entire detection pipeline and not only the final classification stage. 

\paragraph{Counting with raters' uncertainty.} Counting objects using datasets containing multiple annotations per image remains a fairly unexplored problem. In \ref{ch:counting-with-uncertainty}, we proposed a two-stage methodology for counting biological structures in microscopy images under weak-labeled conditions. However, the proposed solution still has some limitations that we believe should be tackled in the future. Specifically, the two stages presented are implemented by two separate models, increasing the computational cost of the solution. In fact, although the scorer model is small compared to the localization model, formulating a unique model, still trained in two separate stages, could deliver the same counting performance while sharing parameters among stages and decreasing the overall computation needed. In light of our results, the detection-based and the segmentation-based models, whose scores exhibited a fair correlation with the raters' agreement in general, are the most promising solution to be extended for integrating the rescoring stage. Additional improvements can be made in scenarios where severe occlusions and overlaps between objects to be counted occur. Along this path, an interesting research direction concerns the extension of density-based methods to model also the objectness or detection confidence of objects. This could increase the counting performance in crowded scenarios while permitting confidence-based filtering of objects. Finally, another line of research could be to apply similar strategies also in the crowd counting task. Indeed, as outlined by \cite{composition_loss}, also the datasets belonging to this field suffer from mistaken ground-truth labels since there are cases where annotations are not done on parts of the images as it is virtually impossible to distinguish heads of neighboring people; a small set of images labeled by multiple annotators could be helpful, by fully exploiting the redundancy of the information represented by the different raters' opinions. 

\paragraph{Synthetic datasets for multi-camera multi-target vehicle tracking and counting.} Synthetic \\ datasets are helpful in contrast to the lack of training data, as we have already seen in previous chapters of this dissertation. The more the task is greedy of complex labels, the more synthetic data are appealing. The multi-camera multi-target vehicle tracking and counting task aims to identify and track vehicles in a multi-camera system, counting the instances across different places during a time interval. The annotation procedure is costly and error-prone since it requires that annotators examine various video sources identifying and localizing the same vehicle instances. To this end, we believe that a synthetic dataset gathered from a virtual environment by putting multiple cameras placed at multiple intersections and where annotations are automatically collected could be a promising research line.

\paragraph{Improvements for human activity monitoring in critical environments.} In \ref{ch:counting-for-covid}, we presented a modular Computer Vision-based framework helpful for monitoring human activities and able to carry out functionalities such as counting people in two different modalities, localizing \acrshort{ppe} and computing social distance. We plan to develop an algorithm that can automatically select the best counting modality between instancing and density map, which is currently chosen manually by the user. At the same time, we will try to integrate and expand modules with further visual analyses,  like gesture/posture recognition and the assessment of appropriate \acrshort{ppe} wearing, as already addressed in \cite{DBLP:journals/corr/abs-2103-08773}. Finally, we will attempt to apply a transfer learning approach to predict physical distances among people by using an automatically labeled computer-generated training set based on a rendering engine simulation.

\cleardoublepage
\phantomsection
\addcontentsline{toc}{chapter}{\bibname}
\small
\bibliographystyle{plainnat}
\bibliography{thesis}

\end{document}